\documentclass[10pt,twocolumn,letterpaper]{article}

\PassOptionsToPackage{table}{xcolor}

\usepackage[pagenumbers]{sty/cvpr}     %

\usepackage{algorithm}
\usepackage{algorithmic}
\usepackage{xcolor}
\usepackage{array}   %
\usepackage{makecell}
\usepackage{colortbl} %
\usepackage{caption}
\usepackage{calc} %
\usepackage{adjustbox}
\usepackage{tikz}
\usepackage{relsize}
\usepackage{lipsum} %
\usepackage{longtable}
\usepackage{tabularx}
\usepackage{subcaption}
\usepackage{dashrule}
\usepackage{afterpage}
\renewcommand{\afterpage}[1]{#1}
\usepackage{pgfplots}
\usepackage{multirow}

\pgfplotsset{compat=1.18}

\definecolor{tabfirst}{rgb}{0.769, 0.628, 0.843} %
\definecolor{tabsecond}{rgb}{0.898, 0.847, 0.930} %
\definecolor{tabthird}{rgb}{0.962, 0.945, 0.973} %
\definecolor{bordeaux}{RGB}{128,0,32} %
\definecolor{darkgreen}{rgb}{0,0.3,0}

\definecolor{mypurple}{HTML}{B47CC7}    %
\definecolor{myblue}{HTML}{4878CF}
\definecolor{mygreen}{HTML}{009E73}
\definecolor{mybrown}{HTML}{95848E}
\definecolor{mylightblue}{HTML}{56B4E9}
\definecolor{mypink}{HTML}{DFB3CC}
\definecolor{mypurple}{HTML}{B04080}
\definecolor{anothergreen}{HTML}{009e4b}
\definecolor{ablresplat}{HTML}{009E73}  %
\definecolor{ablg3r}{HTML}{8C564B}      %
\definecolor{abladam}{HTML}{AE72CA}     %
\definecolor{ablcbor}{HTML}{983DC2}     %
\definecolor{ablscale}{HTML}{762999}    %
\definecolor{ablours}{HTML}{531471}     %

\newcommand{\ours}{Learn2Splat}
\newcommand{\ourshort}{L2S}
\newcommand{\inner}{\text{inner}}
\newcommand{\meta}{\text{meta}}
\newcommand{\tinner}{t}
\newcommand{\tmeta}{t_{\text{meta}}}
\newcommand{\gaus}{\boldsymbol{\cG}}
\newcommand{\gaust}{\boldsymbol{\cG}_t}
\newcommand{\gaustt}[1]{\boldsymbol{\cG}_{#1}}
\newcommand{\ggaus}{\nabla_{\gaus_{\tinner}}}
\newcommand{\ggaust}[1]{\nabla_{\gaus_{#1}}}
\newcommand{\aggaust}[1]{\nabla_{\gaus_{#1}}^{\text{adam}}}
\newcommand{\gtheta}{\nabla_{\btheta_{\tmeta}}}

\usepackage{gensymb}
\usepackage{wrapfig}
\usepackage{soul}

\makeatletter
\DeclareRobustCommand{\mysparse}{%
\begin{tikzpicture}[baseline=-0.35ex]
    \ifdim\f@size pt<9pt
      \node at (0,0) [inner sep=0pt, font=\sffamily\bfseries] {\scalebox{0.5}{S}};
    \else
      \node at (0,0) [inner sep=0pt, font=\sffamily\bfseries] {\scalebox{0.65}{S}};
    \fi
  \end{tikzpicture}%
}
\makeatother

\makeatletter
\DeclareRobustCommand{\mydense}{%
  \begin{tikzpicture}[baseline=-0.35ex]
    \ifdim\f@size pt<9pt
      \node at (0,0) [inner sep=0pt, font=\sffamily\bfseries] {\scalebox{0.5}{D}};
    \else
      \node at (0,0) [inner sep=0pt, font=\sffamily\bfseries] {\scalebox{0.65}{D}};
    \fi
  \end{tikzpicture}%
}
\makeatother

\newcommand{\ourssparse}{$\text{L2S}^{\mysparse}$}
\newcommand{\oursdense}{$\text{L2S}^{\mydense}$}

\newlength{\cellradius}
\newif\ifrounded\roundedtrue
\newcommand{\rcell}[2][width=\linewidth]{%
  \ifrounded
    \tikz\node[inner sep=0pt,outer sep=0pt,rounded corners=\cellradius,clip]{\includegraphics[#1]{#2}};%
  \else
    \includegraphics[#1]{#2}%
  \fi}

\newlength{\cellborderwidth}
\newlength{\bcellradius}
\definecolor{cellborderadam}{RGB}{230,159,0}   %
\definecolor{cellborderours}{RGB}{158,94,187}  %
\definecolor{cellborderref}{RGB}{188,92,82}    %
\newcommand{\bcell}[3][width=\linewidth]{%
  \ifrounded\setlength{\bcellradius}{\cellradius}\else\setlength{\bcellradius}{0pt}\fi
  \begin{tikzpicture}[inner sep=0pt]%
    \node[inner sep=0pt,outer sep=0pt] (bcellim) {\rcell[#1]{#3}};%
    \draw[#2, line width=\cellborderwidth, rounded corners=\bcellradius]
      ([shift={( 0.5\cellborderwidth, 0.5\cellborderwidth)}]bcellim.south west) rectangle
      ([shift={(-0.5\cellborderwidth,-0.5\cellborderwidth)}]bcellim.north east);%
  \end{tikzpicture}}

\newif\ifshowpsnr\showpsnrtrue
\newcommand{\psnrbadgetextfrac}{0.055}
\newcommand{\psnrbadgepadfrac}{0.3}
\newcommand{\psnrbadgemarginfrac}{0.00625}
\newlength{\pcellfont}\newlength{\pcellpad}\newlength{\pcellmargin}
\newcommand{\pcell}[3][width=\linewidth]{%
  \ifshowpsnr
    \setlength{\pcellfont}{\psnrbadgetextfrac\linewidth}%
    \setlength{\pcellpad}{\psnrbadgepadfrac\pcellfont}%
    \setlength{\pcellmargin}{\psnrbadgemarginfrac\linewidth}%
    \begin{tikzpicture}[inner sep=0pt]%
      \node[inner sep=0pt] (pcellim) {\rcell[#1]{#3}};%
      \node[anchor=north west, inner xsep=\pcellpad, inner ysep=0.65\pcellpad,
            rounded corners=0.35\pcellfont, fill=black!86, fill opacity=0.71,
            text opacity=1, text=white,
            font=\fontsize{\pcellfont}{\pcellfont}\selectfont\sffamily]
        at ([shift={(\pcellmargin,-\pcellmargin)}]pcellim.north west) {#2};%
    \end{tikzpicture}%
  \else
    \rcell[#1]{#3}%
  \fi}

\newcommand{\adam}{Adam}
\newcommand{\threedgslm}{3DGS-LM\xspace}
\newcommand{\adamtuned}{\texorpdfstring{Adam$^{\text{tuned}}$}{Adam-tuned}}
\newcommand{\lob}{\texorpdfstring{G3R$^{\star}$}{G3R*}}
\newcommand{\supmat}{sup.~mat.}
\newcommand{\tbud}{\ensuremath{T_{4\text{k}}}}

\newcommand{\bc}{\mathbf{c}}
\newcommand{\bd}{\mathbf{d}}

\newcommand{\bH}{\mathbf{H}}
\newcommand{\bI}{\mathbf{I}}
\newcommand{\bJ}{\mathbf{J}}
\newcommand{\bK}{\mathbf{K}}

\newcommand{\bm}{\mathbf{m}}

\newcommand{\bo}{\mathbf{o}}\newcommand{\bO}{\mathbf{O}}
\newcommand{\bp}{\mathbf{p}}
\newcommand{\bq}{\mathbf{q}}
\newcommand{\bR}{\mathbf{R}}
\newcommand{\bs}{\mathbf{s}}\newcommand{\bS}{\mathbf{S}}
\newcommand{\bt}{\mathbf{t}}
\newcommand{\bu}{\mathbf{u}}
\newcommand{\bv}{\mathbf{v}}

\newcommand{\bx}{\mathbf{x}}

\newcommand{\btheta}{\boldsymbol{\theta}}

\newcommand{\bmu}{\boldsymbol{\mu}}

\newcommand{\brho}{\boldsymbol{\rho}}
\newcommand{\bSigma}{\boldsymbol{\Sigma}}

\newcommand{\nR}{\mathbb{R}}

\newcommand{\cB}{\mathcal{B}}

\newcommand{\cG}{\mathcal{G}}

\newcommand{\cL}{\mathcal{L}}

\newcommand{\cV}{\mathcal{V}}

\DeclareMathOperator*{\argmin}{argmin~}

\makeatletter
\DeclareRobustCommand\onedot{\futurelet\@let@token\@onedot}
\def\@onedot{\ifx\@let@token.\else.\null\fi\xspace}
\def\eg{e.g\onedot} 
\def\ie{i.e\onedot}

\makeatother

\newcommand{\boldparagraph}[1]{\vspace{0.2cm}\noindent{\bf #1} }

\newcommand{\sh}{\bS\bH}

\definecolor{cvprblue}{rgb}{0.21,0.49,0.74}
\definecolor{cvprred}{rgb}{0.78,0.18,0.22}  %
\usepackage[pagebackref,breaklinks,colorlinks,citecolor=cvprblue,linkcolor=cvprred,urlcolor=cvprblue,bookmarksnumbered]{hyperref}

\usepackage{etoc}

\title{\ours{}: Extending the Horizon of \\ Learned 3DGS Optimization}

\author{
Naama Pearl$^{1,*}$ \quad
Stefano Esposito$^{1,*}$ \quad
Haofei Xu$^{1,2}$ \quad
Amit Peleg$^{1}$ \quad
Patricia Gscho{\ss}mann$^{1}$\\
Lorenzo Porzi$^{3}$ \quad
Peter Kontschieder$^{3}$ \quad
Gerard Pons-Moll$^{1}$ \quad
Andreas Geiger$^{1}$\\[5pt]
{\normalsize $^{1}$University of T\"ubingen, T\"ubingen AI Center \quad
$^{2}$ETH Zurich \quad
$^{3}$Meta Reality Labs}\\[2pt]
{\normalsize $^{*}$Equal contribution.}
}

\begin{document}

\twocolumn[{%
  \renewcommand\twocolumn[1][]{#1}
    \maketitle
    \begin{center}
  \captionsetup{type=figure}

\resizebox{1.0\linewidth}{!}{%

\begin{minipage}[t]{0.64467\linewidth}
\centering
\vspace{0pt}
\setlength{\tabcolsep}{0.001551\linewidth} %
\renewcommand{\arraystretch}{0} %
\begin{tabular}{@{}%
    >{\raggedleft\arraybackslash}m{0.04033\linewidth}@{}
    >{\centering\arraybackslash}m{0.23759\linewidth}
    >{\centering\arraybackslash}m{0.23759\linewidth}
    >{\centering\arraybackslash}m{0.23759\linewidth}
    >{\centering\arraybackslash}m{0.23759\linewidth}
@{}}
     & \smaller[2]{$t = 2$} & \smaller[2]{$t = 10$} & \smaller[2]{$t = 100$} & \smaller[2]{$t = 1000$} \\
    \noalign{\vskip 1pt}
    \rotatebox{90}{\makecell{\smaller[3]{\lob{}}}} &
    \pcell[width=\linewidth]{21.4\,dB}{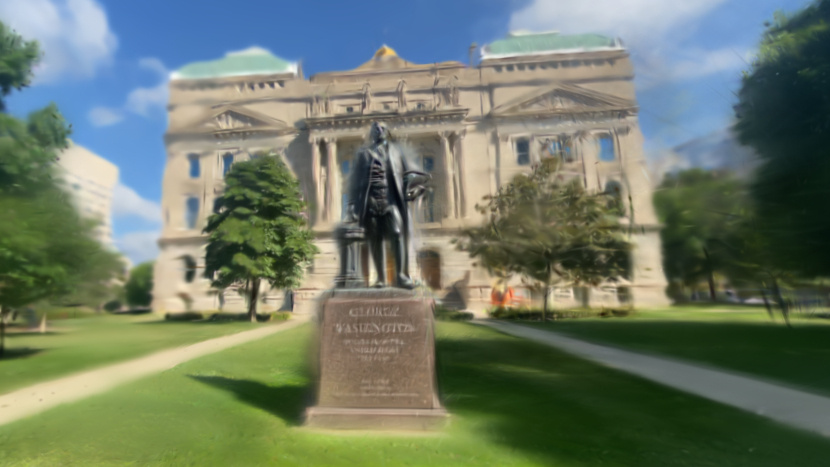} &
    \pcell[width=\linewidth]{23.3\,dB}{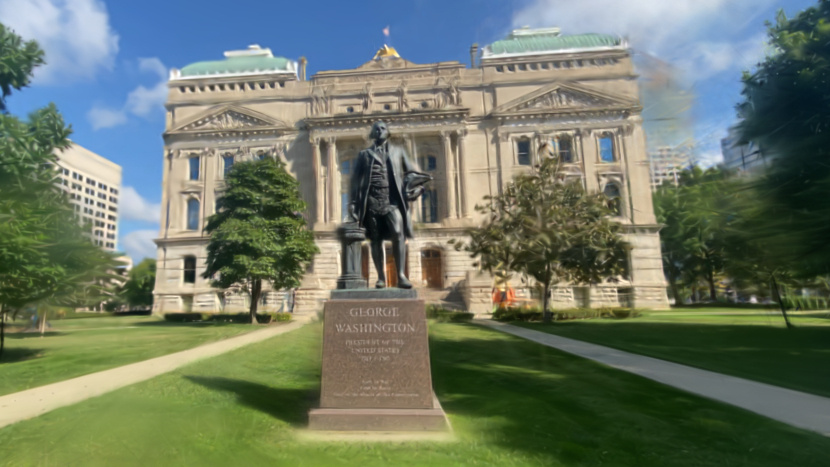} &
    \pcell[width=\linewidth]{20.9\,dB}{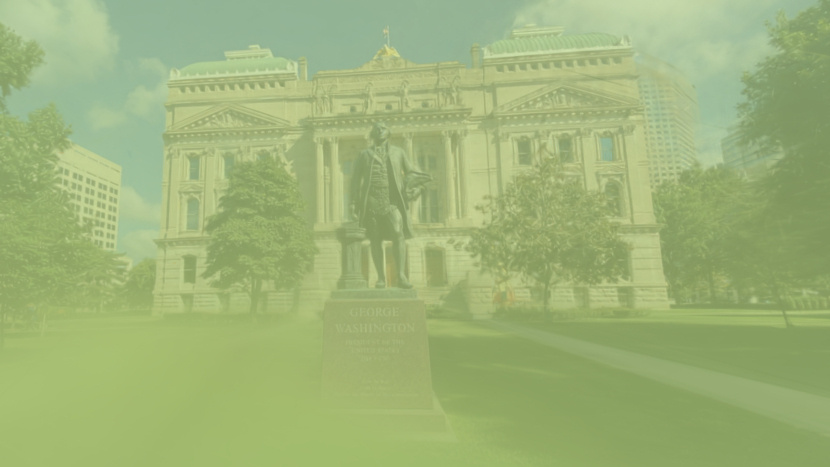} &
    \pcell[width=\linewidth]{22.3\,dB}{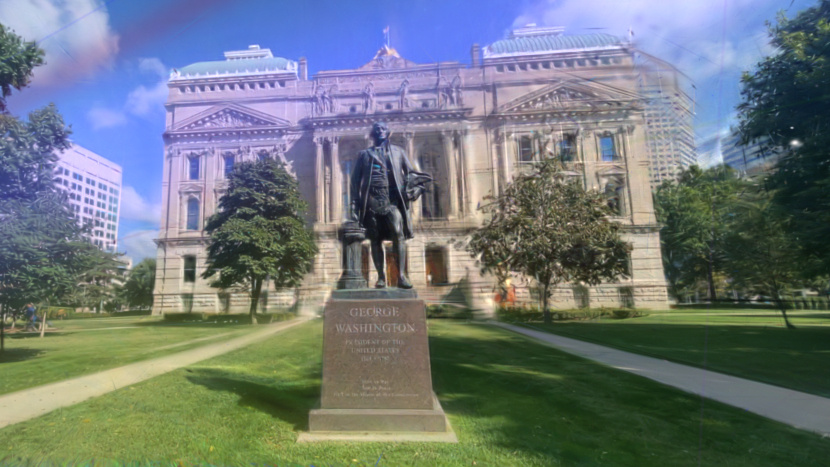} \\
    \noalign{\vskip 2\tabcolsep}
    \rotatebox{90}{\makecell{\smaller[3]{\adam{}}}} &
    \pcell[width=\linewidth]{19.0\,dB}{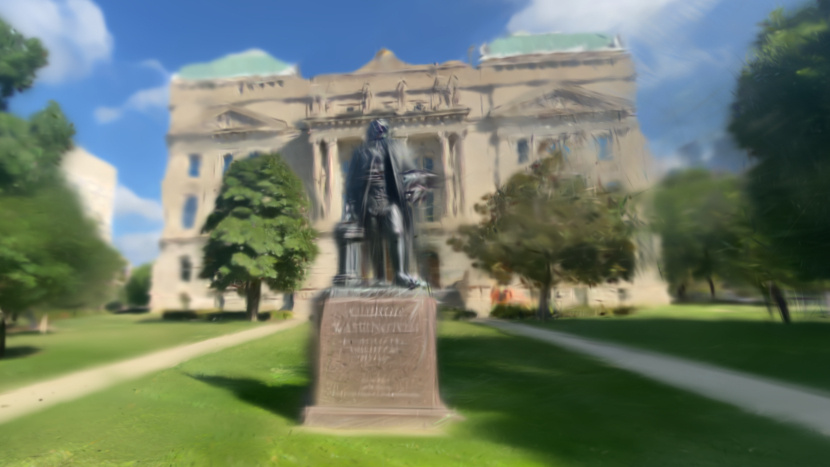} &
    \pcell[width=\linewidth]{20.0\,dB}{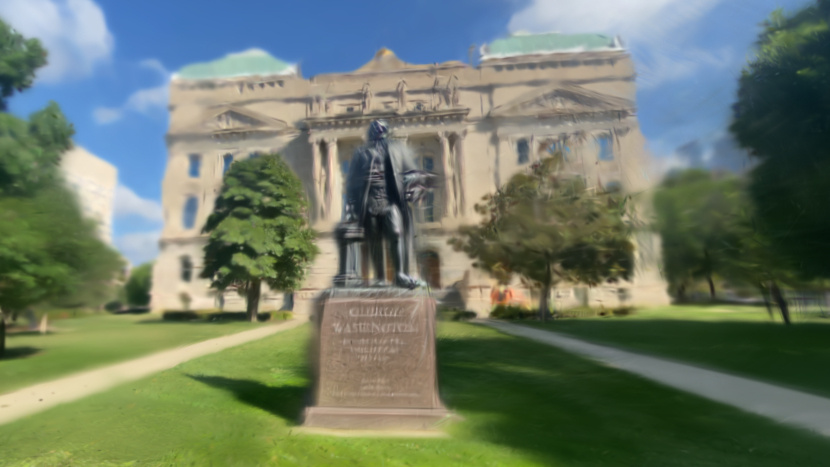} &
    \pcell[width=\linewidth]{22.9\,dB}{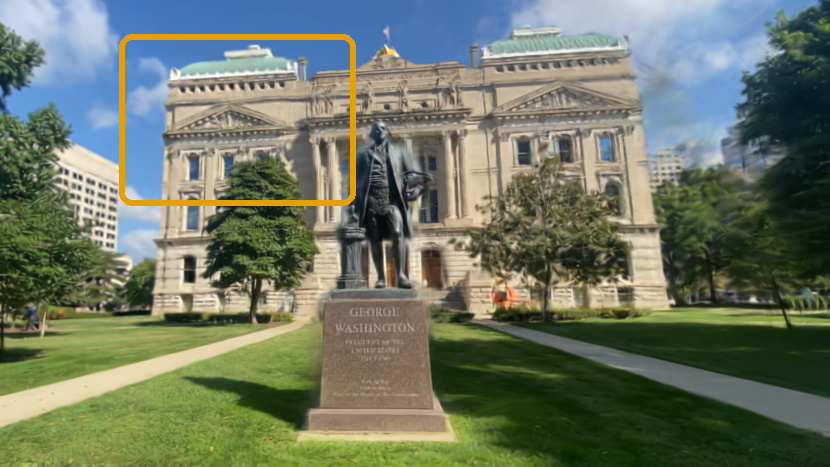} &
    \pcell[width=\linewidth]{23.0\,dB}{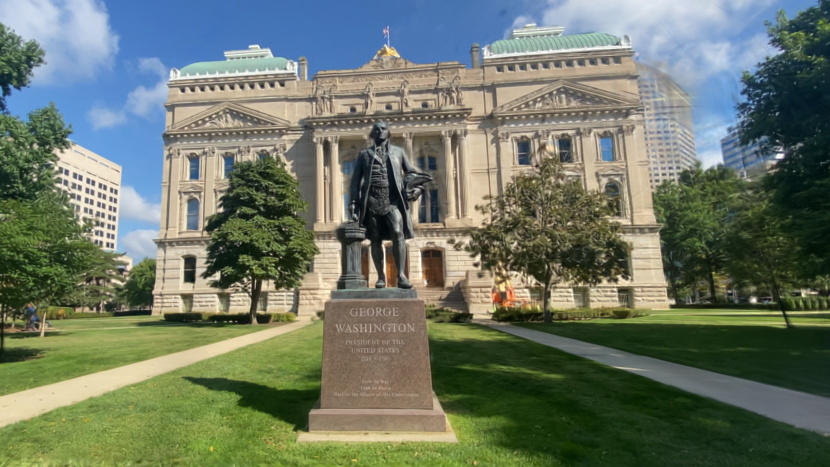} \\
    \noalign{\vskip 2\tabcolsep}
    \rotatebox{90}{\makecell{\smaller[3]{\ourshort{}}}} &
    \pcell[width=\linewidth]{21.4\,dB}{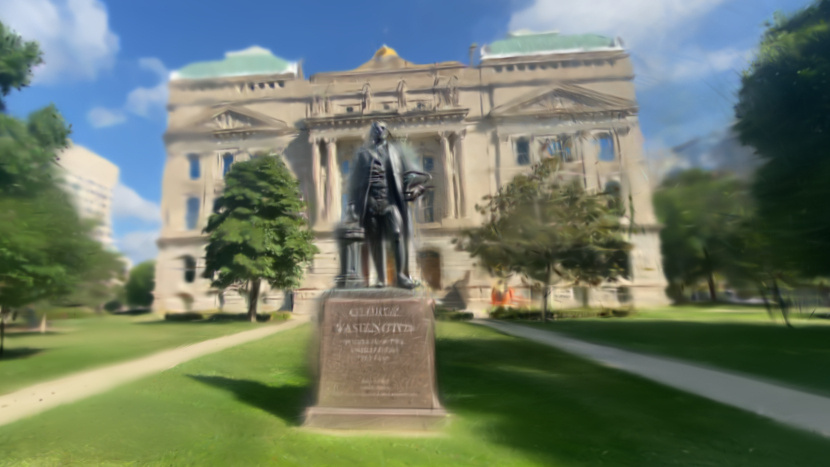} &
    \pcell[width=\linewidth]{23.1\,dB}{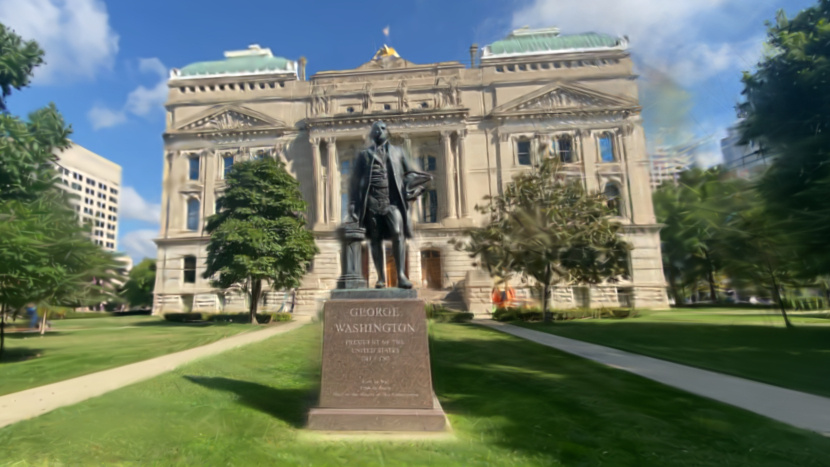} &
    \pcell[width=\linewidth]{23.8\,dB}{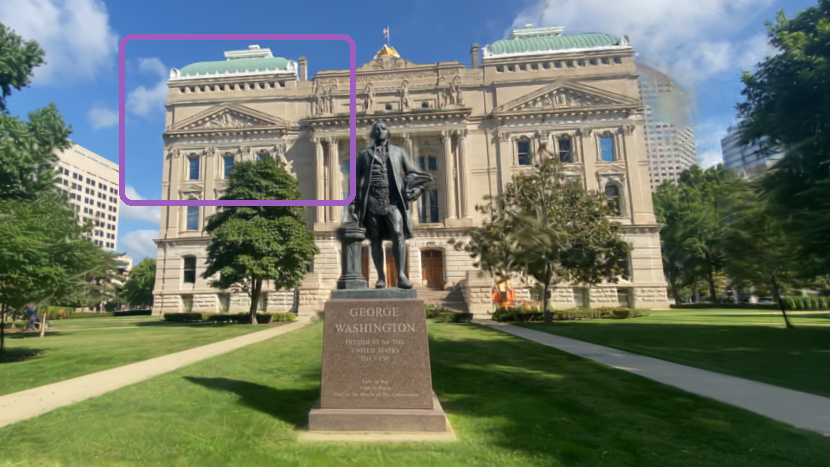} &
    \pcell[width=\linewidth]{23.8\,dB}{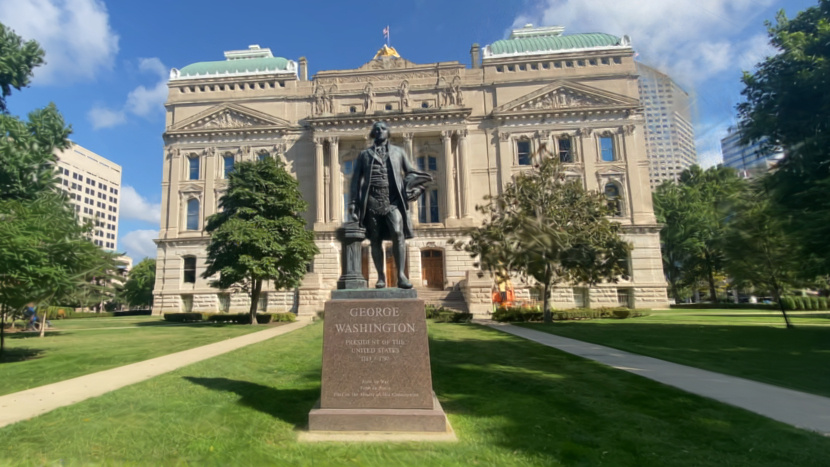} \\
\end{tabular}
\end{minipage}%
\hspace{0.002\linewidth}
\begin{minipage}[t]{0.30833\linewidth}
\centering
\vspace{0pt}
\setlength{\tabcolsep}{0.003243\linewidth}%
\renewcommand{\arraystretch}{0}%
\begin{tabular}{@{}%
    >{\centering\arraybackslash}m{0.49676\linewidth}
    >{\centering\arraybackslash}m{0.49676\linewidth}
@{}}
    \smaller[2]{Initialization} & \smaller[2]{Reference} \\[1pt]
    \pcell[width=\linewidth]{18.6\,dB}{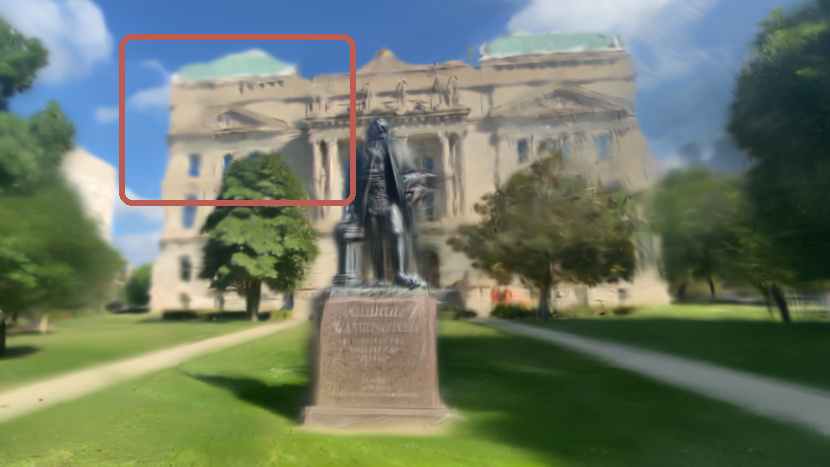} &
    \rcell[width=\linewidth]{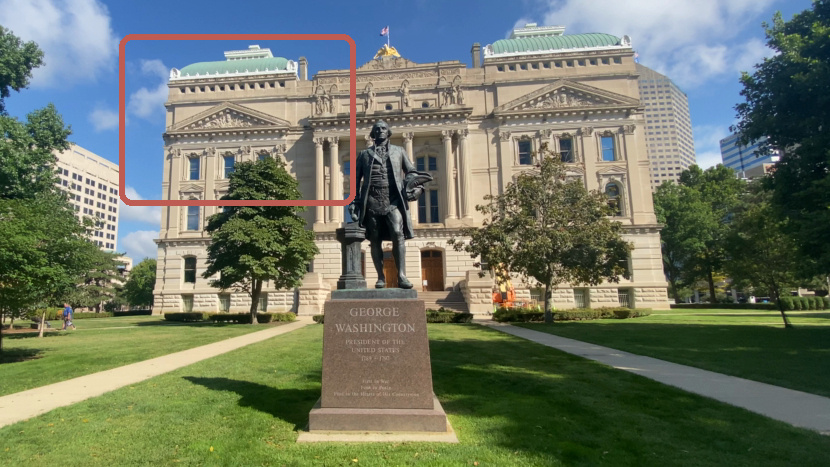} \\
\end{tabular}
\par\vspace*{0.3mm}
\setlength{\tabcolsep}{0.016216\linewidth}%
\renewcommand{\arraystretch}{0}%
\begin{tabular}{@{}%
    >{\centering\arraybackslash}m{0.23378\linewidth}
    >{\centering\arraybackslash}m{0.23378\linewidth}
    >{\centering\arraybackslash}m{0.23378\linewidth}
@{}}
    \smaller[4]{Init} & \smaller[4]{\adam{}\,{\tiny @100}} & \smaller[4]{\ourshort{}\,{\tiny @100}} \\[1pt]
    \bcell{cellborderref}{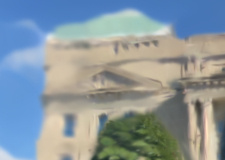} &
    \bcell{cellborderadam}{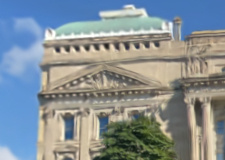} &
    \bcell{cellborderours}{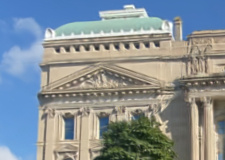} \\
\end{tabular}
\par\vspace*{0.3mm}
\setlength{\tabcolsep}{0pt}%
\renewcommand{\arraystretch}{0}%
\begin{tabular}{@{}%
    >{\centering\arraybackslash}m{0.66\linewidth}
    @{\hspace{0mm}}
    >{\centering\arraybackslash}m{0.31\linewidth}
@{}}
    \includegraphics[width=\linewidth]{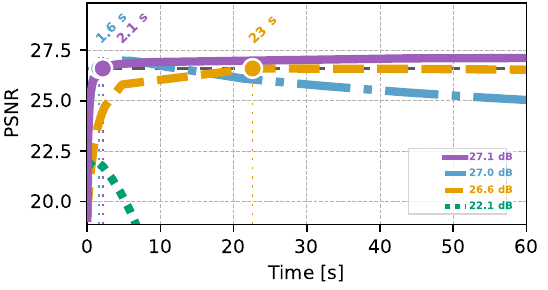} &
    \hspace*{-1mm}\raisebox{7pt}{\includegraphics[width=\linewidth]{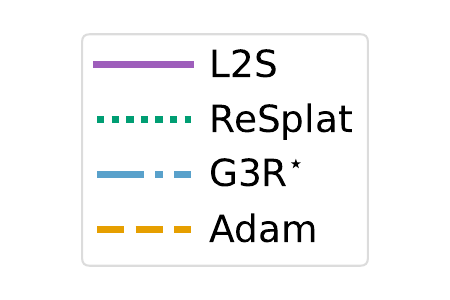}} \\
\end{tabular}
\end{minipage}

}

  \captionof{figure}{
    \textbf{\ours{} (\ourshort{}) is a learned optimizer for 3DGS that stays above \adam{} throughout optimization and keeps improving over long horizons, where prior learned optimizers degrade}.
    Prior learned optimizers~\cite{xu2025resplat,Chen2024g3r,Liu2025quicksplat} rely on learning rate (LR) schedules or time encodings tied to a predefined iteration count, beyond which reconstruction degrades~(\cref{subsec:results}).
    ~\ourshort{} works on SfM point-cloud or feed-forward initializations and maintains performance over long optimization horizons through a meta-training scheme and architectural modifications~(\cref{sec:method}).
    Although trained on low-resolution scenes, it generalizes zero-shot to other datasets and settings.
    Here, we show a high-resolution (zero-shot) sparse-view reconstruction on DL3DV~\cite{ling2024dl3dv}, comparing to \adam{}~\cite{Kingma2015ICLR}, ReSplat~\cite{xu2025resplat} and \lob{}, our implementation of the G3R~\cite{Chen2024g3r} training scheme.
    The right panel marks where each run first reaches \adam{}'s peak PSNR, and lists each method's best PSNR.
    Renders are shown at iterations ($t$) along the optimization, and the crops compare \adam{} and \ourshort{} at iteration $100$.
  }
  \label{fig:teaser}
\end{center}

}]

    \etocsettocdepth.toc{none}
    \begin{abstract}
3D Gaussian Splatting (3DGS) optimization is most commonly performed using general-purpose first-order optimizers such as Adam or SGD.
Although robust across scenes, they update each parameter independently without exploiting the structural and spatial relationships among Gaussians, which slows convergence.
Recent works introduced learned optimizers that predict correlated updates informed by inter-parameter and inter-Gaussian dependencies.
However, those are trained for a fixed number of optimization iterations and rely on manually scheduled learning rates to avoid degradation.
In this paper, we introduce \ours{}, a learned optimizer for 3DGS that avoids degradation over extended optimization horizons without auxiliary mechanisms.
To enable this, we propose a meta-learning scheme that extends the optimization horizon via a checkpoint buffer and an optimizer rollout strategy, combined with an architecture that encodes gradient scale information in its latent states.
Results show higher novel view synthesis quality at equal wall-clock time, while remaining stable over long horizons, with zero-shot generalization to unseen datasets and settings.
To support our findings, we build a unified framework to train and evaluate learned and standard optimizers across sparse and dense view settings.
Code and models will be released publicly.
Our project page is available at \texttt{\url{https://autonomousvision.github.io/learn2splat}}.

\end{abstract}

    \section{Introduction}
\begin{figure}[!t]

    \centering

    \centering
    \resizebox{1.0\linewidth}{!}{%
        \begin{subfigure}[b]{0.315\linewidth}
            \centering
            \includegraphics[
                width=\textwidth,
                trim=25pt 0pt 15pt 0pt,
                clip
            ]{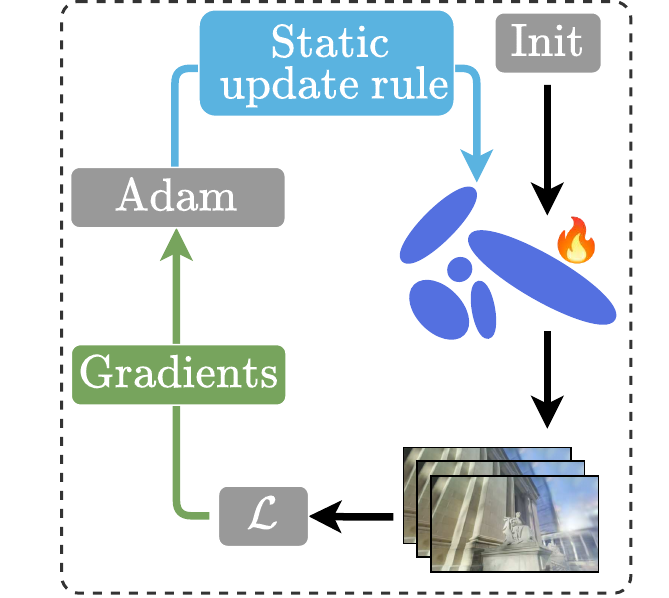}
            \caption{Standard optimizer}
            \label{fig:standard-opt}
        \end{subfigure}
        \begin{subfigure}[b]{0.2\linewidth}
            \centering
            \includegraphics[
                width=0.9\textwidth,
                trim=0pt 0pt 0pt 0pt,
                clip
            ]{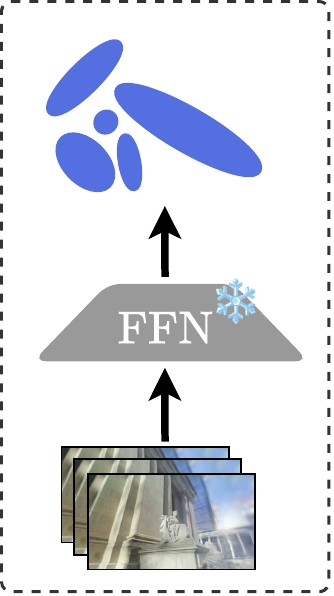}
            \caption{FFN}
            \label{fig:ffn}
        \end{subfigure}
        \begin{subfigure}[b]{0.4059\linewidth}
            \centering
            \includegraphics[
                width=\textwidth,
                trim=15pt 0pt 11pt 0pt,
                clip
            ]{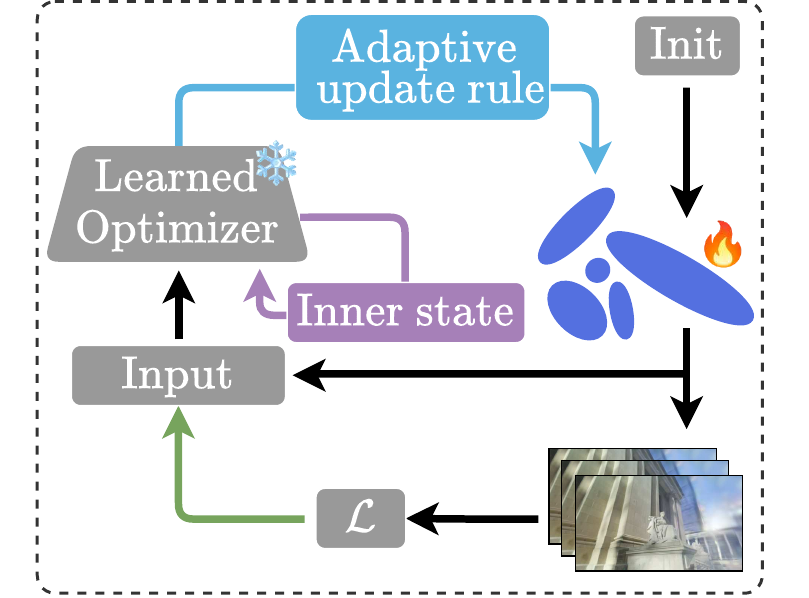}
            \caption{Learned optimizer}
            \label{fig:learned-opt}
        \end{subfigure}
    }
    \caption{\textbf{3DGS optimization paradigms.} (a) In per-scene optimization~(\cref{subsec:standard-optimizer}), the scene representation is learned through iterative updates based on loss evaluation, gradient backpropagation, and standard optimizer rules. (b) In feed-forward networks (FFN), the scene representation is predicted in a single forward pass using a pre-trained model. (c) Learned optimizers~(\cref{subsec:learned-optimizer}) iteratively update the scene representation with a frozen pre-trained model that predicts update steps from input signals such as rendering errors or loss gradients.}

\end{figure}

3D Gaussian Splatting (3DGS)~\cite{Kerbl2023SIGGRAPH} has become an efficient, expressive representation for novel view synthesis (NVS), modeling scenes as a set of 3D Gaussians rendered via differentiable rasterization.
However, 3DGS relies on per-scene optimization that typically runs for thousands of iterations for every new scene (\cref{fig:standard-opt}).
General-purpose optimizers such as Adam~\cite{Kingma2015ICLR} and SGD~\cite{Robbins1951AMS, Kiefer1952StochasticEO}, while robust and widely used, are not tailored to the 3DGS loss landscape and training dynamics.
They update each parameter and Gaussian independently, without leveraging priors about scene geometry, appearance, or parameter dynamics. %

A natural attempt to bypass iterative optimization is to directly predict Gaussian parameters from input images using feed-forward networks (FFN, \cref{fig:ffn})~\cite{Charatan2024CVPR, chen2024mvsplat, liu2024mvsgaussian, xu2025depthsplat, xu2025resplat,jiang2025anysplat,liu2025worldmirror}.
These approaches typically operate in sparse-view settings, predicting a set of Gaussians for each input view, commonly one Gaussian per pixel. 
However, capturing fine-grained geometric and photometric details in a single forward pass often exceeds the representational capacity of current architectures. 
Consequently, to match the quality of per-scene optimization methods, these models require additional scene-specific, iterative fine-tuning.
In this work, we refer to this type of model as \emph{learned initializers}, as their predicted set of Gaussians can serve as initialization to any 3DGS optimization strategy.

A complementary strategy preserves the iterative optimization paradigm but replaces the standard optimizer with a \emph{learned optimizer} that predicts parameter updates~(\cref{fig:learned-opt}).
General-purpose learned optimizers operate on the gradients of a predefined loss, typically after some form of preprocessing or normalization~\cite{Andrychowicz2016NEURIPS, Wichrowska2017ICML, bello2017neural}.
By meta-learning across many tasks, they exploit shared patterns while adapting to each instance, yielding flexible, data-driven updates that can outperform hand-crafted rules.
Throughout this work, we use \emph{learned optimizer} to refer to any learning-based iterative method that updates the parameters of another model.

Recent works~\cite{Chen2024g3r, Liu2025quicksplat, xu2025resplat} explore learned optimization strategies within the 3DGS framework. 
While they accelerate optimization and improve visual fidelity, they rely on auxiliary mechanisms such as learning rate (LR) schedules that confine them to a fixed iteration budget, beyond which they saturate or diverge.
We argue that a desirable property of any learned optimizer is long-horizon robustness: as the loss decreases, the optimizer's predicted updates should naturally decay. %
In contrast to prior work, our approach learns this behavior directly.

To this end, we propose \ours{} (\ourshort{}), a learned optimizer for 3D Gaussian Splatting that remains effective across long optimization trajectories.
Our loss formulation provides effective supervision on the predicted updates (\cref{subsec:meta-learning}), while our architecture enables the optimizer to encode the magnitude of the input gradients (\cref{subsec:model_overview}).
Furthermore, our training approach integrates a \emph{checkpoint buffer} (CB) and an \emph{optimizer rollout} (OR) (\cref{subsec:long-horizon-training}) that expose the network to diverse optimization states~(see \cref{fig:teaser}).

To show that our approach is not tied to a specific initialization or view setting, we train a single model jointly across sparse and dense view settings, initialized from FFN predictions and SfM points~\cite{Schoenberger2016CVPR}, respectively.
In each setting, we evaluate zero-shot generalization to unseen resolutions, view counts and datasets, including RealEstate10K~\cite{Zhou2018SIGGRAPH}, LLFF~\cite{mildenhall2019llff}, DTU~\cite{Aanes2016IJCV}, and Mip-NeRF360~\cite{barron2022mipnerf360}.
\ourshort{} stays above \adam{} throughout the optimization, reaching its quality in less wall-clock time and continuing to improve to the end, unlike prior learned optimizers that peak early and degrade.
To support this study, we introduce a unified framework for 3DGS optimization that supports both learned and standard optimizers.

    \section{Related work}
\boldparagraph{Meta-learning and learned optimizers.}
Standard optimizers such as SGD~\cite{Robbins1951AMS} and Adam~\cite{Kingma2015ICLR} are general-purpose methods that require hyperparameter tuning. 
Meta-learning~\cite{chen2022learning}, by contrast, seeks to learn optimizers that generalize across related tasks by encoding priors about the optimization process itself. 
Early work on learning update rules dates back to~\cite{Bengio1990CITESEER, Bengio1992Optimality}, while~\cite{Andrychowicz2016NEURIPS} introduced the first deep learned optimizer for tasks such as MNIST classification and style transfer.
Subsequent research~\cite{Wichrowska2017ICML, Metz2020ARXIV, Metz2022ARXIV} investigated training stability and scalability, proposing strategies for more effective meta-optimization. 
Despite these advancements, general learned optimizers are usually trained on an immense number of tasks for only modest speed-ups.
Several works apply learned optimization to 3D tasks: rigid-body motion estimation~\cite{Lv2019CVPR}, 3D human-model fitting~\cite{Corona2022ECCV}, face fitting~\cite{xiong2013supervised}, and NVS via learned updates to multiplane images~\cite{Flynn2019CVPR, Deng2023ARXIV}.
In this paper, we train a learned optimizer for NVS with 3DGS such that it generalizes across datasets and settings.

\boldparagraph{Standard 3DGS optimization.}
High-quality 3DGS~\cite{Kerbl2023SIGGRAPH} still relies on expensive per-scene optimization with standard gradient-based optimizers, often taking minutes to hours per scene.
To accelerate it, Taming 3DGS~\cite{mallick2024taming3dgs} and~\cite{RotaBulo2024ECCV} make the primitive count deterministic and add low-level speedups; ScaffoldGS~\cite{lu2024scaffoldgs} uses an anchor-based primitive distribution for efficiency on dense scenes; EDGS~\cite{kotovenko2025edgs} shows densification may be unnecessary given strong initialization; and second-order optimizers~\cite{hoellein20253dgslm,lan20253dgs2,pehlivan2026lmrs} replace Adam with Levenberg--Marquardt / Gauss--Newton updates to cut the iteration count, at the cost of higher memory.
However, these methods still rely on heuristics and hyperparameter tuning, limiting scalability.
A parallel line improves densification and adaptive density control~\cite{kheradmand20243dgsmcmc, fang2024minisplatting, fang2024minisplatting2, ning2026heuristicslearnabledensitycontrol,ren2025fastgs}, changing the number and placement of primitives rather than the optimizer itself.

\boldparagraph{Single-step feed-forward 3DGS.}
To address the limitations of per-scene optimization, feed-forward 3DGS models~\cite{Charatan2024CVPR,szymanowicz2024splatter} predict a set of Gaussians from input images in a single inference, and have steadily improved on standard benchmarks~\cite{liu2024mvsgaussian,chen2024mvsplat,xu2025depthsplat,wang2025zpressor,jiang2025anysplat,wang2025volsplat,liu2025worldmirror,depthanything3,Jeong2026viewsplat,Moreau_2026_CVPR}.
However, their quality and generalization remain inherently constrained by this single forward pass~\cite{xu2025resplat}, so they typically serve as \emph{learned initializers} for subsequent optimization.

\boldparagraph{Learned optimizers for 3DGS.}
Recent works explore learned optimization for 3DGS: G3R~\cite{Chen2024g3r} and QuickSplat~\cite{Liu2025quicksplat} employ gra\-dient-con\-di\-tioned networks that predict parameter updates from gradients of the rendering loss.
In particular, G3R~\cite{Chen2024g3r} uses a sparse 3D CNN to infer update steps for a Gaussian scene representation conditioned on Gaussian parameters and their gradients \wrt input views. 
QuickSplat~\cite{Liu2025quicksplat} focuses on surface reconstruction rather than view synthesis, combining an initialization network with interleaved optimization and densification to refine 3DGS scenes from SfM initialization.
ReSplat~\cite{xu2025resplat}, in contrast, predicts updates from rendering errors rather than gradients, and performs well under sparse-view conditions. 
The concurrent work GIFSplat~\cite{chen2026gifsplat} shares a similar idea with ReSplat, but focuses on pose-free settings.

Despite their differences, these learned optimizers share a key limitation: tied to fixed training horizons, they saturate or diverge beyond the iteration count they were trained on.
For instance, QuickSplat applies only five learned steps followed by 2000 regular gradient updates to improve reconstruction quality.
G3R is trained and evaluated for 24 steps using a LR schedule.
ReSplat, trained for four steps, peaks within roughly ten iterations and then diverges.
In contrast, we propose a meta-learning framework that adapts to long horizons in 3DGS optimization, resulting in a stable learned optimizer.
Furthermore, it works from both SfM point-cloud and feed-forward initializations and in sparse and dense-view settings.

    \section{Preliminaries}

In this section, we review the fundamentals of the 3DGS framework (\cref{subsec:3dgs}), standard optimizers (\cref{subsec:standard-optimizer}), and learned optimizers (\cref{subsec:learned-optimizer}) in 3DGS.
A summary of the notation is provided in the \supmat

\subsection{3D Gaussian Splatting}
\label{subsec:3dgs}
The goal of NVS is to generate novel views from a set of training images of a scene.
For a given scene, the training data consists of a set of views $\cV=\{\cV_i\}_{i=1}^N$, each comprising the RGB image $\bI_i$, the corresponding intrinsic matrix $\bK_i$, a translation vector $\bt_i$, and a rotation matrix $\bR_i$. 

3DGS models the 3D world as a set of $G$ 3D Gaussians $\gaus = \{\gaus_m\}_{m=1}^G$.
Each Gaussian $\gaus_m$ is parametrized by its center $\bp_m \in \nR^3$, a rotation represented as a quaternion $\bq_m \in \nR^4$, and a scaling vector $\bs_m \in \nR^3$.
The contribution of each Gaussian in 3D space is determined by an opacity value $\alpha_m \in [0,1]$, and its color is defined by a set of spherical harmonic coefficients $\sh_m \in \nR^{d \times 3}$.
For simplicity, we refer to the set of Gaussians as a matrix $\gaus \in \nR^{G \times p}$, where $p$ is the number of parameters in each Gaussian.
We denote by $\tilde{\bI}_i(\gaus)$ the rendered image of the 3D Gaussians $\gaus$ given a viewpoint $\cV_i$.
The \supmat~contains the full rendering derivation.
The general optimization problem of 3DGS is to fit the Gaussian parameters to the scene by minimizing a loss term,
\begin{equation}
    \argmin_{\gaus} \frac{1}{N} \sum_{i=1}^{N} \cL(\bI_i, \tilde{\bI}_i(\gaus)),
\end{equation}
where $\cL({\bI_i}, \tilde{\bI_i}(\gaus))$ measures the difference between the target image $\bI_i$ and the rendered image $\tilde{\bI}_i(\gaus)$.

\subsection{Standard optimizers}
\label{subsec:standard-optimizer}
In standard 3DGS~\cite{Kerbl2023SIGGRAPH}, gradient-based optimizers update the Gaussian parameters iteratively for each scene~(\cref{fig:standard-opt}). 
At each iteration $t$, one or more training views are sampled, and the gradient of the loss \wrt the current Gaussian parameters is computed.
Omitting the view indices for brevity, we denote this gradient as
\begin{equation}
\nabla_{\gaus_{t}} = \nabla_{\gaus_{t}} \mathcal{L}({\bI}, \tilde{\bI}(\gaus_{t})).
\end{equation}
The update rule for a general optimizer can then be written as
\begin{equation}
\label{eq:update_step_standard}
\gaus_{t+1} = \gaus_{t} - \eta_t f\left(\nabla_{\gaus_{t}}\right),
\end{equation}
where $f(\cdot)$ is a specific optimizer (e.g., SGD or Adam) and $\eta_t$ is a learning rate (LR), possibly varying over iterations.
The Adam optimizer, for example, adaptively rescales each parameter’s gradient using first- and second-moment estimates (see \supmat).
Given per-parameter-group LRs tuned to the parameters' varying scales and contributions, first-order updates yield a conservative yet reliable trajectory for each Gaussian parameter.

\subsection{Learned optimizers}
\label{subsec:learned-optimizer}
A standard optimizer $f$ can be replaced by a learned optimizer $f_{\btheta}$, a neural network parameterized by $\btheta$ and trained via meta-learning~(\cref{fig:learned-opt}).
Meta-learning consists of two nested updates.
The \emph{inner loop} updates the scene-specific Gaussian parameters at inner step $\tinner$ using the weights of the learned optimizer $\btheta_{\tmeta}$ at meta-iteration $\tmeta$,
\begin{equation}
    \label{eq:update_step}
    \gaus_{{\tinner}+1} = \gaus_{\tinner} + f_{\btheta_{\tmeta}} \left( \ggaus, \gaus_{\tinner}\right).
\end{equation}
Unlike \cref{eq:update_step_standard}, $f_{\btheta}$ directly predicts the signed parameter update, absorbing the step size $\eta_t$ and its sign.
It can also exploit inputs beyond the gradient, such as the current Gaussian state $\gaus_{\tinner}$, rendering-error measures, or higher-level context.
The \emph{meta loop} updates the weights of the optimizer based on the performance of the inner updates, measured by $\cL_{\meta}$, across scenes and reconstruction states,
\begin{equation}
    \btheta_{\tmeta + 1} = \btheta_{\tmeta} - \eta_{\tmeta} \gtheta \cL_{\meta}.
\end{equation}

    \section{Method}
\label{sec:method}
Long-horizon stability arises from three complementary strategies: we identify and address instability via dedicated losses (\cref{subsec:meta-learning}), design an architecture whose updates diminish as the loss gradients diminish (\cref{subsec:model_overview}), and employ a data-centric scheme that exposes the optimizer to intermediate states along real optimization trajectories (\cref{subsec:long-horizon-training}). 
Additional architecture and hyperparameter details are in the \supmat

\subsection{Meta-training}
\label{subsec:meta-learning}
\begin{figure*}[t]
    \centering
    \includegraphics[
        width=\linewidth,
        trim=7mm 15mm 5mm 11mm,
        clip
    ]{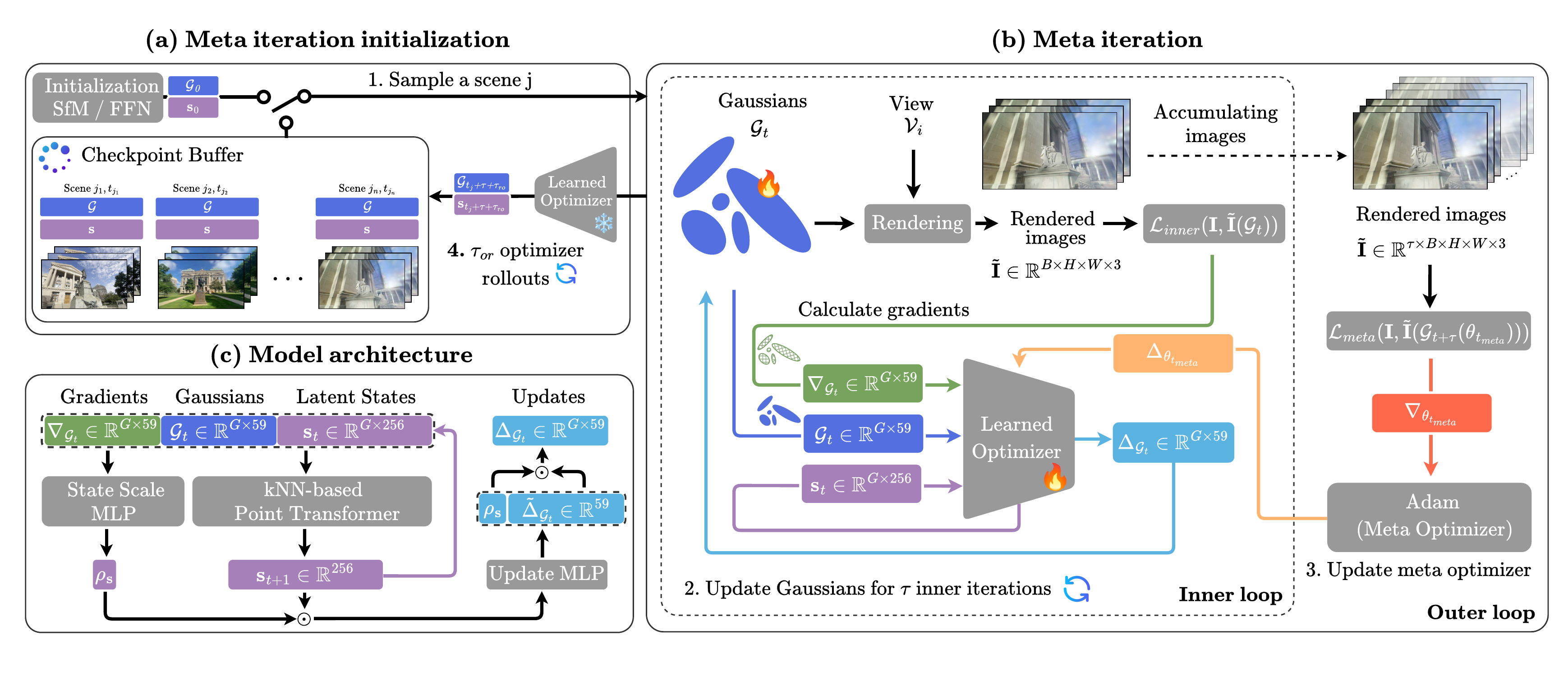}
    \caption{\textbf{\ours{} meta-training and architecture.}
    \textbf{(a)} \textbf{Meta-iteration initialization~(\cref{subsec:long-horizon-training}).} 
    During meta-training, we iterate over different 3D scenes. 
    At each meta-iteration, a scene is sampled and its Gaussians are initialized using either (1) SfM or FFN points at $t=0$, or (2) an intermediate state drawn from the \emph{Checkpoint Buffer}.
    At the end of each meta-iteration, the updated scene is randomly pushed back into the buffer, after applying additional rollout steps using a frozen version of the learned optimizer.
    \textbf{(b)} \textbf{Meta-iteration~(\cref{subsec:meta-learning}).} Starting from the scene sampled in (a), the \emph{inner loop} (dashed block) rolls out the learned optimizer for $\tau$ iterations, predicting Gaussian parameter updates that iteratively refine the scene representation.
    Once the inner loop completes, the meta-iteration (outer block) evaluates the reconstruction performance and backpropagates the resulting meta-gradients to update the optimizer's parameters.
    By observing a different scene at each meta-iteration, the optimizer learns update rules that generalize across scenes.
    \textbf{(c)} \textbf{Model architecture~(\cref{subsec:model_overview}).} 
    Our model comprises two parallel branches: a \emph{State Scale MLP}, which predicts state-scaling coefficients from the concatenated input (Adam-normalized gradients, Gaussian parameters, and latent state), and a \emph{kNN-based Point Transformer}, which predicts updated per-primitive latent states.
    The scaled latent states are fed into the \emph{Update MLP}, which predicts the final Gaussian parameter updates.
    $\odot$ denotes element-wise multiplication; dashed lines indicate concatenation.
}
    \label{fig:overview}

\end{figure*}

A learned optimizer is meta-trained across a collection of 3D scenes $\{\cV^j\}_{j=1}^V$, where each scene $\cV^j$ consists of a set of context 
views used during optimization and a fixed set of target views used for evaluation.
For simplicity, we omit the index $j$ as all quantities are per scene.
\cref{fig:overview} illustrates the training procedure. 

\boldparagraph{Inner loop.} 
At each inner step $\tinner$, the current Gaussians are rendered for a batch of context views, and the inner loss $\cL_{\inner}(\bI, \tilde{\bI}(\gaus_{\tinner}))$ is computed.
The resulting gradients \wrt{} the Gaussian parameters $\ggaus$ are passed as input to the optimizer, which predicts parameter updates $\Delta_{\gaus_{\tinner}}$.
These updates are applied to obtain $\gaus_{\tinner+1}$ (\cref{eq:update_step}), the next set of Gaussians for the following inner iteration.
After $\tau$ inner iterations, where $\tau$ is sampled uniformly from $[1,6]$, the scene reaches the updated state $\gaus_{{\tinner}+\tau}$.

\boldparagraph{Meta loop.}
At each of the $\tau$ inner iterations, both context and target views are rendered, encouraging updates that generalize to novel views rather than overfit the context views.
These renderings are used to compute the meta-loss,
\begin{equation}
    \cL_{\meta}({\bI}, \tilde{\bI}(\gaus_{{\tinner}+\tau}(\btheta_{\tmeta}))),
\end{equation}
where $\btheta_{\tmeta}$ marks the Gaussians' dependence on the learned optimizer's weights through the $\tau$ inner steps.
The meta-loss gradients \wrt{} these weights, $\gtheta = \nabla_{\btheta_{\tmeta}} \cL_{\meta}$, are used by a general-purpose optimizer (Adam~\cite{Kingma2015ICLR}) to update $\btheta_{\tmeta}$.

\boldparagraph{Losses.}
The inner loss follows the 3DGS formulation,
\begin{equation}
\cL_{\inner} = 0.8 \, \ell_1(\bI, \tilde{\bI}) + 0.2 \, \text{D-SSIM}(\bI, \tilde{\bI}).
\end{equation}
For the meta-loss, we combine three components: (1) a rendering loss ($\cL_{\text{render}}$) that supervises reconstruction quality, (2) a low-visibility supervision term ($\cL_{\text{lvs}}$) that regularizes weakly supervised or unsupervised Gaussians, and (3) a stability term ($\cL_{\text{stab}}$) that encourages monotonic improvement,
\begin{equation}
    \cL_{\meta} = \lambda_{\text{render}} \, \cL_{\text{render}} + \lambda_{\text{lvs}} \, \cL_{\text{lvs}} + \lambda_{\text{stab}} \, \cL_{\text{stab}},
    \label{eq:meta-loss}
\end{equation}
where $\lambda_{\text{render}}$, $\lambda_{\text{lvs}}$ and $\lambda_{\text{stab}}$ are the corresponding weights.
The rendering loss is defined following ReSplat~\cite{xu2025resplat} as an exponentially weighted sum over $\tau$ inner steps,
\begin{equation}
    \cL_{\text{render}} =
    \sum_{\tinner=0}^{\tau-1} \gamma^{\tau-1-\tinner}
    \Big[ \ell_1(\bI_{\tinner}, \tilde{\bI}_{\tinner})
    + 0.5 \, \text{LPIPS}(\bI_{\tinner}, \tilde{\bI}_{\tinner}) \Big],
    \label{eq:meta-render-loss}
\end{equation}
where $\gamma = 0.9$.
Although effective, this loss provides limited supervision for Gaussians with negligible contributions to the rendered images, allowing the optimizer to produce unconstrained updates for them.
Such updates may be harmless in the short term, but they accumulate over longer optimization horizons, ultimately degrading rendering quality.
To address this, we introduce a low-visibility supervision loss ($\cL_{\text{lvs}}$) that penalizes updates for weakly supervised Gaussians.
Specifically, we add an $\ell_1$ loss on the predicted updates whenever the gradient magnitude falls below a small threshold $(\varepsilon=10^{-8})$, or when the update's sign differs from that of the Adam update (more details in \supmat).
Finally, we introduce a \emph{stability} loss that encourages the optimizer to produce monotonically improving reconstructions.
This loss is only computed on target views, as they remain constant throughout a meta-iteration. 
Specifically, we penalize inner iterations where the $\ell_1$ error increases relative to the previous step ($\text{sg}[\cdot]$ denotes the stop-gradient operation),
\begin{equation}
    \mathcal{L}_{\text{stab}} = \sum_{\tinner=1}^{\tau - 1} \max\left(0, \ell_1(\bI_{\tinner}, \tilde{\bI}_{\tinner}) - \text{sg}[\ell_1(\bI_{{\tinner}-1}, \tilde{\bI}_{{\tinner}-1})]\right).
    \label{eq:stability-loss}
\end{equation} 

\subsection{Model overview}
\label{subsec:model_overview}
Our learned optimizer maintains an internal per-primitive state and updates the Gaussian parameters at each iteration (\cref{fig:overview}); further details in the \supmat

\boldparagraph{Model input.}
We build upon ReSplat~\cite{xu2025resplat}, which refines pixel-aligned Gaussian parameters through iterative updates driven by per-pixel feature-space errors.
While effective, this limits flexibility across different initializations and ignores inter-pixel and multi-view dependencies. 
To address these issues, we replace the error signal with image-loss gradients computed with respect to the Gaussian parameters.
Feeding such gradients to a network, however, is not effective. They span a huge dynamic range (down to $10^{-8}$),
due to parameters' inherently different scales and their varying magnitudes across scenes and optimization phases.
This drives the activations into saturated or vanishing regimes, motivating a dedicated gradient encoding.

Prior methods~\cite{Chen2024g3r,Liu2025quicksplat} normalize each gradient entry by the maximum absolute value of that parameter within the current scene.
Although this stabilizes training, it induces a counterintuitive effect compared to standard optimization: as the maximum values decrease over training, the effective gradient magnitudes grow.
Inspired by learned optimizers~\cite{Andrychowicz2016NEURIPS, Wichrowska2017ICML}, our network applies an Adam-style transformation, whose running moments encode optimization history and normalize each parameter adaptively~\cite{bello2017neural}.

At iteration $t$, the model operates on all $G$ Gaussians jointly, taking as input their Adam-style gradients $\aggaust{t} \in~\nR^{G \times p}$, their parameters $\gaust \in \nR^{G \times p}$, and latent states $\bs_t \in \nR^{G \times 256}$.
These are concatenated along the feature dimension into a per-Gaussian input vector $\bx = [\aggaust{t}, \gaust, \bs_t] \in \nR^{G \times (2p+256)}$.

\boldparagraph{Latent state predictions.}
The optimizer consists of two parallel branches operating on the per-Gaussian input vector $\bx$.
The first branch is a kNN-based \emph{Point Transformer}~\cite{zhao2021point, wu2022point, wu2024point}, which applies self-attention to each Gaussian by attending to its $k$ nearest neighbors (kNN) in 3D space.
The output is then the updated latent states $\bs_{t+1} \in \nR^{G \times 256}$, encoding per-primitive contextual and temporal information. 
However, using the standard normalization layers within the transformer suppresses input gradient scale information, crucial for ensuring that predicted updates diminish as the loss decreases.
To restore this information, the \emph{State Scale MLP} branch predicts per-Gaussian scaling coefficients $\brho_\bs \in \nR^{G}$ (non-negative), which modulate the magnitude of the state updates.
The resulting scaled states are computed as $\tilde{\bs}_{t+1} = \brho_\bs \odot \bs_{t+1}$, where $\odot$ denotes element-wise multiplication, scaling each Gaussian's state vector by its corresponding scalar in $\brho_\bs$.
The unscaled states $\bs_{t+1}$ are preserved for the next optimization step, while the scaled states $\tilde{\bs}_{t+1}$ are forwarded to the parameter update module.
The latent states $\bs_0$ can be initialized by an FFN or sampled from a standard normal distribution (see \supmat).

\boldparagraph{Update predictions.}
To obtain the actual parameter updates, we employ an \emph{Update MLP} that maps the scaled states $\tilde{\bs}_{t+1}$ to a set of per-Gaussian output vectors $\bO_{\gaust} \in \nR^{G \times (p+1)}$, split into two parts per Gaussian: a $p$-dimensional unit-length vector $\tilde{\Delta}_{\gaust}$, normalized to represent the update direction, and a single non-negative scalar $\brho_{\Delta_t}$, controlling the update magnitude.
The final parameter updates are computed as $\Delta_{\gaust} = \brho_{\Delta_t} \odot \tilde{\Delta}_{\gaust}$, with independent control over the direction and magnitude of each update.
Finally, the Gaussians of the next iteration $t+1$ are computed based on \cref{eq:update_step}, as $\gaustt{t+1} = \gaust + \Delta_{\gaust}$.

\subsection{Long-horizon training}
\label{subsec:long-horizon-training}
A key design principle of our learned optimizer is \emph{long-horizon stability}, requiring that predicted parameter updates diminish as the loss gradients diminish.
This prevents quality degradation over extended optimization horizons and allows the optimizer to be applied beyond its training distribution.
We aim for this behavior to emerge as an intrinsic property rather than enforced by external mechanisms, such as time encoding or LR schedules tied to a predefined number of iterations.

\boldparagraph{Checkpoint buffer.}
Learning long-horizon 3DGS reconstruction requires exposing the optimizer to diverse states, from large early gradients to fine late-stage refinements.  
However, naively extending the inner loop is computationally prohibitive and violates the i.i.d.\ assumption, as the optimizer would repeatedly encounter large gradients early and small ones later.
To address this, we introduce a \emph{checkpoint buffer} that stores intermediate scene states from previous meta-iterations.
At each meta-iteration (\cref{fig:overview}(a)), we either sample a new scene (probability $1 - p_{\text{buffer}}$) or resample a previously optimized scene (checkpoint).
Each checkpoint stores the Gaussian parameters and optimizer states, including Adam moments and per-primitive latent vectors, so optimization resumes seamlessly across meta-iterations.
We then unroll the optimizer for a few inner steps; the updated state is pushed back with probability $p_{\text{push}}$ for newly initialized scenes and $p_{\text{push-back}}$ for resampled ones.
This exposes the optimizer to a balanced distribution of optimization states without extending the inner loop.

\boldparagraph{Optimizer rollout.}
Inspired by policy rollout in reinforcement learning~\cite{bertsekas2021multiagent}, we introduce an \emph{optimizer rollout} strategy to further extend observed optimization horizons.
Before storing a scene in the buffer, we further optimize its parameters using the currently learned optimizer, while keeping the model frozen.
This drives Gaussians along trajectories resembling those encountered during inference, helping the learned optimizer recover from its own mistakes.
The number of rollout iterations is sampled at random in $[1, \tau_a]$, where $\tau_a$ is linearly increased from $1$ to $50$ during the first $10{,}000$ meta-training iterations, effectively implementing easy-to-hard curriculum learning.

    \section{Experiments}
\label{sec:exps}
\ourshort{} can be applied across different initialization and view-sampling settings.
We demonstrate this by jointly training a single model over two configurations of the DL3DV dataset~\cite{ling2024dl3dv} (\cref{subsec:training-details}) and evaluating it on both (\cref{subsec:results}).

\subsection{Training details}
\label{subsec:training-details}
The two configurations we consider are
(i) a \textbf{\emph{sparse-view}}, \emph{forward-facing} setup initialized from ReSplat~\cite{xu2025resplat} and using a fixed set of context views at each inner iteration, and
(ii) a \textbf{\emph{dense-view}}, \emph{large-baseline} setup initialized from SfM point clouds and using sampled context views via farthest point sampling.
During training, both settings use low-resolution images ($256 \times 448$), 8 context views, and 6 target views per optimization trajectory.
For the dense setting, context views are sampled from a per-scene pool of 64 views, which is resampled at the start of each optimization trajectory.
We compare against dedicated per-setting models in the \supmat

At each inner iteration, Gaussian parameters and gradients are detached before being fed to the network, while the latent state remains differentiable, allowing gradients to flow between iterations only through the states.
Our framework is implemented on \texttt{gsplat}~\cite{ye2025gsplat}.
The learned optimizer is trained end-to-end in PyTorch with mixed precision for 150k meta-iterations on 4~NVIDIA~A100 GPUs using Adam as the meta-optimizer (LR $10^{-4}$).
During training, the kNN is recomputed at every iteration; at test time, every 100 iterations for efficiency.

\subsection{Results}
\label{subsec:results}
\begin{table}[t]
\centering
\footnotesize
\setlength{\tabcolsep}{1.2pt}
\renewcommand{\arraystretch}{1.05}
\setlength{\aboverulesep}{0.2ex}\setlength{\belowrulesep}{0.3ex}
\resizebox{\linewidth}{!}{%
\begin{tabular}{@{}l @{\hspace{7pt}} l @{\hspace{3pt}}|@{\hspace{3pt}} r r r @{\hspace{6pt}}|@{\hspace{6pt}} r r @{\hspace{3pt}}|@{\hspace{3pt}} r@{}}
\toprule
& {\scriptsize\emph{Optimizer}} & {\scriptsize LPIPS\hspace{0.6pt}$\downarrow$} & {\scriptsize SSIM\hspace{0.6pt}$\uparrow$} & {\scriptsize PSNR\hspace{0.6pt}$\uparrow$} & {\scriptsize PSNR$_{\tbud}$\hspace{0.6pt}$\uparrow$} & {\scriptsize Iters} & {\scriptsize Mem\,[GB]\hspace{0.6pt}$\downarrow$} \\
\midrule
\multicolumn{8}{@{}l}{\textbf{Dense-view, SfM init}} \\
\midrule
\multirow{3}{*}{\rotatebox[origin=c]{90}{\makebox[1.05cm][c]{\shortstack{DL3DV\\{\scriptsize $\sim$300}}}}} & \lob{} & 0.218 & 0.838 & 25.81 & 25.12 & 1608 & 1.6 \\
 & \adam{} & \cellcolor{tabsecond}0.185 & \cellcolor{tabsecond}0.879 & \cellcolor{tabsecond}27.78 & \cellcolor{tabsecond}27.78 & 4000 & \cellcolor{tabfirst}\textbf{1.5} \\
 & \ourshort{} & \cellcolor{tabfirst}\textbf{0.157} & \cellcolor{tabfirst}\textbf{0.896} & \cellcolor{tabfirst}\textbf{29.04} & \cellcolor{tabfirst}\textbf{29.04} & 1713 & \cellcolor{tabfirst}\textbf{1.5} \\
\arrayrulecolor{gray!40}\midrule\arrayrulecolor{black}
\multirow{3}{*}{\rotatebox[origin=c]{90}{\makebox[1.05cm][c]{\shortstack{DTU\\{\scriptsize $\sim$50}}}}} & \lob{} & 0.350 & 0.869 & 27.19 & \multicolumn{1}{c}{DIV} & \multicolumn{1}{c}{--} & 1.0 \\
 & \adam{} & \cellcolor{tabsecond}0.309 & \cellcolor{tabsecond}0.899 & \cellcolor{tabsecond}28.96 & \cellcolor{tabsecond}28.96 & 4000 & 1.0 \\
 & \ourshort{} & \cellcolor{tabfirst}\textbf{0.308} & \cellcolor{tabfirst}\textbf{0.900} & \cellcolor{tabfirst}\textbf{29.25} & \cellcolor{tabfirst}\textbf{29.25} & 1919 & 1.0 \\
\arrayrulecolor{gray!40}\midrule\arrayrulecolor{black}
\multirow{3}{*}{\rotatebox[origin=c]{90}{\makebox[1.05cm][c]{\shortstack{LLFF\\{\scriptsize 20--60}}}}} & \lob{} & 0.355 & 0.723 & 22.77 & 22.68 & 1975 & 1.4 \\
 & \adam{} & \cellcolor{tabsecond}0.354 & \cellcolor{tabsecond}0.731 & \cellcolor{tabsecond}22.82 & \cellcolor{tabsecond}22.82 & 4000 & \cellcolor{tabfirst}\textbf{1.3} \\
 & \ourshort{} & \cellcolor{tabfirst}\textbf{0.303} & \cellcolor{tabfirst}\textbf{0.767} & \cellcolor{tabfirst}\textbf{24.56} & \cellcolor{tabfirst}\textbf{24.56} & 2855 & \cellcolor{tabfirst}\textbf{1.3} \\
\arrayrulecolor{gray!40}\midrule\arrayrulecolor{black}
\multirow{3}{*}{\rotatebox[origin=c]{90}{\makebox[1.05cm][c]{\shortstack{Mip360\\{\scriptsize 100+}}}}} & \lob{} & 0.423 & 0.658 & 23.62 & 22.68 & 1441 & 4.4 \\
 & \adam{} & \cellcolor{tabsecond}0.372 & \cellcolor{tabfirst}\textbf{0.715} & \cellcolor{tabsecond}25.62 & \cellcolor{tabsecond}25.62 & 4000 & \cellcolor{tabfirst}\textbf{3.4} \\
 & \ourshort{} & \cellcolor{tabfirst}\textbf{0.367} & \cellcolor{tabfirst}\textbf{0.715} & \cellcolor{tabfirst}\textbf{25.75} & \cellcolor{tabfirst}\textbf{25.75} & 2013 & \cellcolor{tabsecond}3.5 \\
\midrule
\multicolumn{8}{@{}l}{\textbf{Sparse-view, ReSplat init}} \\
\midrule
\multirow{4}{*}{\rotatebox[origin=c]{90}{\makebox[1.05cm][c]{\shortstack{DL3DV\\{\scriptsize 8, low}}}}} & ReSplat & 0.144 & 0.854 & 26.59 & 5.99 & 1252 & 1.5 \\
 & \lob{} & \cellcolor{tabsecond}0.093 & \cellcolor{tabsecond}0.916 & \cellcolor{tabsecond}30.34 & 27.93 & 1688 & \cellcolor{tabsecond}0.9 \\
 & \adam{} & 0.097 & 0.909 & 30.02 & \cellcolor{tabsecond}29.91 & 4000 & \cellcolor{tabfirst}\textbf{0.6} \\
 & \ourshort{} & \cellcolor{tabfirst}\textbf{0.090} & \cellcolor{tabfirst}\textbf{0.917} & \cellcolor{tabfirst}\textbf{30.53} & \cellcolor{tabfirst}\textbf{30.53} & 1924 & \cellcolor{tabsecond}0.9 \\
\arrayrulecolor{gray!40}\midrule\arrayrulecolor{black}
\multirow{4}{*}{\rotatebox[origin=c]{90}{\makebox[1.05cm][c]{\shortstack{DL3DV\\{\scriptsize 8, high}}}}} & ReSplat & 0.325 & 0.722 & 22.08 & \multicolumn{1}{c}{DIV} & \multicolumn{1}{c}{--} & 20.2 \\
 & \lob{} & \cellcolor{tabsecond}0.192 & \cellcolor{tabfirst}\textbf{0.846} & \cellcolor{tabsecond}26.99 & 23.83 & 1697 & \cellcolor{tabsecond}3.0 \\
 & \adam{} & 0.197 & 0.837 & 26.60 & \cellcolor{tabsecond}26.32 & 4000 & \cellcolor{tabfirst}\textbf{1.8} \\
 & \ourshort{} & \cellcolor{tabfirst}\textbf{0.185} & \cellcolor{tabfirst}\textbf{0.846} & \cellcolor{tabfirst}\textbf{27.16} & \cellcolor{tabfirst}\textbf{27.16} & 2063 & \cellcolor{tabsecond}3.0 \\
\arrayrulecolor{gray!40}\midrule\arrayrulecolor{black}
\multirow{4}{*}{\rotatebox[origin=c]{90}{\makebox[1.05cm][c]{\shortstack{DL3DV\\{\scriptsize 32, low}}}}} & ReSplat & 0.258 & 0.743 & 21.33 & 5.51 & 1797 & 12.0 \\
 & \lob{} & \cellcolor{tabsecond}0.100 & \cellcolor{tabsecond}0.911 & \cellcolor{tabsecond}29.86 & 26.75 & 2279 & \cellcolor{tabsecond}4.2 \\
 & \adam{} & 0.103 & 0.904 & 29.78 & \cellcolor{tabsecond}29.70 & 4000 & \cellcolor{tabfirst}\textbf{4.1} \\
 & \ourshort{} & \cellcolor{tabfirst}\textbf{0.095} & \cellcolor{tabfirst}\textbf{0.913} & \cellcolor{tabfirst}\textbf{30.26} & \cellcolor{tabfirst}\textbf{30.26} & 2334 & \cellcolor{tabsecond}4.2 \\
\arrayrulecolor{gray!40}\midrule\arrayrulecolor{black}
\multirow{4}{*}{\rotatebox[origin=c]{90}{\makebox[1.05cm][c]{\shortstack{RE10K\\{\scriptsize 8, high}}}}} & ReSplat & 0.338 & 0.745 & 19.90 & \multicolumn{1}{c}{DIV} & \multicolumn{1}{c}{--} & 20.2 \\
 & \lob{} & \cellcolor{tabfirst}\textbf{0.214} & \cellcolor{tabfirst}\textbf{0.855} & \cellcolor{tabsecond}26.01 & \multicolumn{1}{c}{DIV} & \multicolumn{1}{c}{--} & \cellcolor{tabsecond}3.0 \\
 & \adam{} & 0.223 & 0.844 & 25.57 & \cellcolor{tabsecond}25.26 & 4000 & \cellcolor{tabfirst}\textbf{1.8} \\
 & \ourshort{} & \cellcolor{tabfirst}\textbf{0.214} & \cellcolor{tabsecond}0.852 & \cellcolor{tabfirst}\textbf{26.08} & \cellcolor{tabfirst}\textbf{26.08} & 1818 & \cellcolor{tabsecond}3.0 \\
\bottomrule
\end{tabular}%
}%
\caption{%
\textbf{\ourshort{} reaches the best PSNR and keeps improving until the end of the run.} Each dataset is labelled with the number of views per scene and resolution. Metrics are reported at the best-PSNR iteration within a fixed runtime $\tbud$, which we set to \adam{}'s runtime of $4000$ iterations. PSNR$_{\tbud}$ is taken at the last iteration of each method inside this runtime (Iters). DIV marks a run whose Gaussian parameters diverged to non-finite values, with no quality to report beyond that point. Only \ourshort{} keeps improving to the end on every benchmark, while \lob{} loses up to $3.2$~dB, ReSplat over $15$~dB, and \adam{} stays within $0.4$~dB. \colorbox{tabfirst}{\textbf{Best}} and \colorbox{tabsecond}{second best} values are highlighted.%
}
\label{tab:main-results}
\end{table}

We assess in-domain and zero-shot generalization in the regimes introduced in \cref{subsec:training-details}, across diverse datasets, resolutions, and view counts.
We compare against \adam{} with standard 3DGS LRs and \lob{}, our implementation of the G3R~\cite{Chen2024g3r} meta-training scheme (no public code available).
\lob{} is built on our optimizer architecture to isolate G3R's training strategy from its original sparse 3D CNN and voxelization pipeline.
It retains G3R's short-horizon meta-training ($24$ steps) and iteration-based scheduling (time-step conditioning, cosine LR) but omits our long-horizon components (CB, OR, predicted scaling, stability losses).
We compare against ReSplat~\cite{xu2025resplat} only in the sparse setting, as it is restricted to pixel-aligned Gaussians, which are incompatible with the SfM initialization.

All optimizers run under the same controlled protocol, with spherical-harmonic degree~3 from the start.
Each optimizer is given the same runtime, which we set to \adam{}'s wall-clock of $4000$ iterations, and runs as many iterations as fit within it.
For \adam{} and \lob{},  the LR schedule is scaled to the run's own length.
We disable adaptive density control to compare optimizers under a fixed-capacity representation, with additional densification results in the \supmat{}
All metrics are read from scene-averaged curves against both iteration and wall-clock time.

\begin{figure}[!t]
    \centering
    \includegraphics[width=\linewidth]{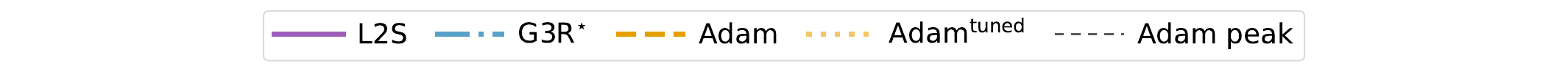}\\
    \includegraphics[width=\linewidth]{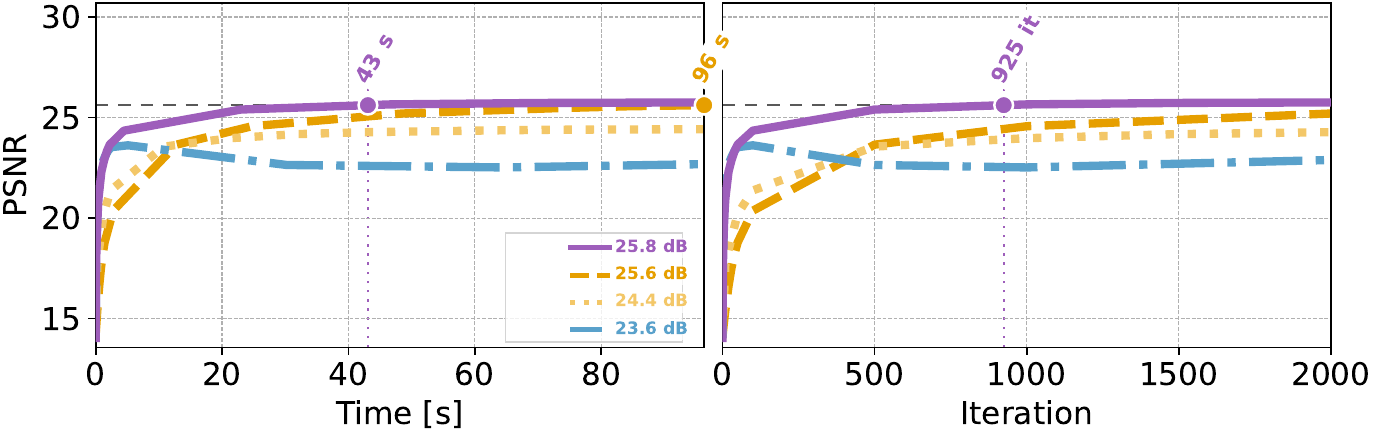}
    \caption{
    \textbf{\ourshort{} stays above \adam{} and remains stable over long horizons with higher final PSNR.}
    We show detailed results on the Mip-NeRF360 dense-view setting (\cref{subsec:results}).
    \textit{\textbf{Left:}} PSNR vs.\ wall-clock time, over the runtime of \cref{tab:main-results}.
    \textit{\textbf{Right:}} PSNR vs.\ optimization iteration.
    Dots and labels mark where each run reaches \adam{}'s peak PSNR. The left panel lists each method's best PSNR.
    \ourshort{} reaches each quality level of \adam{} faster.
    }
    \label{fig:speed}
\end{figure}

\begin{figure}[t]
    \centering
    \setlength{\tabcolsep}{0pt} %
    \renewcommand{\arraystretch}{0.3} %
    \newcommand{\imagevshift}{-3pt} %

    \begin{subfigure}[b]{\linewidth}
        \centering
        \setlength{\tabcolsep}{0.002\linewidth} %
        \renewcommand{\arraystretch}{0} %
        \begin{tabular}{@{}%
        >{\centering\arraybackslash}m{0.035\linewidth}
        @{\hspace{3pt}}
        >{\centering\arraybackslash}m{0.181\linewidth}
        >{\centering\arraybackslash}m{0.181\linewidth}
        >{\centering\arraybackslash}m{0.181\linewidth}
        >{\centering\arraybackslash}m{0.181\linewidth}
        >{\centering\arraybackslash}m{0.025\linewidth}
        >{\centering\arraybackslash}m{0.181\linewidth}
        @{}}
             & {\tiny $t=2$} & {\tiny $t=10$} & {\tiny $t=100$} & {\tiny $t=1000$} &  &  \\[1pt]

                \rotatebox{90}{\makecell{\tiny ReSplat}} &
                \raisebox{\imagevshift}{\pcell[width=\linewidth]{22.8}{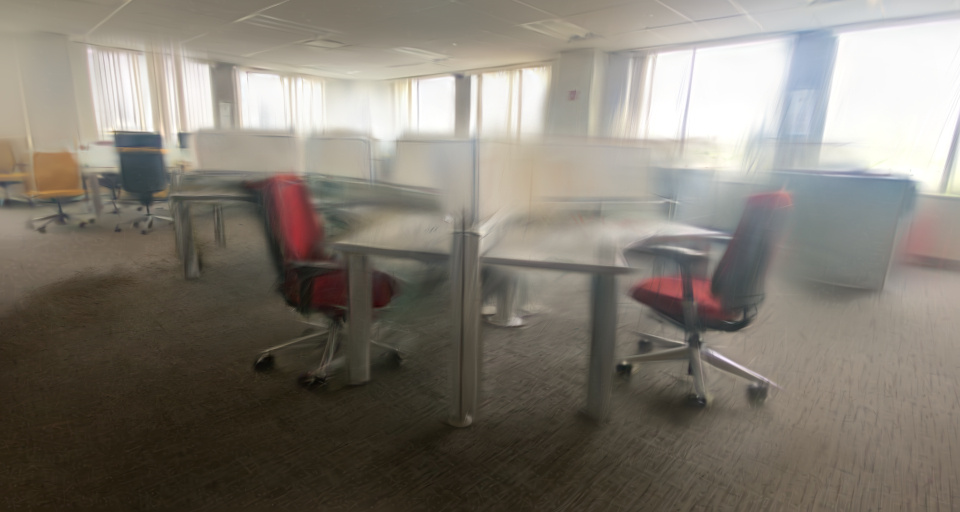}} &
                \raisebox{\imagevshift}{\pcell[width=\linewidth]{23.2}{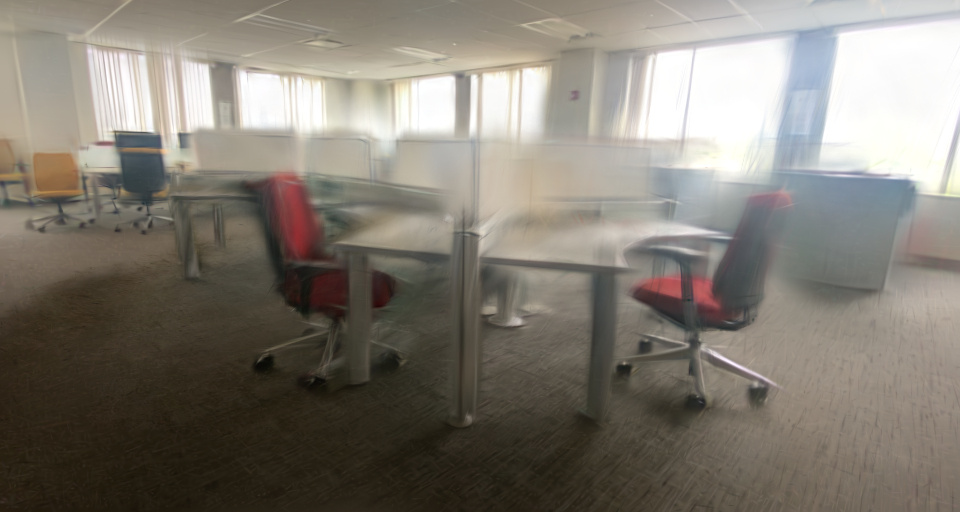}} &
                \raisebox{\imagevshift}{\pcell[width=\linewidth]{14.7}{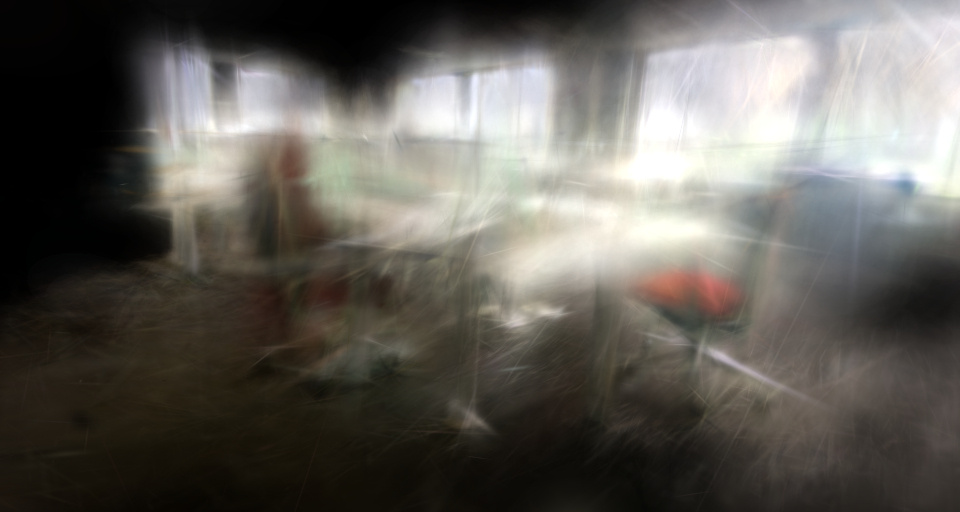}} &
                \raisebox{\imagevshift}{\pcell[width=\linewidth]{5.2}{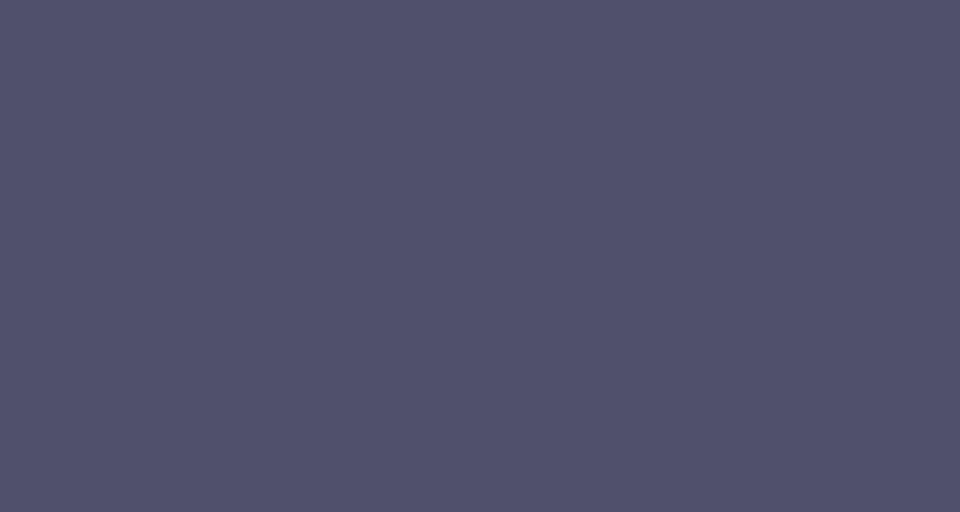}} &
                \rotatebox{90}{\tiny Init} &
                \raisebox{\imagevshift}{\pcell[width=\linewidth]{21.4}{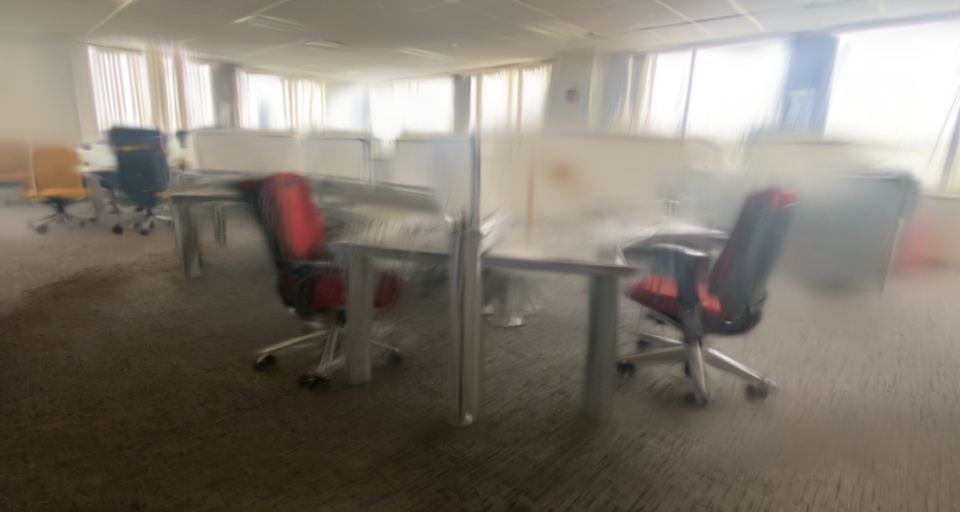}} \\
                \noalign{\vskip 2\tabcolsep}
                \rotatebox{90}{\makecell{\tiny \lob{}}} &
                \raisebox{\imagevshift}{\pcell[width=\linewidth]{24.7}{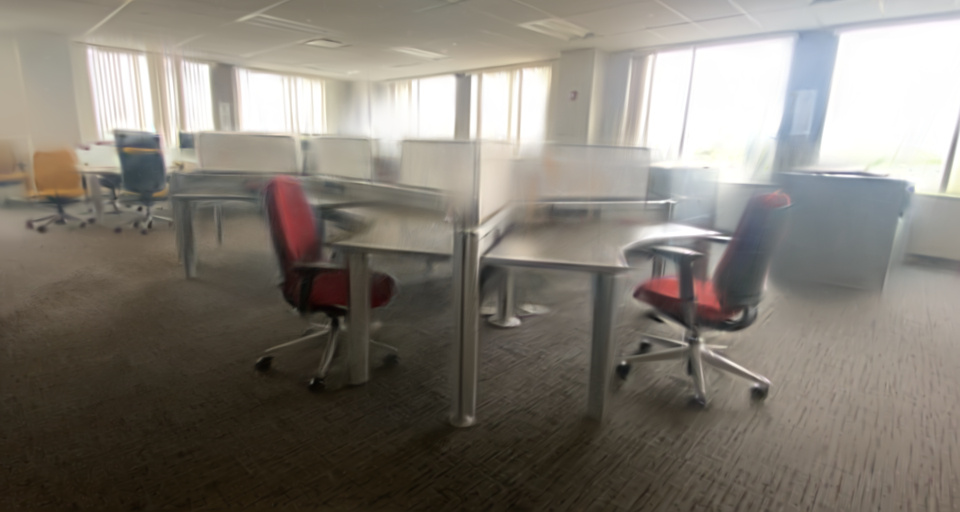}} &
                \raisebox{\imagevshift}{\pcell[width=\linewidth]{26.7}{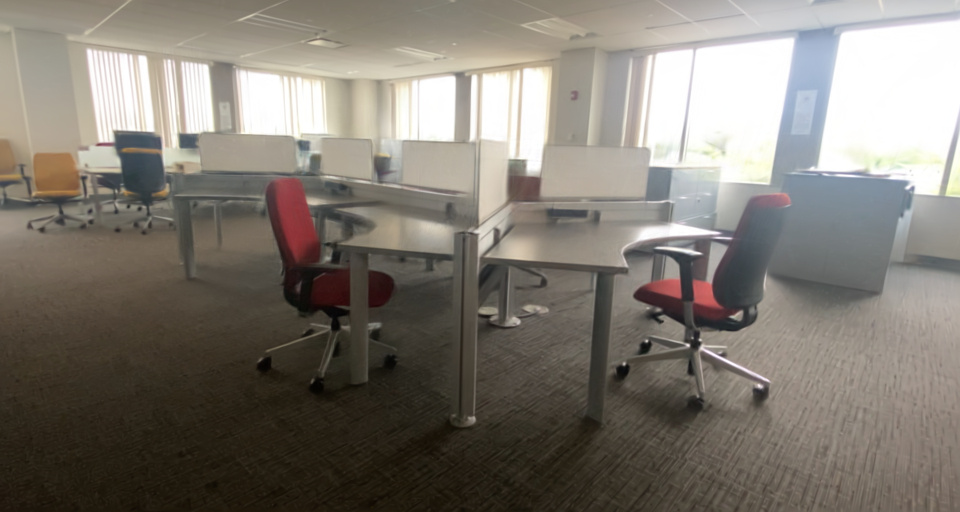}} &
                \raisebox{\imagevshift}{\pcell[width=\linewidth]{26.1}{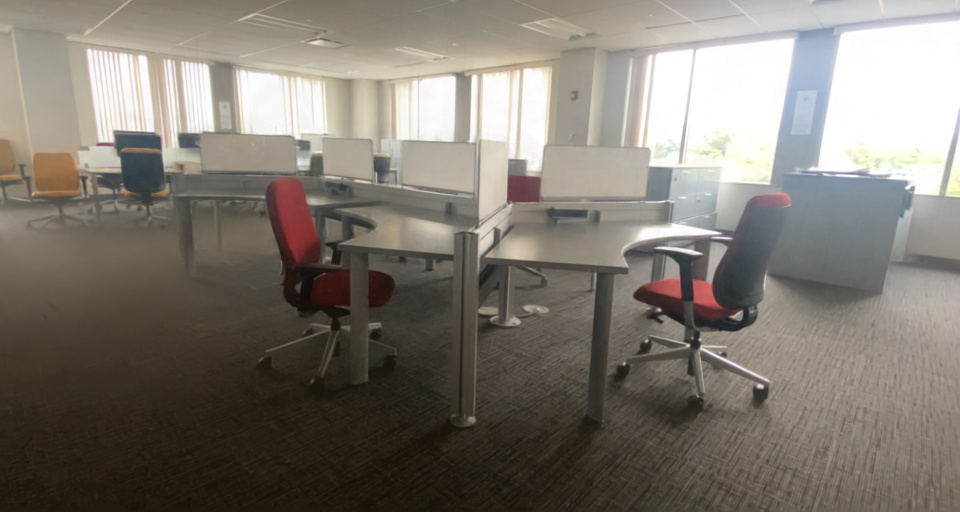}} &
                \raisebox{\imagevshift}{\pcell[width=\linewidth]{22.1}{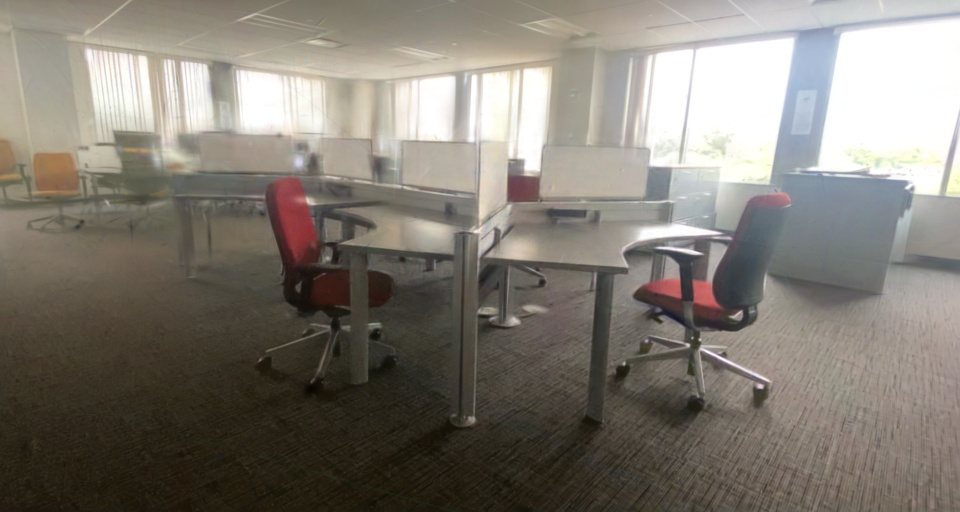}} &
                \rotatebox{90}{\tiny Ref} &
                \raisebox{\imagevshift}{\rcell[width=\linewidth]{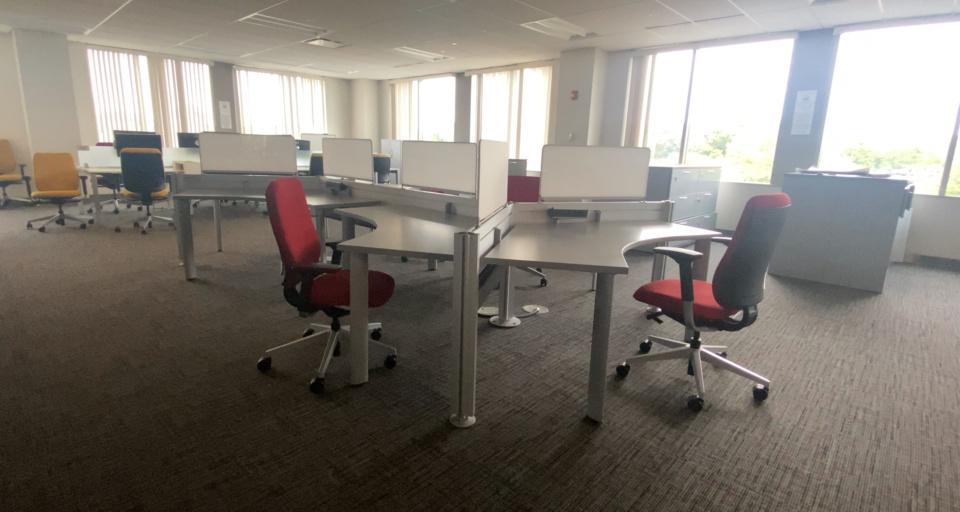}} \\
                \noalign{\vskip 2\tabcolsep}
                \rotatebox{90}{\makecell{\tiny \adam{}}} &
                \raisebox{\imagevshift}{\pcell[width=\linewidth]{21.8}{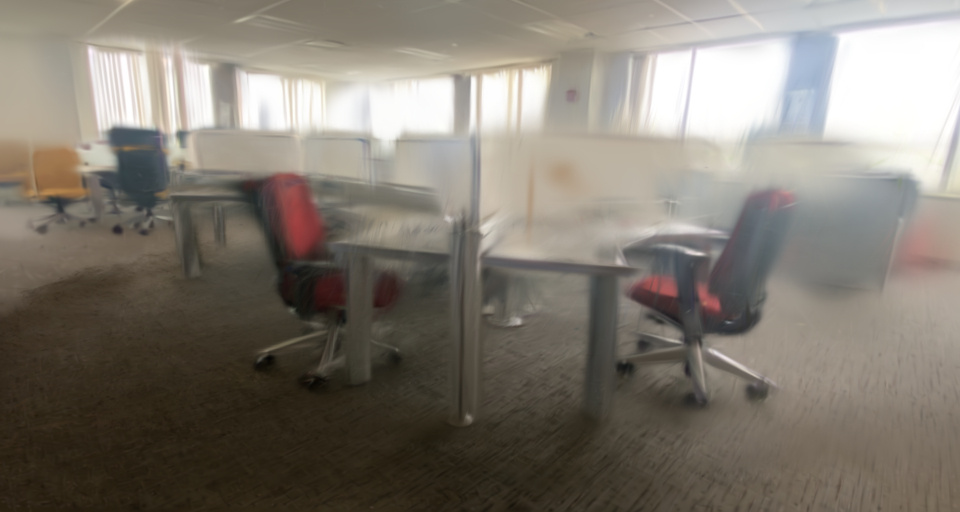}} &
                \raisebox{\imagevshift}{\pcell[width=\linewidth]{23.1}{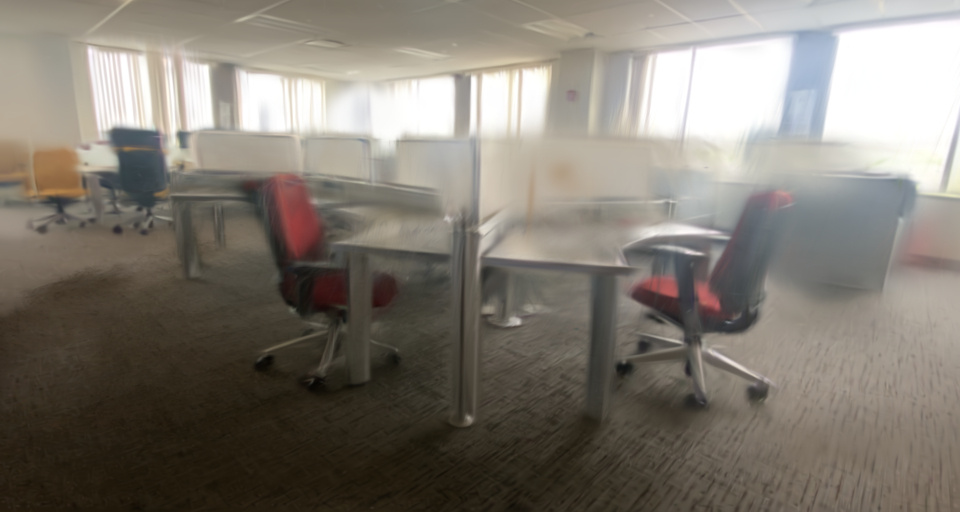}} &
                \raisebox{\imagevshift}{\pcell[width=\linewidth]{26.4}{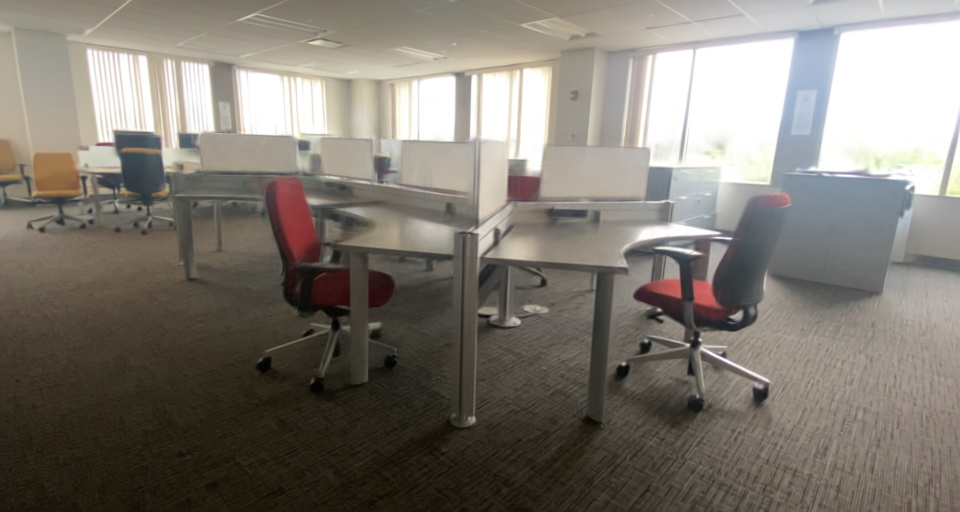}} &
                \raisebox{\imagevshift}{\pcell[width=\linewidth]{26.5}{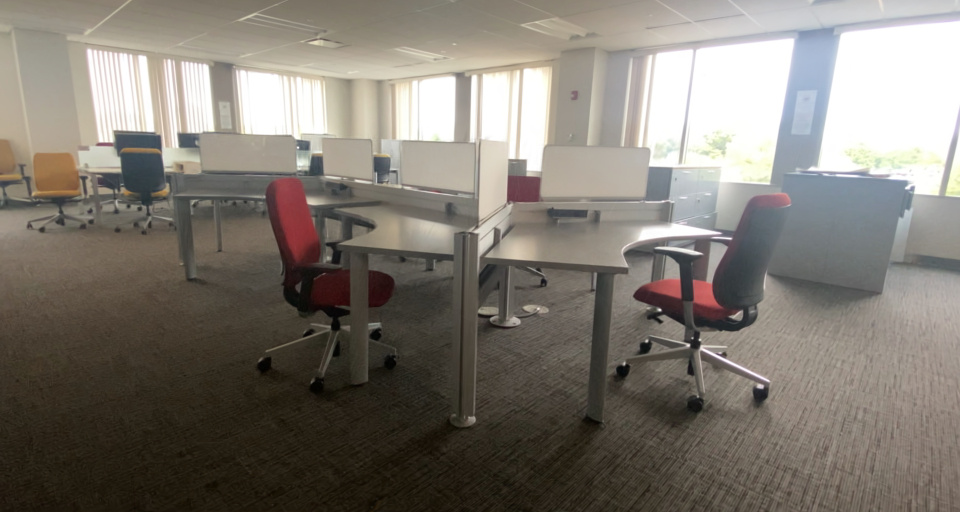}} &
                 &
                 \\
                \noalign{\vskip 2\tabcolsep}
                \rotatebox{90}{\makecell{\tiny \ourshort{}}} &
                \raisebox{\imagevshift}{\pcell[width=\linewidth]{24.8}{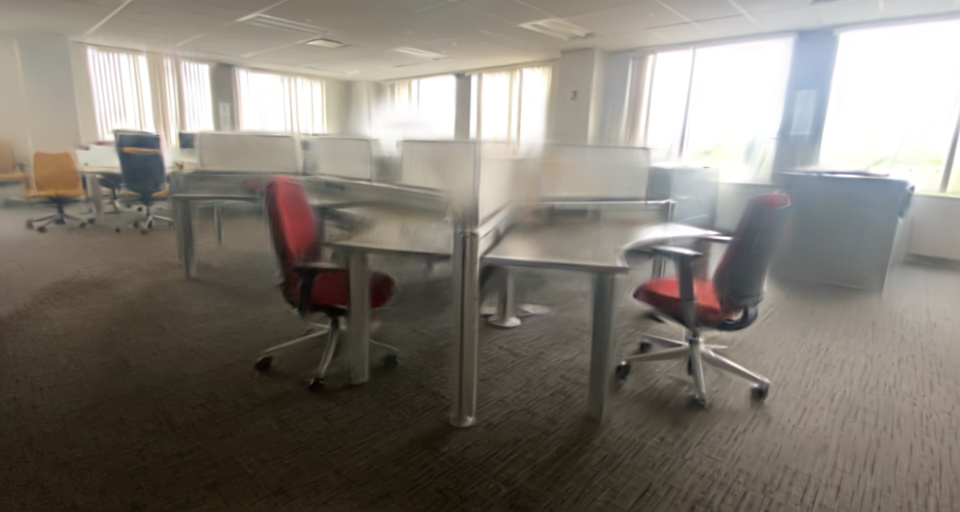}} &
                \raisebox{\imagevshift}{\pcell[width=\linewidth]{26.6}{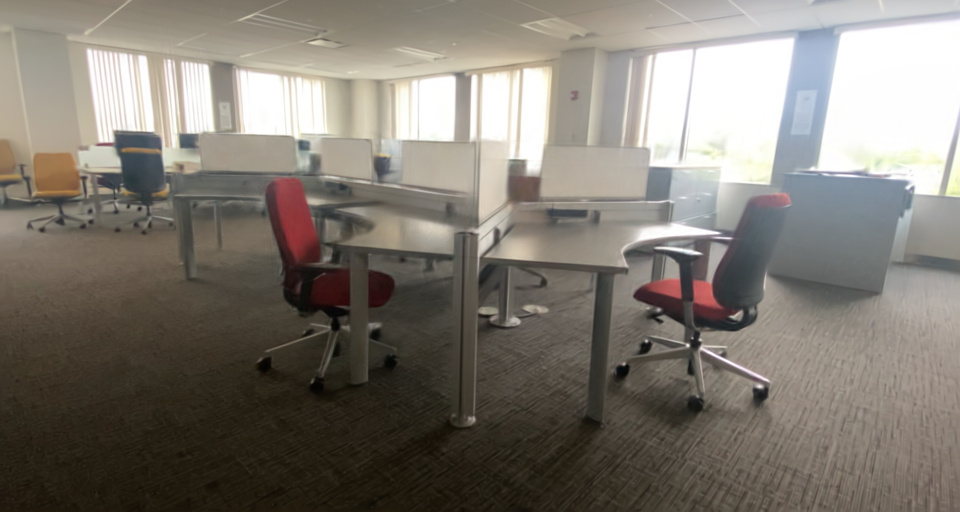}} &
                \raisebox{\imagevshift}{\pcell[width=\linewidth]{27.0}{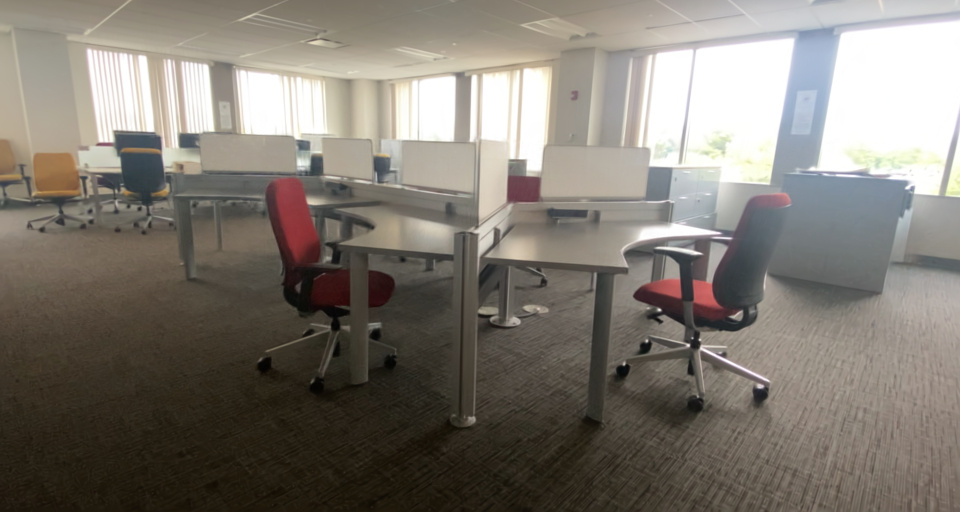}} &
                \raisebox{\imagevshift}{\pcell[width=\linewidth]{27.2}{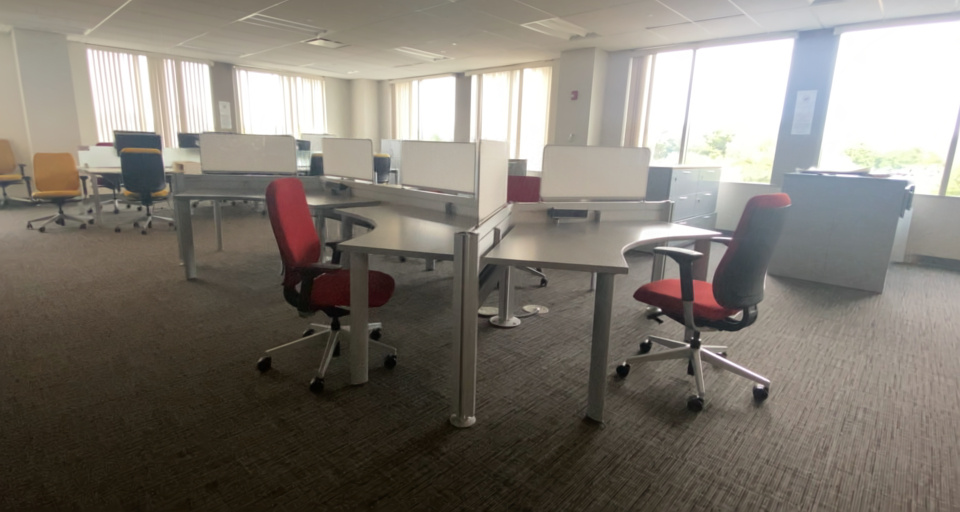}} &
                 &
                 \\
        \end{tabular}
        \caption{DL3DV (high resolution)~\cite{ling2024dl3dv} scene in a sparse-view setting: ReSplat initialization, using the same 8 views in every iteration.}
        \label{fig:comparison_top}
    \end{subfigure}

    \vspace{5pt}

    \begin{subfigure}[b]{\linewidth}
        \centering
        \setlength{\tabcolsep}{0.002\linewidth} %
        \renewcommand{\arraystretch}{0} %
        \begin{tabular}{@{}%
        >{\centering\arraybackslash}m{0.035\linewidth}
        @{\hspace{3pt}}
        >{\centering\arraybackslash}m{0.181\linewidth}
        >{\centering\arraybackslash}m{0.181\linewidth}
        >{\centering\arraybackslash}m{0.181\linewidth}
        >{\centering\arraybackslash}m{0.181\linewidth}
        >{\centering\arraybackslash}m{0.025\linewidth}
        >{\centering\arraybackslash}m{0.181\linewidth}
        @{}}
             & {\tiny $t=2$} & {\tiny $t=10$} & {\tiny $t=100$} & {\tiny $t=1000$} &  &  \\[1pt]

                \rotatebox{90}{\makecell{\tiny \lob{}}} &
                \raisebox{\imagevshift}{\pcell[width=\linewidth]{19.4}{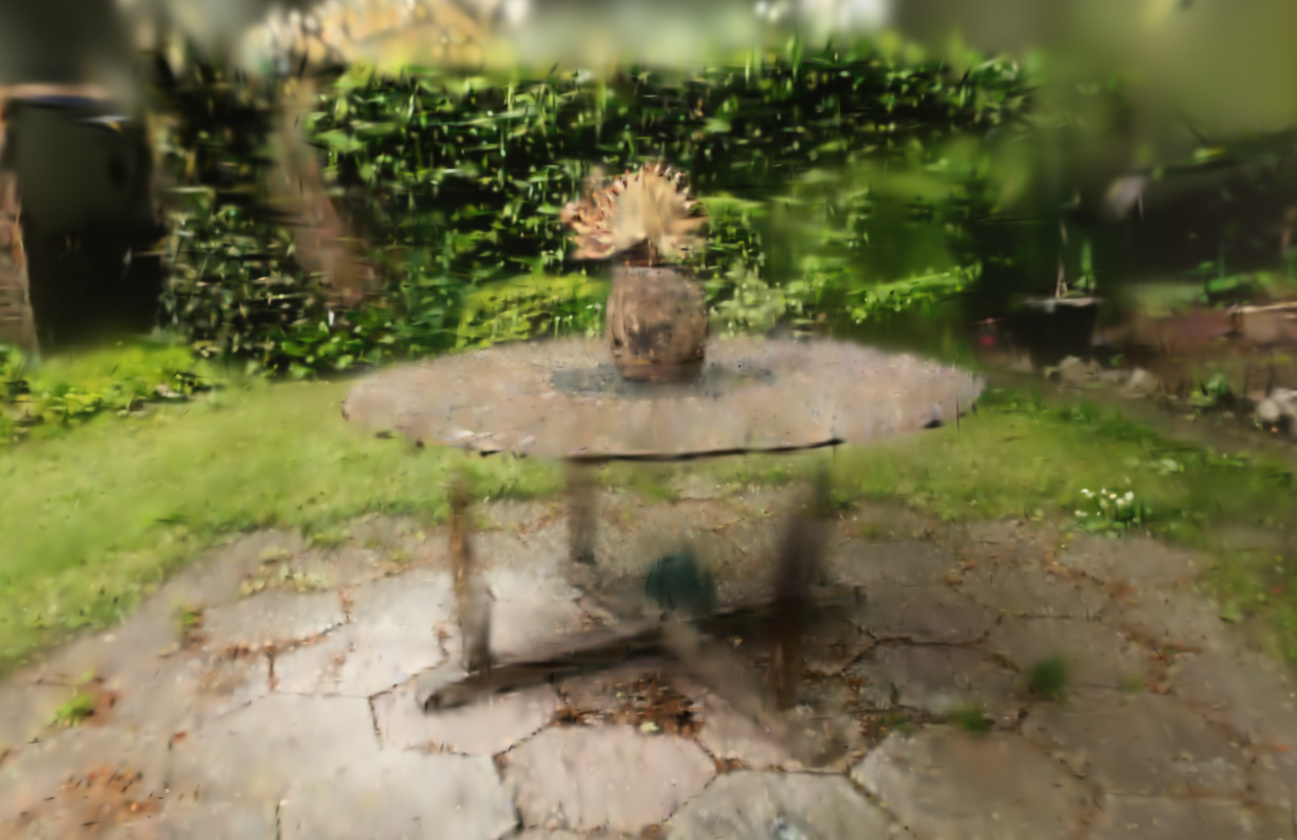}} &
                \raisebox{\imagevshift}{\pcell[width=\linewidth]{22.1}{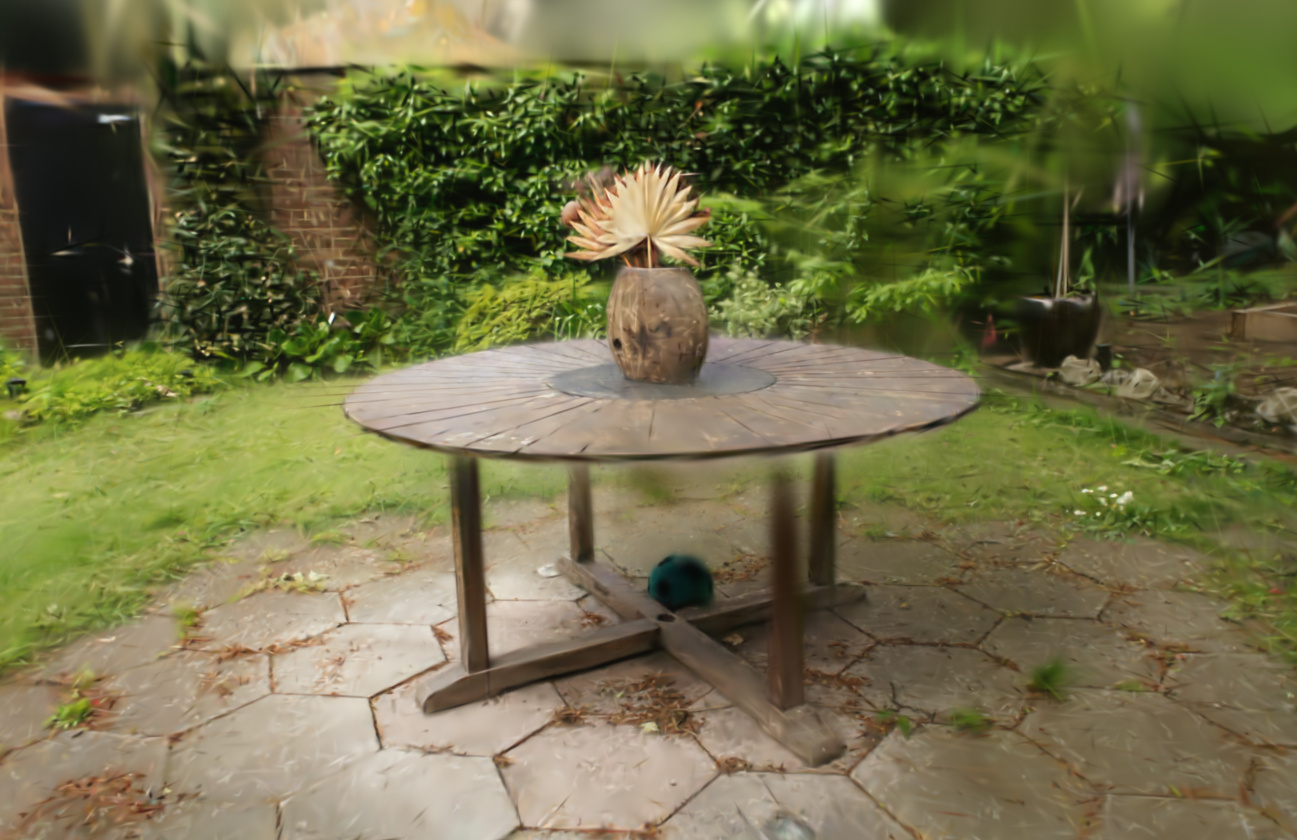}} &
                \raisebox{\imagevshift}{\pcell[width=\linewidth]{23.4}{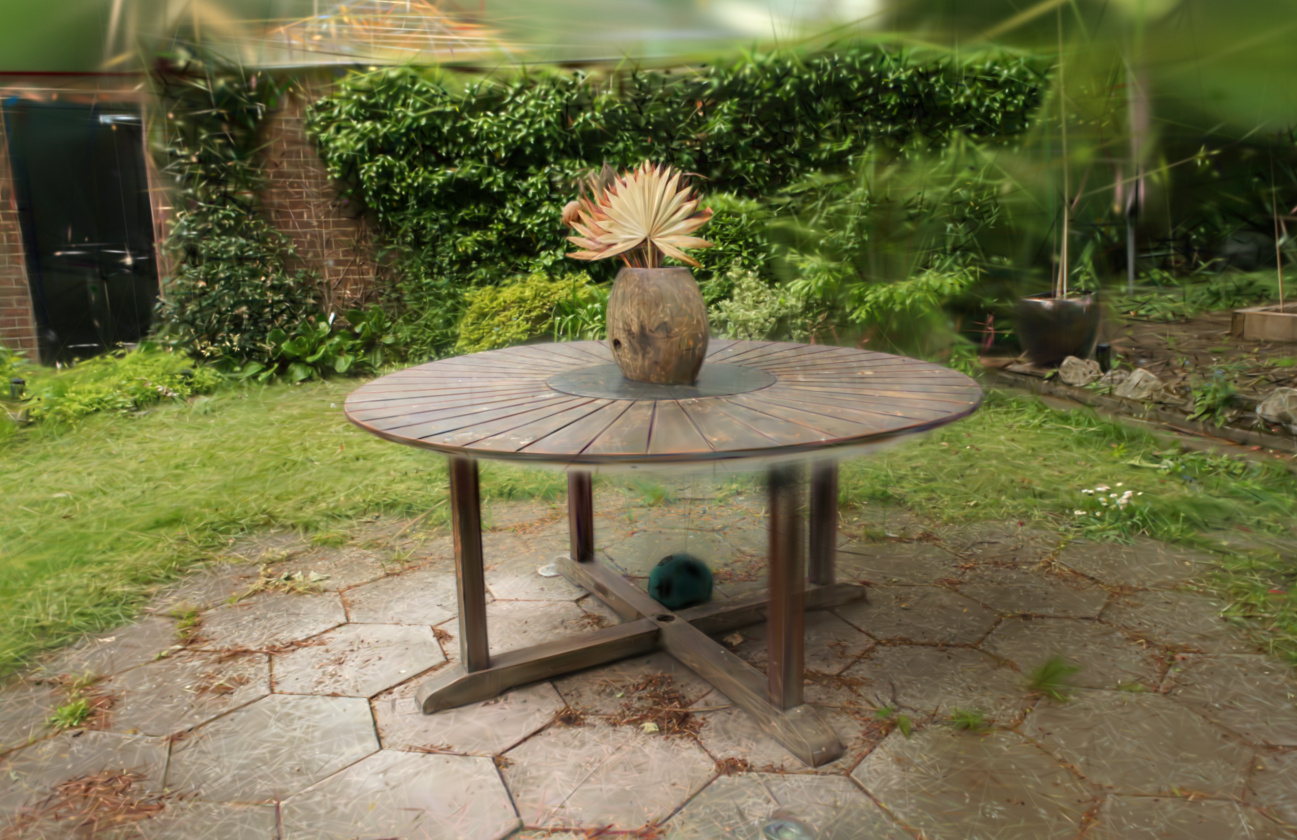}} &
                \raisebox{\imagevshift}{\pcell[width=\linewidth]{22.9}{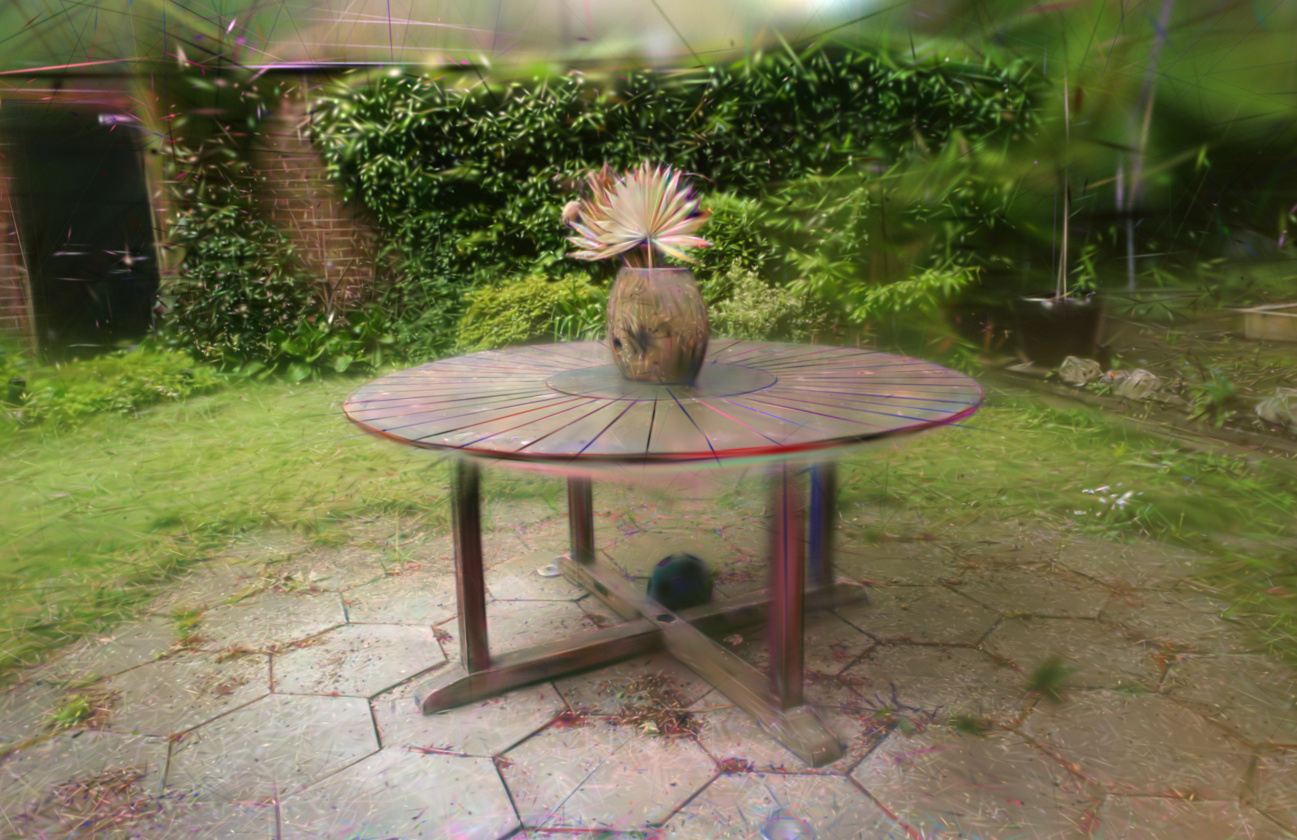}} &
                \rotatebox{90}{\tiny Init} &
                \raisebox{\imagevshift}{\pcell[width=\linewidth]{13.9}{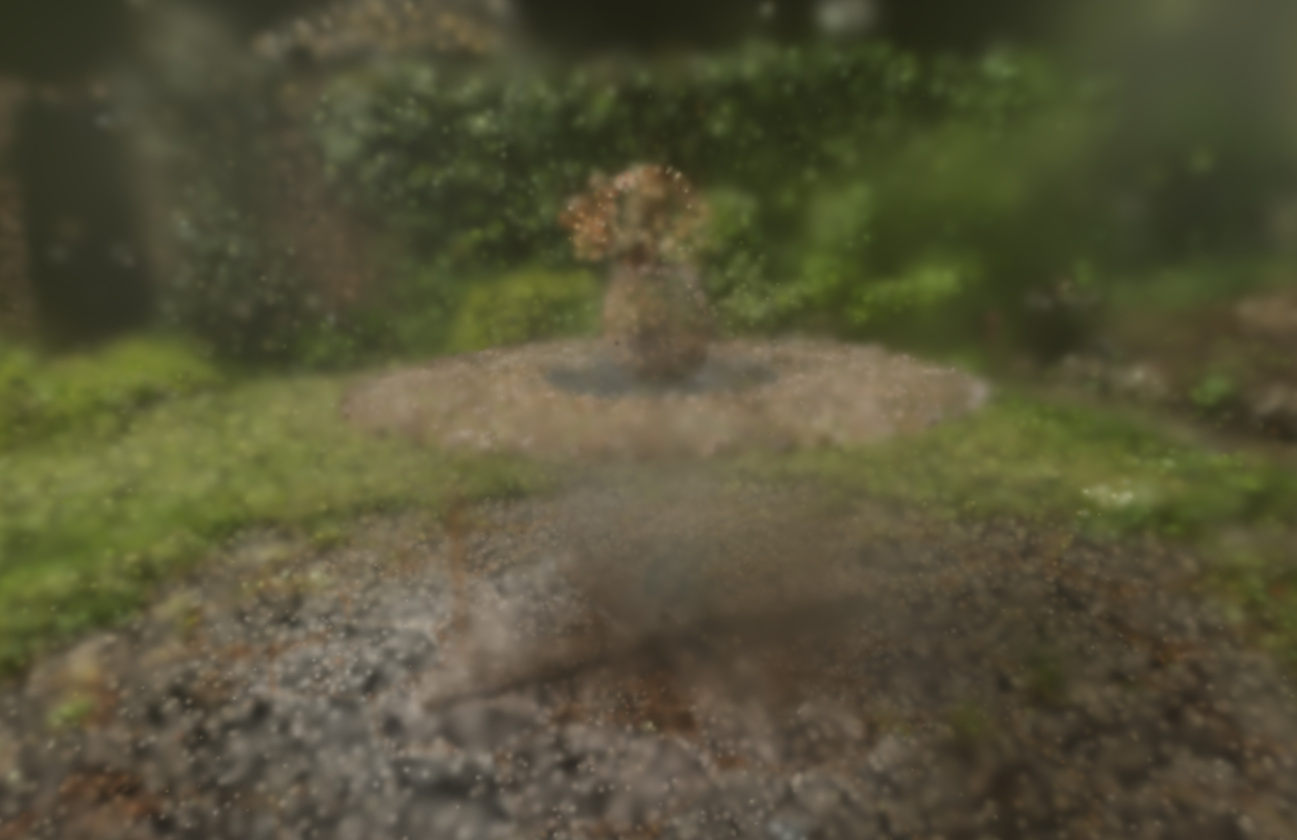}} \\
                \noalign{\vskip 2\tabcolsep}
                \rotatebox{90}{\makecell{\tiny \adam{}}} &
                \raisebox{\imagevshift}{\pcell[width=\linewidth]{14.4}{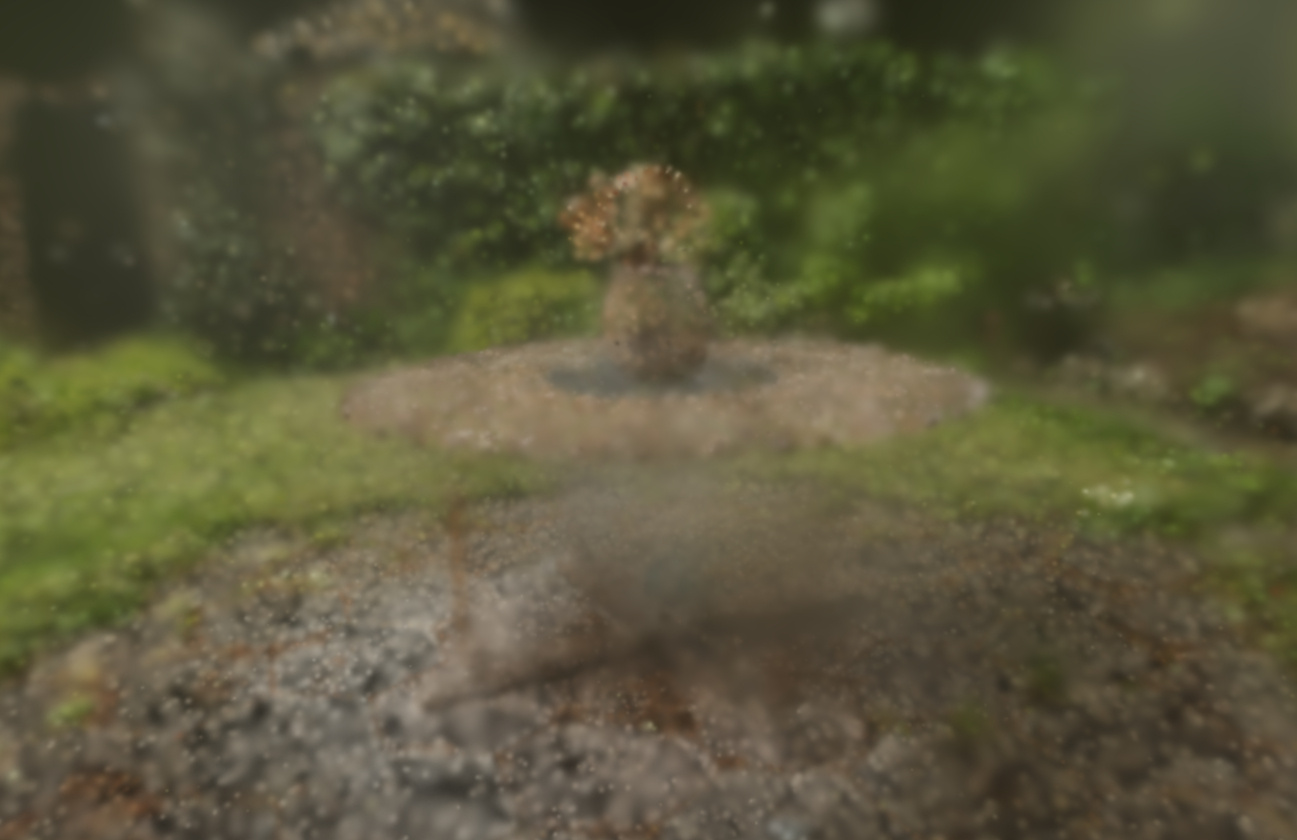}} &
                \raisebox{\imagevshift}{\pcell[width=\linewidth]{16.5}{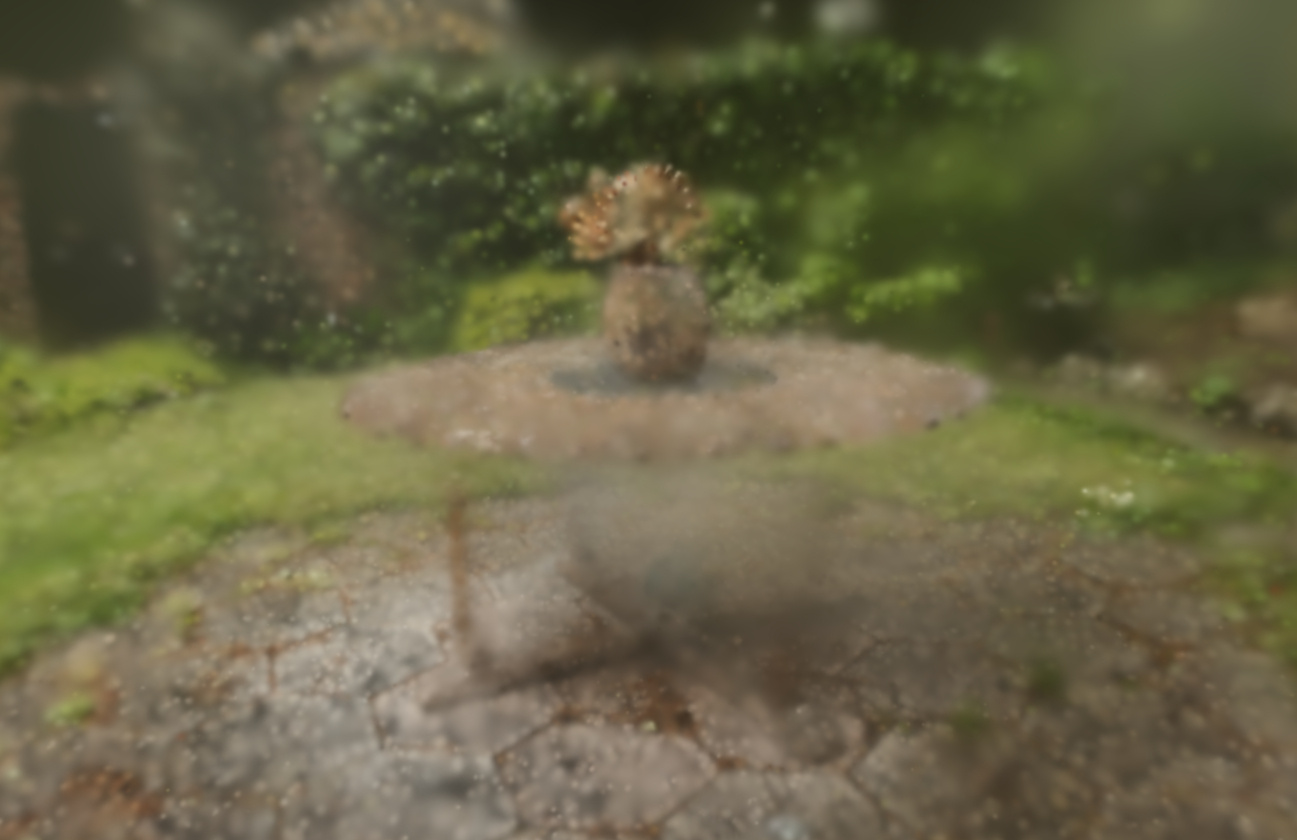}} &
                \raisebox{\imagevshift}{\pcell[width=\linewidth]{20.7}{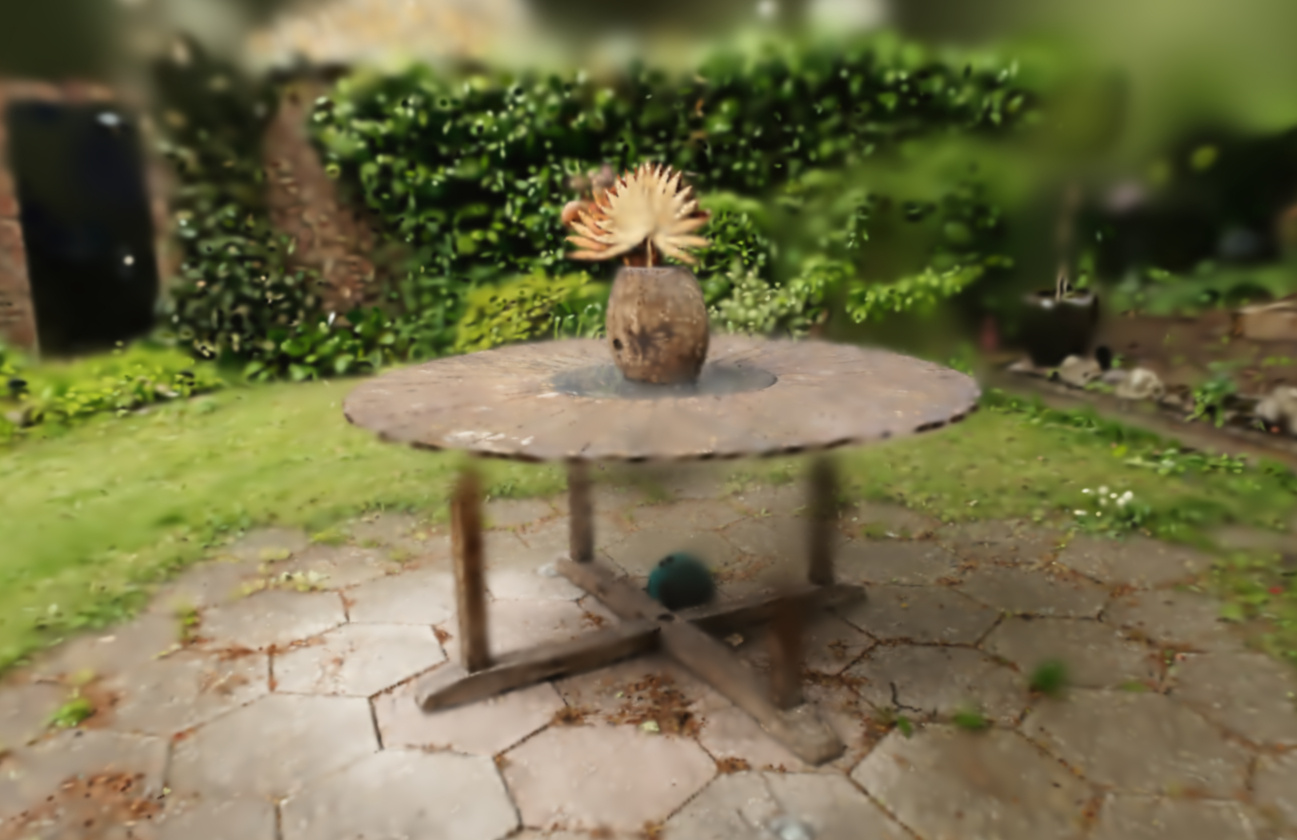}} &
                \raisebox{\imagevshift}{\pcell[width=\linewidth]{24.1}{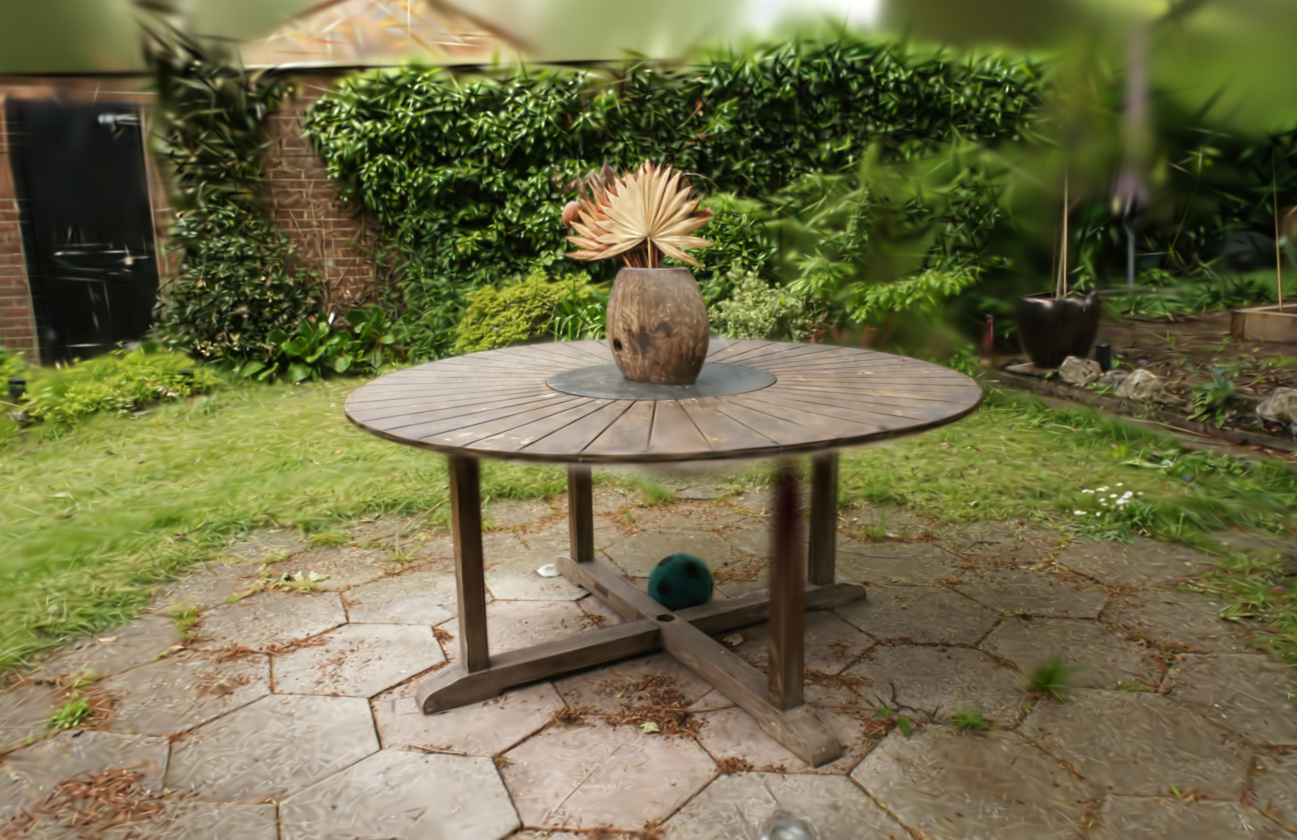}} &
                \rotatebox{90}{\tiny Ref} &
                \raisebox{\imagevshift}{\rcell[width=\linewidth]{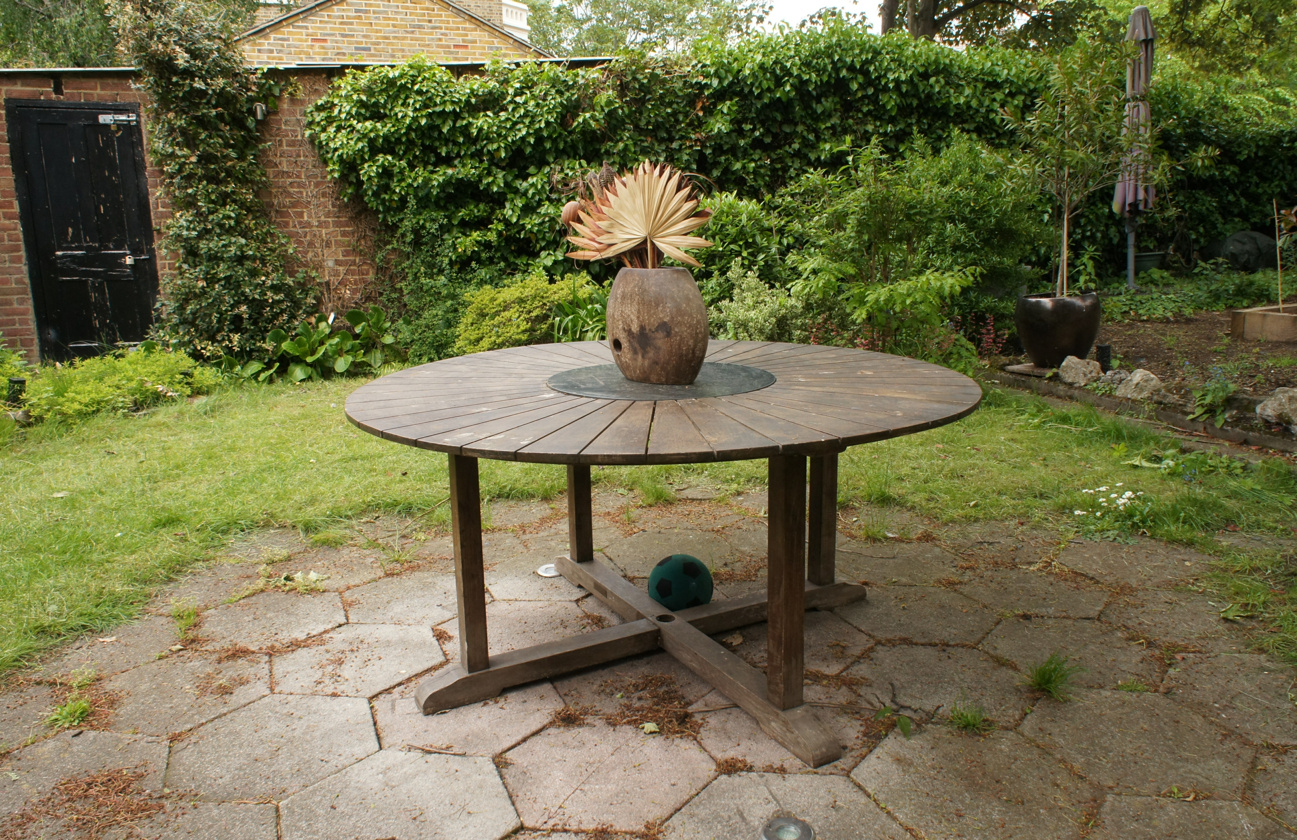}} \\
                \noalign{\vskip 2\tabcolsep}
                \rotatebox{90}{\makecell{\tiny \ourshort{}}} &
                \raisebox{\imagevshift}{\pcell[width=\linewidth]{18.9}{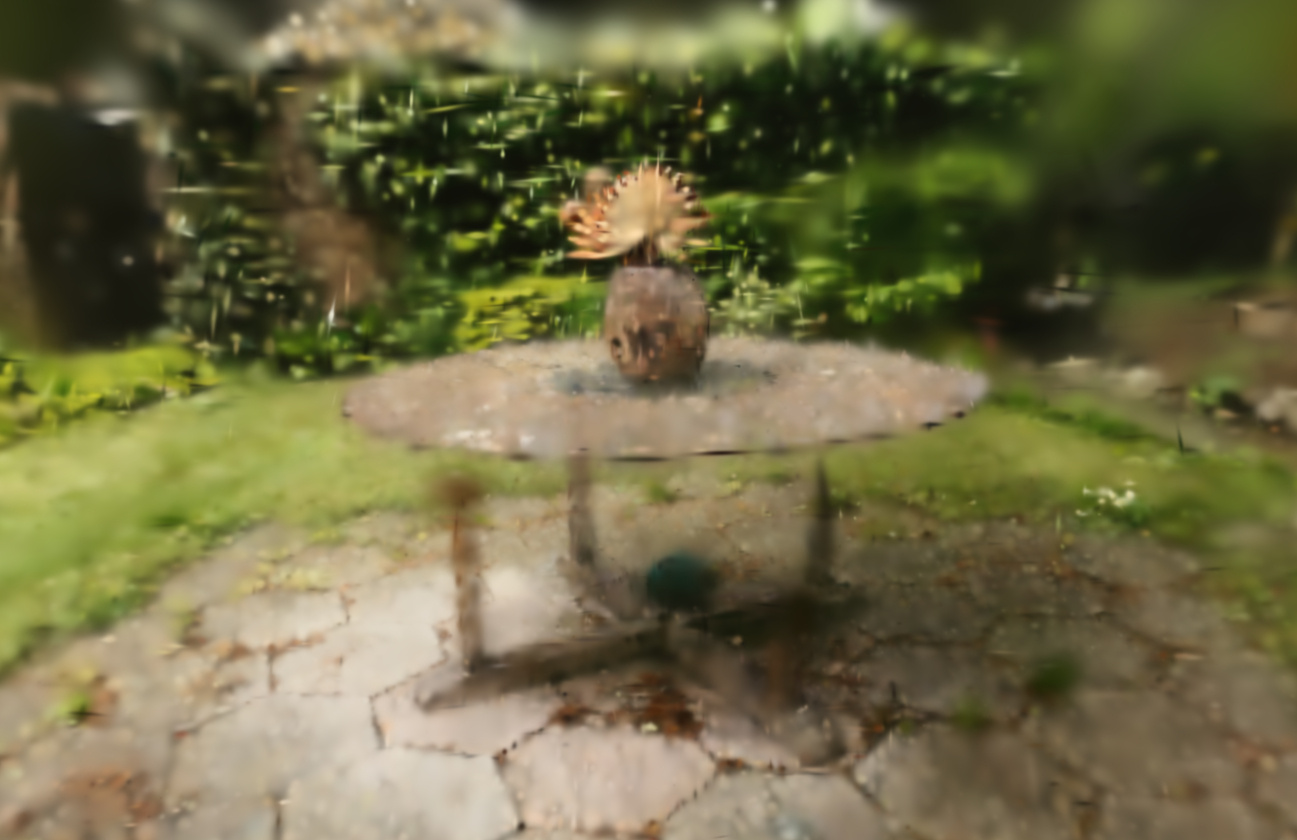}} &
                \raisebox{\imagevshift}{\pcell[width=\linewidth]{21.3}{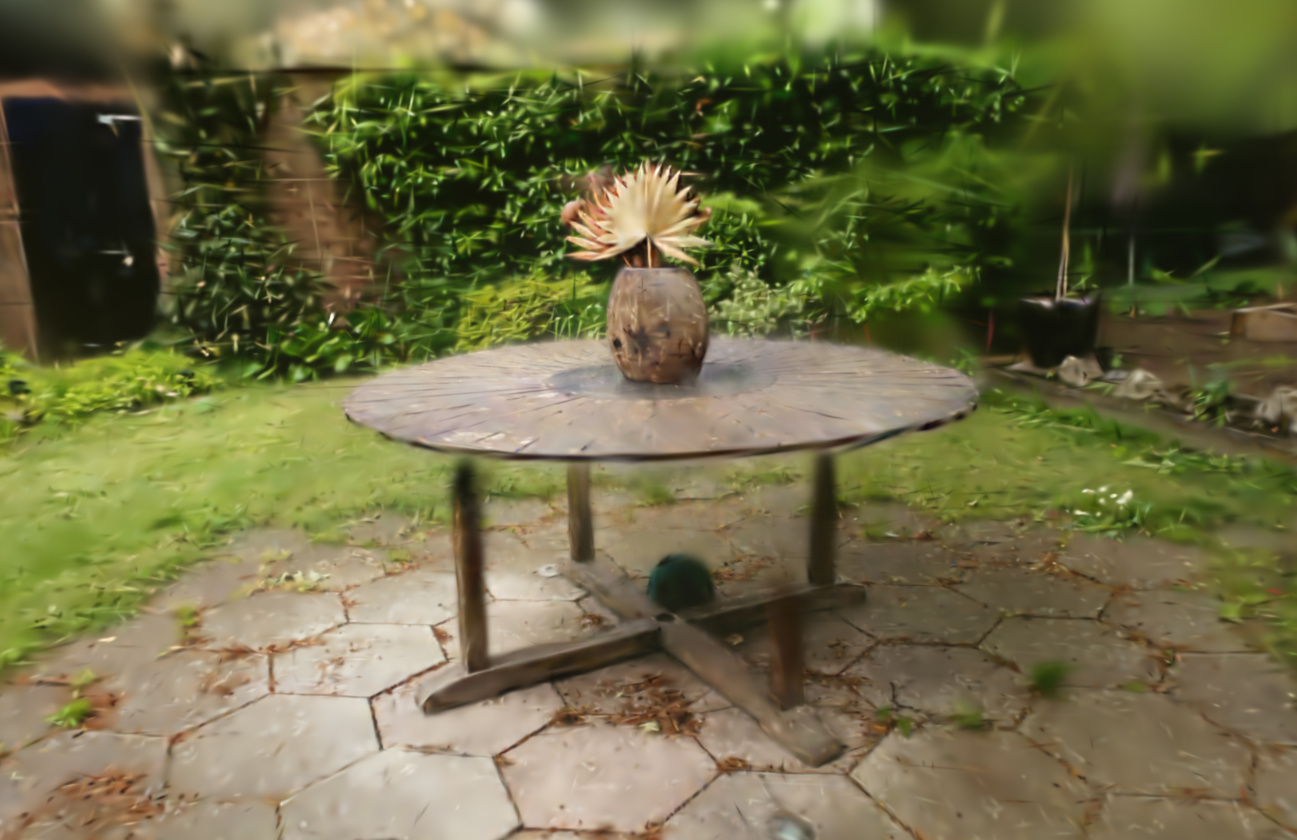}} &
                \raisebox{\imagevshift}{\pcell[width=\linewidth]{23.9}{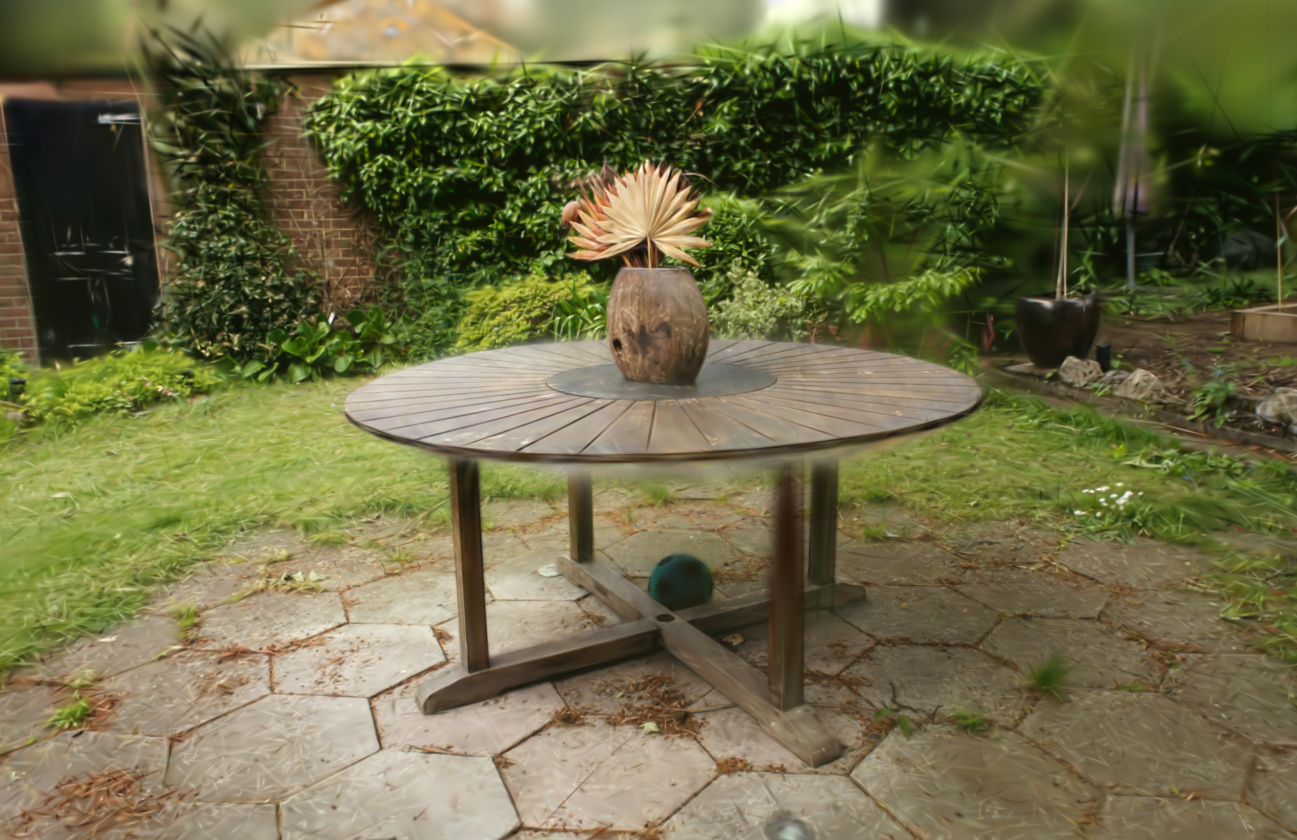}} &
                \raisebox{\imagevshift}{\pcell[width=\linewidth]{24.7}{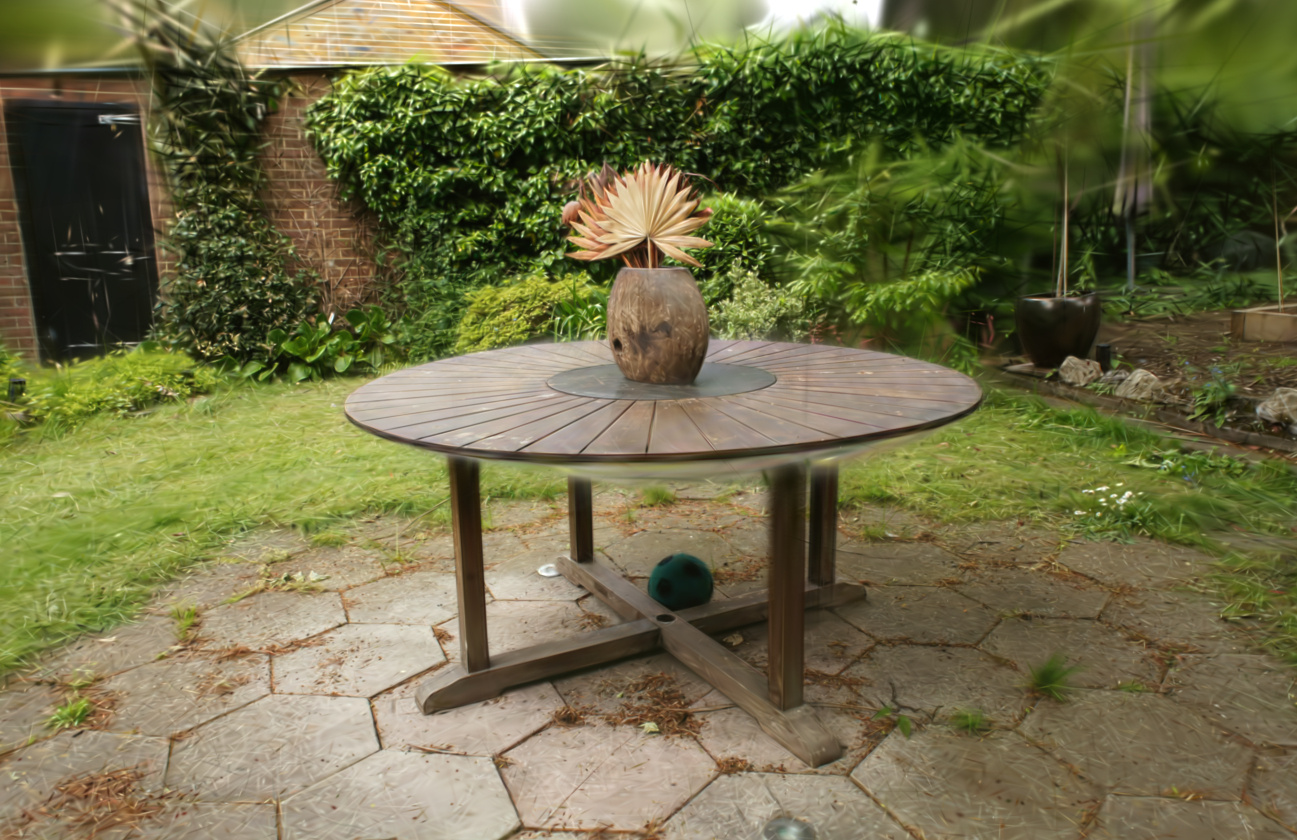}} &
                 &
                 \\
        \end{tabular}
        \caption{Mip-NeRF360~\cite{barron2022mipnerf360} scene in a dense-view setting: SfM initialization, sampling 8 views per iteration from all available views.}
        \label{fig:comparison_bottom}
    \end{subfigure}

    \caption{\textbf{Qualitative results.} \ourshort{} demonstrates zero-shot generalization to higher resolutions and different datasets.}
    \label{fig:qualitative_results}
\end{figure}

Within the fixed runtime, \ourshort{} reaches the best PSNR on all eight benchmarks, while still improving until the end of the trajectory (\cref{tab:main-results}).
It also reaches the best SSIM and LPIPS on all but RealEstate10K, where \lob{} peaks marginally higher on SSIM before diverging.
\ourshort{}'s updates yield higher quality per step, staying above \adam{} at every \emph{iteration} (\cref{fig:speed}, right, and \cref{fig:qualitative_results}).
Although each \ourshort{} iteration is ${\sim}2\times$ slower than an \adam{} step, it still reaches any PSNR level in less wall-clock \emph{time} (\cref{fig:speed}, left).
Peak-memory overhead remains $\leq\!1.1\times$ in the dense setting, where the shared renderer dominates VRAM, and $\leq\!1.7\times$ in the sparse setting.
Optimization trajectories, runtime analyses and extended tables are in the \supmat{}

\boldparagraph{Dense setting.}
We evaluate the dense-view regime with SfM-initialized Gaussians, sampling $8$ of all available context views at every iteration.
Experiments are conducted on $140$ low-resolution DL3DV~\cite{ling2024dl3dv} scenes (in-domain), $15$ DTU~\cite{Aanes2016IJCV}, $8$ LLFF~\cite{mildenhall2019llff} and $9$ Mip-NeRF360~\cite{barron2022mipnerf360} scenes, which differ in resolution, scene layout and view count.
\ourshort{} improves over \adam{} on all four datasets within the fixed runtime by $0.1$--$1.7$~dB.
\lob{} peaks fast, within $100$ steps, as it is tuned for early performance without accounting for later optimization.
It then degrades, ending $0.1$--$0.9$~dB below its own peak, and $1.8$--$3.2$~dB below \ourshort{}, while diverging to infinite values on DTU.
\cref{fig:qualitative_results} shows the same trends in images rendered at matched iterations, where \lob{} visibly degrades over the trajectory while \ourshort{} keeps improving.

\boldparagraph{Sparse setting.}
We evaluate the sparse-view regime on $140$ low- and high-resolution DL3DV~\cite{ling2024dl3dv} scenes and $72$ high-resolution RealEstate10K~\cite{Zhou2018SIGGRAPH} scenes, using the ReSplat initialization and all available context views ($8$ or $32$) at each inner step.
As in the dense setting, \ourshort{} stays above \adam{} throughout the run (\cref{fig:teaser}), improving on all four configurations by close to $0.5$~dB.
\adam{} instead drops by $0.1$--$0.3$~dB below its peak.
The gain is smaller here, as the feed-forward initialization already provides a good reconstruction.
For the same reason, \lob{} reaches \adam{}'s peak before degrading, ending $2.6$--$3.5$~dB below \ourshort{} on DL3DV and diverging to infinite values on RealEstate10K.
ReSplat improves only during the first $\sim$10 iterations before diverging, with a substantially lower peak quality.

\boldparagraph{Comparison to second-order optimization.}
We compare against second-order optimization on the nine Mip-NeRF360 scenes in LMRS's~\cite{pehlivan2026lmrs} regime (SH$_0$, $L_2$ inner loss; \cref{tab:lmrs_budget}).
LMRS's second-order step uses $16$--$32$ views, which dominates its memory.
We therefore run \adam{} and \ourshort{} at $16$ views.
We retrain \ourshort{} for SH$_0$ but meta-train it on the $L_1{+}$SSIM inner loss with $8$ views, so the $L_2$ loss and $16$-view batch it optimizes here are both out of distribution.
Within the same runtime, \ourshort{} attains the best PSNR, SSIM, and LPIPS, over $0.3$~dB above both \adam{} and LMRS, at half of LMRS's memory.

Following LMRS's protocol, we also compare to \threedgslm{}~\cite{hoellein20253dgslm}, running from both the shared SfM initialization and a $2000$ \adam{} run.
From the SfM initialization, \threedgslm{} exceeds $80$~GB on indoor scenes and is $2.3$--$4.4$~dB below \adam{} on outdoor scenes.
Warm-started from a $2000$ \adam{} run, \threedgslm{} spends most of the runtime on that initialization and fits only $13$ refinement steps, ending $0.3$~dB below \adam{} at equal time.
We discuss the batch size, iteration counts, and memory costs in the \supmat
\begin{table}[t]
\centering
\footnotesize
\setlength{\tabcolsep}{1.2pt}
\renewcommand{\arraystretch}{1.05}
\setlength{\aboverulesep}{0.2ex}\setlength{\belowrulesep}{0.3ex}
\begin{tabular}{@{}l @{\hspace{3pt}}|@{\hspace{3pt}} r r r @{\hspace{6pt}}|@{\hspace{6pt}} r @{\hspace{3pt}}|@{\hspace{3pt}} r@{}}
\toprule
{\scriptsize\emph{Optimizer}} & {\scriptsize LPIPS\,$\downarrow$} & {\scriptsize SSIM\,$\uparrow$} & {\scriptsize PSNR\,$\uparrow$} & {\scriptsize Iters} & {\scriptsize Mem\,[GB]\,$\downarrow$} \\
\midrule
\multicolumn{6}{@{}l}{\footnotesize\emph{Second-order}} \\
\arrayrulecolor{gray!40}\midrule\arrayrulecolor{black}
LMRS~\cite{pehlivan2026lmrs} & \cellcolor{tabsecond}0.443 & \cellcolor{tabsecond}0.677 & \cellcolor{tabsecond}25.43 & 335 & 14.2 \\
\threedgslm{}~\cite{hoellein20253dgslm}$^{\text{SfM init}}$ & \multicolumn{5}{c}{{\color{black!30}\leavevmode\leaders\hrule height 0.55ex depth -0.45ex\hfill\kern0pt}~\emph{OOM} on 4/9 scenes~{\color{black!30}\leavevmode\leaders\hrule height 0.55ex depth -0.45ex\hfill\kern0pt}} \\
\threedgslm{}~\cite{hoellein20253dgslm}$^{\text{\adam{}-2k init}}$ & 0.462 & 0.664 & 25.09 & 2k+13 & 33.9 \\
\addlinespace[2pt]
\arrayrulecolor{gray!70}\midrule\arrayrulecolor{black}
\addlinespace[2pt]
\multicolumn{6}{@{}l}{\footnotesize\emph{First-order}} \\
\arrayrulecolor{gray!40}\midrule\arrayrulecolor{black}
\adam{} & 0.447 & 0.674 & 25.41 & 4000 & \cellcolor{tabfirst}\textbf{6.3} \\
\ourshort{}-SH$_0$ & \cellcolor{tabfirst}\textbf{0.432} & \cellcolor{tabfirst}\textbf{0.682} & \cellcolor{tabfirst}\textbf{25.77} & 2000 & \cellcolor{tabsecond}6.5 \\
\bottomrule
\end{tabular}
\caption{\textbf{\ourshort{} outperforms the second-order methods on every metric at equal runtime.} Mip-NeRF360 scenes in the SH$_0$ and $L_2$-loss regime of LMRS. Columns follow \cref{tab:main-results}. For \threedgslm{}, \emph{Iters} counts the refinement steps taken on top of its 2k-\adam{} warm start. \colorbox{tabfirst}{\textbf{Best}} and \colorbox{tabsecond}{second best} values are highlighted.}
\label{tab:lmrs_budget}
\end{table}

\begin{figure}[!t]
    \centering
    \vspace{-10pt} %
    \includegraphics[width=\linewidth]{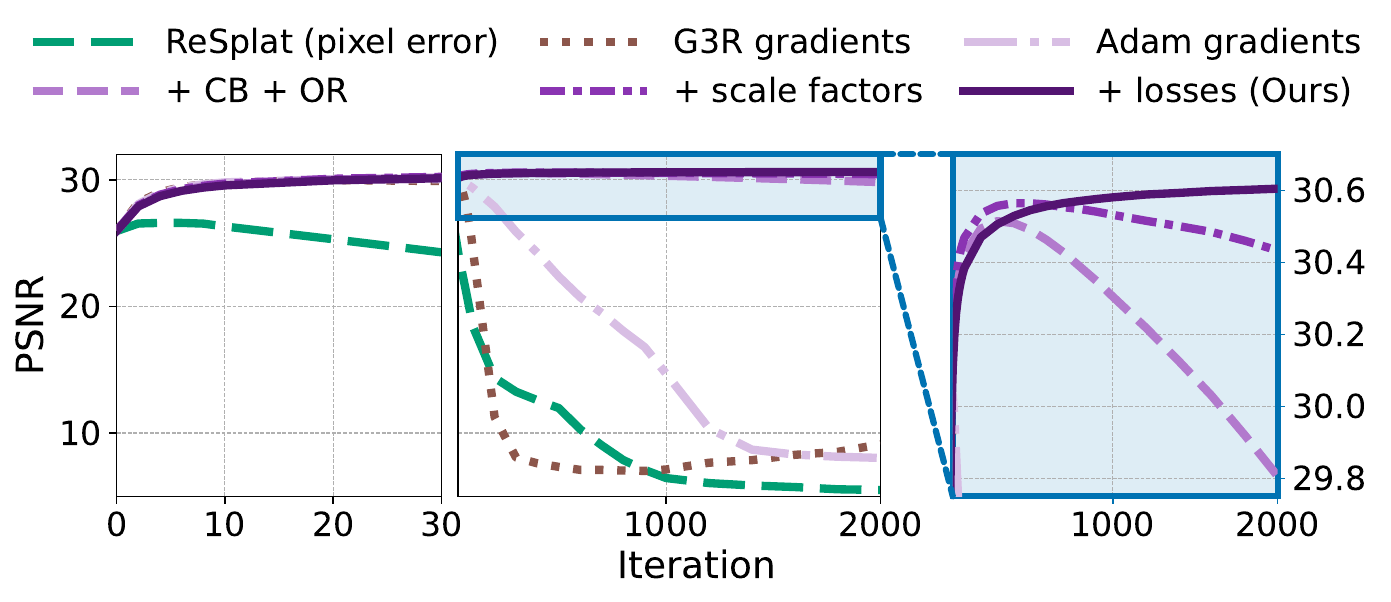}
    \caption{\textbf{Ablation study in the sparse-view training.} The shaded region is magnified on the right. A leading ``+'' marks a component added on top of the previous variant, while rows without one are single changes to ReSplat.}
    \label{fig:ablation}
\end{figure}

\boldparagraph{Ablation study.}
Replacing {\color{ablresplat}ReSplat}'s~\cite{xu2025resplat} per-pixel error inputs with {\color{ablg3r}G3R gradients}~\cite{Chen2024g3r} or {\color{abladam}Adam gradients} raises the peak from $26.6$ to $30.0$~dB but does not prevent degradation (\cref{fig:ablation}).
Starting from the Adam-gradient variant, we add our meta-training components one at a time, each retaining the peak and further reducing the degradation.
{\color{ablcbor}CB + OR} stay within $0.7$~dB of the peak and the {\color{ablscale}scale factors} within $0.13$~dB, while our {\color{ablours}losses} remove degradation entirely, with PSNR still rising at $2000$ iterations instead of peaking and declining.
The LR schedule and time encoding of prior work~\cite{Chen2024g3r, Liu2025quicksplat} are not sufficient on their own as \lob{} still loses $2.4$~dB over the same horizon (see \supmat).

    \section{Conclusions}
\label{sec:conclusion}

We introduced \ourshort{}, a learned optimizer for 3DGS that surpasses \adam{}'s reconstruction quality and reaches it faster, and unlike prior learned optimizers, remains effective across long optimization horizons.
Its effectiveness stems from its architecture, losses, and meta-learning scheme.
\ourshort{} generalizes zero-shot to higher resolutions and unseen datasets.
This points toward task-specialized optimizers for efficient, robust 3DGS reconstruction.

\boldparagraph{Future work.}
In complex scenes, our optimizer can saturate early, likely due to overly conservative updates. 
In these cases, longer optimization horizons favor \adam{}.
Future work should address this by making the learned optimizer explicitly view-aware and by introducing per-parameter-group adaptive scaling instead of a global scaling factor.
We also plan to incorporate learnable densification and pruning, jointly controlling parameter updates and scene density.

    \boldparagraph{Acknowledgements.}
    The authors affiliated with the T\"ubingen AI Center acknowledge the support of the Deutsche Forschungsgemeinschaft (DFG, German Research Foundation) under Germany's Excellence Strategy (EXC number 2064/1, project number 390727645).
    Gerard Pons-Moll is endowed by the Carl Zeiss Foundation.
    Stefano Esposito acknowledges travel support from the European Union's Horizon 2020 research and innovation program under ELISE Grant Agreement No.\ 951847.
    We thank the International Max Planck Research School for Intelligent Systems (IMPRS-IS) for supporting Patricia Gschoßmann and Amit Peleg.

    {
        \small
        \bibliography{bib/bibliography_long,bib/bibliography,bib/bibliography_custom}
    }
    \clearpage
    \setcounter{section}{0}
\renewcommand{\thesection}{\Alph{section}}
\renewcommand{\thesubsection}{\thesection.\arabic{subsection}}
\renewcommand{\thesubsubsection}{\thesubsection.\arabic{subsubsection}}

\renewcommand{\theHsection}{supp.\Alph{section}}
\renewcommand{\theHsubsection}{supp.\Alph{section}.\arabic{subsection}}
\renewcommand{\theHsubsubsection}{supp.\Alph{section}.\arabic{subsection}.\arabic{subsubsection}}

\section*{Supplementary Material}

\etocsettocdepth.toc{subsection}
\etocsetnexttocdepth{subsection}
\etocsettocstyle{\subsection*{Contents}}{}
\begingroup
\hypersetup{linkcolor=cvprblue}
\tableofcontents
\endgroup
\vspace{0.8em}

We provide additional technical details and results complementing the main paper. 
\cref{tab:notation} summarizes the mathematical notation used throughout the paper. 
We first review the Adam~\cite{Kingma2015ICLR} update rule (\cref{app:subsec-adam}) and derive the 3DGS rendering formulation (\cref{sec:app-rendering}). 
\cref{sec:app-initializations} discusses scene initialization strategies, \cref{sec:app-datasets} introduces the datasets used in our experiments.
\cref{sec:app-training} details the meta-training procedure, including the checkpoint buffer, optimizer rollout strategy, low-visibility loss, stability loss, and additional training details (\cref{subsec:supp-training-details}). 
Further architectural details are given in \cref{sec:app-impl}, while \cref{sec:app-baselines} describes the baselines and their implementations. 
\cref{sec:app-results} presents additional qualitative and quantitative results. 
Finally, \cref{sec:app-analysis} analyzes the optimization dynamics of our optimizer compared to Adam.

\section{Preliminaries}
For completeness, we provide the full Adam~\cite{Kingma2015ICLR} algorithm in \cref{app:subsec-adam} and the full 3DGS~\cite{Kerbl2023SIGGRAPH} rendering derivation in \cref{sec:app-rendering}.

\subsection{Adam}
\label{app:subsec-adam}
The general update rule (\cref{eq:update_step_standard}) can be written as follows,
\begin{equation}
\begin{aligned}
\bm_{t} &= \beta_1 \bm_{{t}-1} + (1 - \beta_1) \nabla_{\gaus_{t}}, \\
\bv_{t} &= \beta_2 \bv_{{t}-1} + (1 - \beta_2) \left(\nabla_{\gaus_{t}}\right)^2, \\
\hat{\bm}_{t} &= \frac{\bm_{t}}{1 - \beta_1^{t}}, \quad
\hat{\bv}_{t} = \frac{\bv_{t}}{1 - \beta_2^{t}}, \\
f_{\text{Adam}}\left(\nabla_{\gaus_{t}}\right) &= \frac{\hat{\bm}_{t}}{\sqrt{\hat{\bv}_{t}} + \epsilon},
\end{aligned}
\label{eq:adam-normalization}
\end{equation}
where $\epsilon$ is a small constant for numerical stability, $\beta_1$ and $\beta_2$ are decay rates for the moment estimates, with the moments initialized to $0$.

\subsection{Rendering derivation}
\label{sec:app-rendering}
Each 3D Gaussian $\gaus_m$ is defined in world space by its center $\bp_m$, scale $\bs_m$, and rotation $\bR_m$ (parameterized by a quaternion),
\begin{equation}
    \gaus_m(\bx) = e^{-\frac12 (\bx-\bp_m)^\top \bSigma_m^{-1} (\bx-\bp_m)},
\end{equation}
with 
\begin{equation}
    \bSigma_m = \bR_m\,\mathrm{diag}(\bs_m^2)\,\bR_m^\top.
\end{equation}
Each Gaussian also has an opacity $\alpha_m$ controlling its contribution to the rendered image, and spherical harmonics coefficients $\sh_m$ modeling its view-dependent appearance.

Given a viewpoint $\cV_i=(\bK_i,\bR_i,\bt_i)$, the ray passing through a pixel $\bu = (u,v)$ is expressed in camera coordinates as
\begin{equation}
    \bx_i(t)=\bo_i + t\bd_i(\bu),
\end{equation}
where $\bo_i=\mathbf 0$ is the camera origin and
\begin{equation}
    \bd_i(\bu) = \frac{\bK_i^{-1}[u,v,1]^\top}{\|\bK_i^{-1}[u,v,1]^\top\|}
\end{equation}
is the viewing direction.
To render $\gaus_m$ from viewpoint $\cV_i$, it is first transformed into the camera coordinate frame,
\begin{equation}
    \bmu_{m,i} = \bR_i(\bp_m - \bt_i),
    \quad
    \bSigma_{m,i} = \bR_i \bSigma_m \bR_i^\top.
\end{equation}
Next, the transformed Gaussian is projected onto the image plane using a local affine approximation of the perspective projection around $\bmu_{m, i}$.
The resulting 2D Gaussian footprint $\gaus_{m,i}^{2D}$ is defined by
\begin{equation}
    \bmu_{m,i}^{2D} = \pi(\bmu_{m, i}),
    \quad
    \bSigma_{m,i}^{2D} = \bJ_{m,i} \bSigma_{m,i} \bJ_{m, i}^\top,
\end{equation}
where $\bJ_{m,i}$ is the Jacobian of the projection $\pi(\cdot)$ evaluated at $\bmu_{m, i}$.

Each Gaussian $\gaus_m$ contributes to pixel $\bu = (u, v)$ with weight 
\begin{equation}
    w_{m, i} = \alpha_m\gaus_{m,i}^{2D}(\bu),
\end{equation}
and view-dependent color $\bc_{m, i}$ obtained by evaluating $\sh_m$ in the direction 
\begin{equation}
   \bv_{m, i} = -\frac{\bmu_{m, i}}{\|\bmu_{m, i}\|}.
\end{equation}
The final pixel color is obtained via front-to-back alpha blending of all Gaussians along the ray, sorted by depth,
\begin{equation}
    \bc_i(\bu) = \sum_{k=1}^K \bc_{k, i} w_{k, i} \prod_{j=1}^{k-1}(1 - w_{j, i}).
\end{equation}

In total, each Gaussian is described by $p = 59$ optimizable parameters: $3$ for the center $\bp_m$, $3$ for the scale $\bs_m$, $4$ for the rotation quaternion, $1$ for the opacity $\alpha_m$, and $48$ spherical-harmonic coefficients ($d = 16$ basis functions of degree~$3$, one set per RGB channel).
Consequently, $p = 59$ is also the dimension of the per-Gaussian parameter vector $\gaust$, the corresponding per-Gaussian gradient $\ggaust{t}$ used as input to the learned optimizer, and the per-Gaussian update vector $\Delta_{\gaust}$ produced by the optimizer (\cref{sec:app-impl}).

\section{Initializations}
\label{sec:app-initializations}
We consider two scene initialization strategies. 
In all experiments, we use the same initialization for all compared methods to ensure that performance differences reflect the effectiveness of the optimizer rather than the initialization procedure.

\boldparagraph{SfM init.}
We consider structure-from-motion (SfM) initialization using a COLMAP~\cite{Schoenberger2016CVPR} point cloud, following the standard 3D Gaussian Splatting initialization procedure~\cite{Kerbl2023SIGGRAPH}. 
In this case, per-Gaussian latent state vectors for the learned optimizer are initialized by sampling from a standard normal distribution.

\boldparagraph{ReSplat init.}
When initializing a scene with the ReSplat feed-forward network~\cite{xu2025resplat}, we run inference on the context views to produce a set of Gaussians and per-Gaussian latent state vectors derived from ReSplat's pixel-aligned features.
For the 8-view setting, the initialization contains $57$K primitives at low resolution and $245$K primitives at high resolution.
For the 32-view setting at low resolution, the initialization contains $229$K primitives.

\section{Datasets}
\label{sec:app-datasets}

\boldparagraph{DL3DV~\cite{ling2024dl3dv}.}
A large-scale dataset for deep learning-based 3D vision with 51.2M posed frames from 10,510 videos.
Testing is done on a standard 140 scenes split.
For view selection within each scene, we follow the split of ReSplat~\cite{xu2025resplat}, where context views are selected using farthest-point sampling (FPS) within a frame window, and target views are chosen evenly from the remaining frames.
We use this dataset for training and testing.

\boldparagraph{RealEstate10K~\cite{Zhou2018SIGGRAPH}.}
RealEstate10K is a large-scale dataset of real estate videos with camera poses. 
For evaluation, we use scenes from the official test split, subsampled by selecting every 25th scene. 
Scenes with an insufficient number of frames are discarded, resulting in a subset of 72 scenes over which all metrics are averaged. 
View selection within each scene follows the same procedure used for the DL3DV dataset.
We only use this dataset at test time.

\boldparagraph{DTU~\cite{Aanes2016IJCV}.}
We use a subset of 15 scenes commonly used for NVS from the DTU dataset, a large-scale dataset for 3D reconstruction and editing. 
Every 8th view is used as a target view, while the remaining views serve as context views, following the standard protocol.
We only use this dataset at test time.

\boldparagraph{LLFF~\cite{mildenhall2019llff}.}
A dataset of 8 front-facing scenes with 20 to 60 handheld images each.
Every 8th view is used as a target view, while the remaining views serve as context views, following the standard protocol.
We only use this dataset at test time.

\boldparagraph{Mip-NeRF360~\cite{barron2022mipnerf360}.}
A dataset of 9 scenes with mixed indoor and outdoor environments, each containing more than 100 views. 
We downscale all scenes by a factor of 4 and use the same context/target view split as in 3DGS~\cite{Kerbl2023SIGGRAPH}.
The second-order comparison (\cref{subsubsec:supp-lmrs}) instead adopts the standard Mip-NeRF360 protocol of 3DGS~\cite{Kerbl2023SIGGRAPH} and LMRS~\cite{pehlivan2026lmrs}, downscaling outdoor scenes by 4 and indoor scenes by 2.
We only use this dataset at test time.

\section{Meta-training}
\label{sec:app-training}

\begin{algorithm}[t]
\caption{Learning a 3DGS Optimizer $f_{\theta}$}
\label{alg:training}
\small
\begin{flushleft}  %
\begin{algorithmic}[1]
    \STATE \textbf{Input:} Dataset of 3D scenes
    \STATE \textbf{Init:} Optimizer parameters $\btheta$, meta optimizer $f$ (Adam)
    
    \FOR{$\tmeta=1,2,...$}
        \STATE Sample Gaussian set $\cG_{\tinner}^j$ of a scene $j$
        
        \FOR{$t = \tinner$ \textbf{to} $\tinner + \tau$}
            \STATE Compute inner loss gradients $\ggaust{t}$
            \STATE Update Gaussians: $\cG_{t+1}^j \gets \cG_{t}^j + f_{\btheta}(\ggaust{t}, \cG_{t}^j)$
        \ENDFOR
        
        \STATE Compute meta loss gradients $\gtheta$
        \STATE Update optimizer: $\btheta_{\tmeta+1} \gets \btheta_{\tmeta} - f(\gtheta, \btheta_{\tmeta})$
    \ENDFOR
\end{algorithmic}
\end{flushleft}
\end{algorithm}

We summarize the meta-training details in \cref{subsec:supp-training-details}.
A general meta-training scheme is described in \cref{alg:training}.
Additional details on the checkpoint buffer and the optimizer rollout strategy are provided in \cref{app:checkpoint-buffer}. 
Further details on the low-visibility loss and the stability loss can be found in \cref{subsec:supp-losses}.

\subsection{Training details}
\label{subsec:supp-training-details}
We summarize the meta-training setup of the \ourshort{} model used in the main paper.
The model is meta-trained jointly on the sparse and dense configurations described below, alternating between the two across meta-iterations.
The dedicated \oursdense{} and \ourssparse{} checkpoints discussed in \cref{subsec:supp-dedicated-vs-unified} are meta-trained using only the dense and sparse configurations, respectively.

\boldparagraph{Meta-loss supervision and weighting.}
At each inner iteration, eight context views are rendered and used to compute the inner loss, whose gradients are provided to the learned optimizer to update the Gaussians.
After each update, the resulting Gaussians are evaluated using a fixed set of six target views and four of the eight context views, which are rendered to compute the meta-loss.
The eight context views are selected differently across configurations, as described below.
For the meta-loss (\cref{eq:meta-loss}), we set the weights to $\lambda_{\text{render}} = 1$, $\lambda_{\text{lvs}} = 2$ and
$\lambda_{\text{stab}} = 1$, with the low-visibility term enabled only after the first $100$ meta-iterations.

\boldparagraph{Dataset and splits.}
We train on the DL3DV training set~\cite{ling2024dl3dv}, consisting of real-world scenes with diverse environments and camera trajectories.
We use the default DL3DV train/test scenes split throughout all experiments.

\boldparagraph{Dense configuration.}
In the dense setting, Gaussians are initialized from SfM point clouds reconstructed using all available views, which are a free by-product of camera pose estimation and are standard in 3DGS pipelines.
We observed that SfM reconstructions from DL3DV contain 10--20\% of points with RGB values of exactly [0,0,0] (black), most likely an artifact of the COLMAP reconstruction.
Empirically, a model trained on initializations containing these black Gaussians learns to exploit the additional representational capacity they provide.
However, this leads to degraded performance when evaluated on unseen datasets that lack such points, since the assumed extra capacity is absent.
For instance, fewer than 1\% of initialization points are black in the Mip-NeRF360 dataset.
Furthermore, we found that Adam struggles to recover from these uninformative Gaussians and performs substantially better when applied on filtered initializations (see also the discussion on the LLFF dataset in \cref{subsec:supp-results-dense}).
For these reasons, we exclude black points from training.

Latent states are randomly initialized at the start of each trajectory.
For each new scene, 64 context views are selected from all available frames via farthest point sampling (FPS) to maximize scene coverage.
At each inner iteration, the 8 context views are sampled (also via FPS) from this fixed pool of 64.
Note that the SfM point cloud is reconstructed from the full set of available frames, whereas the optimizer only observes the sampled subset of context views during training.
This introduces a distribution gap, as the initialization may include scene content not visible in the sampled views.
To improve robustness to varying point cloud densities and this view–initialization mismatch, we apply data augmentation by randomly retaining 10--100\% of the initial SfM points for each new scene, encouraging the optimizer to recover from sparse or incomplete initializations.
The augmentation range is chosen empirically such that its lower bound roughly matches the number of Gaussians observed in typical sparse initializations.

\boldparagraph{Sparse configuration.}
In the sparse-views forward-facing setting, Gaussians are initialized from the feed-forward predictions of ReSplat~\cite{xu2025resplat}, which provides a dense initialization from a small set of posed images.
Latent states are initialized from the feature vectors produced by the initialization network via a linear projection layer.
For each new scene, the 8 context views are sampled from a window of frames within the scene video, ensuring a relatively short baseline between views.
These 8 views also serve as the input to the ReSplat initialization.
Unlike the dense setting, the same fixed set of 8 context views is used across all inner optimization steps, reflecting the constrained view availability in sparse-views scenarios.

\boldparagraph{Image resolution and compute.}
Both configurations use low-resolution images ($256 \times 448$ pixels) during meta-training to reduce memory consumption.
All experiments are conducted in PyTorch with mixed precision (fp16/fp32) enabled.
The learned optimizer is trained end-to-end for 150{,}000 meta-iterations on 4 NVIDIA A100 GPUs (40GB), using Adam as the meta-optimizer with a learning rate of $10^{-4}$, requiring approximately 3 days.

\boldparagraph{Gradient flow and parameterization.}
At each inner iteration, Gaussian parameters and their gradients are detached before being passed to the optimizer network $f_{\btheta}$, preventing direct gradient flow through the Gaussian parameters across steps.
The latent state $\bs_t$ remains fully differentiable, serving as the sole pathway for gradients to propagate between inner iterations during meta-training.
This design choice reduces memory consumption during unrolling while preserving the optimizer's ability to learn long-horizon update strategies through the recurrent state.

\subsection{Checkpoint buffer and optimizer rollout}
\label{app:checkpoint-buffer}
\begin{figure}[t]
    \centering
        \centering
        \includegraphics[width=\linewidth]{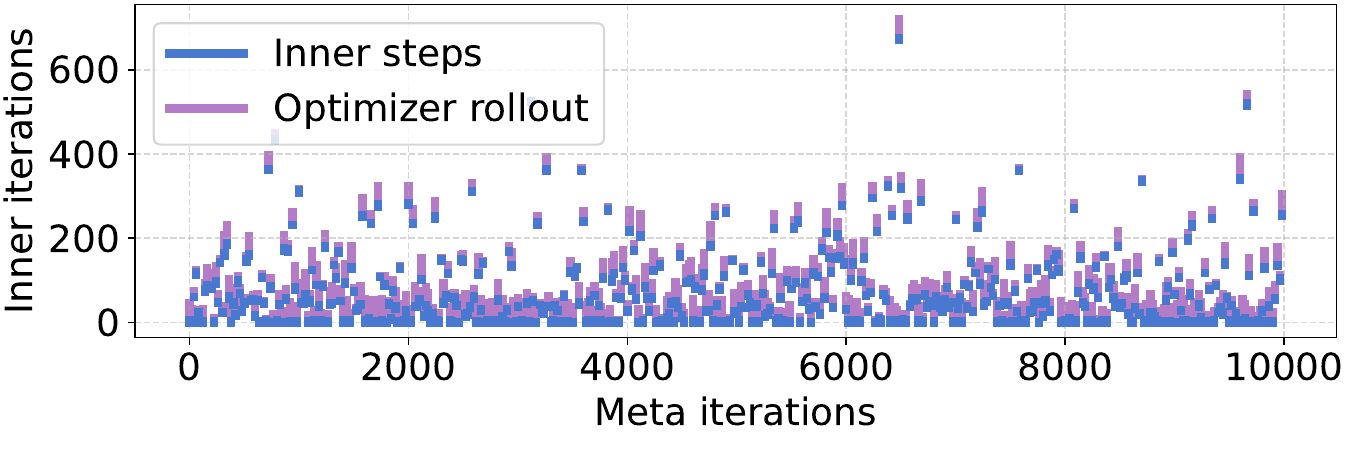}
        \caption{
        \textbf{Extended training horizon} enabled by the checkpoint buffer and optimizer rollout~(\cref{subsec:long-horizon-training}) along an entire epoch (10{,}000 meta-iterations).
        {\color{myblue}Blue lines} are inner iterations on which the learned optimizer is trained.
        {\color{mypurple} Purple lines} are optimizer rollouts, in which the learned optimizer is frozen.
        }
        \label{fig:replay_buffer_sim_full}
\end{figure}
\begin{figure}[t]
        \centering
        \includegraphics[width=\linewidth]{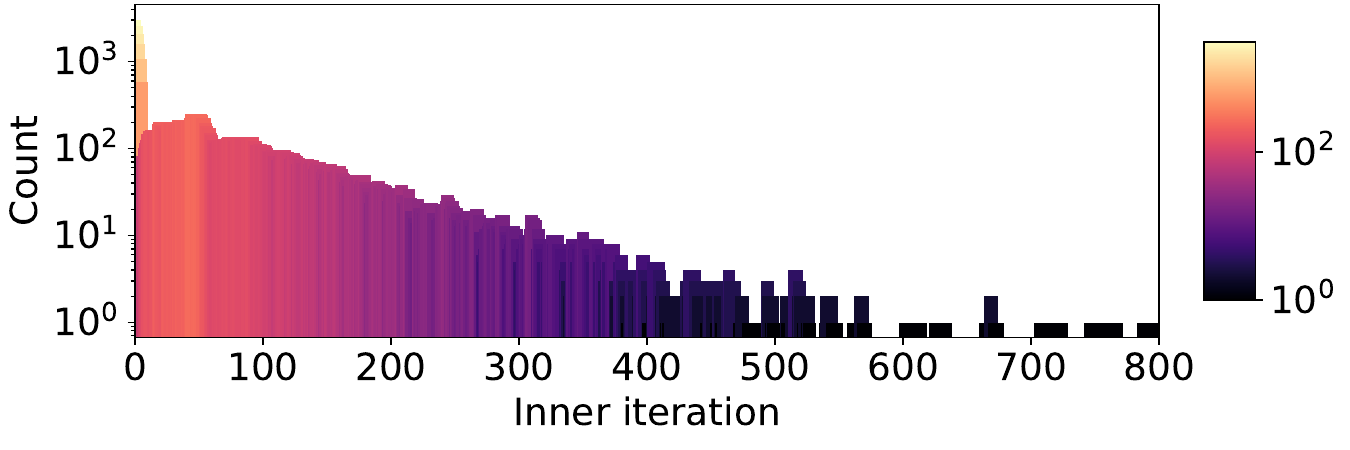}
        \caption{
        \textbf{Distribution of inner steps} encountered by the learned optimizer during meta-training using the checkpoint buffer.
        If, at a given meta-iteration, the Gaussians start at inner step 20 and are updated for 6 timesteps, the range [20, ..., 25] is considered as inner steps observed once by the optimizer.
        }
        \label{fig:replay_buffer_hist}
\end{figure}

During meta-training, we employ a checkpoint buffer with a capacity of 20 scenes.
Each scene contains the intermediate Gaussian parameters after applying $\tinner$ optimization steps ($\gaus^j_{\tinner}$), the optimizer state (\eg, Adam statistics), and auxiliary information required for consistent continuation across meta-iterations (\eg, scene id and total number of inner iterations applied).  

At the start of each meta-iteration, the optimizer resamples a stored scene with probability $p_{\text{buffer}} = 0.7$, or samples a new scene (post initialization) with probability $1 - p_{\text{buffer}}=0.3$.
At the end of each meta-iteration, we first \emph{rollout} the optimizer for additional inner steps without updating the meta-learner. 
The number of rollout steps is drawn uniformly between 1 and $\tau_a$, where $\tau_a$ increases linearly from 1 to 50 over the first 10{,}000 meta-iterations (\ie, the first epoch).
Then, the optimizer pushes the current scene to the checkpoint buffer with a probability $p_{\text{push}} = 0.7$ for new scenes and $p_{\text{push-back}} = 0.99$ for resampled scenes.
If a push occurs when the buffer is already full, the oldest scene is removed from the buffer.
Note that the checkpoint buffer imposes no fixed limit on the number of inner optimization steps a scene may undergo.
Our configuration ensures that the meta-learner is exposed to a balanced mixture of newly sampled and replayed episodes while also observing additional simulated trajectories.  

\cref{fig:replay_buffer_sim_full} shows the simulation of a single epoch of meta-training ($10{,}000$ meta-iterations). 
In each meta-iteration the rollout length is sampled uniformly between 1 and $\tau_a=50$.
{\color{myblue}Blue lines} indicate inner iterations on which the learned optimizer is actively trained (sampled between 1 and $\tau=6$), while {\color{mypurple}purple lines} denote optimizer rollouts during which the learned optimizer is frozen. 
As shown, the meta-learner encounters trajectories from both early and late inner timesteps (e.g., $t_{\text{inner}} = 300$) throughout the entire meta-training process, providing a more faithful representation of its eventual deployment on novel scenes.

\cref{fig:replay_buffer_hist} further shows the distribution of inner timesteps sampled during a single meta-epoch.
Since a fresh scene is drawn with probability $0.3$ and always begins at inner step $0$, the early timesteps (1--6) are covered by approximately $3{,}000$ of the $10{,}000$ meta-iterations, whereas later timesteps are distributed over a much wider range and are therefore each visited less often.
This sampling bias aligns with the observation that the optimizer predicts its largest updates near initialization, consistent with \ourshort{} achieving $90\%$ of the PSNR gain over \adam{} within the first inner iterations (\cref{fig:supp-timing-dense,fig:supp-timing-sparse}).
The buffer nonetheless retains later, near-saturated timesteps in the training distribution, exposing the optimizer to the late stages of optimization encountered during long optimization runs.
For completeness, we provide pseudo-code for the checkpoint-buffer algorithm in \cref{alg:checkpoint-buffer}.
\begin{algorithm}[t]
\caption{Checkpoint Buffer Training for a Learned Optimizer in 3DGS }
\label{alg:checkpoint-buffer}
\small
\begin{flushleft}  %
\begin{algorithmic}[1]
    \STATE \textbf{Init:} Checkpoint buffer $\mathcal{B} \gets \emptyset$
    \STATE \textbf{Input:} Learned optimizer $f_\theta$; probabilities $p_{\text{buffer}}$, $p_{\text{push}}$, $p_{\text{push-back}}$, $\tau_a$
    \FOR{each meta-iteration $\tmeta$}
        \STATE \textcolor{bordeaux}{\footnotesize\texttt{\# Sample new scene 
        or from buffer}}
        \STATE $\gaus^j_{t_{\text{inner}}} \gets
            \begin{cases}
                \text{Sample}(\mathcal{B}) & \text{if rand() $< p_{\text{buffer}}$ and $\mathcal{B}\neq\emptyset$} \\
                \text{Initialize}() & \text{otherwise}
            \end{cases}$
        
        \STATE \textcolor{bordeaux}{\footnotesize\texttt{\# Inner optimization}}
        \STATE $\tau \gets U(1, 6) $
        
        \FOR{$t = t_{\text{inner}}$ \textbf{to} $t_{\text{inner}}+\tau$}
            \STATE $\gaus^j_{t} \gets \text{Update}(\gaus^j_{t-1}, f_{\btheta})$
        \ENDFOR
        \STATE Update optimizer parameters $\btheta_{t_{\meta}}$ with accumulated loss

        \STATE \textcolor{bordeaux}{\footnotesize\texttt{\# Optimizer rollout ($f_{\btheta}$ frozen)}}
        \STATE $\tau_{\text{rollout}} \gets U(1, \tau_a) $
        \FOR{$t = t_{\text{inner}}+\tau$ \textbf{to} $t_{\text{inner}}+\tau + \tau_{\text{rollout}}$}
            \STATE $\gaus^j_t \gets \text{Update}(\gaus^j_{t-1}, f_{\btheta})$
        \ENDFOR

        \STATE \textcolor{bordeaux}{\footnotesize\texttt{\# Push to buffer}}
        \STATE $\text{Push}(\mathcal{B}, \gaus^j_t, 
            p_{\text{push}} \text{ if $\gaus^j_{t_{\text{inner}}}$ new else } p_{\text{push-back}})$
    \ENDFOR
\end{algorithmic}
\end{flushleft}
\end{algorithm}

\subsection{Losses}\label{subsec:supp-losses}

\boldparagraph{Low-visibility loss.}
As discussed in~\cref{subsec:meta-learning}, Gaussians with low visibility receive only weak supervision from the rendering loss. 
To mitigate this, we introduce a per-parameter \emph{low-visibility} loss. 
For clarity, we describe the loss for a single Gaussian out of the $G$ Gaussians in the scene.   
For each Gaussian parameter update $\Delta_{\gaust} \in \nR^{p}$, with its corresponding raw gradient $\ggaust{t} \in \nR^{p}$ and normalized Adam-style gradient $\aggaust{t} \in \nR^{p}$, %
we impose a penalty on parameter $i$ whenever one of the following conditions holds:

\noindent
    (i) \textbf{Vanishing gradient:} the gradient magnitude is \emph{extremely} small,
    \begin{equation}
        |{\ggaust{t}}_{,i}| < \varepsilon, \qquad \varepsilon = 10^{-8}.
    \end{equation}
    (ii) \textbf{Sign disagreement:} the predicted update direction disagrees with the Adam update direction $-\aggaust{t}$,
    \begin{equation}
        \operatorname{sign}({\Delta_{\gaust}}_{,i}) \neq \operatorname{sign}(-{\aggaust{t}}_{,i}).
    \end{equation}
    
\noindent
We define a binary mask for each parameter as
\begin{equation}
m_i = \mathbf{1}\!\left[\, |{\ggaust{t}}_{,i}| < \varepsilon
\;\;\lor\;\;
\operatorname{sign}({\Delta_{\gaust}}_{,i}) \neq \operatorname{sign}(-{\aggaust{t}}_{,i})
\right],
\end{equation}
and express the low-visibility loss in vector form for a single Gaussian as
\begin{equation}
    \mathcal{L}_{\text{lvs}}
    = \sum_{i=1}^{p} m_i \, |\Delta_{\gaust,i}|.
\end{equation}
This applies an $\ell_1$ penalty exclusively to parameters that satisfy the low-visibility conditions, thereby providing explicit feedback to the meta-optimizer on how to update parameters that would otherwise receive weak or inconsistent gradient signals.

\boldparagraph{Stability loss.}
\begin{figure}[t]
    \centering
    \includegraphics[width=0.5\linewidth]{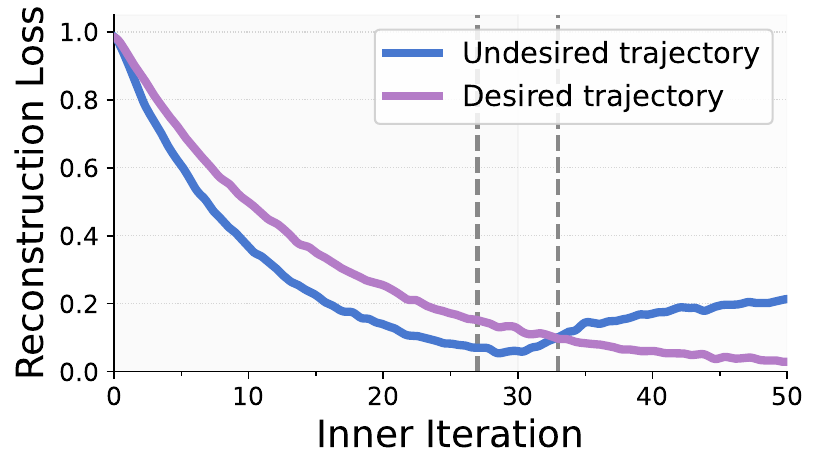}
    \caption{
\textbf{Mitigating optimization degradation via stability constraints.} 
Within a meta-iteration (dashed lines), the blue trajectory achieves a lower instantaneous loss than the purple one even though it has already entered a degraded regime that will lead to worse reconstructions later. 
Relying solely on absolute reconstruction error would incorrectly favor this unstable path. The stability loss instead penalizes local increases in error, promoting the more robust, consistently improving trajectory shown in purple.
}
    \label{fig:supp-stability-loss}
\end{figure}
The stability loss addresses a common failure mode where learned optimizers discover ``shortcuts'' that yield a low instantaneous error but lead to long-term divergence. In practice, an optimizer may exploit local minima by overfitting to current views or prematurely optimizing higher-order spherical harmonics. As illustrated in \cref{fig:supp-stability-loss}, evaluating progress based only on ground-truth loss at a fixed point fails to distinguish these unstable spikes from sustainable improvement. By penalizing increases in reconstruction error between consecutive inner iterations, this loss term encourages the discovery of update directions that maintain consistent optimization behavior over time.

\section{Architectural details}
\label{sec:app-impl}
This section provides additional architectural details of our learned optimizer.

\boldparagraph{kNN-based Point Transformer.}
The Point Transformer branch takes as input the per-Gaussian input vector $\bx = [\aggaust{t}, \gaust, \bs_t]$, consisting of the $p$-dimensional Adam-style gradients, the $p$-dimensional Gaussian parameters, and the $256$-dimensional latent state.
With the per-Gaussian parameter count $p = 59$ (\cref{sec:app-rendering}), the total input dimension is $2p + 256 = 374$.
The branch begins with a linear layer $[374, 256]$, followed by four sequential Transformer blocks.
Each block applies LayerNorm to its input and performs $k$-nearest neighbors (kNN) attention over the 3D point cloud of Gaussian means.
In each kNN attention layer, the $256$-dim input is linearly mapped to $192$ dimensions and split into query, key, and value vectors of $64$ dimensions each. 
Attention is computed for each Gaussian using its $k$ nearest neighbors, yielding a $64$-dim output per point, which is then projected back to $256$ dimensions via a linear layer $[64, 256]$.
The attention output and the block input are combined via a residual (skip) connection before being passed through LayerNorm and an MLP consisting of two linear layers $[256, 1024]$ and $[1024, 256]$, with a GeLU activation in between, followed by a second skip connection.
The output of the final Transformer block produces the updated Gaussian states $\bs_{t+1}$ before scaling.

We analyze the kNN transformer scalability in a controlled setup with varying numbers of primitives ($G \in [10^{4}, \ldots, 10^{6}]$).
Empirically, both peak memory and runtime scale approximately linearly in $G$.
In absolute terms, at the scale of our undensified Gaussian scenes ($G\!\sim\!100$K), the kNN transformer adds $\sim\!1$~GB and $\sim\!22$~ms per inner iteration.
In our end-to-end experiments, the total memory footprint depends on the setting.
In the dense setting, the Gaussian count is fixed by the SfM point cloud and VRAM is dominated by the shared renderer, so the peak-memory overhead over \adam{} stays within $1.1\times$.
In the sparse setting, the initialization is pixel-aligned, so $G$ grows with both the image resolution and the number of context views, and at $8$ views and high resolution ($G\!\approx\!245$K) the overhead reaches $1.7\times$.
Following ReSplat, we initially use $k = 16$ neighbors, which accounts for $28\%$ of the inner-step runtime.
Ablating $k$ shows it can be reduced to $4$, lowering total runtime by $15.7\%$ without affecting quality; our final models therefore use $k = 4$.
Finally, for additional efficiency, we run the kNN operation only every 100 optimization iterations at test time.

\boldparagraph{State Scale MLP.} 
The \emph{State Scale MLP} is a lightweight network that takes as input the same $374$-dimensional vector $\bx$ as the Point Transformer branch, i.e.\ the full concatenated vector.
It consists of two linear layers $[374, 187]$ and $[187, 1]$, each followed by a ReLU activation.

\boldparagraph{Update MLP.}
The \emph{Update MLP} includes two linear layers $[256, p{+}1]$ and $[p{+}1, p{+}1]$, with a GeLU activation in between.
The first $p$ elements of the output vector $\bO_{\gaust}$ form the update direction $\tilde{\Delta}_{\gaust}$, which is normalized to unit length.
The last element is activated with ReLU and represents a learned update scaling factor.

\section{Baselines}
\label{sec:app-baselines}
We detail here every baseline we compare against, including the first-order references \adam{} and \adamtuned{}, the learned optimizers ReSplat and \lob{}, and the second-order methods LMRS and \threedgslm{}.

\subsection{\adam{}}
\label{subsec:app-baselines-adam}

For the \adam{} baseline, we use separate learning rates for each group of parameters, as in the original implementation \cite{Kerbl2023SIGGRAPH} (see ~\cref{tab:3dgs-lrs}).
The learning rate for the Gaussian means is scaled based on the number of optimization steps performed (log-linear interpolation).
Adam's $\beta$ hyperparameters are kept at their default values
($\beta_1=0.9, \beta_2=0.999$).
Unlike the standard setting, we always use a batch size of 8 views for rendering and loss computation. 
In the sparse setting, this corresponds to using all available views at each iteration, while in the dense setting we sample a different set of views at each iteration.
Adaptive density control is not applied.
Spherical harmonics are optimized at degree~3 from the first iteration, with no SH warm-up schedule, matching the controlled protocol used for all optimizers.
\begin{table}[t]
    \vspace{0pt} %
    \centering
    \small %
    \begin{tabular}{lc}
    \toprule
    \textbf{Parameter} & \textbf{LR} \\
    \midrule
    means (init)     & 1.6e-4 \\
    means (final)    & 1.6e-6 \\
    scales           & 5e-3   \\
    rotations        & 1e-3   \\
    opacities        & 5e-2   \\
    sh0s             & 2.5e-3 \\
    shNs             & 1.25e-4 \\
    \bottomrule
    \end{tabular}
    \captionof{table}{Learning rates used for the \adam{} baseline.}
    \label{tab:3dgs-lrs}
\end{table}

\subsection{\adamtuned{}}
\label{subsec:app-baselines-adam-tuned}
In addition to the standard \adam{} hyperparameters (\cref{tab:3dgs-lrs}), we grid-search the best-performing configuration for the sparse setting.
Specifically, we tune the learning rates and the Adam optimizer parameters using 10 scenes from the DL3DV test set, rendered at low resolution with 8 views per scene and ReSplat initialization (the setup of \cref{fig:supp-sparse-curves}(a)).
We find that the best-performing learning rates are approximately $5\times$ larger than those listed in ~\cref{tab:3dgs-lrs}, with no decay applied to the mean learning rate. 
The optimal Adam hyperparameters are $\beta_1 = 0.99$ and $\beta_2 = 0.999$.
These values are used in all experiments that reference \adamtuned{}.
Since \adamtuned{} is grid-searched specifically for the sparse setting, it improves results in that regime but does not provide consistent gains for the standard dense-view SfM initialization (\cref{subsec:supp-adam-vs-tuned}).
We therefore use default \adam{} in the main paper and report \adamtuned{} in the supplement for completeness.

\subsection{ReSplat}
We use the 8 views, low resolution setting, matching the sparse training configuration of \ourshort{}.
As ReSplat operates on pixel-aligned Gaussians, we compare it only in the sparse settings, given the ReSplat initialization.

\subsection{\lob{}}
\label{subsec:supp-lob}
As discussed in \cref{subsec:results}, \lob{} is a learned-optimizer baseline that shares our architecture but adopts G3R's training strategy in place of our long-horizon components, isolating the effect of the training strategy from the network architecture.
Like our method, \lob{} builds on the ReSplat~\cite{xu2025resplat} backbone, replacing its pixel-aligned error input with a gradient-based signal, and uses the same SH dimensionality and Adam-style gradient computation.
It follows G3R's short-horizon scheme (the cosine LR scheduler of~\cite{Chen2024g3r} and the time positional encoding of~\cite{Chen2024g3r,Liu2025quicksplat}, with the optimizer updated at every inner step) and omits our long-horizon components ($\mathcal{L}_{\text{lvs}}$, $\mathcal{L}_{\text{stab}}$, the checkpoint buffer, the optimizer rollout, and state and update scaling).
It is meta-trained on the same sparse and dense configurations as \ourshort{}.

With this baseline, we establish a \emph{fair and controlled reference} for evaluating the impact of our long-horizon components.
The comparison shows that the auxiliary mechanisms of prior work (time encoding, LR scheduling) only partially mitigate optimization degradation and cannot fully resolve it across training horizons.

\boldparagraph{Time positional encoding.}
We follow QuickSplat's implementation of the time encoding mentioned by both G3R~\cite{Chen2024g3r} and QuickSplat~\cite{Liu2025quicksplat}.
It embeds the iteration step with the sinusoidal timestep encoding common to diffusion models~\cite{nichol2021improved}, in the form introduced for sequence positions~\cite{Vaswani2017NIPS},
\begin{equation}
\begin{aligned}
\gamma(t) &= \left(\sin\left(\frac{t}{T}\,\omega_k\right), \cos\left(\frac{t}{T}\,\omega_k\right)\right)_{k=0}^{m-1}, \\
\omega_k &= 10000^{-k/(m-1)},
\end{aligned}
\end{equation}
where $T$ is the horizon of the run, $10000$ is the maximum period and we set $m=6$ frequencies.
Normalizing the step by $T$ rather than by a fixed constant keeps the conditioning within the range seen during meta-training when the schedule is stretched to longer runs.

\boldparagraph{Learning rate scheduler.}
The baseline employs the DDPM cosine learning rate scheduler~\cite{nichol2021improved}, following~\cite{Chen2024g3r}.
The learning rate at step $t$ is
\begin{equation}
\eta_t = \eta_{\min} + (\eta_{\max}-\eta_{\min})\cos \left( \frac{t/T+s}{1+s} \cdot \frac{\pi}{2} \right)^2,
\end{equation}
where $\eta_{\max}=1$ and $\eta_{\min}=0$ denote the initial and minimum learning rates, respectively, and $T$ denotes the horizon the schedule is defined over.
We use the default offset $s=0.008$. 
This schedule allows for large updates early in training and gradual decay as optimization progresses.

For \lob{}, we follow the training setup of G3R\@.
The learned optimizer is updated at every inner iteration during meta-training, using 24 inner steps.
This differs from our approach, which aggregates information across multiple inner iterations before performing an update.
During meta-training we therefore set the maximum timestep to $T=100$, rather than to the $24$ steps the optimizer is unrolled for.
The larger $T$ keeps the relative time the optimizer is conditioned on, and the learning rate that scales its updates, closer to the stage of the optimization they correspond to at test time.
We found that this choice delays the degradation.
At test time, $T$ is the length of the run, which we choose so that it roughly covers the fixed runtime we evaluate in, which is $2000$ iterations in most cases.

\subsection{LMRS}
\label{subsec:app-baselines-lmrs}
LMRS~\cite{pehlivan2026lmrs} is a second-order optimizer that does not form the normal-equation system explicitly, but rather solves it matrix-free while sampling a subset of the rendering residuals at each step.
We run it in the authors' own implementation, so it uses their rasterizer while \adam{} and \ourshort{} use ours.
Since the method is limited to SH$_0$ and their main comparison uses an $L_2$ loss, we restrict our second-order comparison to this regime.
We evaluate on the Mip-NeRF360 scenes with their default configuration for real-world scenes, where each iteration takes $16$ views, raised to $32$ after the $50$th iteration.
When reproducing their reported setting, we keep the per-scene iteration counts of $200$ to $280$ (\cref{tab:lmrs_indoor}).
In our comparison (\cref{tab:lmrs_budget,tab:lmrs_budget_r4}) we run it for the iteration count equivalent to the shared runtime.
We also sweep a fixed iteration count shared by all scenes (\cref{tab:lmrs_supp}).
The \adam{} reference reported in their evaluation uses a single view per iteration.
In our comparison (\cref{tab:lmrs_budget}), we instead run \adam{} at a matched batch size of $16$ views, isolating the optimizer from the number of views each step sees.

\subsection{\threedgslm{}}
\label{subsec:app-baselines-3dgslm}
\threedgslm{}~\cite{hoellein20253dgslm} refines an existing reconstruction with Levenberg--Marquardt steps.
In contrast to LMRS's matrix-free formulation, it materializes the Jacobian of the per-pixel rendering residuals explicitly, so its memory grows both with the number of Gaussians and with the number of pixels each one covers.
Each Levenberg--Marquardt step solves over four subsampled sets of $25$ views in turn.
We evaluate it using the authors' public implementation, keeping the SH$_0$ and $L_2$ loss to match LMRS's evaluation.
As with LMRS, it therefore renders with the authors' own rasterizer rather than ours.
Their default configuration targets the \adam{} warm start, since \threedgslm{} is never run from an SfM initialization in the original paper.
Following LMRS~\cite{pehlivan2026lmrs}, we evaluate it from both initializations.
LMRS reports $50$~GB for \threedgslm{} in both initializations on the indoor scenes.
In our reproduction the \adam{} warm start uses $40.4$~GB on those scenes.
In the SfM initialization, the initial Gaussians are larger, so more pixels contribute to the Jacobian.
\threedgslm{} then exceeds the $80$~GB of an NVIDIA H100 on the four indoor scenes, which carry higher Gaussian counts, while the outdoor scenes still fit.
In \cref{tab:lmrs_indoor}, which reproduces LMRS's table, we follow \threedgslm{}'s released protocol of $5$ refinement steps, $10$ on counter and kitchen scenes (Mip-NeRF360), on top of its $10$k \adam{} run.
In our \cref{tab:lmrs_budget,tab:lmrs_budget_r4}, we follow \cref{tab:main-results} and run every method to the same runtime.

\begin{table}[t]
\centering
\small
\setlength{\tabcolsep}{6pt}
\renewcommand{\arraystretch}{1.15}
\begin{tabular}{@{}l c c@{}}
\toprule
\textbf{Evaluation} & \textbf{Dense} & \textbf{Sparse} \\
\midrule
\adam{} vs.\ \adamtuned{}           & \multicolumn{2}{c}{\cref{tab:adam-vs-tuned}} \\
Dedicated vs.\ unified    & \multicolumn{2}{c}{\cref{tab:dedicated-vs-mixed}} \\
\midrule
Main tables      & \cref{tab:dense-all}                 & \cref{tab:sparse-all} \\
Metrics vs.\ time      & \cref{fig:supp-timing-dense} & \cref{fig:supp-timing-sparse} \\
Metrics vs.\ iteration & \cref{fig:supp-dense-curves} & \cref{fig:supp-sparse-curves} \\
\midrule
Densification            & \cref{tab:mcmc-adc}                  & -- \\
Second-order                & \cref{tab:lmrs_budget_r4,tab:lmrs_indoor,tab:lmrs_supp,fig:supp-lmrs-curves} & -- \\
\midrule
Qualitative comparisons              & \cref{fig:mipnerf-dtu-comp,fig:llff-comp}                 & \cref{fig:re10k-comp,fig:dl3dv-comp} \\
\bottomrule
\end{tabular}
\caption{\textbf{Overview of the supplementary results}.}
\label{tab:supp-results-overview}
\end{table}

\section{Additional results}
\label{sec:app-results}
\cref{tab:supp-results-overview} maps every evaluation in this section to its table or figure.
We start by evaluating the two Adam baseline configurations used throughout these results (\cref{subsec:supp-adam-vs-tuned}) and compare the joint training vs dedicated configuration training (\cref{subsec:supp-dedicated-vs-unified}).
We then provide additional results in the dense setting (\cref{subsec:supp-results-dense}), covering integration with a densification scheme (\cref{subsec:supp-mcmc}) and the comparison to second-order optimization (\cref{subsubsec:supp-lmrs}).
We then show detailed results on the sparse setting (\cref{subsec:supp-results-sparse}).
Both settings report trajectories against optimization iteration and against wall-clock time.
Finally, we show qualitative comparisons (\cref{subsec:supp-qualitative}).

\subsection{\adam{} vs.\ \adamtuned{}}
\label{subsec:supp-adam-vs-tuned}
Our experiments in the supplementary use two Adam configurations.
The standard \adam{} (\cref{subsec:app-baselines-adam}) keeps the default 3DGS learning rates, while \adamtuned{} (\cref{subsec:app-baselines-adam-tuned}) uses learning rates and momentum that were grid-searched on sparse DL3DV scenes.
As \cref{tab:adam-vs-tuned} shows, \adamtuned{} improves final PSNR on the sparse benchmarks over \adam{} ($+0.1$ to $+0.5$~dB), but since it was grid-searched over the sparse configuration, it underperforms \adam{} on the dense DL3DV, LLFF, and Mip-NeRF360.
DTU is the one dense exception, where \adamtuned{} still improves over \adam{} ($+0.35$~dB).
Since no configuration is best across both regimes, the main results use the default \adam{}, and we report \adamtuned{} here for completeness.
Nonetheless, \ourshort{} outperforms \adam{} on all eight benchmarks, and \adamtuned{} on all but DTU, where \adamtuned{} is ahead by $0.06$~dB.
\begin{table}[t]
\centering
\small
\setlength{\tabcolsep}{3pt}
\renewcommand{\arraystretch}{1.05}
\setlength{\aboverulesep}{0.2ex}\setlength{\belowrulesep}{0.3ex}
\begin{tabular}{@{}l l @{\hspace{4pt}}|@{\hspace{4pt}} r r r@{}}
\toprule
 & & \adam{} & \adamtuned{} & \ourshort{} \\
\midrule
\multirow{4}{*}{Dense} & DL3DV & \cellcolor{tabsecond}27.78 & 27.18 & \cellcolor{tabfirst}\textbf{29.04} \\
 & DTU & 28.96 & \cellcolor{tabfirst}\textbf{29.31} & \cellcolor{tabsecond}29.25 \\
 & LLFF & \cellcolor{tabsecond}22.82 & 20.98 & \cellcolor{tabfirst}\textbf{24.56} \\
 & Mip360 & \cellcolor{tabsecond}25.62 & 24.42 & \cellcolor{tabfirst}\textbf{25.75} \\
\midrule
\multirow{4}{*}{Sparse} & DL3DV 8, low-res & 30.02 & \cellcolor{tabsecond}30.29 & \cellcolor{tabfirst}\textbf{30.53} \\
 & DL3DV 8, hi-res & 26.60 & \cellcolor{tabsecond}26.72 & \cellcolor{tabfirst}\textbf{27.16} \\
 & DL3DV 32, low-res & 29.78 & \cellcolor{tabsecond}30.02 & \cellcolor{tabfirst}\textbf{30.26} \\
 & RE10K 8 & 25.57 & \cellcolor{tabsecond}26.02 & \cellcolor{tabfirst}\textbf{26.08} \\
\bottomrule
\end{tabular}
\caption{%
\textbf{\adam{} vs.\ \adamtuned{} (\cref{subsec:supp-adam-vs-tuned}).} Best PSNR within the runtime of Adam with default learning rates (\adam{}), Adam with learning rates grid-searched on sparse DL3DV scenes (\adamtuned{}), and our learned optimizer (\ourshort{}).
The tuned learning rates improve every sparse setting but degrade most dense ones (DTU excepted); as no single Adam configuration is best across regimes, the main results use the default \adam{}. \ourshort{} outperforms \adam{} on every benchmark, and \adamtuned{} on all but DTU, where \adamtuned{} is ahead by $0.06$~dB.
\colorbox{tabfirst}{\textbf{Best}} and \colorbox{tabsecond}{second best} values are highlighted.%
}
\label{tab:adam-vs-tuned}
\end{table}

\begin{table}[t]
\centering
\small
\setlength{\tabcolsep}{3pt}
\renewcommand{\arraystretch}{1.05}
\setlength{\aboverulesep}{0.2ex}\setlength{\belowrulesep}{0.3ex}
\begin{tabular}{@{}l l @{\hspace{4pt}}|@{\hspace{4pt}} r r r@{}}
\toprule
 & & \ourssparse{} & \oursdense{} & \ourshort{} \\
\midrule
\multirow{4}{*}{Dense} & DL3DV & 17.03 & 28.92$^{*}$ & \cellcolor{tabfirst}\textbf{29.07} \\
 & DTU & 21.61 & 29.10$^{*}$ & \cellcolor{tabfirst}\textbf{29.26} \\
 & LLFF & 14.21 & 24.33$^{*}$ & \cellcolor{tabfirst}\textbf{24.46} \\
 & Mip360 & 18.26 & 25.67$^{*}$ & \cellcolor{tabfirst}\textbf{25.78} \\
\midrule
\multirow{4}{*}{Sparse} & DL3DV 8, low-res & \cellcolor{tabfirst}\textbf{30.60}$^{*}$ & 30.24 & 30.53 \\
 & DL3DV 8, hi-res & 27.10$^{*}$ & 26.90 & \cellcolor{tabfirst}\textbf{27.17} \\
 & DL3DV 32, low-res & 30.18$^{*}$ & 30.05 & \cellcolor{tabfirst}\textbf{30.25} \\
 & RE10K 8 & \cellcolor{tabfirst}\textbf{26.28}$^{*}$ & 26.11 & 26.08 \\
\bottomrule
\end{tabular}
\caption{%
\textbf{Dedicated vs.\ unified checkpoint (\cref{subsec:supp-dedicated-vs-unified}).} Best PSNR over a $2000$-iteration trajectory of the single jointly-trained \ourshort{} model against the two dedicated per-regime checkpoints, \ourssparse{} and \oursdense{}, each meta-trained on the sparse or dense configuration. $^{*}$ marks the dedicated checkpoint evaluated inside its own training regime. We highlight the \colorbox{tabfirst}{\textbf{best}} value per row.
The unified model matches or exceeds the dedicated dense checkpoint on every dense benchmark and stays within $0.2$~dB of the dedicated sparse checkpoint.%
}
\label{tab:dedicated-vs-mixed}
\end{table}

\begin{table*}[t]
  \setlength{\tabcolsep}{2pt}
  \centering

  \begin{subtable}{0.49\linewidth}
    \centering
    \centering
{\footnotesize%
\setlength{\tabcolsep}{0.7pt}%
\renewcommand{\arraystretch}{1.05}%
\setlength{\aboverulesep}{0.2ex}\setlength{\belowrulesep}{0.3ex}%
\begin{tabular}{@{}l @{\hspace{1.5pt}}|@{\hspace{1.5pt}} r r r @{\hspace{1.5pt}}|@{\hspace{1.5pt}} r r @{\hspace{1.5pt}}|@{\hspace{1.5pt}} r@{}}
\toprule
 & LPIPS\,$\downarrow$ & SSIM\,$\uparrow$ & PSNR\,$\uparrow$ & PSNR$_{\tbud}$\,$\uparrow$ & Iters & Mem\,$\downarrow$ \\
\midrule
\multicolumn{7}{@{}l}{\emph{Initializer}} \\
\arrayrulecolor{gray!40}\midrule\arrayrulecolor{black}
SfM & 0.743 & 0.392 & 13.38 & -- & -- & -- \\
\addlinespace[2pt]
\arrayrulecolor{gray!70}\midrule\arrayrulecolor{black}
\addlinespace[2pt]
\multicolumn{7}{@{}l}{\emph{Optimizer}} \\
\arrayrulecolor{gray!40}\midrule\arrayrulecolor{black}
\adam{} & \cellcolor{tabsecond}0.185 & \cellcolor{tabsecond}0.879 & \cellcolor{tabsecond}27.78 & \cellcolor{tabsecond}27.78 & 4000 & \cellcolor{tabfirst}\textbf{1.5} \\
\adamtuned{} & 0.208 & 0.865 & 27.18 & 27.18 & 5068 & \cellcolor{tabfirst}\textbf{1.5} \\
\lob{} & 0.218 & 0.838 & 25.81 & 25.12 & 1608 & 1.6 \\
\ourshort{} & \cellcolor{tabfirst}\textbf{0.157} & \cellcolor{tabfirst}\textbf{0.896} & \cellcolor{tabfirst}\textbf{29.04} & \cellcolor{tabfirst}\textbf{29.04} & 1713 & \cellcolor{tabfirst}\textbf{1.5} \\
\bottomrule
\end{tabular}%
}
\small
\caption{\textbf{In-domain:} DL3DV~\cite{ling2024dl3dv}, $\sim$300 views, $256 \times 448$.}
\label{tab:supp-dl3dv_dense_low_res}

  \end{subtable}
  \hfill %
  \begin{subtable}{0.49\linewidth}
    \centering
    \centering
{\footnotesize%
\setlength{\tabcolsep}{0.7pt}%
\renewcommand{\arraystretch}{1.05}%
\setlength{\aboverulesep}{0.2ex}\setlength{\belowrulesep}{0.3ex}%
\begin{tabular}{@{}l @{\hspace{1.5pt}}|@{\hspace{1.5pt}} r r r @{\hspace{1.5pt}}|@{\hspace{1.5pt}} r r @{\hspace{1.5pt}}|@{\hspace{1.5pt}} r@{}}
\toprule
 & LPIPS\,$\downarrow$ & SSIM\,$\uparrow$ & PSNR\,$\uparrow$ & PSNR$_{\tbud}$\,$\uparrow$ & Iters & Mem\,$\downarrow$ \\
\midrule
\multicolumn{7}{@{}l}{\emph{Initializer}} \\
\arrayrulecolor{gray!40}\midrule\arrayrulecolor{black}
SfM & 0.657 & 0.480 & 13.13 & -- & -- & -- \\
\addlinespace[2pt]
\arrayrulecolor{gray!70}\midrule\arrayrulecolor{black}
\addlinespace[2pt]
\multicolumn{7}{@{}l}{\emph{Optimizer}} \\
\arrayrulecolor{gray!40}\midrule\arrayrulecolor{black}
\adam{} & \cellcolor{tabsecond}0.309 & \cellcolor{tabsecond}0.899 & 28.96 & 28.96 & 4000 & 1.0 \\
\adamtuned{} & 0.326 & 0.893 & \cellcolor{tabfirst}\textbf{29.31} & \cellcolor{tabfirst}\textbf{29.31} & 4182 & 1.0 \\
\lob{} & 0.350 & 0.869 & 27.19 & \multicolumn{1}{c}{DIV} & \multicolumn{1}{c}{--} & 1.0 \\
\ourshort{} & \cellcolor{tabfirst}\textbf{0.308} & \cellcolor{tabfirst}\textbf{0.900} & \cellcolor{tabsecond}29.25 & \cellcolor{tabsecond}29.25 & 1919 & 1.0 \\
\bottomrule
\end{tabular}%
}
\small
\caption{\textbf{Zero-shot:} DTU~\cite{Aanes2016IJCV}, $\sim$ 50 views, $1162 \times 1554$.}
\label{tab:supp-dtu_dense}

  \end{subtable}

  \vspace{10pt}

  \begin{subtable}{0.49\linewidth}
    \centering
    \centering
{\footnotesize%
\setlength{\tabcolsep}{0.7pt}%
\renewcommand{\arraystretch}{1.05}%
\setlength{\aboverulesep}{0.2ex}\setlength{\belowrulesep}{0.3ex}%
\begin{tabular}{@{}l @{\hspace{1.5pt}}|@{\hspace{1.5pt}} r r r @{\hspace{1.5pt}}|@{\hspace{1.5pt}} r r @{\hspace{1.5pt}}|@{\hspace{1.5pt}} r@{}}
\toprule
 & LPIPS\,$\downarrow$ & SSIM\,$\uparrow$ & PSNR\,$\uparrow$ & PSNR$_{\tbud}$\,$\uparrow$ & Iters & Mem\,$\downarrow$ \\
\midrule
\multicolumn{7}{@{}l}{\emph{Initializer}} \\
\arrayrulecolor{gray!40}\midrule\arrayrulecolor{black}
SfM & 0.741 & 0.214 & 9.17 & -- & -- & -- \\
\addlinespace[2pt]
\arrayrulecolor{gray!70}\midrule\arrayrulecolor{black}
\addlinespace[2pt]
\multicolumn{7}{@{}l}{\emph{Optimizer}} \\
\arrayrulecolor{gray!40}\midrule\arrayrulecolor{black}
\adam{} & \cellcolor{tabsecond}0.354 & \cellcolor{tabsecond}0.731 & \cellcolor{tabsecond}22.82 & \cellcolor{tabsecond}22.82 & 4000 & \cellcolor{tabfirst}\textbf{1.3} \\
\adamtuned{} & 0.443 & 0.661 & 20.98 & 20.98 & 3787 & 1.4 \\
\lob{} & 0.355 & 0.723 & 22.77 & 22.68 & 1975 & 1.4 \\
\ourshort{} & \cellcolor{tabfirst}\textbf{0.303} & \cellcolor{tabfirst}\textbf{0.767} & \cellcolor{tabfirst}\textbf{24.56} & \cellcolor{tabfirst}\textbf{24.56} & 2855 & \cellcolor{tabfirst}\textbf{1.3} \\
\bottomrule
\end{tabular}%
}
\small
\caption{\textbf{Zero-shot:} LLFF~\cite{mildenhall2019llff}, $\sim$ 20 to 60 views, $756{\times}1008$.}
\label{tab:supp-llff_dense}

  \end{subtable}
  \hfill
  \begin{subtable}{0.49\linewidth}
    \centering
    \centering
{\footnotesize%
\setlength{\tabcolsep}{0.7pt}%
\renewcommand{\arraystretch}{1.05}%
\setlength{\aboverulesep}{0.2ex}\setlength{\belowrulesep}{0.3ex}%
\begin{tabular}{@{}l @{\hspace{1.5pt}}|@{\hspace{1.5pt}} r r r @{\hspace{1.5pt}}|@{\hspace{1.5pt}} r r @{\hspace{1.5pt}}|@{\hspace{1.5pt}} r@{}}
\toprule
 & LPIPS\,$\downarrow$ & SSIM\,$\uparrow$ & PSNR\,$\uparrow$ & PSNR$_{\tbud}$\,$\uparrow$ & Iters & Mem\,$\downarrow$ \\
\midrule
\multicolumn{7}{@{}l}{\emph{Initializer}} \\
\arrayrulecolor{gray!40}\midrule\arrayrulecolor{black}
SfM & 0.754 & 0.358 & 14.04 & -- & -- & -- \\
\addlinespace[2pt]
\arrayrulecolor{gray!70}\midrule\arrayrulecolor{black}
\addlinespace[2pt]
\multicolumn{7}{@{}l}{\emph{Optimizer}} \\
\arrayrulecolor{gray!40}\midrule\arrayrulecolor{black}
\adam{} & \cellcolor{tabsecond}0.372 & \cellcolor{tabfirst}\textbf{0.715} & \cellcolor{tabsecond}25.62 & \cellcolor{tabsecond}25.62 & 4000 & \cellcolor{tabfirst}\textbf{3.4} \\
\adamtuned{} & 0.430 & 0.665 & 24.42 & 24.42 & 4504 & \cellcolor{tabfirst}\textbf{3.4} \\
\lob{} & 0.423 & 0.658 & 23.62 & 22.68 & 1441 & 4.4 \\
\ourshort{} & \cellcolor{tabfirst}\textbf{0.367} & \cellcolor{tabfirst}\textbf{0.715} & \cellcolor{tabfirst}\textbf{25.75} & \cellcolor{tabfirst}\textbf{25.75} & 2013 & 3.5 \\
\bottomrule
\end{tabular}%
}
\small
\caption{\textbf{Zero-shot:} Mip-NeRF360~\cite{barron2022mipnerf360}, 100+ views, $520 \times 780$.}
\label{tab:supp-mip360_dense}

  \end{subtable}

\caption{
  \textbf{Quantitative evaluation on the dense datasets.}
  All methods use the SfM initialization.
  Columns follow \cref{tab:main-results}.
  Every optimizer is given a fixed runtime $\tbud$, set to \adam{}'s wall-clock of $4000$ iterations, and runs as many of its own iterations as fit within it.
  PSNR, SSIM and LPIPS are reported at the best-PSNR iteration within that runtime, and PSNR$_{\tbud}$ at its last step (\emph{Iters}).
  The \emph{Initializer} row reports the SfM point cloud before any optimization.
  \adamtuned{} uses the learning rates grid-searched on sparse scenes (\cref{subsec:supp-adam-vs-tuned}).
  \ourshort{} initializes each Gaussian's latent state vector by sampling from a standard normal distribution.
  Input gradients are computed for a subset of $8$ views sampled at each iteration with furthest point sampling from the training views set.
  This matches the dense-view configuration of our joint training.
  \textbf{In-domain} results match that configuration in dataset, number of views, and resolution, while \textbf{zero-shot} results differ in all of these factors.
  \colorbox{tabfirst}{\textbf{Best}} and \colorbox{tabsecond}{second best} values are highlighted.
    See \cref{subsec:supp-results-dense} for discussion.
}
\label{tab:dense-all}
\end{table*}

\begin{figure}[!tb]
    \centering
    \captionsetup[subfigure]{labelformat=simple}
    \renewcommand\thesubfigure{(\alph{subfigure})}

    \includegraphics[width=0.8353\linewidth]{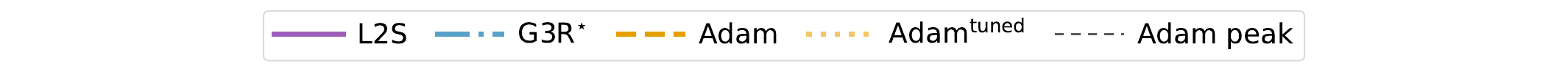}

    \medskip

    \begin{subfigure}[b]{0.49\linewidth}
        \centering
        \includegraphics[width=\linewidth, trim=0 0 0 0, clip]{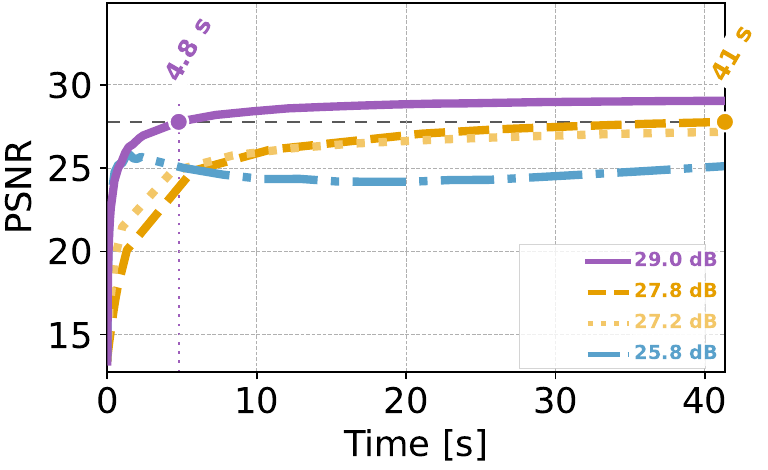}
        \caption{DL3DV~\cite{ling2024dl3dv}, $\sim$300 views, $256 \times 448$.}
        \label{fig:supp-dl3dv_sfm_timing}
    \end{subfigure}
    \hfill
    \begin{subfigure}[b]{0.49\linewidth}
        \centering
        \includegraphics[width=\linewidth, trim=0 0 0 0, clip]{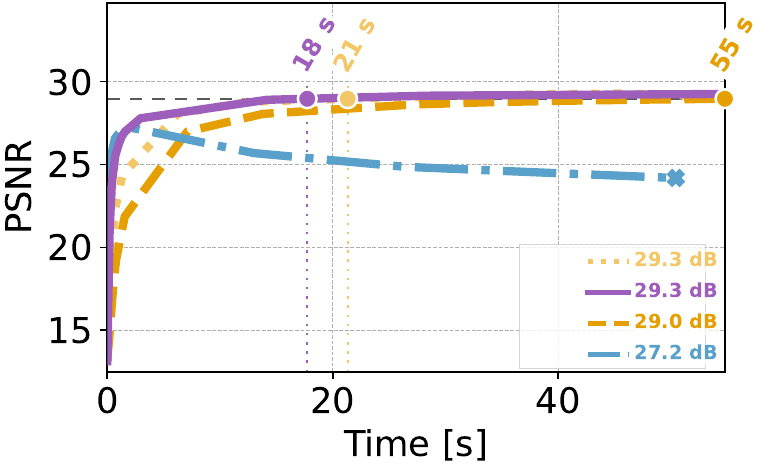}
        \caption{DTU~\cite{Aanes2016IJCV}, $\sim$ 50 views, $1162 \times 1554$.}
        \label{fig:supp-dtu_sfm_timing}
    \end{subfigure}

    \medskip

    \begin{subfigure}[b]{0.49\linewidth}
        \centering
        \includegraphics[width=\linewidth, trim=0 0 0 0, clip]{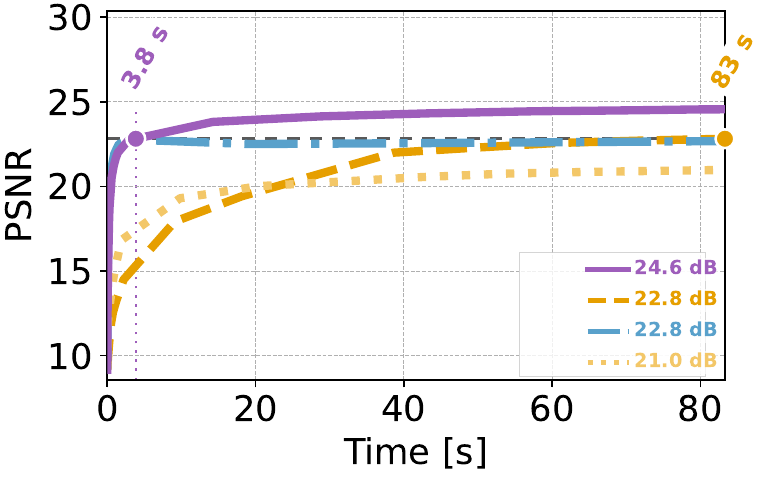}
        \caption{LLFF~\cite{mildenhall2019llff}, $20$-$60$ views, $756 \times 1008$.}
        \label{fig:supp-llff_sfm_timing}
    \end{subfigure}
    \hfill
    \begin{subfigure}[b]{0.49\linewidth}
        \centering
        \includegraphics[width=\linewidth, trim=0 0 0 0, clip]{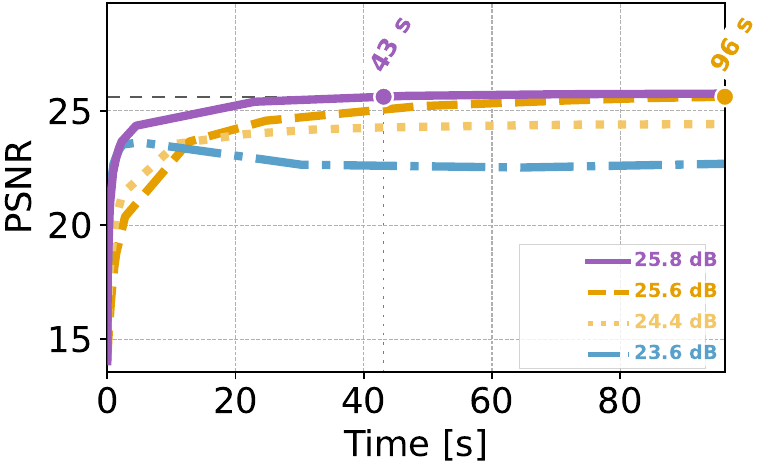}
        \caption{Mip-NeRF360~\cite{barron2022mipnerf360}, 100+ views, $520 \times 780$.}
        \label{fig:supp-mip360_sfm_timing}
    \end{subfigure} %

    \caption{\textbf{Optimization against wall-clock time, dense setting.}
    PSNR averaged across scenes, plotted against elapsed optimization time.
    Dots and labels mark where each run reaches \adam{}'s peak PSNR, and each panel lists the methods' best PSNR. A cross marks where a run's Gaussian parameters diverged to non-finite values.
    \ourshort{} reaches every quality level faster than both \adam{} variants and ends higher, except on DTU, where \adamtuned{} ends marginally above it. See \cref{subsec:supp-results-dense} for discussion.}
    \label{fig:supp-timing-dense}
\end{figure}

\begin{figure*}[!t]
    \centering
    \includegraphics[width=0.45\linewidth]{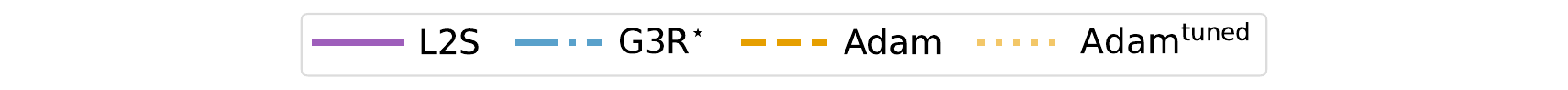}

    \medskip

    \makebox[0.245\linewidth][c]{\shortstack{\textbf{Context Views}\\[3pt]\rule{0.23\linewidth}{0.4pt}}}%
    \hfill%
    \makebox[0.748\linewidth][c]{\shortstack{\textbf{Target Views}\\[3pt]\rule{0.735\linewidth}{0.4pt}}}

    \smallskip

    \noindent
    \newcommand{\tinier}{\fontsize{7.5}{7.5}\selectfont}
    \makebox[0.245\linewidth][c]{\tinier PSNR $\uparrow$}%
    \hfill%
    \makebox[0.245\linewidth][c]{\tinier PSNR $\uparrow$}%
    \hfill%
    \makebox[0.245\linewidth][c]{\tinier SSIM $\uparrow$}%
    \hfill%
    \makebox[0.245\linewidth][c]{\tinier LPIPS $\downarrow$}

    \setcounter{subfigure}{0}
    \renewcommand{\thesubfigure}{\alph{rownum}}

    \setcounter{rownum}{1}
    \setcounter{subfigure}{0}
    \begin{subfigure}[b]{\linewidth}
        \centering
        \includegraphics[width=0.245\linewidth]{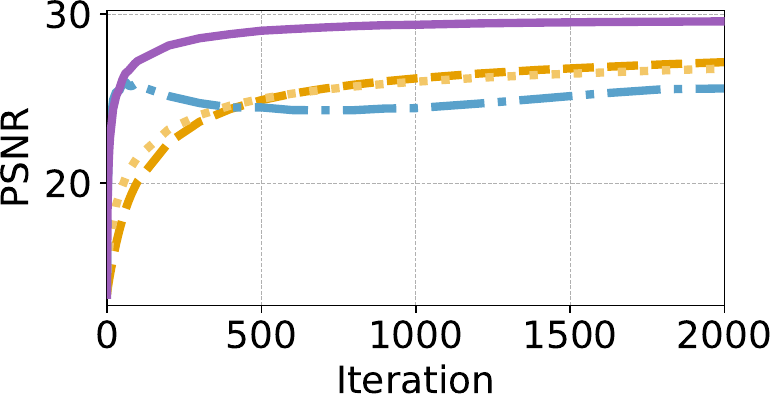}%
        \hfill%
        \includegraphics[width=0.245\linewidth]{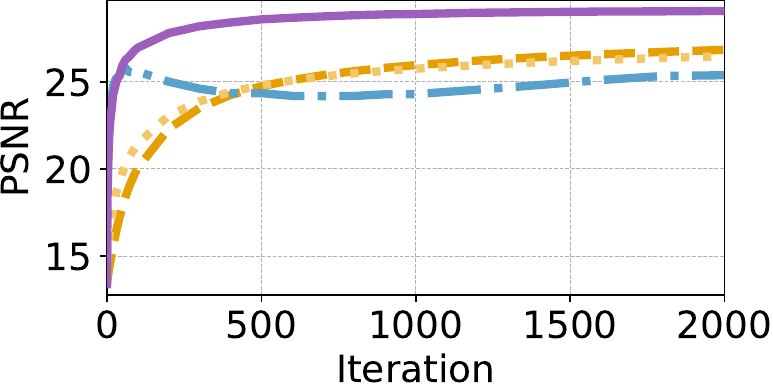}%
        \hfill%
        \includegraphics[width=0.245\linewidth]{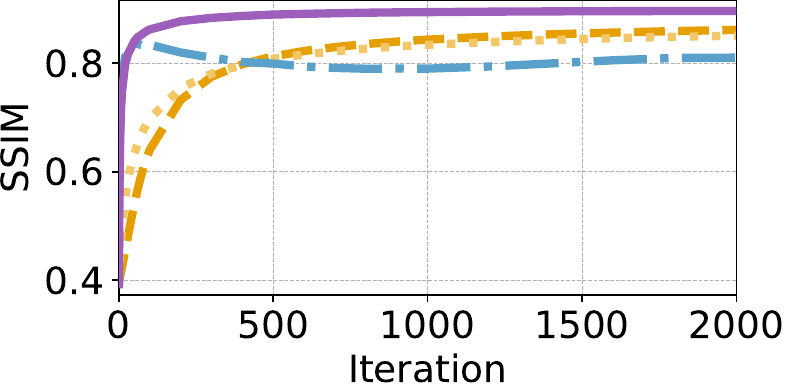}%
        \hfill%
        \includegraphics[width=0.245\linewidth]{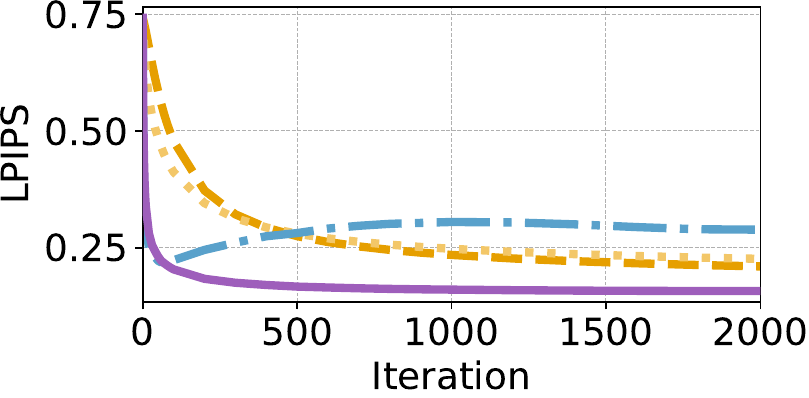}
        \caption{\textbf{In-domain}: DL3DV, $\sim$300 views, $256{\times}448$}
        \label{fig:supp-dl3dv_dense_low_res_context}
    \end{subfigure}

    \medskip

    \setcounter{rownum}{2}
    \setcounter{subfigure}{0}
    \begin{subfigure}[b]{\linewidth}
        \centering
        \includegraphics[width=0.245\linewidth]{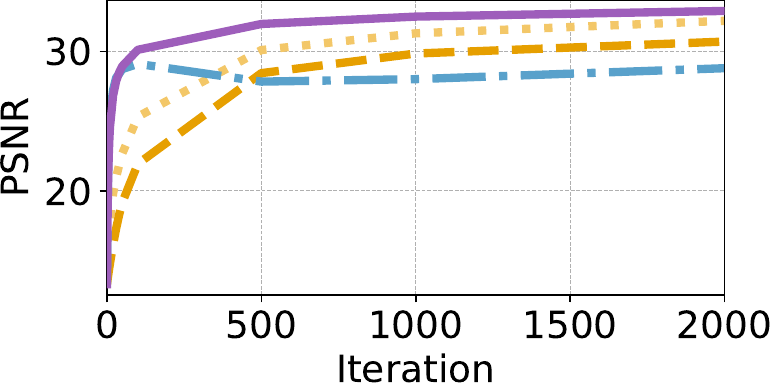}%
        \hfill%
        \includegraphics[width=0.245\linewidth]{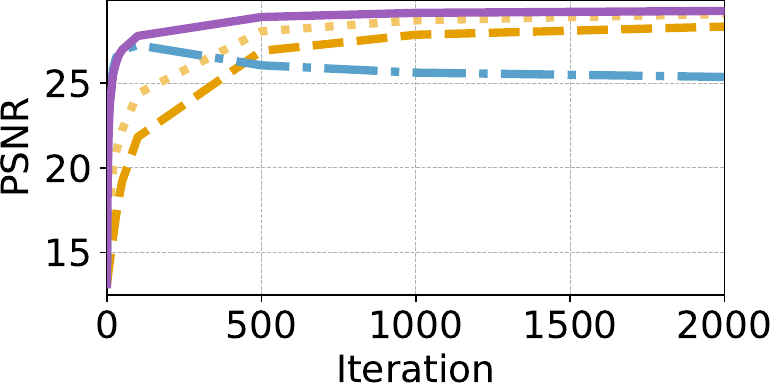}%
        \hfill%
        \includegraphics[width=0.245\linewidth]{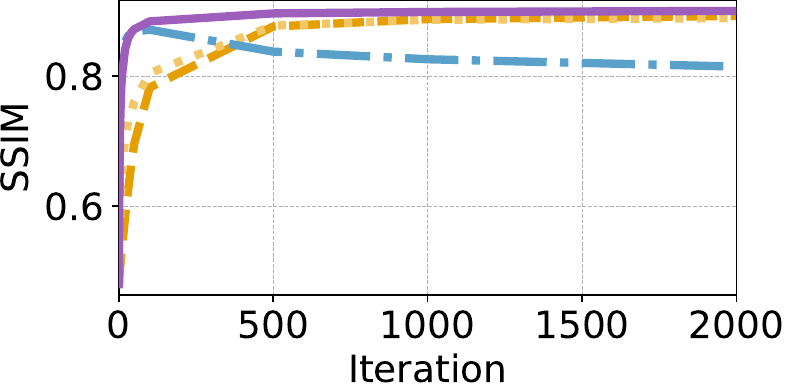}%
        \hfill%
        \includegraphics[width=0.245\linewidth]{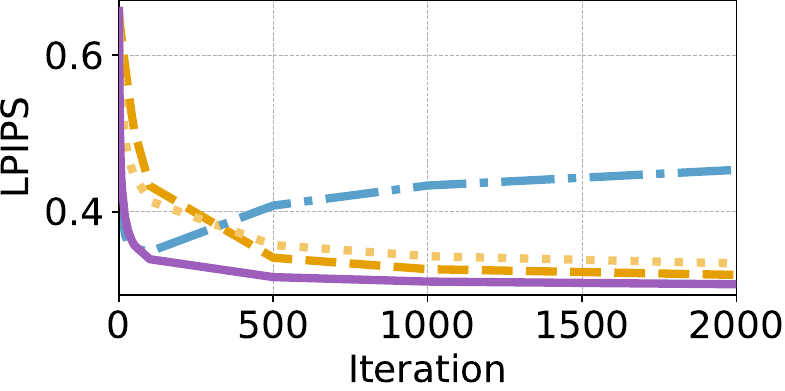}
        \caption{\textbf{Zero-shot}: DTU, $\sim$ 50 views, $1162{\times}1554$}
        \label{fig:supp-dtu_context}
    \end{subfigure}

    \medskip

    \setcounter{rownum}{3}
    \setcounter{subfigure}{0}
    \begin{subfigure}[b]{\linewidth}
        \centering
        \includegraphics[width=0.245\linewidth]{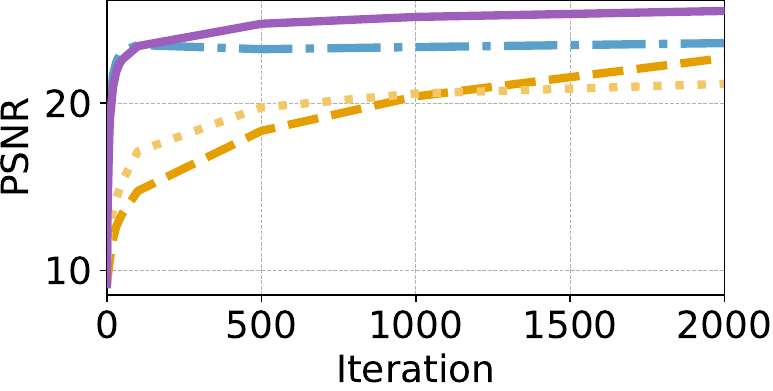}%
        \hfill%
        \includegraphics[width=0.245\linewidth]{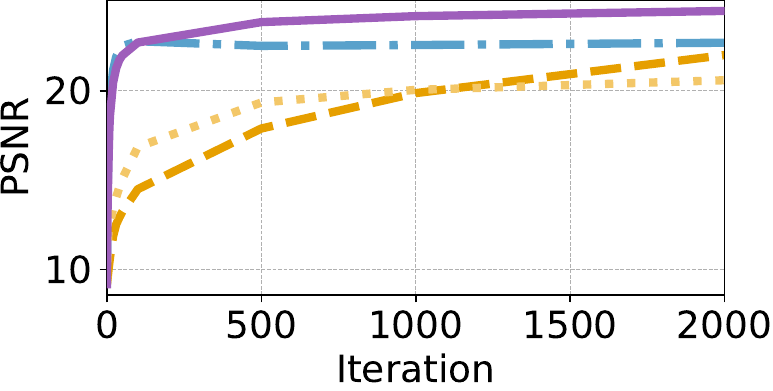}%
        \hfill%
        \includegraphics[width=0.245\linewidth]{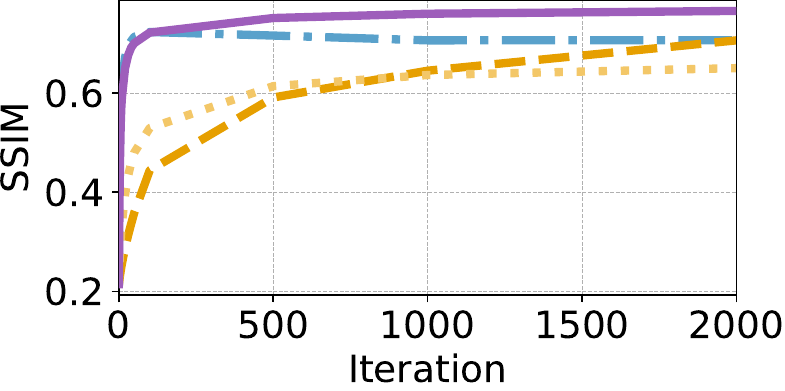}%
        \hfill%
        \includegraphics[width=0.245\linewidth]{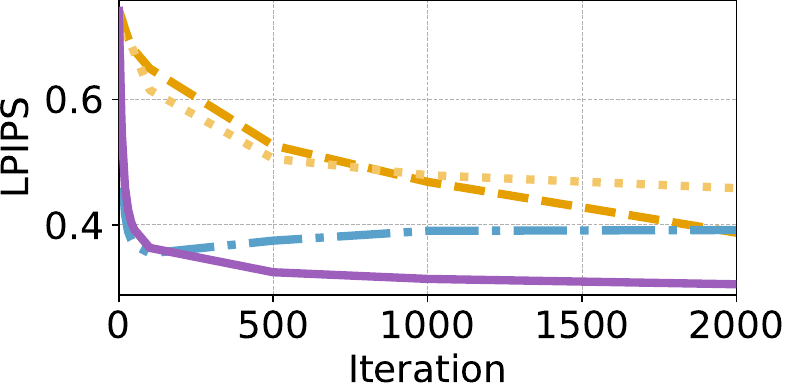}
        \caption{\textbf{Zero-shot}: LLFF, $\sim$ 20 to 60 views, $756{\times}1008$}
        \label{fig:supp-llff_context}
    \end{subfigure}

    \medskip

    \setcounter{rownum}{4}
    \setcounter{subfigure}{0}
    \begin{subfigure}[b]{\linewidth}
        \centering
        \includegraphics[width=0.245\linewidth]{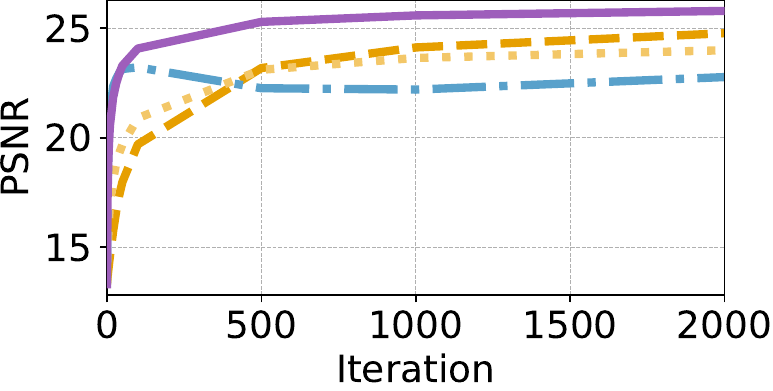}%
        \hfill%
        \includegraphics[width=0.245\linewidth]{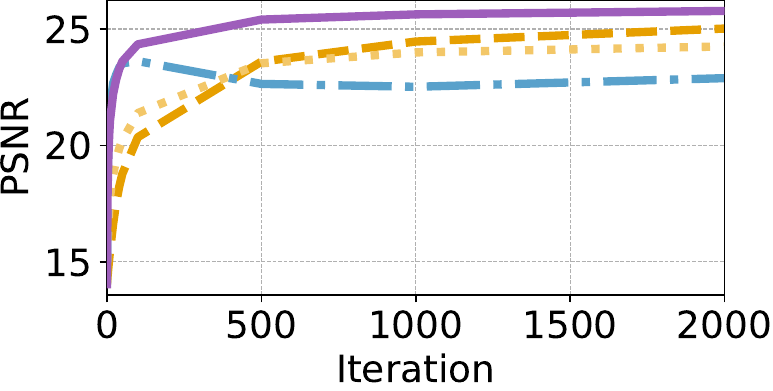}%
        \hfill%
        \includegraphics[width=0.245\linewidth]{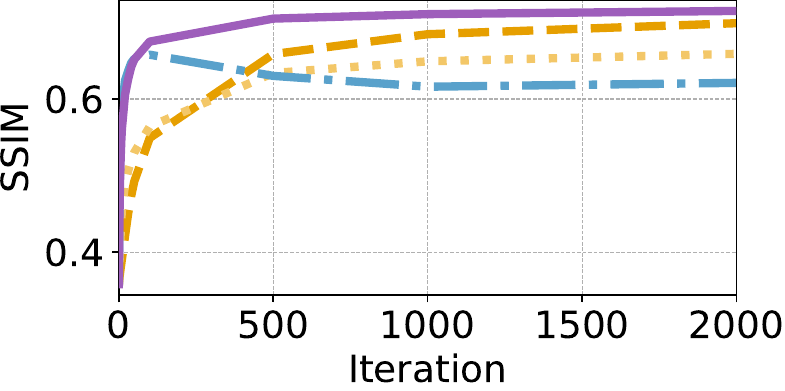}%
        \hfill%
        \includegraphics[width=0.245\linewidth]{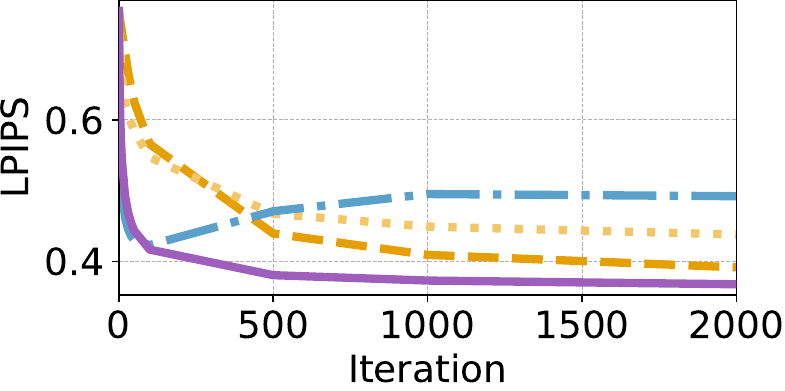}
        \caption{\textbf{Zero-shot}: Mip-NeRF360, 100+ views, $520{\times}780$}
        \label{fig:supp-mip360_context}
    \end{subfigure}

    \caption{\textbf{Optimization trajectories on SfM initialization, dense setting.} Results on DL3DV, DTU, LLFF and Mip-NeRF360. We report PSNR on context views and PSNR, SSIM and LPIPS on target views vs the optimization iteration. See \cref{subsec:supp-results-dense} for discussion.}
    \label{fig:supp-dense-curves}
    \label{fig:supp-dense-metrics}
\end{figure*}

\subsection{Dedicated vs.\ unified checkpoint}
\label{subsec:supp-dedicated-vs-unified}
The main paper reports a single \ourshort{} model trained jointly over both the sparse and dense configurations (\cref{subsec:supp-training-details}). 
Here we compare it against two \emph{dedicated} checkpoints, \oursdense{} and \ourssparse{}, each meta-trained on a single configuration (dense or sparse, respectively; \cref{tab:dedicated-vs-mixed}).
As all three share the same architecture, and therefore the same cost per iteration, we compare them at equal iterations rather than at a shared runtime.
The unified model matches the dedicated checkpoints on all benchmarks within $0.2$~dB in either direction, while surpassing \adam{}.
Applying a dedicated checkpoint outside its training regime degrades quality, though the two directions are not symmetric.
\oursdense{} loses $0.13$ to $0.36$~dB on the sparse benchmarks, whereas \ourssparse{} loses between $7.4$ and $11.9$~dB on the dense ones.
We suspect that \ourssparse{} over-specializes to its training regime, in which the same eight context views are used at every inner iteration (\cref{subsec:supp-training-details}), rather than resampled as in the dense configuration.
Consistent with this, we observed that earlier \ourssparse{} checkpoints degrade less on the dense datasets than later ones.
In contrast, \oursdense{} transfers to the sparse benchmarks with only a small degradation despite being trained with resampled views and evaluated under a different initialization.
Joint training over both configurations provides a single checkpoint that performs well across the two evaluated regimes.

\subsection{Dense setting} \label{subsec:supp-results-dense}
We evaluate the zero-shot generalization of \ourshort{} across different datasets, resolutions, and numbers of views in the dense setting.
We compare learned and non-learned optimizers initialized from SfM, with a variable number of Gaussians per scene.
In all experiments, all methods use a different subset of 8 views at each iteration and run for a fixed runtime, set to \adam{}'s wall-clock of $4000$ iterations.
We consider four datasets: (a)~DL3DV with all views at low resolution (in domain), (b)~DTU with $\sim$50 views at high resolution, (c)~LLFF with $\sim$20 to 60 views at high resolution, and (d)~Mip-NeRF360 with 100+ views at high resolution.

\boldparagraph{Target-view performance.}
Across all datasets, \ourshort{} reaches higher PSNR than \adam{} within the fixed runtime and keeps improving to the end~(\cref{tab:dense-all}).
The largest gap is on LLFF, whose SfM initialization is dominated by the black-Gaussian artifact discussed below.
We additionally report here the performance of \adamtuned{} (\cref{subsec:supp-adam-vs-tuned}), which in the dense setting is below the standard 3DGS \adam{} parameters by $0.6$--$1.8$~dB on three of the four datasets.
DTU is the exception, where \adamtuned{} exceeds \adam{} by $0.35$~dB and \ourshort{} by $0.06$~dB on PSNR, while \ourshort{} remains better on SSIM and LPIPS.

Although each \ourshort{} step costs ${\sim}2\times$ an \adam{} step, it stays above \adam{} at every runtime and passes \adam{}'s final quality before \adam{} reaches it (\cref{fig:supp-timing-dense}).
For \lob{}, which is trained over a short horizon, the learning-rate schedule has to be stretched when applied to a longer run at test time.
The stage of the optimization, the relative time the optimizer is conditioned on and the learning rate that scales its updates are then no longer consistent with one another, so it deviates from the desired optimization behavior beyond its trained range (\cref{subsec:supp-lob}).
This is visible in the trajectories, where \lob{} is briefly ahead of \ourshort{}, then peaks early and degrades by $0.1$--$0.9$~dB, diverging on DTU.

We also report the trajectories against iteration rather than wall-clock, over the first $2000$ iterations common to all methods.
This isolates the update rule itself, as all three optimizers see the same gradients and views at every step and optimize a fixed capacity representation (\cref{fig:supp-dense-curves}).
\ourshort{} keeps improving at every iteration and reaches higher final PSNR than both Adam variants, with large gains in the early stages.

\boldparagraph{Context-view performance.}
The image quality gap between context and target views is small in the dense setting. 
We attribute this to the stronger constraints imposed by dense multi-view supervision, which limit overfitting to the context views.

\boldparagraph{LLFF results comments.}
The LLFF dataset is strongly affected by the black-Gaussian COLMAP artifact described in \cref{subsec:supp-training-details}, as the SfM initialization contains a large number of black Gaussians, and in some scenes almost exclusively so (\cref{fig:llff-comp}).
Although we filter out such points during the DL3DV training, \ourshort{} can recover from these uninformative Gaussians, whereas Adam struggles.
This results in a larger PSNR gap in favor of our method compared to both Adam configurations ($1.7$--$3.6$~dB, \cref{fig:supp-llff_context,tab:supp-llff_dense}).
After recoloring the point clouds by reassigning point colors via projection onto image planes and taking the median color across observations, the final PSNR gap reduces to approximately 0.35 dB.

\subsubsection{Densification and pipeline integration} 
\label{subsec:supp-mcmc}
\begin{table}[t]
\centering
\small
\setlength{\tabcolsep}{4pt}
\renewcommand{\arraystretch}{1.05}
\setlength{\aboverulesep}{0.2ex}\setlength{\belowrulesep}{0.3ex}
\begin{tabular}{@{}l c @{\hspace{6pt}}|@{\hspace{6pt}} r r@{}}
\toprule
 & & \multicolumn{2}{c@{}}{PSNR\,$\uparrow$} \\
\cmidrule(l){3-4}
MCMC & \#G & \multicolumn{1}{c}{\adam{}} & \multicolumn{1}{c@{}}{\ourshort{}} \\
\midrule
w/o & 115k & 25.02\phantom{\,\textsubscript{+0.00}} & \cellcolor{tabsecond}25.80\phantom{\,\textsubscript{+0.00}} \\
w/  & 178k & 25.12\,\textsubscript{\textcolor{black!75}{+0.10}} & \cellcolor{tabfirst}\textbf{25.99}\,\textsubscript{\textcolor{black!75}{+0.19}} \\
\bottomrule
\end{tabular}
\caption{%
\textbf{Integration with the 3DGS-MCMC densification scheme} on Mip-NeRF360.
Subscripts give the gain from enabling MCMC over the corresponding w/o-MCMC row;
\#G is the resulting Gaussian count.
Both optimizers benefit from the MCMC scheme, and \ourshort{} stays ahead of \adam{}
with densification enabled.
\colorbox{tabfirst}{\textbf{Best}} and \colorbox{tabsecond}{second best} PSNR are highlighted.
See \cref{subsec:supp-mcmc} for discussion.%
}
\label{tab:mcmc-adc}
\end{table}

\begin{figure}[!t]
    \centering
    \includegraphics[width=0.85\linewidth]{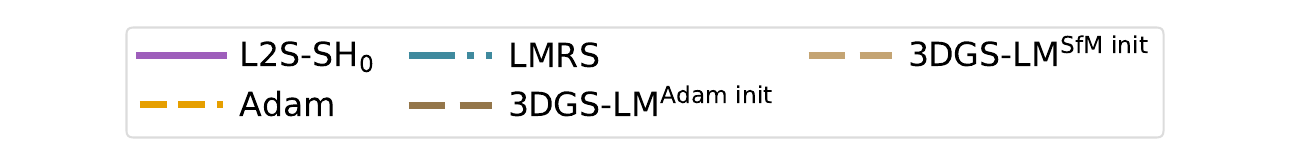}

    \noindent
    \begin{minipage}{0.5\linewidth}\centering{\small\textbf{All nine scenes}}\\[2pt]\includegraphics[width=\linewidth]{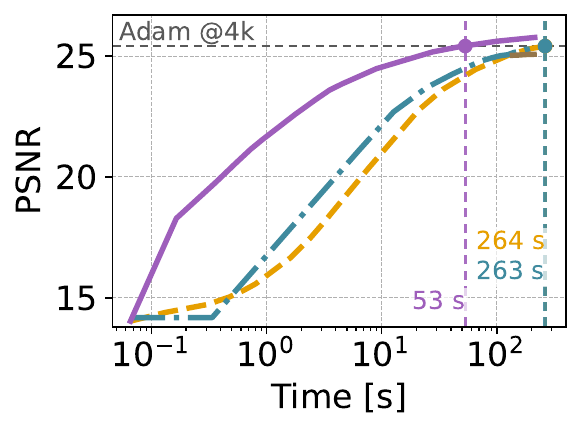}\end{minipage}%
    \begin{minipage}{0.5\linewidth}\centering{\small\textbf{The five outdoor scenes}}\\[2pt]\includegraphics[width=\linewidth]{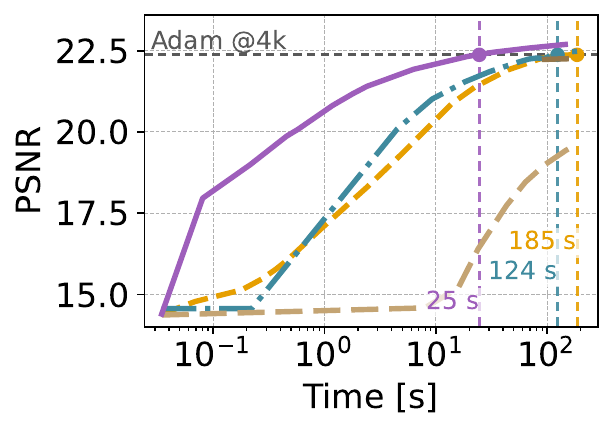}\end{minipage}

    \caption{\textbf{\ourshort{} reaches \adam{}'s quality faster than second-order optimization.}
Scene-averaged PSNR against wall-clock for the runs of \cref{tab:lmrs_budget}.
\textit{\textbf{Left:}} all nine scenes. \textit{\textbf{Right:}} the five outdoor scenes, the scenes \threedgslm{} completes from the SfM initialization.
The dot and label on each run give its time to reach \adam{}'s PSNR at $4$k iterations (dashed line).
\threedgslm{} runs from a $2$k \adam{} run when warm-started and from the SfM initialization otherwise.
See \cref{subsubsec:supp-lmrs} for discussion.}
    \label{fig:supp-lmrs-curves}
\end{figure}

\begin{table}[t]
\centering
\footnotesize
\setlength{\tabcolsep}{1.2pt}
\renewcommand{\arraystretch}{1.05}
\setlength{\aboverulesep}{0.2ex}\setlength{\belowrulesep}{0.3ex}
\begin{tabular}{@{}l @{\hspace{3pt}}|@{\hspace{3pt}} r r r @{\hspace{6pt}}|@{\hspace{6pt}} r @{\hspace{3pt}}|@{\hspace{3pt}} r@{}}
\toprule
{\scriptsize\emph{Optimizer}} & {\scriptsize LPIPS\,$\downarrow$} & {\scriptsize SSIM\,$\uparrow$} & {\scriptsize PSNR\,$\uparrow$} & {\scriptsize Iters} & {\scriptsize Mem\,[GB]\,$\downarrow$} \\
\midrule
\multicolumn{6}{@{}l}{\footnotesize\emph{Second-order}} \\
\arrayrulecolor{gray!40}\midrule\arrayrulecolor{black}
LMRS~\cite{pehlivan2026lmrs} & 0.416 & 0.674 & 25.21 & 238 & 8.2 \\
\threedgslm{}~\cite{hoellein20253dgslm}$^{\text{SfM init}}$ & 0.637 & 0.479 & 20.45 & 21 & 33.1 \\
\threedgslm{}~\cite{hoellein20253dgslm}$^{\text{\adam{}-2k init}}$ & 0.420 & 0.671 & 25.29 & 2k+15 & 18.5 \\
\addlinespace[2pt]
\arrayrulecolor{gray!70}\midrule\arrayrulecolor{black}
\addlinespace[2pt]
\multicolumn{6}{@{}l}{\footnotesize\emph{First-order}} \\
\arrayrulecolor{gray!40}\midrule\arrayrulecolor{black}
\adam{} & \cellcolor{tabsecond}0.404 & \cellcolor{tabsecond}0.680 & \cellcolor{tabsecond}25.58 & 4000 & \cellcolor{tabfirst}\textbf{3.4} \\
\ourshort{}-SH$_0$ & \cellcolor{tabfirst}\textbf{0.389} & \cellcolor{tabfirst}\textbf{0.688} & \cellcolor{tabfirst}\textbf{25.91} & 1978 & \cellcolor{tabsecond}3.7 \\
\bottomrule
\end{tabular}
\caption{\textbf{Reproduction of \cref{tab:lmrs_budget} at a single downscale factor.} Mip-NeRF360 scenes in the SH$_0$ and $L_2$-loss regime of LMRS, with all scenes downscaled by $4$ (same as \cref{tab:main-results}), rather than the Mip-NeRF360 protocol of $4$ for outdoor scenes and $2$ for indoors (\cref{tab:lmrs_budget}). Columns follow \cref{tab:main-results}. For \threedgslm{}, \emph{Iters} counts the refinement steps taken on top of its 2k-\adam{} warm start. At this resolution \threedgslm{} no longer exceeds memory from the SfM initialization, and the trends of \cref{tab:lmrs_budget} otherwise hold. \colorbox{tabfirst}{\textbf{Best}} and \colorbox{tabsecond}{second best} values are highlighted. See \cref{subsubsec:supp-lmrs} for discussion.}
\label{tab:lmrs_budget_r4}
\end{table}

\begin{table}[t]
\centering
\footnotesize
\setlength{\tabcolsep}{1.2pt}
\renewcommand{\arraystretch}{1.05}
\setlength{\aboverulesep}{0.2ex}\setlength{\belowrulesep}{0.3ex}
\begin{tabular}{@{}l @{\hspace{3pt}}|@{\hspace{3pt}} r r r @{\hspace{3pt}}|@{\hspace{3pt}} l @{\hspace{3pt}}|@{\hspace{3pt}} r@{}}
\toprule
{\scriptsize\emph{Optimizer}} & {\scriptsize PSNR\,$\uparrow$} & {\scriptsize SSIM\,$\uparrow$} & {\scriptsize LPIPS\,$\downarrow$} & {\scriptsize Time\,[s]\,$\downarrow$} & {\scriptsize Mem\,$\downarrow$} \\
\midrule
\shortstack[l]{\adam{}\\{\scriptsize\textcolor{black!55}{(1, 10k)}}} & 28.44 & 0.864 & \cellcolor{tabfirst}\textbf{0.271} & 300 & \cellcolor{tabsecond}7.9 \\
\arrayrulecolor{gray!40}\midrule\arrayrulecolor{black}
\shortstack[l]{LMRS~\cite{pehlivan2026lmrs}\\{\scriptsize\textcolor{black!55}{(16/32, 200--280)}}} & \cellcolor{tabsecond}28.76 & \cellcolor{tabsecond}0.868 & \cellcolor{tabthird}0.333 & \cellcolor{tabthird}205 & 17.4 \\
\arrayrulecolor{gray!40}\midrule\arrayrulecolor{black}
\shortstack[l]{\threedgslm{}~\cite{hoellein20253dgslm}$^{\text{SfM init}}$\\{\scriptsize\textcolor{black!55}{($25{\times}4$, 100)}}} & \multicolumn{5}{c}{{\color{black!30}\leavevmode\leaders\hrule height 0.55ex depth -0.45ex\hfill\kern0pt}~\emph{out of memory} on 4/4 scenes ($>$80\,GB)~{\color{black!30}\leavevmode\leaders\hrule height 0.55ex depth -0.45ex\hfill\kern0pt}} \\
\arrayrulecolor{gray!40}\midrule\arrayrulecolor{black}
\shortstack[l]{\threedgslm{}~\cite{hoellein20253dgslm}$^{\text{\adam{} init}}$\\{\scriptsize\textcolor{black!55}{($25{\times}4$, 10k+5/10)}}} & \cellcolor{tabthird}28.59 & \cellcolor{tabthird}0.865 & 0.334 & 293 & 40.4 \\
\arrayrulecolor{gray!40}\midrule\arrayrulecolor{black}
\shortstack[l]{\ourshort{}-SH$_0$\\{\scriptsize\textcolor{black!55}{(16, 1k)}}} & \cellcolor{tabfirst}\textbf{29.40} & \cellcolor{tabfirst}\textbf{0.876} & \cellcolor{tabsecond}0.316 & \cellcolor{tabsecond}139 & \cellcolor{tabthird}9.4 \\
\bottomrule
\end{tabular}
\caption{\textbf{Reproduction of LMRS's Table~1} (SH$_0$, $L_2$-loss) on the \textbf{four indoor scenes}, against their reported single-view $10$k \adam{} baseline.
LMRS reaches that baseline's peak PSNR $1.9\times$ faster and \ourshort{}-SH$_0$ $9.6\times$ faster.
Each row is labelled with the views per iteration and the iteration count it was run at.
\threedgslm{}'s $25{\times}4$ denotes the four sets of $25$ views each of its steps solves over.
PSNR, SSIM and LPIPS are reported at each method's iteration count, with \emph{Time} the corresponding runtime.
\emph{Mem} reports mean peak GPU memory (GB).
\colorbox{tabfirst}{\textbf{Best}}, \colorbox{tabsecond}{second best}, and \colorbox{tabthird}{third best} values are highlighted.
See \cref{subsubsec:supp-lmrs} for discussion.}
\label{tab:lmrs_indoor}
\end{table}

\begin{table}[t]
\centering
\footnotesize
\setlength{\tabcolsep}{1.2pt}
\renewcommand{\arraystretch}{1.05}
\setlength{\aboverulesep}{0.2ex}\setlength{\belowrulesep}{0.3ex}
\begin{tabular}{@{}c @{\hspace{5pt}} c @{\hspace{6pt}} l @{\hspace{4pt}} l @{\hspace{3pt}}|@{\hspace{3pt}} r r r @{\hspace{4pt}} r @{\hspace{3pt}}|@{\hspace{3pt}} r@{}}
\toprule
{\scriptsize Views} & {\scriptsize Iters} & {\scriptsize\emph{Optimizer}} & {\scriptsize ID} & {\scriptsize PSNR\,$\uparrow$} & {\scriptsize SSIM\,$\uparrow$} & {\scriptsize LPIPS\,$\downarrow$} & {\scriptsize Time\,[s]\,$\downarrow$} & {\scriptsize Mem\,$\downarrow$} \\
\midrule
 & \multirow{2}{*}{{\footnotesize 0}} & \adam{} & {\footnotesize\texttt{A0}} & 14.04 & 0.388 & 0.754 & 0 & \multicolumn{1}{c}{--} \\
 &  & \ourshort{}-SH$_0$ & {\footnotesize\texttt{L0}} & 14.04 & 0.388 & 0.754 & 0 & \multicolumn{1}{c}{--} \\
\addlinespace[2pt]
\arrayrulecolor{gray!70}\midrule\arrayrulecolor{black}
\addlinespace[2pt]
{\footnotesize 1} & {\footnotesize 10k} & \adam{} & {\footnotesize\texttt{A1}} & 25.05 & 0.667 & \cellcolor{tabfirst}\textbf{0.407} & 244 & \cellcolor{tabfirst}\textbf{5.6} \\
\addlinespace[2pt]
\arrayrulecolor{gray!70}\midrule\arrayrulecolor{black}
\addlinespace[2pt]
\multirow{4}{*}{{\footnotesize 16}} & \multirow{2}{*}{{\footnotesize 2k}} & \adam{} & {\footnotesize\texttt{A2}} & 25.06 & 0.660 & 0.464 & \cellcolor{tabfirst}\textbf{130} & \cellcolor{tabsecond}6.3 \\
 &  & \ourshort{}-SH$_0$ & {\footnotesize\texttt{L2}} & \cellcolor{tabfirst}\textbf{25.77} & \cellcolor{tabthird}0.682 & 0.432 & 214 & 6.5 \\
\arrayrulecolor{gray!40}\cmidrule(lr){2-4}\arrayrulecolor{black}
 & {\footnotesize 4k} & \adam{} & {\footnotesize\texttt{A3}} & 25.31 & 0.669 & 0.453 & \cellcolor{tabthird}195 & \cellcolor{tabsecond}6.3 \\
\arrayrulecolor{gray!40}\cmidrule(lr){2-4}\arrayrulecolor{black}
 & {\footnotesize 10k} & \adam{} & {\footnotesize\texttt{A4}} & \cellcolor{tabsecond}25.69 & \cellcolor{tabsecond}0.687 & \cellcolor{tabthird}0.429 & 703 & \cellcolor{tabsecond}6.3 \\
\addlinespace[2pt]
\arrayrulecolor{gray!70}\midrule\arrayrulecolor{black}
\addlinespace[2pt]
\multirow{3}{*}{{\footnotesize 16/32}} & {\footnotesize 0} & \multirow{3}{*}{\shortstack[l]{LMRS\\\cite{pehlivan2026lmrs}}} & {\footnotesize\texttt{LM0}} & 14.18 & 0.389 & 0.754 & 0 & \multirow{3}{*}{14.2} \\
 & {\footnotesize 300} &  & {\footnotesize\texttt{LM1}} & 25.38 & 0.675 & 0.445 & 228 &  \\
 & {\footnotesize 1000} &  & {\footnotesize\texttt{LM2}} & \cellcolor{tabthird}25.68 & \cellcolor{tabfirst}\textbf{0.689} & \cellcolor{tabsecond}0.427 & 891 &  \\
\addlinespace[2pt]
\arrayrulecolor{gray!70}\midrule\arrayrulecolor{black}
\addlinespace[2pt]
\multirow{4}{*}{{\footnotesize $25{\times}4$}} & {\footnotesize +0} & \multirow{4}{*}{\shortstack[l]{\threedgslm{}\\\cite{hoellein20253dgslm}\\\adam{} init}} & {\footnotesize\texttt{GL0}} & 25.03 & 0.663 & 0.462 & 130 & \multirow{4}{*}{33.9} \\
 & {\footnotesize +5} &  & {\footnotesize\texttt{GL1}} & 25.07 & 0.664 & 0.462 & \cellcolor{tabsecond}182 &  \\
 & {\footnotesize +20} &  & {\footnotesize\texttt{GL2}} & 25.10 & 0.664 & 0.461 & 337 &  \\
 & {\footnotesize +100} &  & {\footnotesize\texttt{GL3}} & 25.16 & 0.666 & 0.460 & 1156 &  \\
\bottomrule
\end{tabular}
\caption{\textbf{Batch size and iteration count sweep.}
Nine Mip-NeRF360 scenes in LMRS's SH$_0$ and $L_2$-loss regime, with \adam{}, \ourshort{} and LMRS from a shared SfM initialization.
\adam{} at $16$ views and $2$k iterations (\texttt{A2}) reaches the PSNR of LMRS's reported single-view $10$k baseline (\texttt{A1}).
All metrics, including \emph{Time}, are reported at each row's \emph{Iters}.
\threedgslm{}'s \emph{Views} of $25{\times}4$ denotes the four sets of $25$ views each of its steps solves over.
\threedgslm{}'s \emph{Iters} counts only its second-order steps (``$+N$'') on top of a $2$k, $16$-view \adam{} warm start, which \texttt{GL0} reports.
\colorbox{tabfirst}{\textbf{Best}}, \colorbox{tabsecond}{second best}, and \colorbox{tabthird}{third best} values are highlighted.
See \cref{subsubsec:supp-lmrs} for discussion.}
\label{tab:lmrs_supp}
\end{table}

The experiments in \cref{tab:main-results} use the standard 3DGS training pipeline implemented in gsplat, with the standard $\ell_1+\mathrm{D-SSIM}$ loss and adaptive density control (ADC) disabled, so that optimizers are compared at a fixed scene capacity.
Since learned optimizers specialize to the optimization dynamics of their training pipeline, changes to the renderer, losses, update rule, or Gaussian count can induce a distribution shift in the gradients observed by the optimizer.
We therefore evaluate whether the learned optimizer remains effective when applied within 3DGS-MCMC~\cite{kheradmand20243dgsmcmc}, which differs from our training setup through stochastic updates, opacity regularization, relocation of dead Gaussians, and densification.

\cref{tab:mcmc-adc} reports results on Mip-NeRF360 with and without the MCMC scheme enabled, comparing against \adam{} under the same pipeline.
Despite never being trained under these modified optimization dynamics, \ourshort{} remains effective, outperforming \adam{} both without MCMC ($+0.78$ dB) and with MCMC enabled ($+0.87$ dB).
Moreover, \ourshort{} benefits from densification rather than being destabilized by it, improving by $+0.19$ dB as the Gaussian count increases from $115$k to $178$k, compared to $+0.10$ dB for \adam{}.
For these densification experiments, we recompute the $k$-nearest-neighbor graph at every iteration rather than using the lazy every-$100$-iteration update employed elsewhere at test time (\cref{sec:app-impl}), ensuring that neighborhoods remain accurate as Gaussians are added.

We view these results as an initial demonstration of robustness to pipeline changes.
Achieving full adaptivity to densification, and to the Gaussian distributions that arise at higher primitive counts, likely requires meta-training the optimizer jointly with the densification scheme while respecting the kNN-attention budget (\cref{sec:app-impl}).
We leave this direction to future work.

\subsubsection{Comparison to second-order optimization}
\label{subsubsec:supp-lmrs}

\begin{table*}[t]
  \centering
  \setlength{\tabcolsep}{2pt}

  \begin{subtable}[t]{0.49\linewidth}
    \centering
    \centering
{\footnotesize%
\setlength{\tabcolsep}{0.7pt}%
\renewcommand{\arraystretch}{1.05}%
\setlength{\aboverulesep}{0.2ex}\setlength{\belowrulesep}{0.3ex}%
\begin{tabular}{@{}l @{\hspace{1.5pt}}|@{\hspace{1.5pt}} r r r @{\hspace{1.5pt}}|@{\hspace{1.5pt}} r r @{\hspace{1.5pt}}|@{\hspace{1.5pt}} r@{}}
\toprule
 & LPIPS\,$\downarrow$ & SSIM\,$\uparrow$ & PSNR\,$\uparrow$ & PSNR$_{\tbud}$\,$\uparrow$ & Iters & Mem\,$\downarrow$ \\
\midrule
\multicolumn{7}{@{}l}{\emph{Initializer (feed forward)}} \\
\arrayrulecolor{gray!40}\midrule\arrayrulecolor{black}
MVSplat~\cite{chen2024mvsplat} & 0.170 & 0.820 & 24.16 & -- & -- & -- \\
DepthSplat~\cite{xu2025depthsplat} & 0.130 & 0.860 & 25.81 & -- & -- & -- \\
ReSplat~\cite{xu2025resplat} & 0.140 & 0.850 & 25.96 & -- & -- & -- \\
WorldMirror~\cite{liu2025worldmirror} & 0.180 & 0.790 & 23.54 & -- & -- & -- \\
\addlinespace[2pt]
\arrayrulecolor{gray!70}\midrule\arrayrulecolor{black}
\addlinespace[2pt]
\multicolumn{7}{@{}l}{\emph{Optimizer}} \\
\arrayrulecolor{gray!40}\midrule\arrayrulecolor{black}
\adam{} & 0.097 & 0.909 & 30.02 & 29.91 & 4000 & \cellcolor{tabfirst}\textbf{0.6} \\
\adamtuned{} & \cellcolor{tabfirst}\textbf{0.090} & 0.915 & 30.29 & \cellcolor{tabsecond}30.07 & 4000 & \cellcolor{tabfirst}\textbf{0.6} \\
ReSplat~\cite{xu2025resplat} & 0.144 & 0.854 & 26.59 & 5.99 & 1252 & 1.5 \\
\lob{} & 0.093 & \cellcolor{tabsecond}0.916 & \cellcolor{tabsecond}30.34 & 27.93 & 1688 & 0.9 \\
\ourshort{} & \cellcolor{tabfirst}\textbf{0.090} & \cellcolor{tabfirst}\textbf{0.917} & \cellcolor{tabfirst}\textbf{30.53} & \cellcolor{tabfirst}\textbf{30.53} & 1924 & 0.9 \\
\bottomrule
\end{tabular}%
}

    \caption{\textbf{In-domain:} DL3DV, 8 views, $256 \times 448$.}
    \label{tab:supp-dl3dv_8views_480p}
  \end{subtable}
  \hfill
  \begin{subtable}[t]{0.49\linewidth}
    \centering
    \centering
{\footnotesize%
\setlength{\tabcolsep}{0.7pt}%
\renewcommand{\arraystretch}{1.05}%
\setlength{\aboverulesep}{0.2ex}\setlength{\belowrulesep}{0.3ex}%
\begin{tabular}{@{}l @{\hspace{1.5pt}}|@{\hspace{1.5pt}} r r r @{\hspace{1.5pt}}|@{\hspace{1.5pt}} r r @{\hspace{1.5pt}}|@{\hspace{1.5pt}} r@{}}
\toprule
 & LPIPS\,$\downarrow$ & SSIM\,$\uparrow$ & PSNR\,$\uparrow$ & PSNR$_{\tbud}$\,$\uparrow$ & Iters & Mem\,$\downarrow$ \\
\midrule
\multicolumn{7}{@{}l}{\emph{Initializer (feed forward)}} \\
\arrayrulecolor{gray!40}\midrule\arrayrulecolor{black}
MVSplat~\cite{chen2024mvsplat} & 0.460 & 0.530 & 16.42 & -- & -- & -- \\
DepthSplat~\cite{xu2025depthsplat} & 0.430 & 0.550 & 16.83 & -- & -- & -- \\
ReSplat~\cite{xu2025resplat} & 0.260 & 0.710 & 20.28 & -- & -- & -- \\
WorldMirror~\cite{liu2025worldmirror} & OOM & OOM & OOM & -- & -- & -- \\
\addlinespace[2pt]
\arrayrulecolor{gray!70}\midrule\arrayrulecolor{black}
\addlinespace[2pt]
\multicolumn{7}{@{}l}{\emph{Optimizer}} \\
\arrayrulecolor{gray!40}\midrule\arrayrulecolor{black}
\adam{} & 0.103 & 0.904 & 29.78 & 29.70 & 4000 & \cellcolor{tabfirst}\textbf{4.1} \\
\adamtuned{} & \cellcolor{tabsecond}0.096 & \cellcolor{tabsecond}0.912 & \cellcolor{tabsecond}30.02 & \cellcolor{tabsecond}29.79 & 5343 & \cellcolor{tabfirst}\textbf{4.1} \\
ReSplat~\cite{xu2025resplat} & 0.258 & 0.743 & 21.33 & 5.51 & 1797 & 12.0 \\
\lob{} & 0.100 & 0.911 & 29.86 & 26.75 & 2279 & 4.2 \\
\ourshort{} & \cellcolor{tabfirst}\textbf{0.095} & \cellcolor{tabfirst}\textbf{0.913} & \cellcolor{tabfirst}\textbf{30.26} & \cellcolor{tabfirst}\textbf{30.26} & 2334 & 4.2 \\
\bottomrule
\end{tabular}%
}

    \caption{\textbf{Zero-shot:} DL3DV, 32 views, $256 \times 448$.}
    \label{tab:supp-dl3dv_32views_480p}
  \end{subtable}

  \vspace{10pt}  %

  \begin{subtable}[t]{0.49\linewidth}
    \centering
    \centering
{\footnotesize%
\setlength{\tabcolsep}{0.7pt}%
\renewcommand{\arraystretch}{1.05}%
\setlength{\aboverulesep}{0.2ex}\setlength{\belowrulesep}{0.3ex}%
\begin{tabular}{@{}l @{\hspace{1.5pt}}|@{\hspace{1.5pt}} r r r @{\hspace{1.5pt}}|@{\hspace{1.5pt}} r r @{\hspace{1.5pt}}|@{\hspace{1.5pt}} r@{}}
\toprule
 & LPIPS\,$\downarrow$ & SSIM\,$\uparrow$ & PSNR\,$\uparrow$ & PSNR$_{\tbud}$\,$\uparrow$ & Iters & Mem\,$\downarrow$ \\
\midrule
\multicolumn{7}{@{}l}{\emph{Initializer (feed forward)}} \\
\arrayrulecolor{gray!40}\midrule\arrayrulecolor{black}
MVSplat~\cite{chen2024mvsplat} & 0.420 & 0.630 & 19.01 & -- & -- & -- \\
DepthSplat~\cite{xu2025depthsplat} & 0.350 & 0.670 & 20.25 & -- & -- & -- \\
ReSplat~\cite{xu2025resplat} & 0.370 & 0.620 & 19.19 & -- & -- & -- \\
WorldMirror~\cite{liu2025worldmirror} & 0.400 & 0.510 & 17.60 & -- & -- & -- \\
\addlinespace[2pt]
\arrayrulecolor{gray!70}\midrule\arrayrulecolor{black}
\addlinespace[2pt]
\multicolumn{7}{@{}l}{\emph{Optimizer}} \\
\arrayrulecolor{gray!40}\midrule\arrayrulecolor{black}
\adam{} & 0.197 & 0.837 & 26.60 & \cellcolor{tabsecond}26.32 & 4000 & \cellcolor{tabfirst}\textbf{1.8} \\
\adamtuned{} & 0.193 & 0.845 & 26.72 & 26.20 & 5709 & \cellcolor{tabfirst}\textbf{1.8} \\
ReSplat~\cite{xu2025resplat} & 0.325 & 0.722 & 22.08 & \multicolumn{1}{c}{DIV} & \multicolumn{1}{c}{--} & 20.2 \\
\lob{} & \cellcolor{tabsecond}0.192 & \cellcolor{tabfirst}\textbf{0.846} & \cellcolor{tabsecond}26.99 & 23.83 & 1697 & 3.0 \\
\ourshort{} & \cellcolor{tabfirst}\textbf{0.185} & \cellcolor{tabfirst}\textbf{0.846} & \cellcolor{tabfirst}\textbf{27.16} & \cellcolor{tabfirst}\textbf{27.16} & 2063 & 3.0 \\
\bottomrule
\end{tabular}%
}

    \caption{\textbf{Zero-shot:} DL3DV, 8 views, $512 \times 960$.}
    \label{tab:supp-dl3dv_8views_960p}
  \end{subtable}
  \hfill
  \begin{subtable}[t]{0.49\linewidth}
    \centering
    \centering
{\footnotesize%
\setlength{\tabcolsep}{0.7pt}%
\renewcommand{\arraystretch}{1.05}%
\setlength{\aboverulesep}{0.2ex}\setlength{\belowrulesep}{0.3ex}%
\begin{tabular}{@{}l @{\hspace{1.5pt}}|@{\hspace{1.5pt}} r r r @{\hspace{1.5pt}}|@{\hspace{1.5pt}} r r @{\hspace{1.5pt}}|@{\hspace{1.5pt}} r@{}}
\toprule
 & LPIPS\,$\downarrow$ & SSIM\,$\uparrow$ & PSNR\,$\uparrow$ & PSNR$_{\tbud}$\,$\uparrow$ & Iters & Mem\,$\downarrow$ \\
\midrule
\multicolumn{7}{@{}l}{\emph{Initializer (feed forward)}} \\
\arrayrulecolor{gray!40}\midrule\arrayrulecolor{black}
ReSplat~\cite{xu2025resplat} & 0.385 & 0.689 & 18.40 & -- & -- & -- \\
\addlinespace[2pt]
\arrayrulecolor{gray!70}\midrule\arrayrulecolor{black}
\addlinespace[2pt]
\multicolumn{7}{@{}l}{\emph{Optimizer}} \\
\arrayrulecolor{gray!40}\midrule\arrayrulecolor{black}
\adam{} & 0.223 & 0.844 & 25.57 & 25.26 & 4000 & \cellcolor{tabfirst}\textbf{1.8} \\
\adamtuned{} & \cellcolor{tabfirst}\textbf{0.213} & \cellcolor{tabfirst}\textbf{0.858} & \cellcolor{tabsecond}26.02 & \cellcolor{tabsecond}25.55 & 5318 & \cellcolor{tabfirst}\textbf{1.8} \\
ReSplat~\cite{xu2025resplat} & 0.338 & 0.745 & 19.90 & \multicolumn{1}{c}{DIV} & \multicolumn{1}{c}{--} & 20.2 \\
\lob{} & \cellcolor{tabsecond}0.214 & \cellcolor{tabsecond}0.855 & 26.01 & \multicolumn{1}{c}{DIV} & \multicolumn{1}{c}{--} & 3.0 \\
\ourshort{} & \cellcolor{tabsecond}0.214 & 0.852 & \cellcolor{tabfirst}\textbf{26.08} & \cellcolor{tabfirst}\textbf{26.08} & 1818 & 3.0 \\
\bottomrule
\end{tabular}%
}

    \caption{\textbf{Zero-shot:} RealEstate10K, 8 views, $512 \times 960$.}
    \label{tab:supp-re10k_8views_960p}
  \end{subtable}
  
  \caption{%
    \textbf{Quantitative evaluation in the sparse setting.}
    Results on DL3DV~\cite{ling2024dl3dv} and RealEstate10K~\cite{Zhou2018SIGGRAPH} datasets, using the ReSplat initialization.
    Columns follow \cref{tab:main-results}.
    Every optimizer is given a fixed runtime $\tbud$, set to \adam{}'s wall-clock of $4000$ iterations, and runs as many of its own iterations as fit within it.
    PSNR, SSIM and LPIPS are reported at the best-PSNR iteration within that runtime, and PSNR$_{\tbud}$ at its last step (\emph{Iters}).
    The \emph{Initializer} rows report feed-forward predictions before any optimization.
    The \emph{Optimizer} rows refine the ReSplat initialization.
    \emph{OOM} marks an initializer whose prediction did not fit in GPU memory.
    \emph{DIV} marks an optimizer whose Gaussian parameters diverged to infinite values, with no quality to report beyond that point.
    \adamtuned{} uses the learning rates grid-searched on sparse scenes (\cref{subsec:supp-adam-vs-tuned}).
    \ourshort{} initializes per-Gaussian latent state vectors using ReSplat's pixel-aligned features.
    \textbf{In-domain} results match the sparse-view configuration of our training, which covers both sparse and dense views (dataset, number of views, and resolution), while \textbf{zero-shot} results differ in at least one of these factors.
    \protect\colorbox{tabfirst}{\textbf{Best}} and \protect\colorbox{tabsecond}{second best} values are highlighted.
    See \cref{subsec:supp-results-sparse} for discussion.
  }
  \label{tab:sparse-all}
\end{table*}

We further discuss here the details of the second-order comparison, including the memory cost, batch size, and iteration count of each method.
\cref{fig:supp-lmrs-curves} plots the trajectories of \cref{tab:lmrs_budget} for all scenes and the outdoor scenes only (where \threedgslm{} fits the memory from the SfM initialization).
\ourshort{} reaches \adam{}'s $4000$ quality in $53$~s, against $264$~s for \adam{} itself and $263$~s for LMRS, while \threedgslm{} never reaches it.

\threedgslm{} stores the Jacobian of the per-pixel rendering residuals explicitly, a cost that grows both with the number of Gaussians and with the pixels each one covers.
In the shared SfM initialization, Gaussians are larger, each covering more pixels.
Hence, at Mip-NeRF360's default resolution (outdoor scenes at $1/4$, indoor at $1/2$), \threedgslm{} exceeds memory on all four indoor scenes ($>$$80$~GB, \cref{tab:lmrs_budget}).
\cref{tab:lmrs_budget_r4} reproduces this comparison at a uniform $1/4$ resolution (same as \cref{tab:main-results}), at which \threedgslm{} fits in $33.1$~GB.
Within \adam{}'s runtime it fits $21$ steps and reaches $20.5$~dB, $5.1$~dB below \adam{}, at roughly nine times \ourshort{}'s memory.
Running it for $100$ iterations, at $4.8\times$ the fixed runtime, it reaches $22.4$~dB, still $3.2$~dB below \adam{}.
Warm-started from \adam{}, its memory reduces to $18.5$~GB.
At this resolution, the warm start enables only $15$ refinement steps, and it ends $0.3$~dB below \adam{} within the same runtime.
Among the remaining methods the trends hold, and the gap between \ourshort{} and LMRS widens to $0.7$~dB.

\cref{tab:lmrs_indoor} isolates the four indoor Mip-NeRF360 scenes, reproducing LMRS's Table~1~\cite{pehlivan2026lmrs} (under the SH$_0$ and $L_2$-loss regime).
Their reported \adam{} baseline uses a batch size of one, which we reproduce here and further discuss in \cref{tab:lmrs_supp}.
We run each baseline at the iteration counts that LMRS uses, and \emph{Time} reports the corresponding runtime.
We report \ourshort{} at $1$k iterations, where it spends less wall-clock than LMRS ($139$ against $205$~s) and remains $0.6$~dB ahead, at $9.4$ against $17.4$~GB.
\threedgslm{}, warm-started from \adam{}, fits in $40.4$~GB and its released protocol of $5$ refinement steps, $10$ on counter and kitchen, adds $0.21$~dB over its own \adam{} run, ending $0.8$~dB below \ourshort{} while using over twice its runtime and four times its memory.

\cref{tab:lmrs_supp} varies the batch size and iteration counts of each method on the nine Mip-NeRF360 scenes.
LMRS's reported \adam{} baseline (\texttt{A1}) reaches only $25.0$\,dB at $10$k steps, but uses a single view.
LMRS itself uses a batch size of $16$ views, increased to $32$ after the $50$th iteration~\cite{pehlivan2026lmrs}.
Running \adam{} with a matched $16$ views reaches the same PSNR within $2000$ iterations (\texttt{A2}), $25.3$ at the $4000$ (\texttt{A3}), and $25.7$ by $10$k (\texttt{A4}).
LMRS keeps improving, reaching $25.7$~dB at $1000$ steps in $891$~s against \adam{}'s $703$ (\texttt{LM2}).
\ourshort{}-SH$_0$ (\texttt{L2}) exceeds both at $2$k iterations, reaching $25.77$~dB in $214$~s.

We run LMRS and \threedgslm{} in their authors' implementations, so both render with a different rasterizer than ours.
The starting points still match, with LMRS within ${\sim}0.1$~dB of ours from the shared SfM initialization (\texttt{LM0} against \texttt{A0}) and \threedgslm{}'s \adam{} warm start within $0.03$\,dB (\texttt{GL0} against \texttt{A2}).
Its warm start, \adam{} at $2000$ iterations (\texttt{A2}), matches in PSNR the $10$k \adam{} run \threedgslm{} starts from in LMRS's Table~1 (\texttt{A1}).
\threedgslm{}'s $100$ refinement steps, taken on top of that warm start and beyond the fixed runtime, add $0.13$\,dB (\texttt{GL3}), almost all of it in indoor scenes ($0.21$\,dB against $0.06$\,dB on the outdoor scenes).
At its released protocol of $5$ steps (\texttt{GL1}) the gain is $0.04$\,dB, comparable to the gap between the two warm starts.

\begin{figure}[!tb]
    \centering
    \captionsetup[subfigure]{labelformat=simple}
    \renewcommand\thesubfigure{(\alph{subfigure})}

    \includegraphics[width=\linewidth]{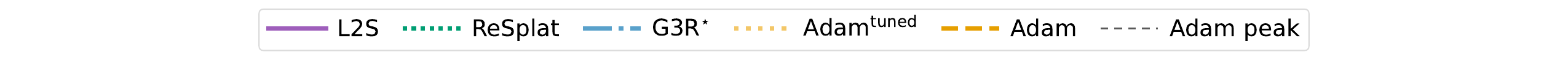}

    \medskip

    \begin{subfigure}[b]{0.49\linewidth}
        \centering
        \includegraphics[width=\linewidth, trim=0 0 0 0, clip]{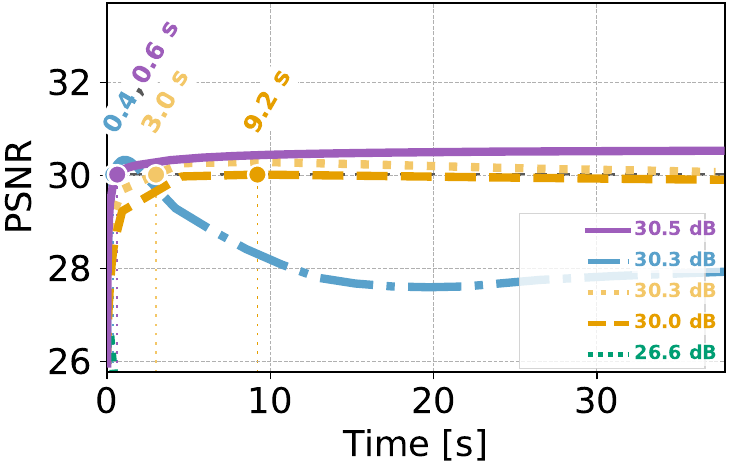}
        \caption{DL3DV~\cite{ling2024dl3dv}, 8 views, $256 \times 448$.}
        \label{fig:supp-timing_dl3dv_8_low_res}
    \end{subfigure}
    \hfill
    \begin{subfigure}[b]{0.49\linewidth}
        \centering
        \includegraphics[width=\linewidth, trim=0 0 0 0, clip]{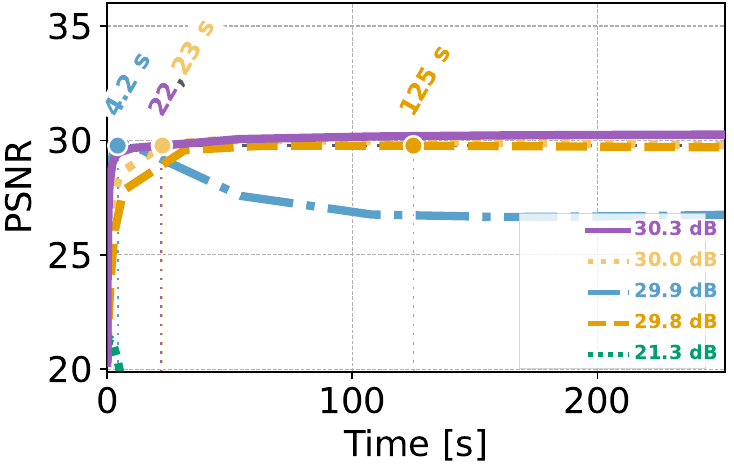}
        \caption{DL3DV~\cite{ling2024dl3dv}, 32 views, $256 \times 448$.}
        \label{fig:supp-timing_dl3dv_32_low_res}
    \end{subfigure}

    \medskip

    \begin{subfigure}[b]{0.49\linewidth}
        \centering
        \includegraphics[width=\linewidth, trim=0 0 0 0, clip]{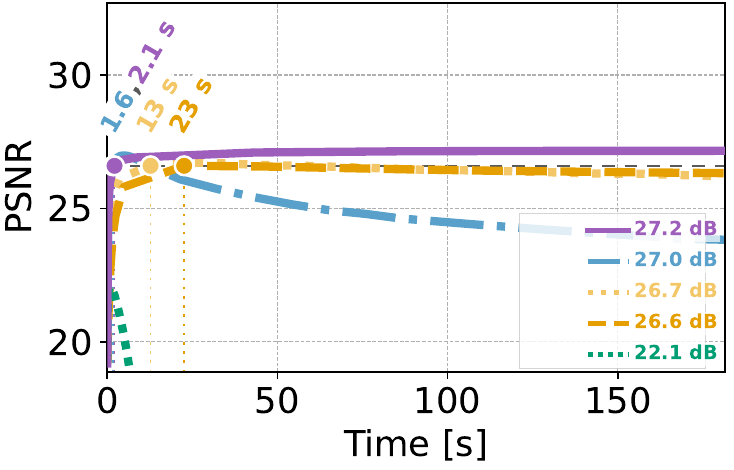}
        \caption{DL3DV~\cite{ling2024dl3dv}, 8 views, $512 \times 960$.}
        \label{fig:supp-timing_dl3dv_8_high_res}
    \end{subfigure}
    \hfill
    \begin{subfigure}[b]{0.49\linewidth}
        \centering
        \includegraphics[width=\linewidth, trim=0 0 0 0, clip]{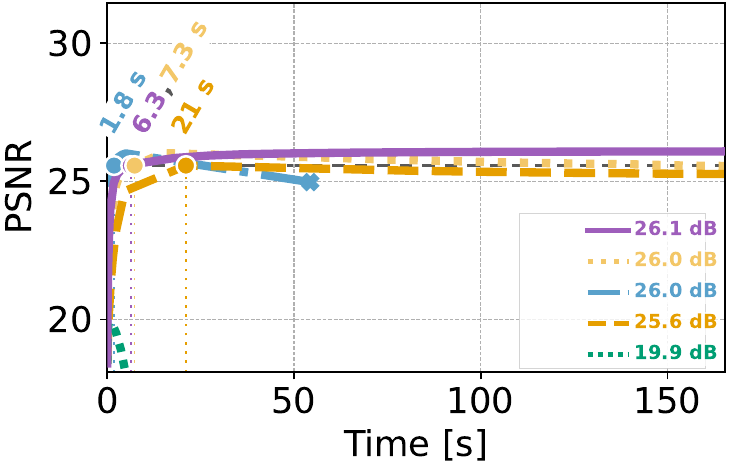}
        \caption{RealEstate10K~\cite{Zhou2018SIGGRAPH}, 8 views, $512 \times 960$.}
        \label{fig:supp-timing-re10k_8_high_res}
    \end{subfigure}

    \caption{\textbf{Optimization against wall-clock time, sparse setting.} PSNR averaged across scenes, plotted against elapsed optimization time. Dots and labels mark where each run reaches \adam{}'s peak PSNR, and each panel lists the methods' best PSNR. A cross marks where a run's Gaussian parameters diverged to non-finite values. \ourshort{} reaches every quality level faster than both \adam{} variants and ends higher, despite a more expensive update. See \cref{subsec:supp-results-sparse} for discussion.}
    \label{fig:supp-timing-sparse}
\end{figure}

\begin{figure*}[!t]
    \centering
    \includegraphics[width=0.56\linewidth]{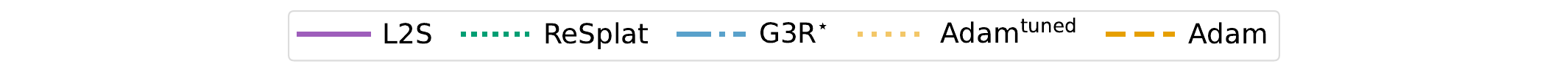}

    \medskip

    \makebox[0.245\linewidth][c]{\shortstack{\textbf{Context Views}\\[3pt]\rule{0.23\linewidth}{0.4pt}}}%
    \hfill%
    \makebox[0.748\linewidth][c]{\shortstack{\textbf{Target Views}\\[3pt]\rule{0.735\linewidth}{0.4pt}}}

    \smallskip

    \noindent
    \newcommand{\tinier}{\fontsize{7.5}{7.5}\selectfont}
    \makebox[0.245\linewidth][c]{\tinier PSNR $\uparrow$}%
    \hfill%
    \makebox[0.245\linewidth][c]{\tinier PSNR $\uparrow$}%
    \hfill%
    \makebox[0.245\linewidth][c]{\tinier SSIM $\uparrow$}%
    \hfill%
    \makebox[0.245\linewidth][c]{\tinier LPIPS $\downarrow$}

    \setcounter{subfigure}{0}
    \renewcommand{\thesubfigure}{\alph{rownum}}

    \setcounter{rownum}{1}
    \setcounter{subfigure}{0}
    \begin{subfigure}[b]{\linewidth}
        \centering
        \includegraphics[width=0.245\linewidth]{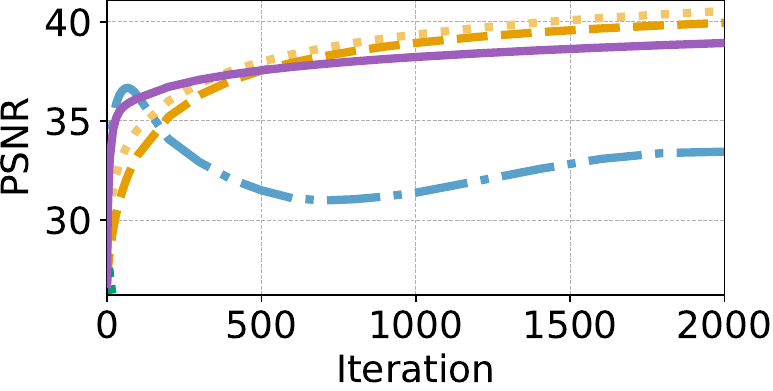}%
        \hfill%
        \includegraphics[width=0.245\linewidth]{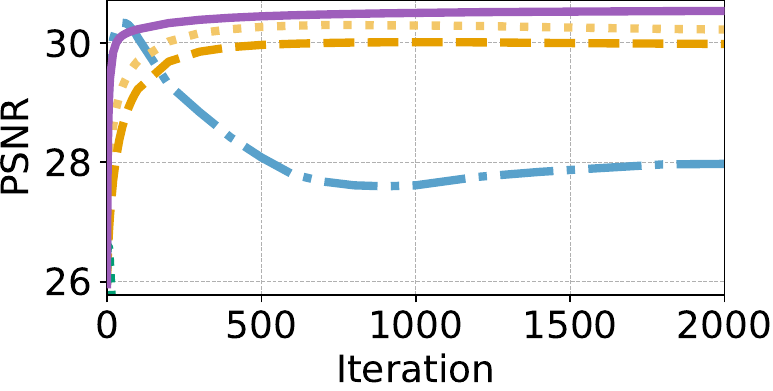}%
        \hfill%
        \includegraphics[width=0.245\linewidth]{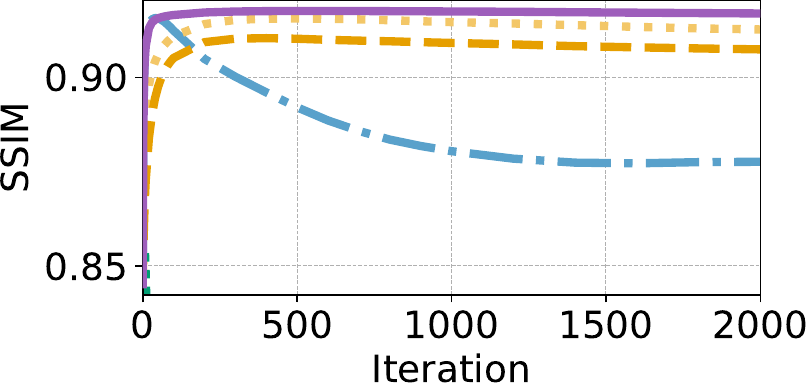}%
        \hfill%
        \includegraphics[width=0.245\linewidth]{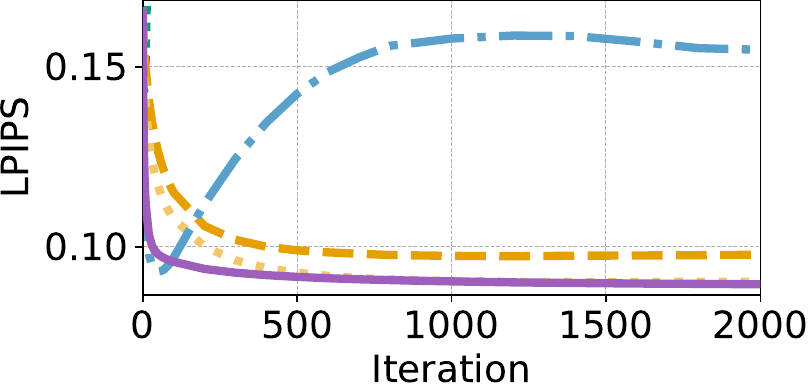}
        \caption{\textbf{In-domain}: DL3DV, 8 views, $256{\times}448$}
        \label{fig:supp-dl3dv_8_low_res_context}
    \end{subfigure}

    \medskip

    \setcounter{rownum}{2}
    \setcounter{subfigure}{0}
    \begin{subfigure}[b]{\linewidth}
        \centering
        \includegraphics[width=0.245\linewidth]{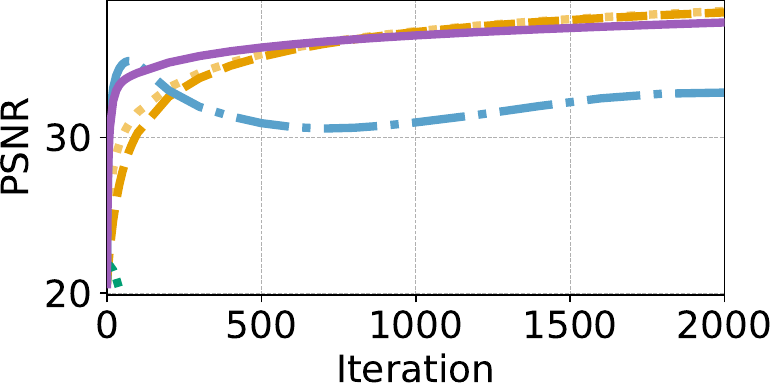}%
        \hfill%
        \includegraphics[width=0.245\linewidth]{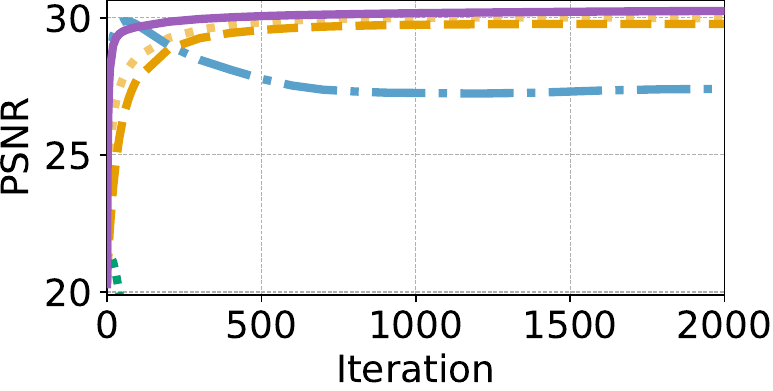}%
        \hfill%
        \includegraphics[width=0.245\linewidth]{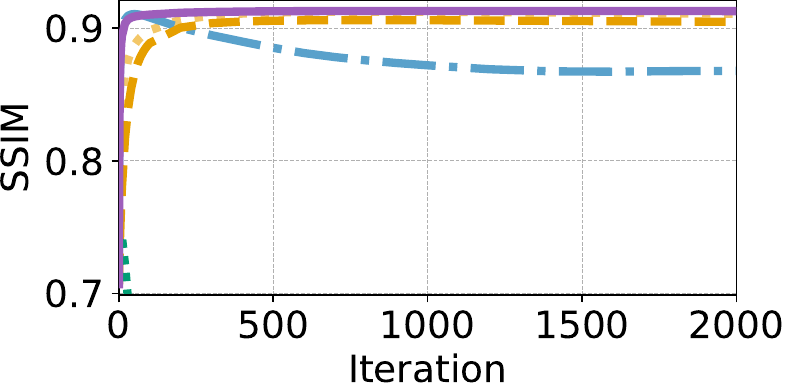}%
        \hfill%
        \includegraphics[width=0.245\linewidth]{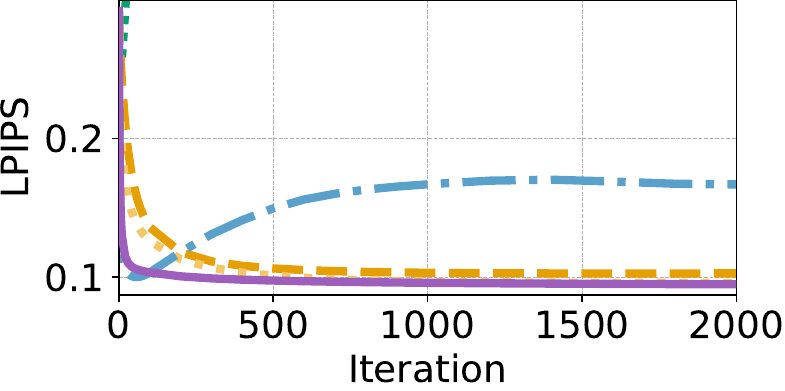}
        \caption{\textbf{Zero-shot}: DL3DV, 32 views, $256{\times}448$}
        \label{fig:supp-dl3dv_32_low_res_context}
    \end{subfigure}

    \medskip

    \setcounter{rownum}{3}
    \setcounter{subfigure}{0}
    \begin{subfigure}[b]{\linewidth}
        \centering
        \includegraphics[width=0.245\linewidth]{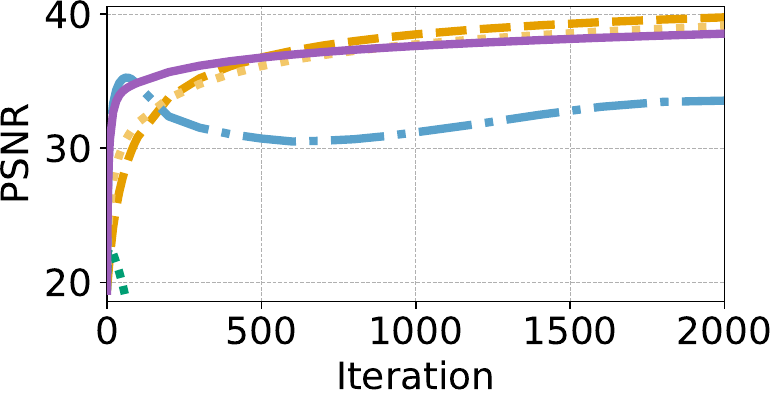}%
        \hfill%
        \includegraphics[width=0.245\linewidth]{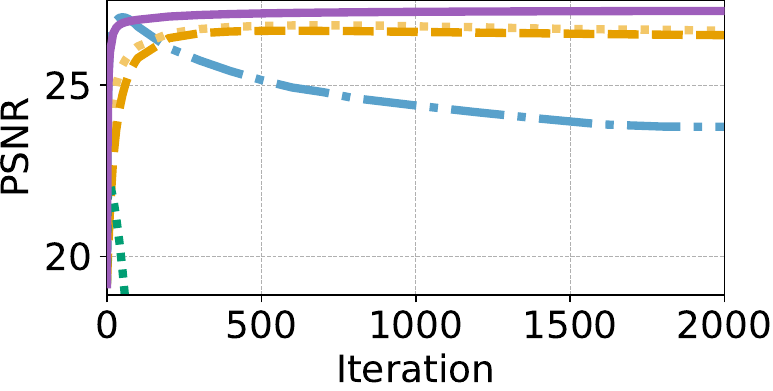}%
        \hfill%
        \includegraphics[width=0.245\linewidth]{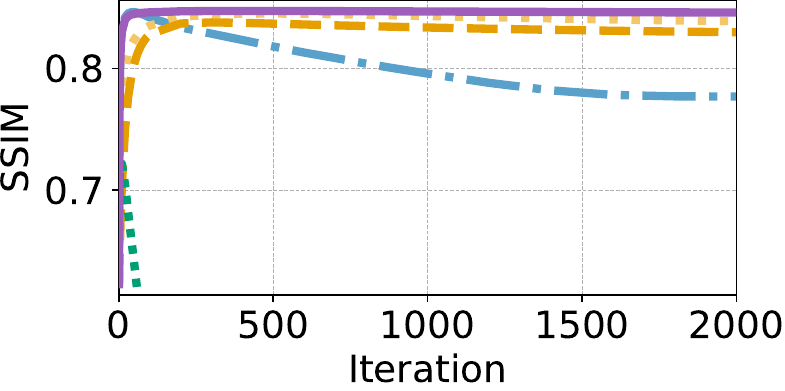}%
        \hfill%
        \includegraphics[width=0.245\linewidth]{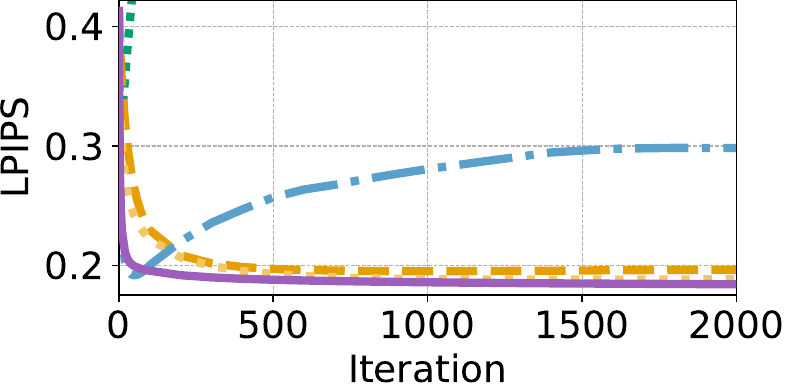}
        \caption{\textbf{Zero-shot}: DL3DV, 8 views, $512{\times}960$}
        \label{fig:supp-dl3dv_8_high_res_context}
    \end{subfigure}

    \medskip

    \setcounter{rownum}{4}
    \setcounter{subfigure}{0}
    \begin{subfigure}[b]{\linewidth}
        \centering
        \includegraphics[width=0.245\linewidth]{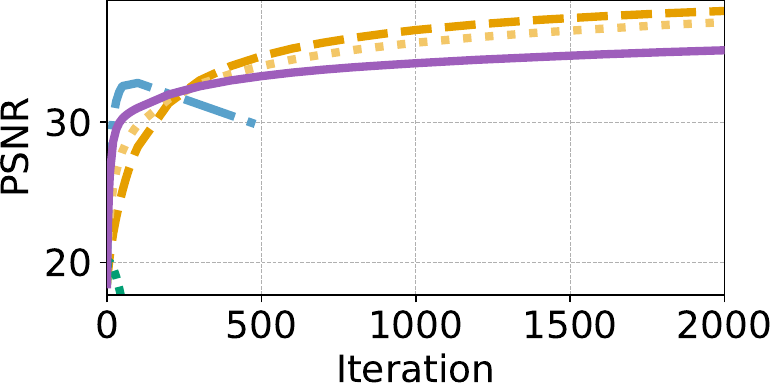}%
        \hfill%
        \includegraphics[width=0.245\linewidth]{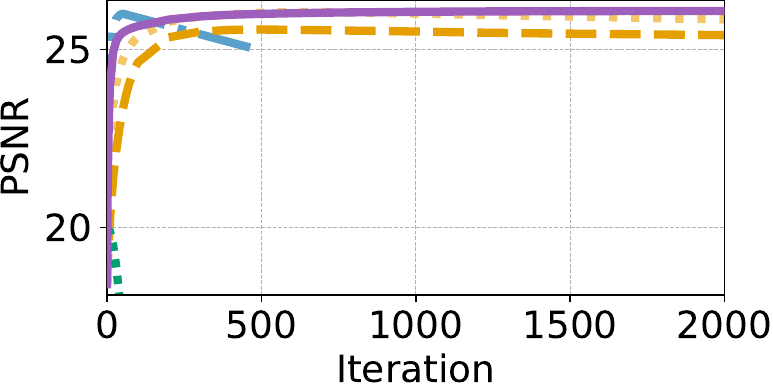}%
        \hfill%
        \includegraphics[width=0.245\linewidth]{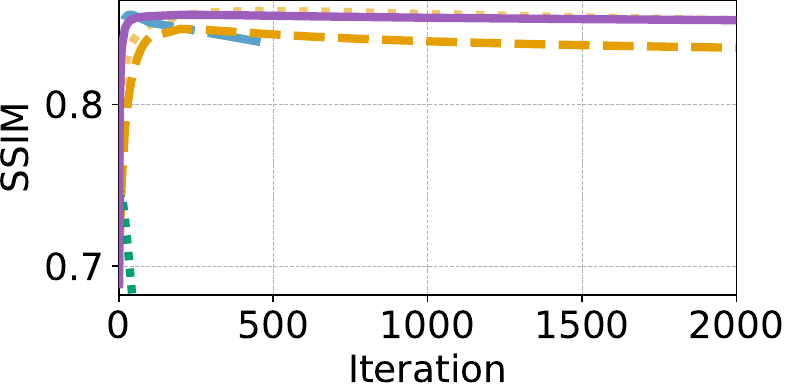}%
        \hfill%
        \includegraphics[width=0.245\linewidth]{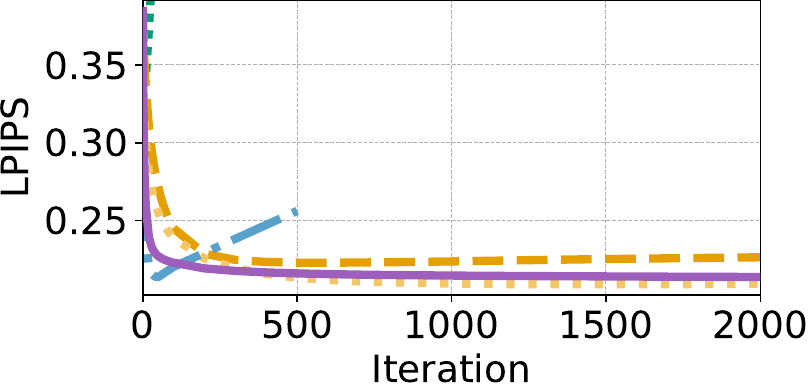}
        \caption{\textbf{Zero-shot}: RealEstate10K, 8 views, $512{\times}960$}
        \label{fig:supp-re10k_context}
    \end{subfigure}

    \caption{\textbf{Optimization trajectories on ReSplat initialization, sparse setting.} Results on DL3DV and RealEstate10K. We report PSNR on context views and PSNR, SSIM and LPIPS on target views vs the optimization iteration. On RealEstate10K, \lob{} diverges on most scenes after roughly 100 iterations; its curve ends where the scene-averaged metric becomes undefined. See \cref{subsec:supp-results-sparse} for discussion.}
    \label{fig:supp-sparse-curves}
    \label{fig:supp-sparse-metrics}
\end{figure*}

\subsection{Sparse setting} \label{subsec:supp-results-sparse}
We further evaluate the zero-shot generalization of \ourshort{} in the sparse-view setting across different datasets, resolutions, and numbers of views, following the evaluation in the dense setting (\cref{subsec:supp-results-dense}).
We compare learned and non-learned optimizers initialized from the feed-forward predictions of ReSplat~\cite{xu2025resplat}.
In all experiments, all methods use all available views at each iteration and are compared within a fixed runtime, set to \adam{}'s wall-clock of $4000$ iterations.
We consider four configurations: (a)~DL3DV with 8 views at low resolution (in domain), (b)~DL3DV with 32 views at low resolution, (c)~DL3DV with 8 views at high resolution, and (d)~RealEstate10K with 8 views at high resolution.

\boldparagraph{Target-view performance.}
Across all configurations, \ourshort{} reaches higher PSNR than \adam{} within the fixed runtime and is still improving to the end~(\cref{tab:sparse-all}).
\adam{} instead ends $0.1$--$0.3$~dB below its peak.
The gap is most pronounced in the high-resolution DL3DV setting.
\adamtuned{}, which was tuned on the sparse scenes, exceeds \adam{} by $0.1$--$0.5$~dB on every configuration.
Still, \ourshort{} improves over it by $0.2$--$0.4$~dB on DL3DV and matches it on RealEstate10K.

Despite the higher per-iteration cost, \ourshort{} stays above both \adam{} variants at every runtime (\cref{fig:supp-timing-sparse}).
RealEstate10K is the exception, where \adamtuned{} reaches \ourshort{}'s quality sooner and stays ahead over the first $35$~s.
It then degrades by $0.2$~dB, while \ourshort{} keeps improving to the end of the run.
\lob{} again improves faster over the first seconds, then degrades by $2.4$--$3.2$~dB on DL3DV and by $1.0$~dB on RealEstate10K until diverging to infinite values.

Per iteration, \ourshort{} again improves at every step and reaches higher final PSNR than both Adam variants, with large gains early (\cref{fig:supp-sparse-curves}).
The same trends hold for SSIM and LPIPS, except again on RealEstate10K, where \adamtuned{} is slightly above \ourshort{} on SSIM from roughly iteration $300$ to $2000$ and ends marginally ahead on LPIPS ($0.209$ against $0.214$).
ReSplat improves during the very early iterations but subsequently degrades.
Note that ReSplat was trained on the sparse configuration used to train \ourshort{} (low resolution, 8 views) and is therefore also evaluated in a zero-shot manner.

\begin{figure}[t]
    \centering
    \begin{subfigure}{0.49\linewidth}
        \centering
        \includegraphics[width=\linewidth]{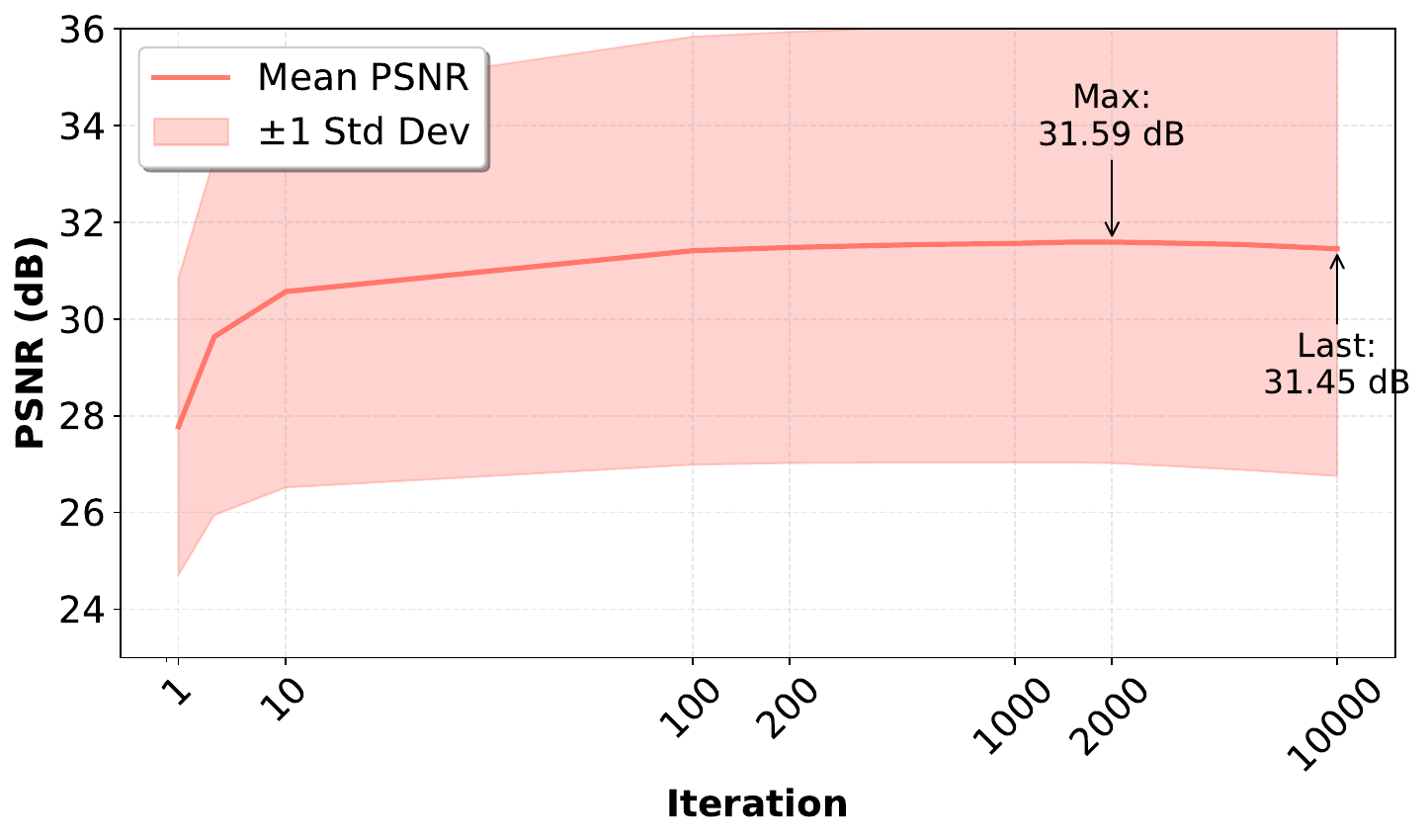}
        \caption{\ourssparse{} (Testing)}
        \label{fig:ours_psnr_test}
    \end{subfigure}
    \begin{subfigure}{0.49\linewidth}
        \centering
        \includegraphics[width=\linewidth]{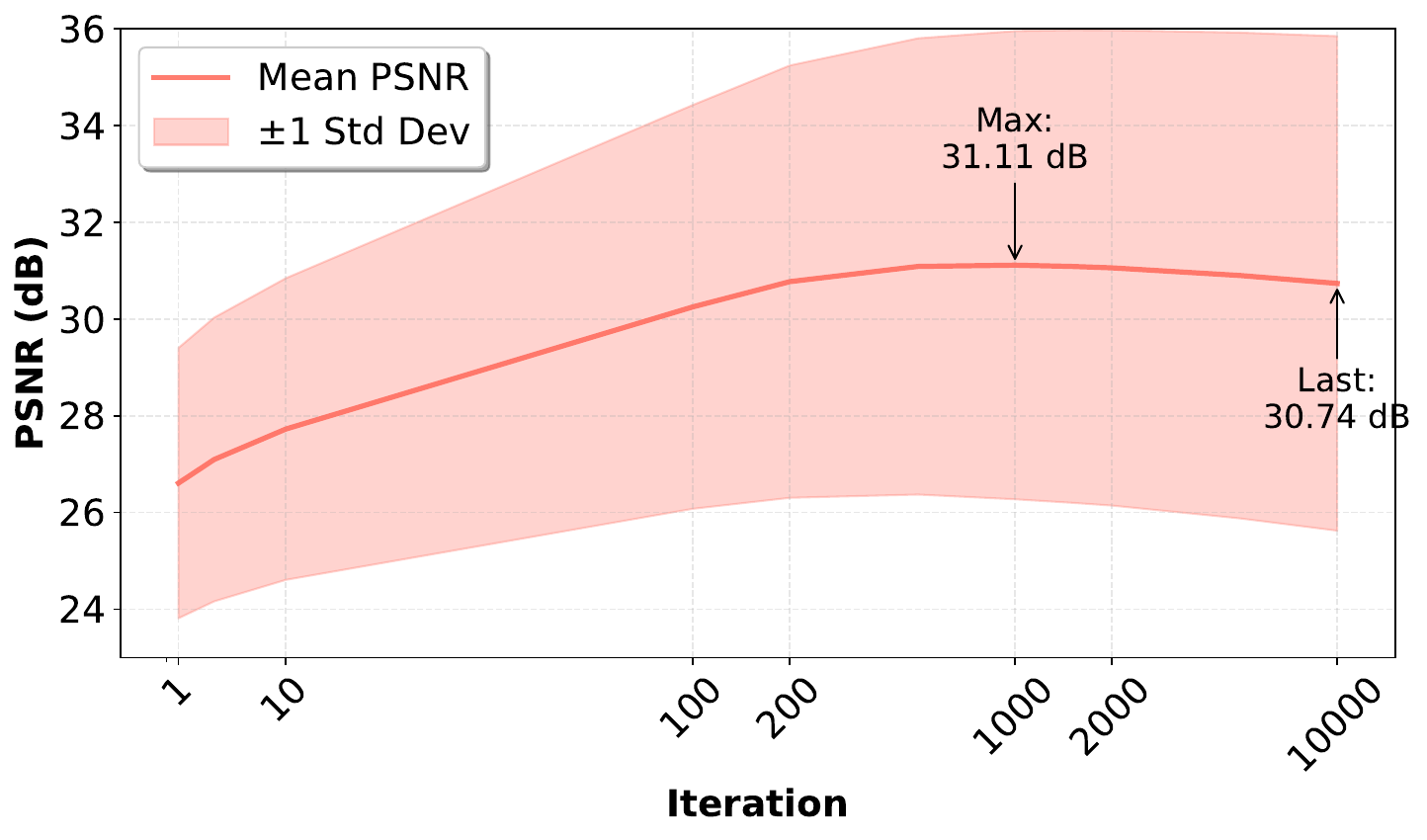}
        \caption{\adam{} (Testing)}
        \label{fig:3dgs_psnr_test}
    \end{subfigure}

    \vspace{1em} %

    \begin{subfigure}{0.49\linewidth}
        \centering
        \includegraphics[width=\linewidth]{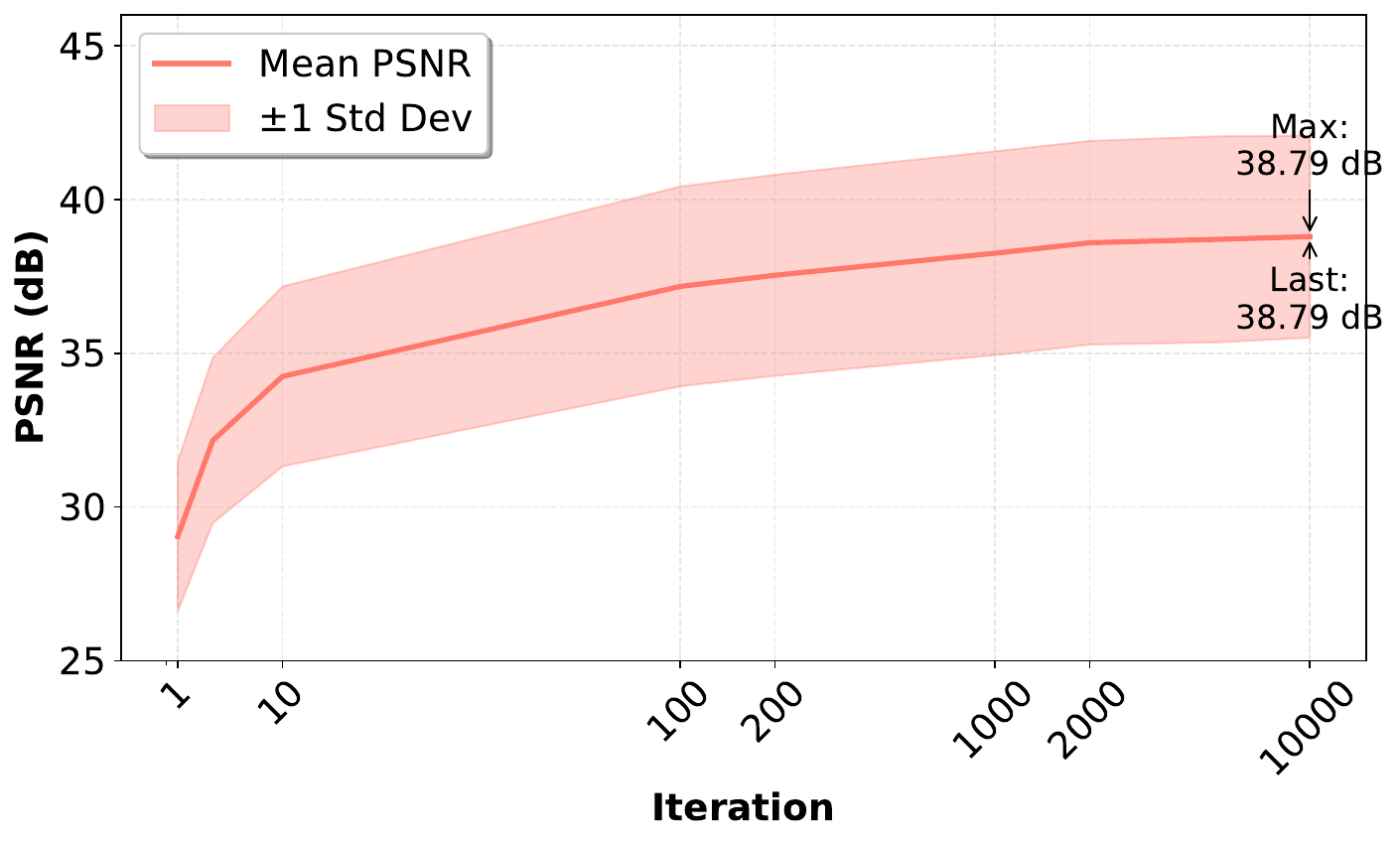}
        \caption{\ourssparse{} (Training)}
        \label{fig:ours_psnr_train}
    \end{subfigure}
    \begin{subfigure}{0.49\linewidth}
        \centering
        \includegraphics[width=\linewidth]{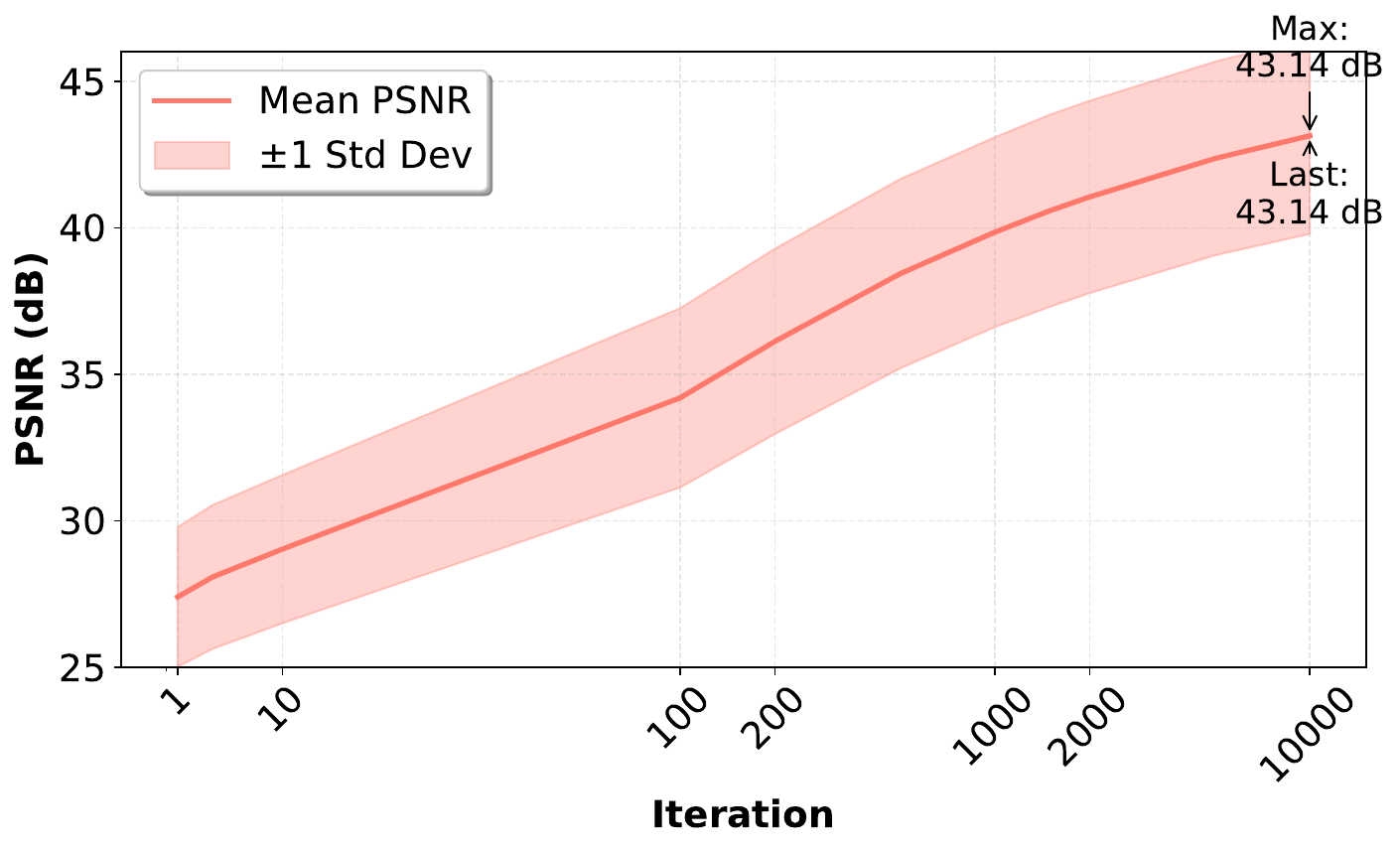}
        \caption{\adam{} (Training)}
        \label{fig:3dgs_psnr_train}
    \end{subfigure}

    \caption{\textbf{PSNR comparison over a $10{,}000$-iteration run.} PSNR against optimization iteration on the testing (top row) and training (bottom row) views, for \ourssparse{} (left) and \adam{} (right).
    Values are computed over 10 scenes from the DL3DV test set in the sparse low-resolution setting (8 views, $256 \times 448$ resolution).
    See \cref{subsec:supp-results-sparse} for discussion.}
    \label{fig:psnr_comparison}
\end{figure}

\boldparagraph{Context-view performance.}
The behavior on context views (left column of \cref{fig:supp-sparse-curves}) reveals a clear difference between learned and non-learned optimization strategies.
For the Adam variants, PSNR on the context views continues to increase even after the performance on the target views has saturated, suggesting overfitting to the training views.
In contrast, \ourshort{} exhibits more moderate improvements on the context views.
This behavior is consistent with the meta-training objective, which includes a loss on the \emph{target views} and therefore discourages overfitting to the context observations.
We illustrate this by optimizing 10 scenes from the DL3DV test set in the sparse, low-resolution setup for 10,000 iterations, as shown in \cref{fig:psnr_comparison}.
This experiment uses the dedicated sparse checkpoint \ourssparse{}, as do the analyses of \cref{sec:app-analysis}.
While Adam continues to improve performance on the training views, its testing performance degrades by approximately 0.37 dB.
In contrast, \ourssparse{} maintains a small improvement on the training set while exhibiting a smaller degradation on the testing set (0.14 dB).
Overall, our method outperforms Adam by approximately 0.4 dB after iteration 1,000, with a larger gap earlier in the optimization.

\subsection{Qualitative results}
\label{subsec:supp-qualitative}
We show reconstructions over the course of optimization for representative Mip-NeRF360 and DTU (\cref{fig:mipnerf-dtu-comp}), LLFF (\cref{fig:llff-comp}), DL3DV (\cref{fig:dl3dv-comp}) and RealEstate10K (\cref{fig:re10k-comp}) scenes.

\afterpage{
    \clearpage

\begin{figure*}[p]
    \centering

    \resizebox{1.0\linewidth}{!}{%
    \setlength{\tabcolsep}{0.002\linewidth} %
    \renewcommand{\arraystretch}{0} %

    \begin{tabular}{@{}%
        >{\centering\arraybackslash}m{0.03\linewidth}
        >{\centering\arraybackslash}m{0.24\linewidth}
        >{\centering\arraybackslash}m{0.24\linewidth}
        >{\centering\arraybackslash}m{0.24\linewidth}
        >{\centering\arraybackslash}m{0.24\linewidth}
        >{\centering\arraybackslash}m{0.24\linewidth}
        >{\centering\arraybackslash}m{0.02\linewidth}
        >{\centering\arraybackslash}m{0.24\linewidth}
    @{}}
         & $t = 2$ & $t = 10$ & $t = 100$ & $t = 1000$ & $t = 2000$ &  &  \\[3pt]

        \rotatebox{90}{\makecell{\smaller[1]{\lob{}}}} &
        \pcell[width=\linewidth]{19.4}{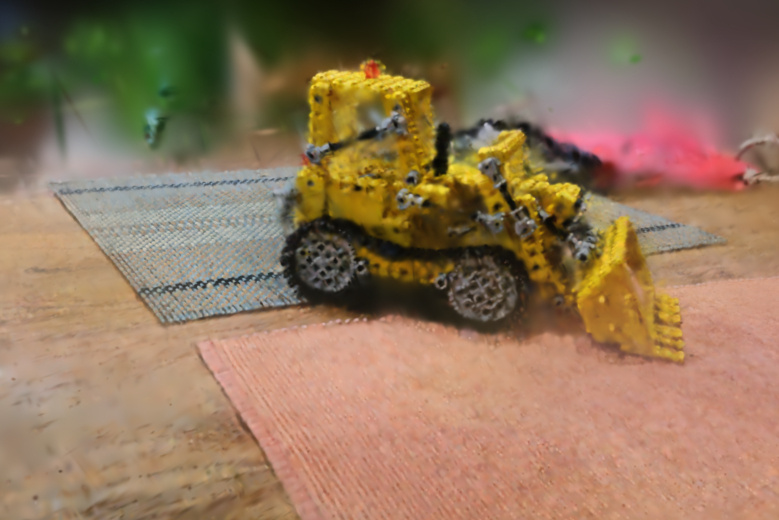} &
        \pcell[width=\linewidth]{23.9}{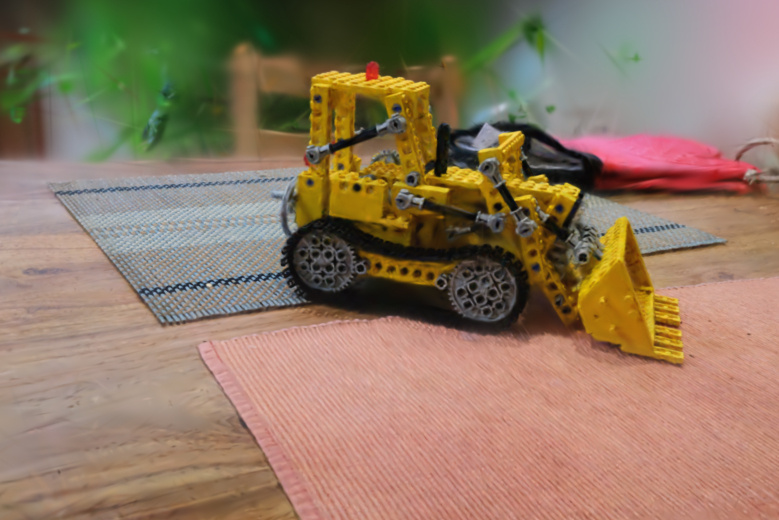} &
        \pcell[width=\linewidth]{26.8}{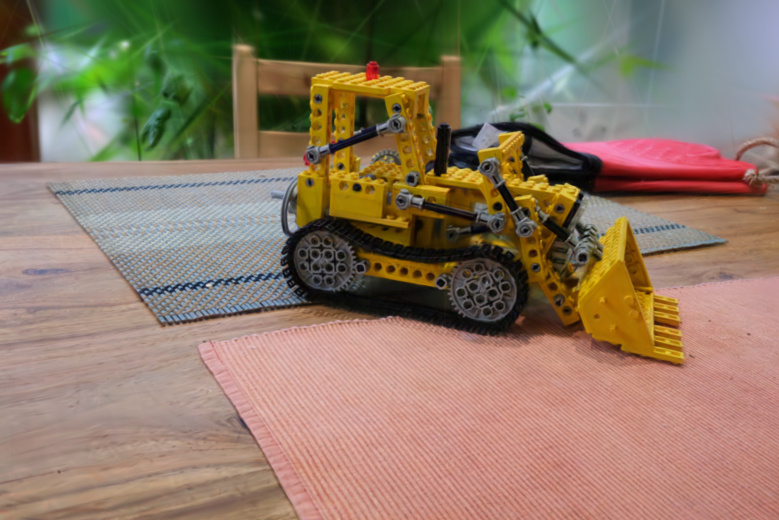} &
        \pcell[width=\linewidth]{25.2}{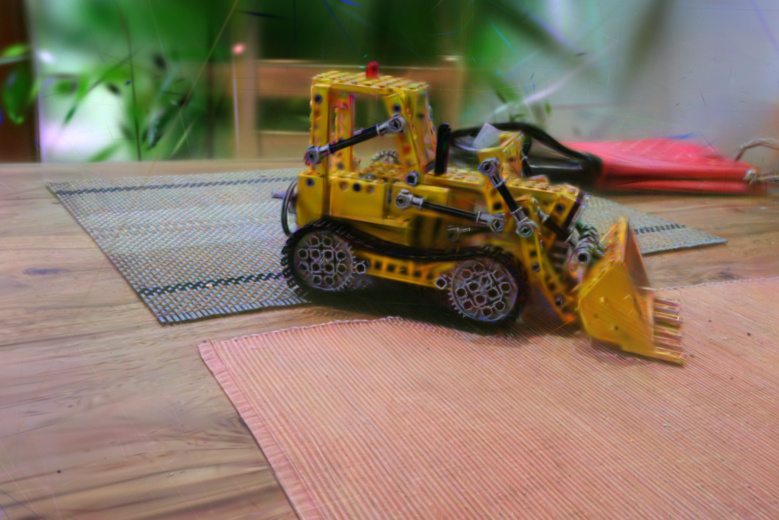} &
        \pcell[width=\linewidth]{25.9}{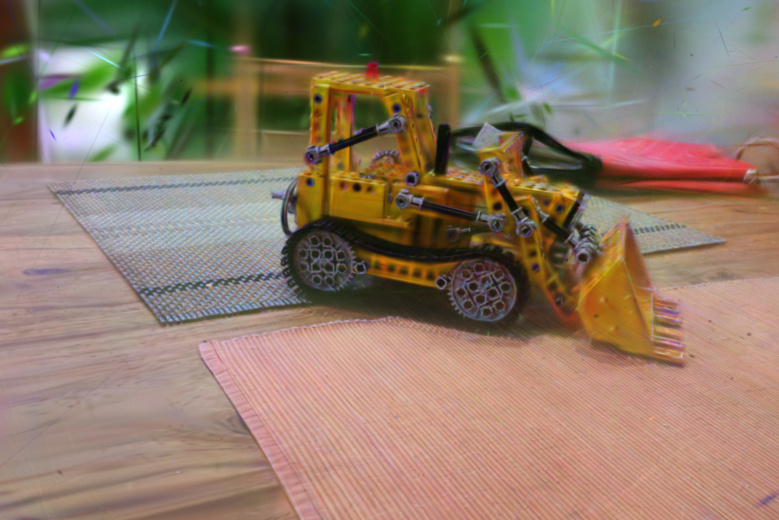} &
        \rotatebox{90}{\smaller[1]{Init}} &
        \pcell[width=\linewidth]{14.2}{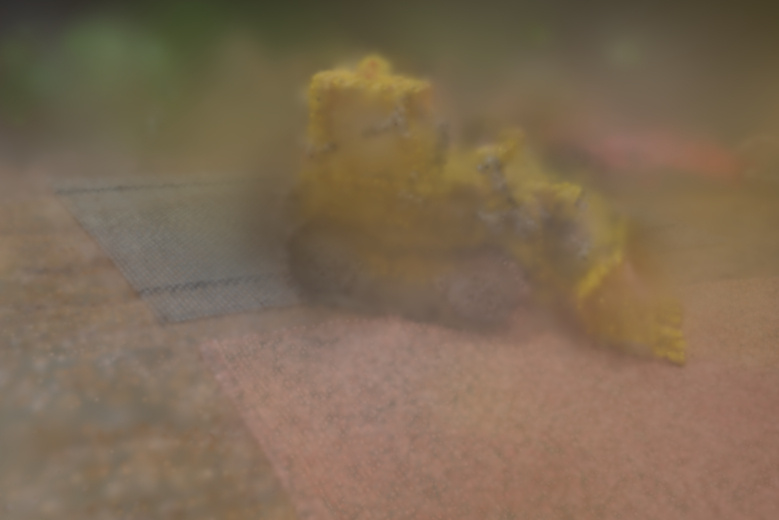} \\
        \noalign{\vskip 2\tabcolsep}
        \rotatebox{90}{\makecell{\smaller[1]{\adam{}}}} &
        \pcell[width=\linewidth]{14.7}{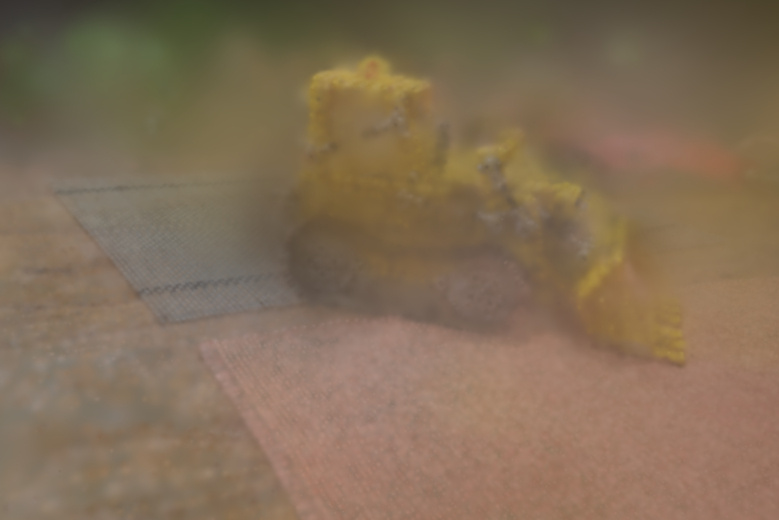} &
        \pcell[width=\linewidth]{16.2}{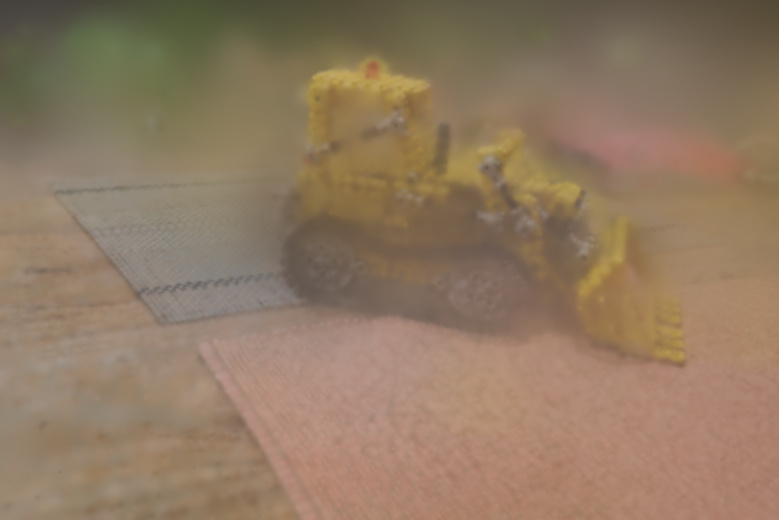} &
        \pcell[width=\linewidth]{22.2}{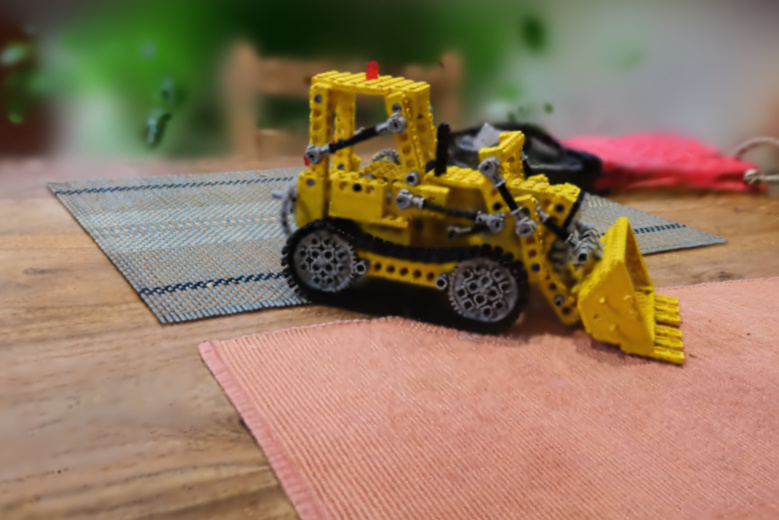} &
        \pcell[width=\linewidth]{27.7}{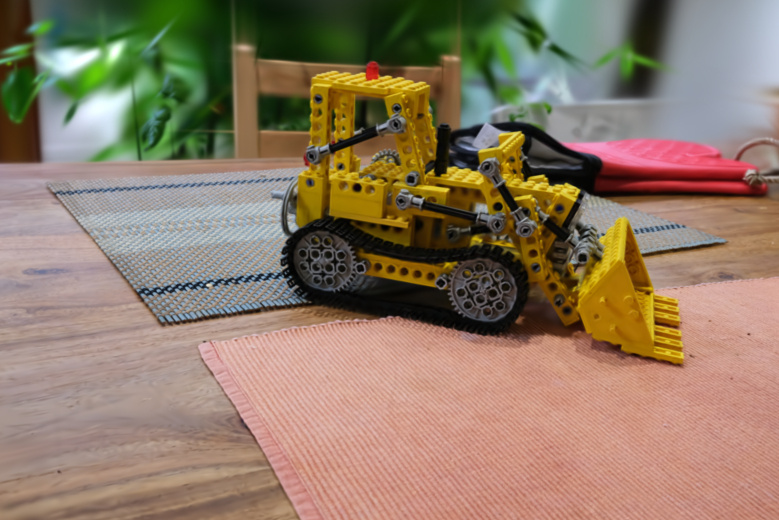} &
        \pcell[width=\linewidth]{28.6}{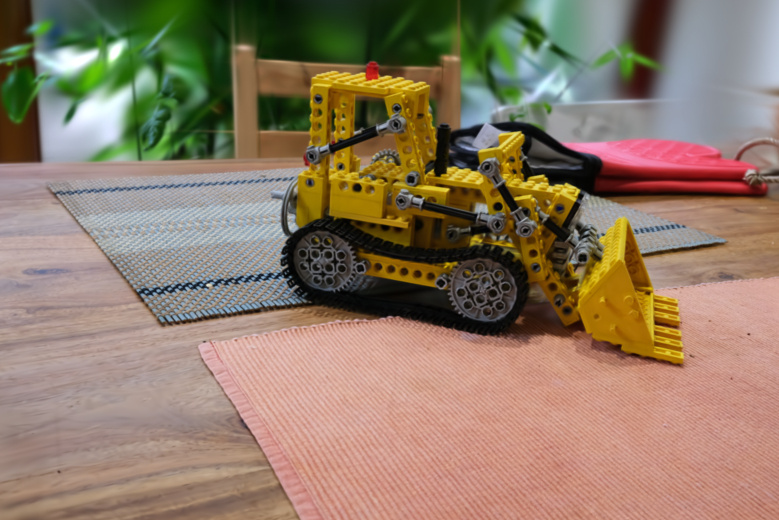} &
        \rotatebox{90}{\smaller[1]{Ref}} &
        \rcell[width=\linewidth]{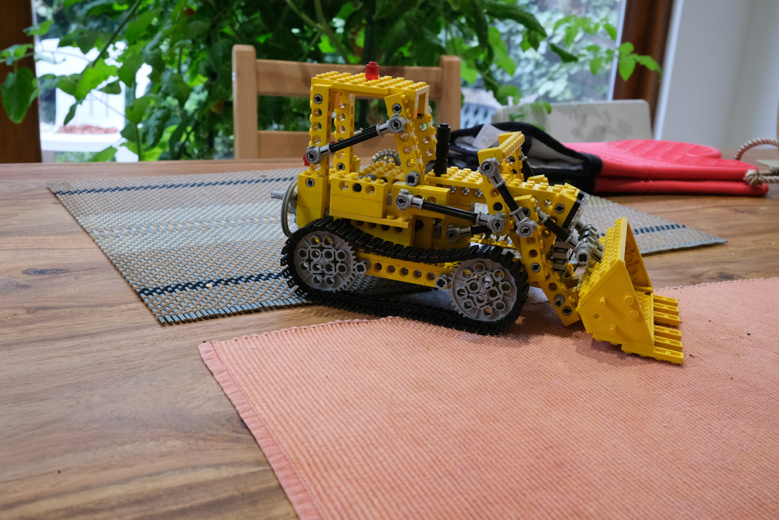} \\
        \noalign{\vskip 2\tabcolsep}
        \rotatebox{90}{\makecell{\smaller[1]{\ourshort{}}}} &
        \pcell[width=\linewidth]{18.9}{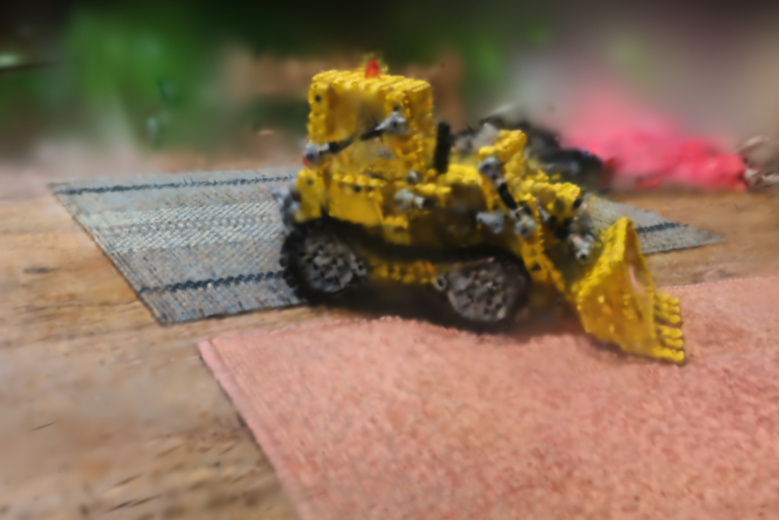} &
        \pcell[width=\linewidth]{23.0}{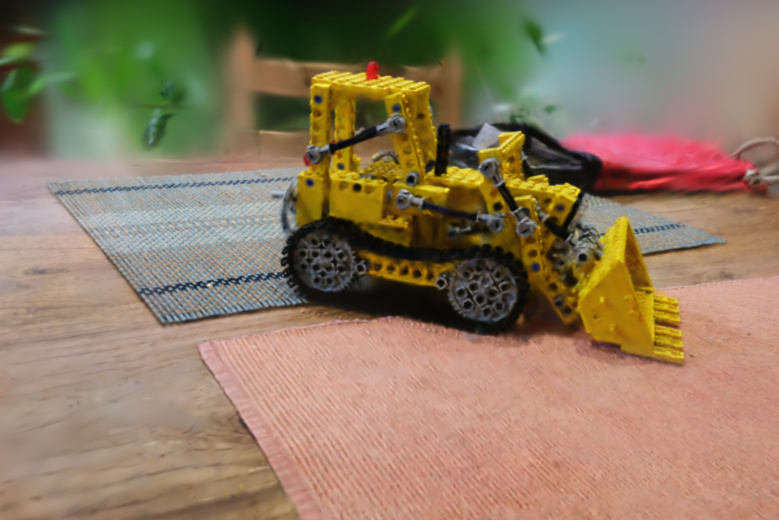} &
        \pcell[width=\linewidth]{27.7}{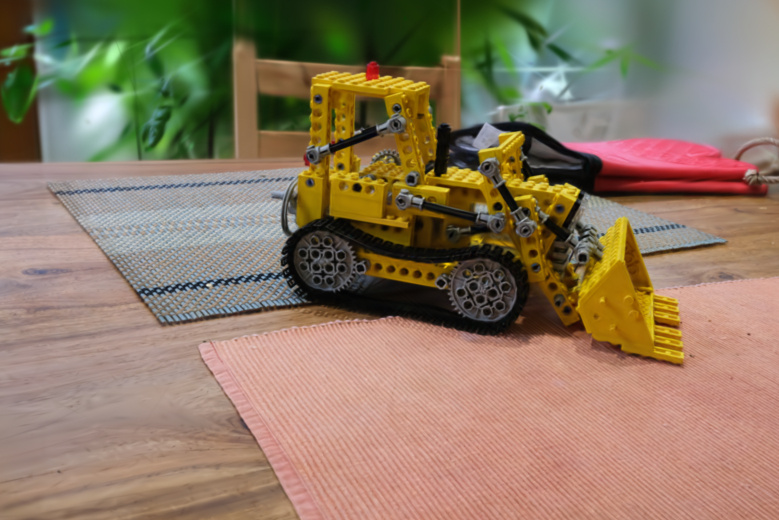} &
        \pcell[width=\linewidth]{30.0}{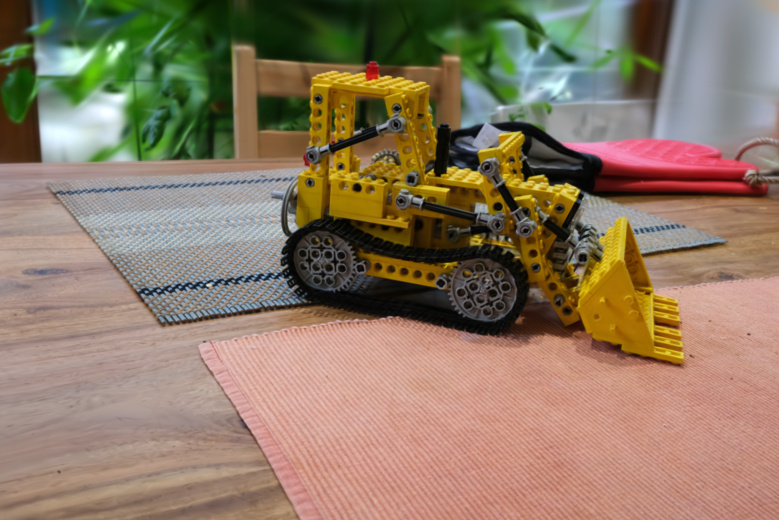} &
        \pcell[width=\linewidth]{30.3}{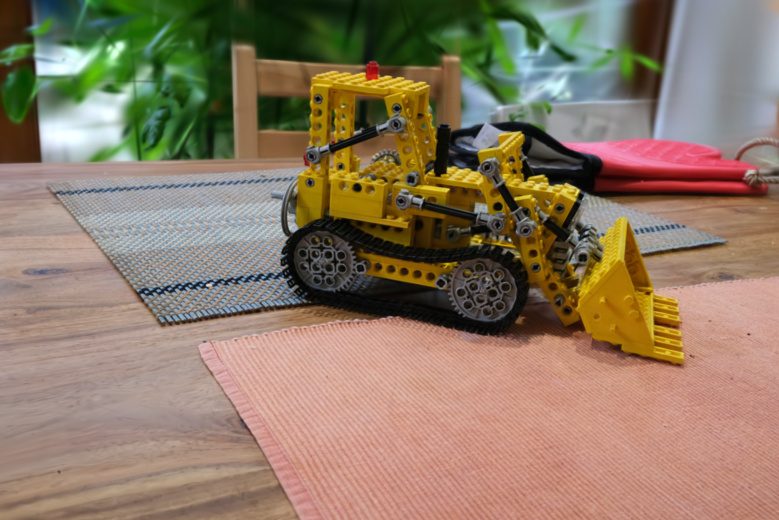} &
         &
         \\
        \noalign{\vskip 2\tabcolsep}
        \rotatebox{90}{\makecell{\smaller[1]{\adam{} error}}} &
        \rcell[width=\linewidth]{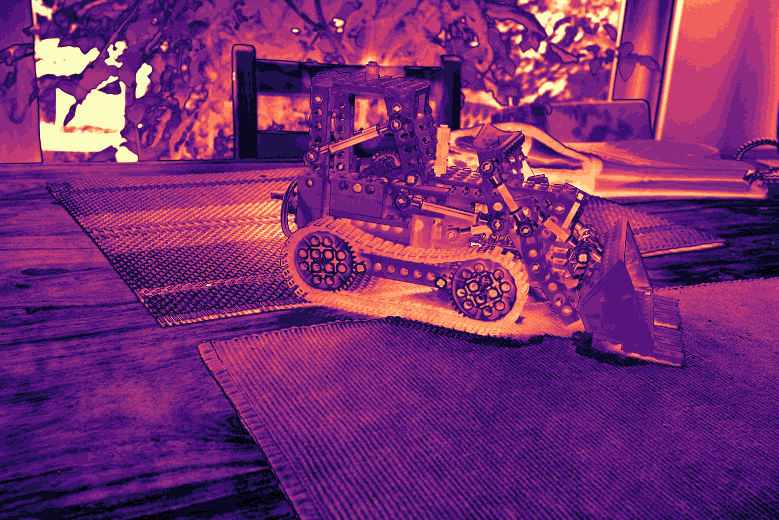} &
        \rcell[width=\linewidth]{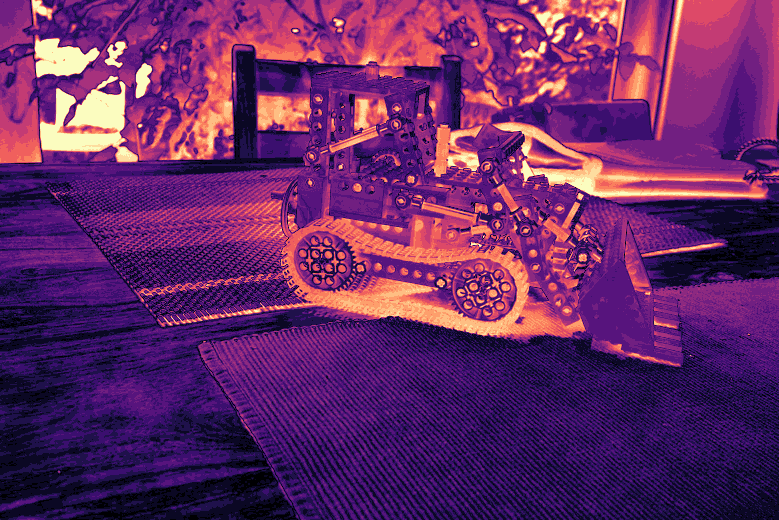} &
        \rcell[width=\linewidth]{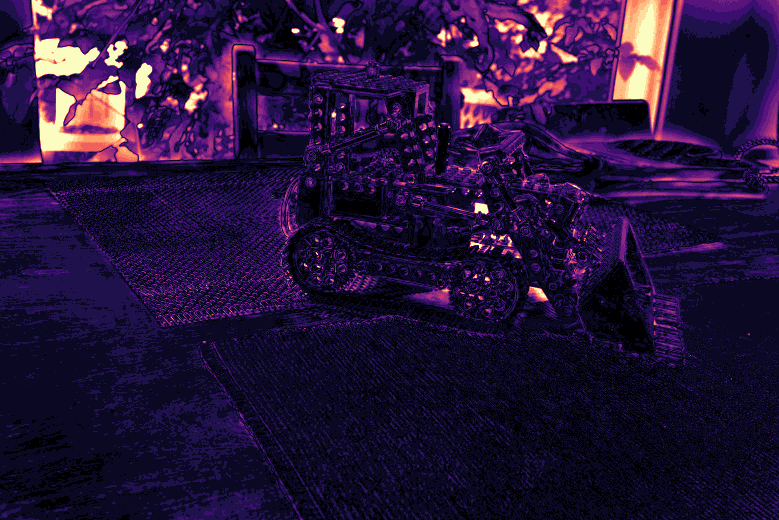} &
        \rcell[width=\linewidth]{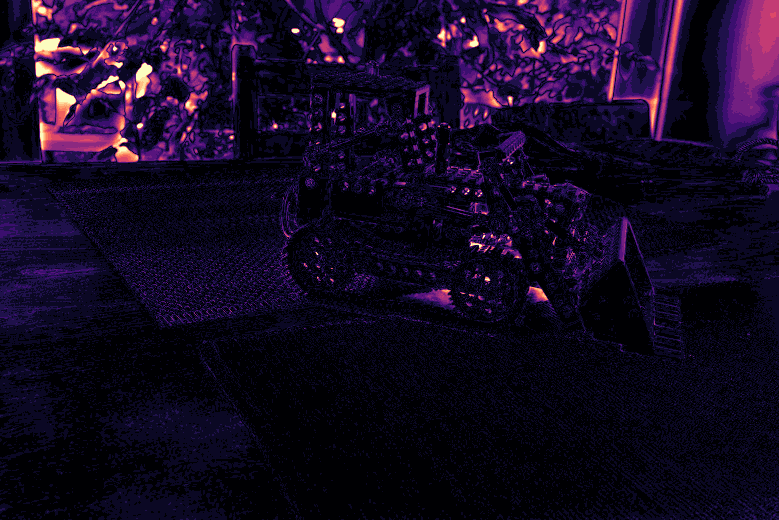} &
        \rcell[width=\linewidth]{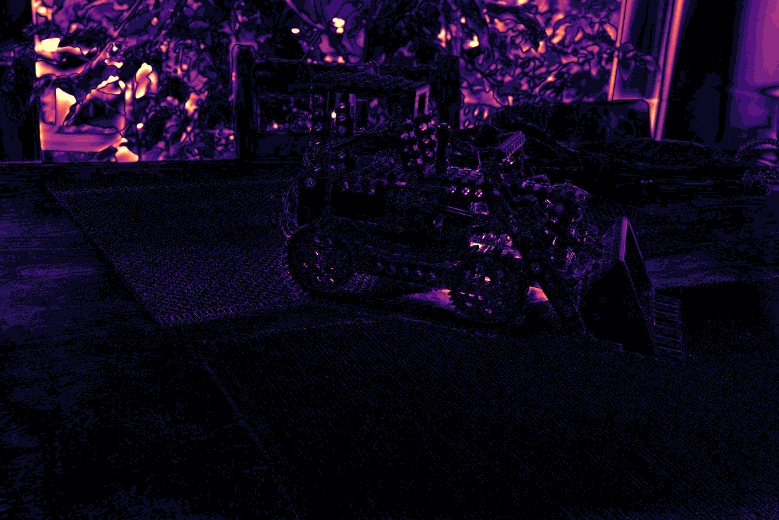} &
        \rotatebox{90}{\smaller[1]{Init error}} &
        \rcell[width=\linewidth]{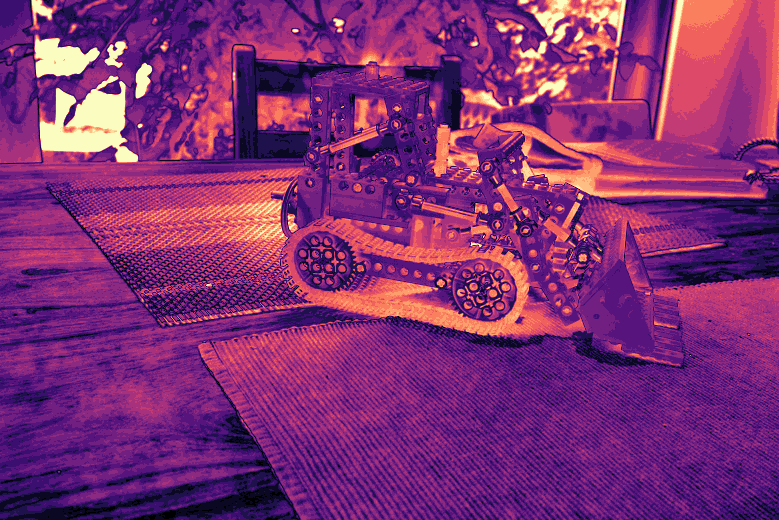} \\
        \noalign{\vskip 2\tabcolsep}
        \rotatebox{90}{\makecell{\smaller[1]{\ourshort{} error}}} &
        \rcell[width=\linewidth]{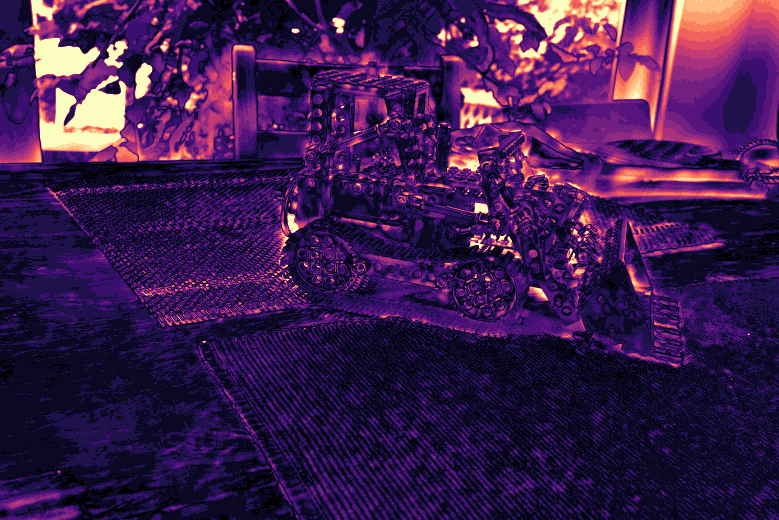} &
        \rcell[width=\linewidth]{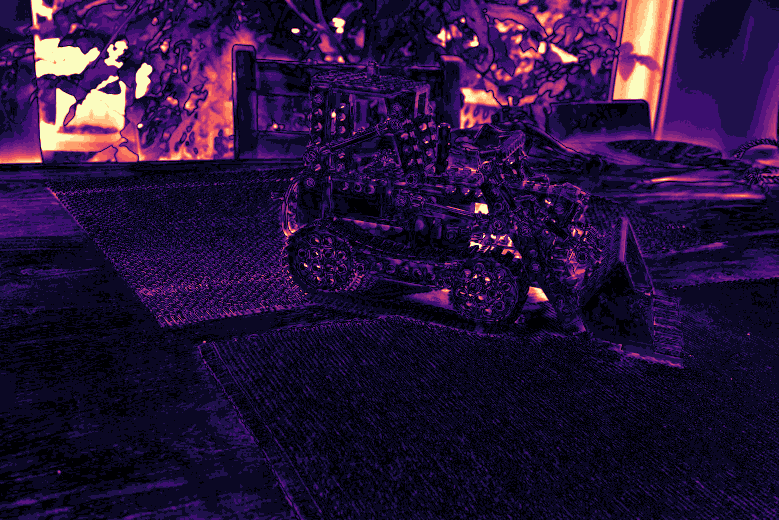} &
        \rcell[width=\linewidth]{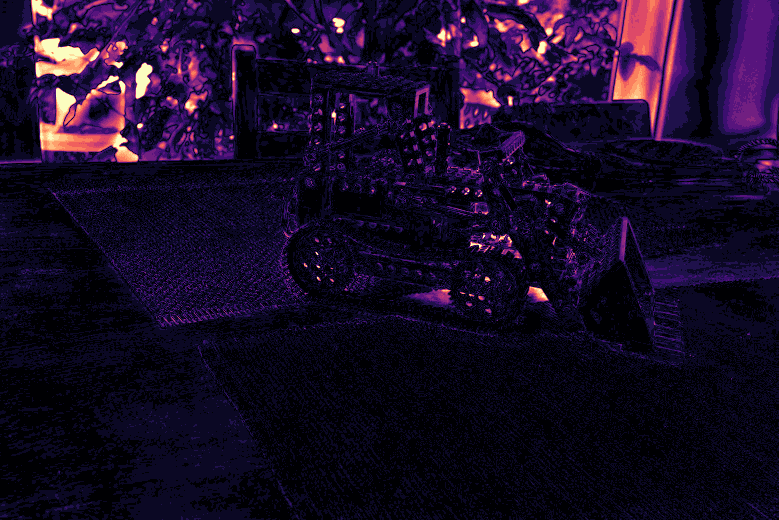} &
        \rcell[width=\linewidth]{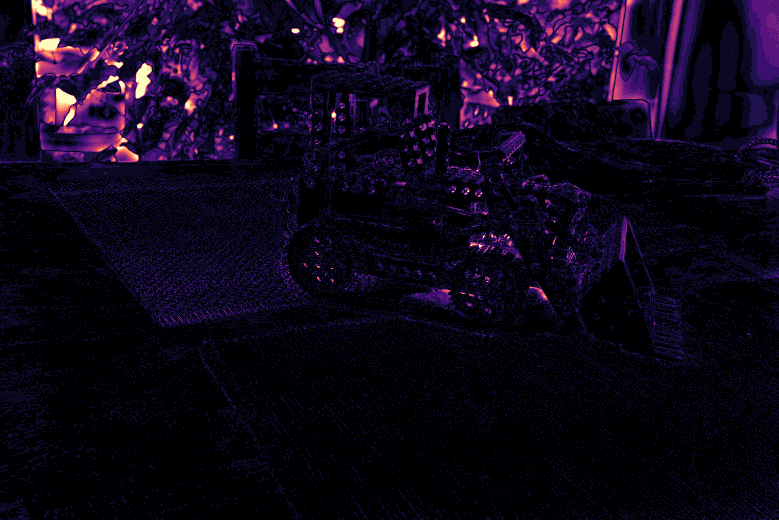} &
        \rcell[width=\linewidth]{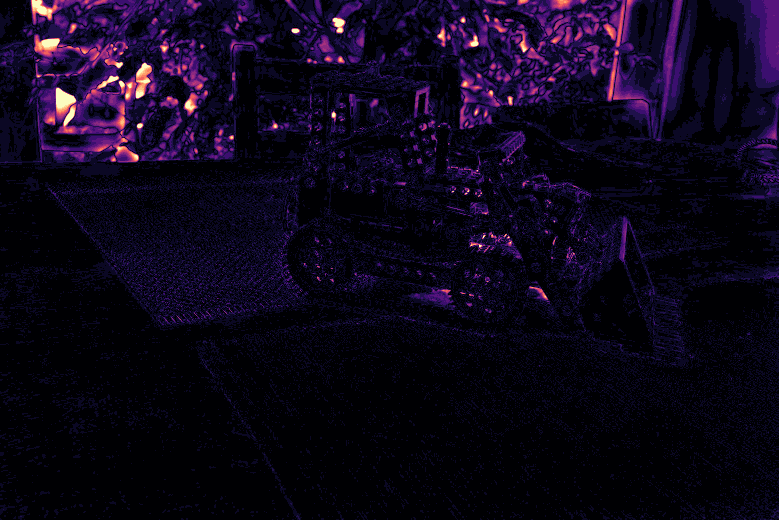} &
         &
         \\
        \noalign{\vskip 2\tabcolsep}
        \rotatebox{90}{\makecell{\smaller[1]{\lob{}}}} &
        \pcell[width=\linewidth]{11.1}{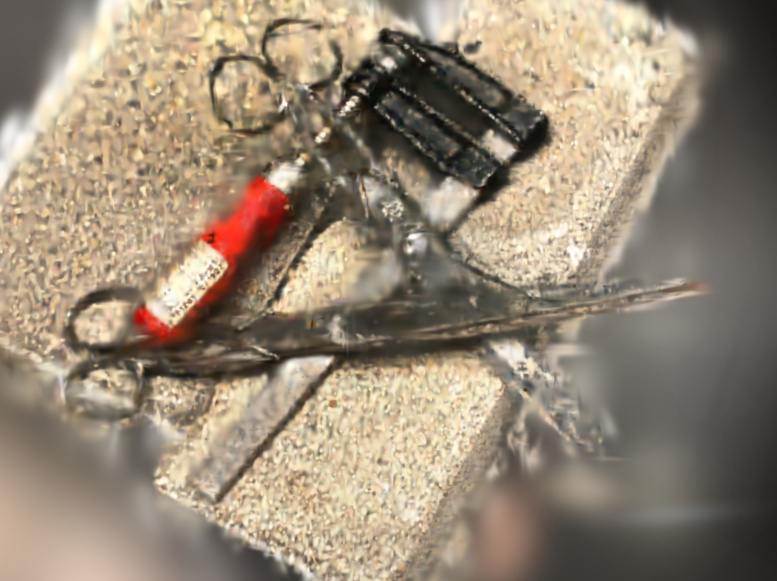} &
        \pcell[width=\linewidth]{19.2}{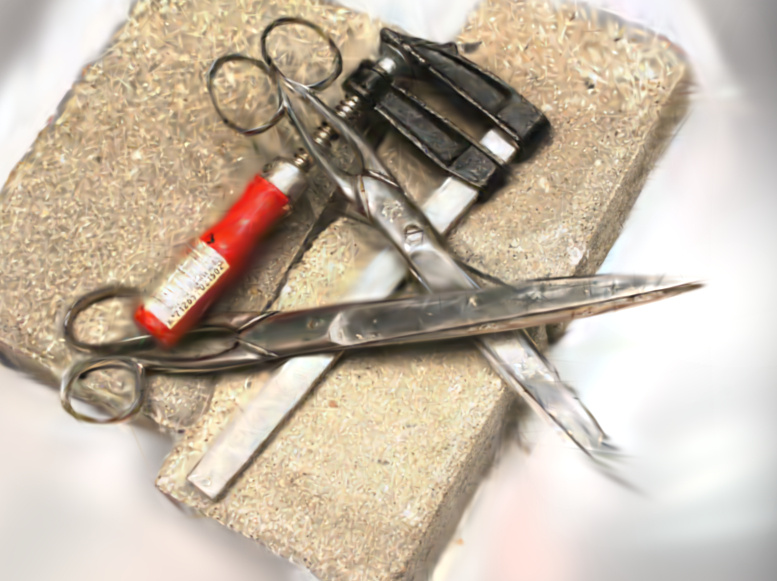} &
        \pcell[width=\linewidth]{22.6}{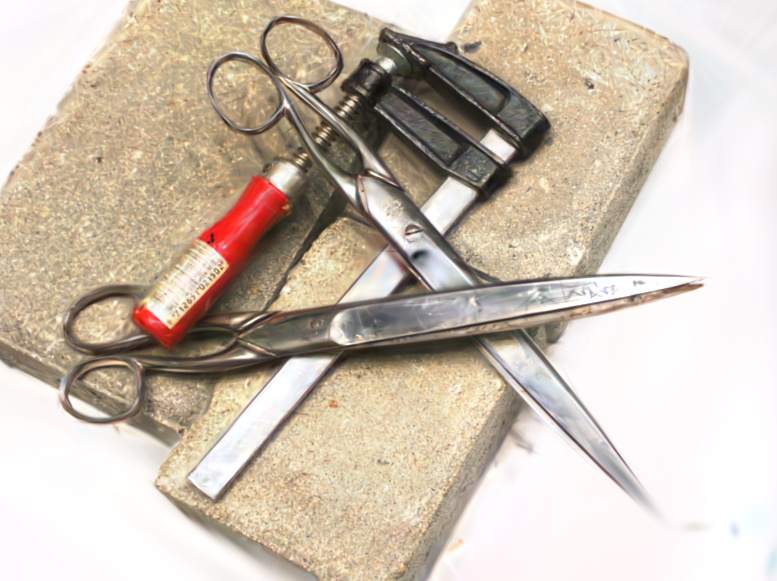} &
        \pcell[width=\linewidth]{22.6}{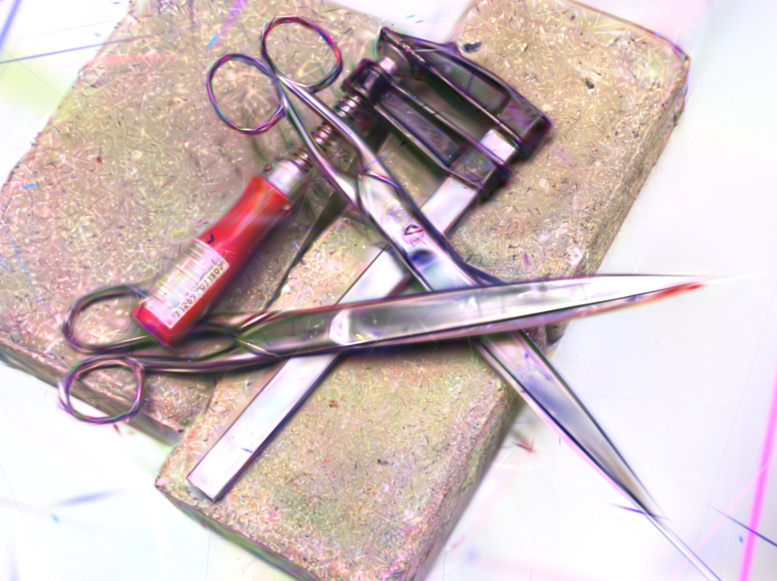} &
        \pcell[width=\linewidth]{22.2}{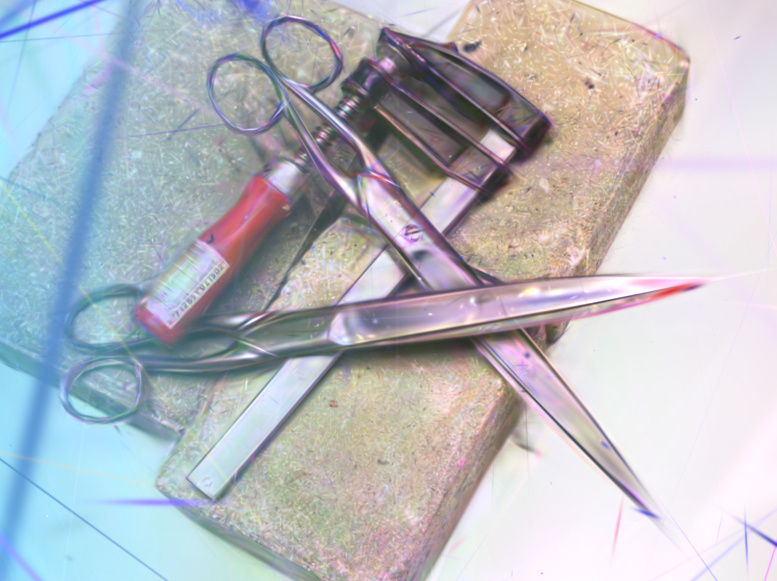} &
        \rotatebox{90}{\smaller[1]{Init}} &
        \pcell[width=\linewidth]{7.7}{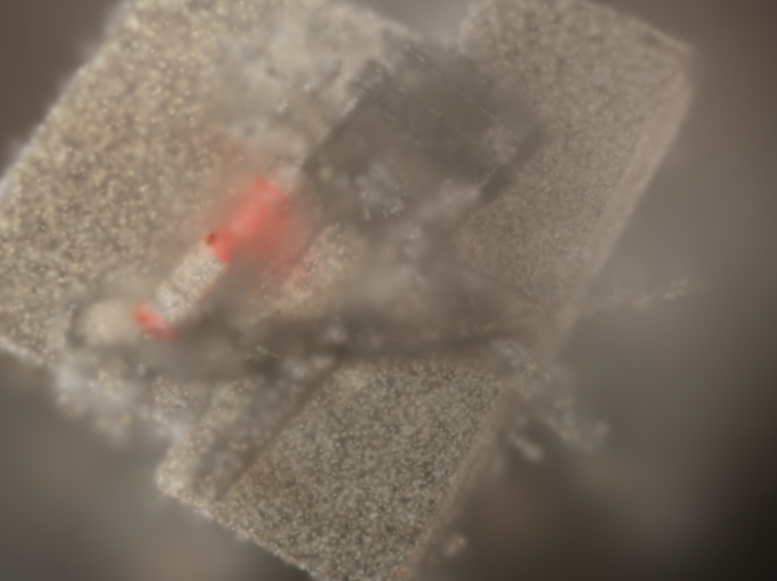} \\
        \noalign{\vskip 2\tabcolsep}
        \rotatebox{90}{\makecell{\smaller[1]{\adam{}}}} &
        \pcell[width=\linewidth]{7.9}{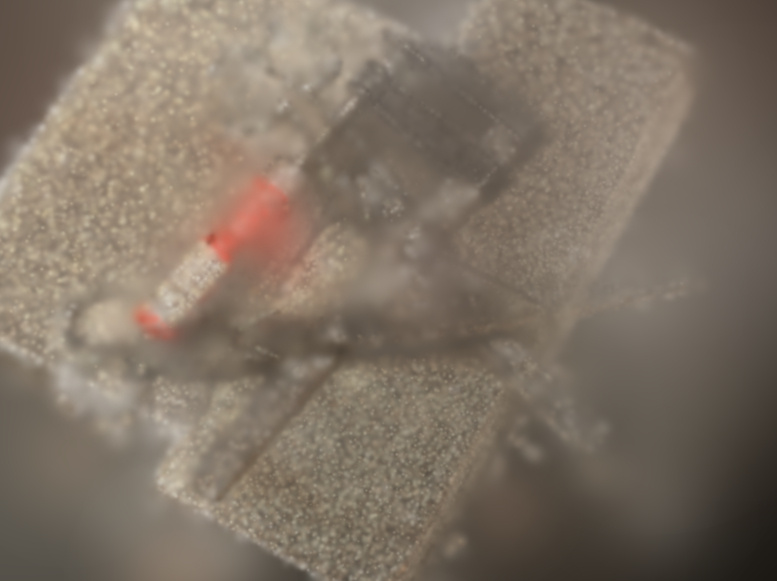} &
        \pcell[width=\linewidth]{8.6}{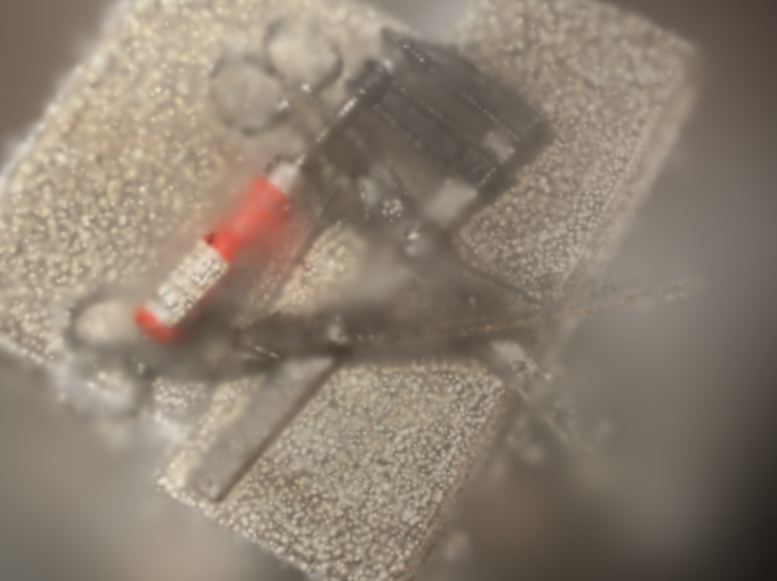} &
        \pcell[width=\linewidth]{14.2}{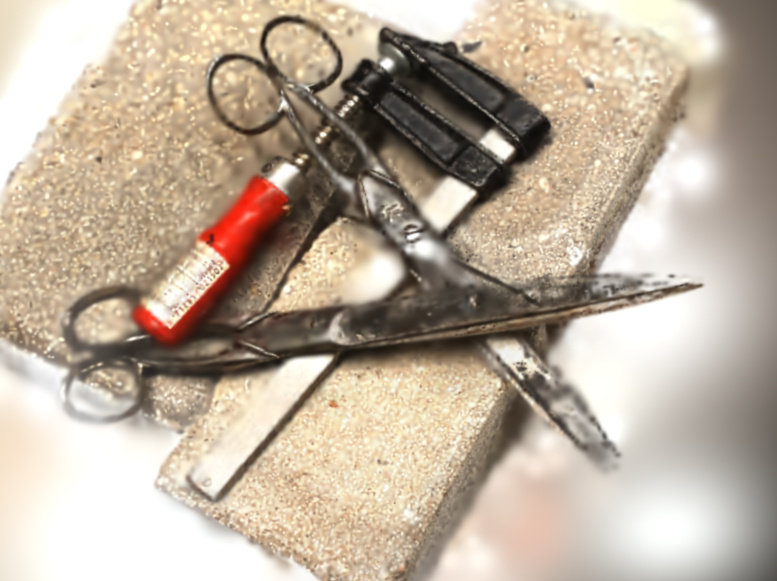} &
        \pcell[width=\linewidth]{21.3}{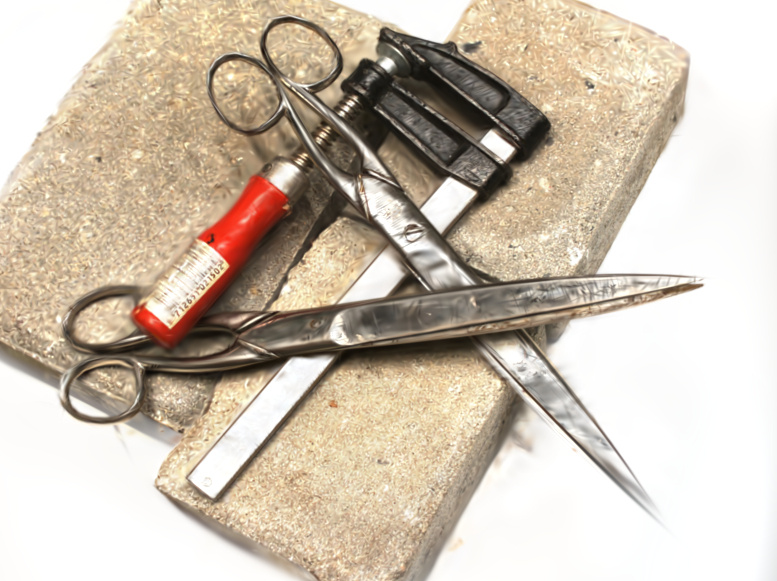} &
        \pcell[width=\linewidth]{22.0}{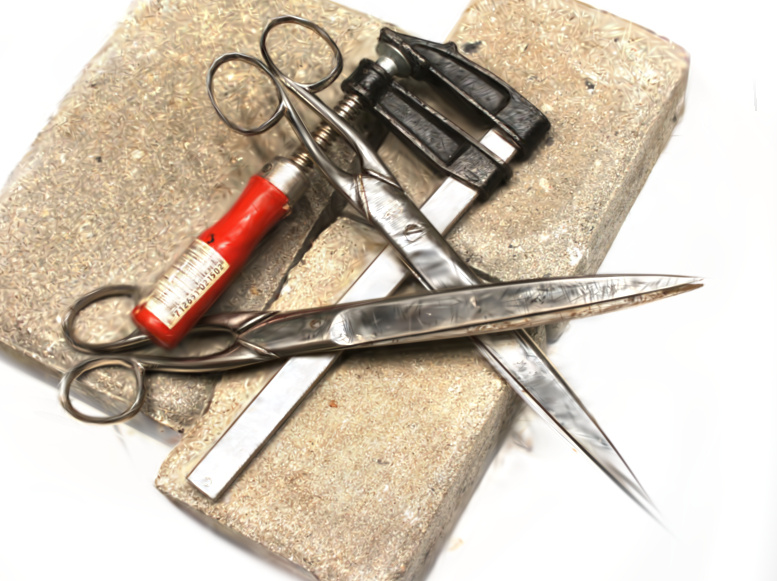} &
        \rotatebox{90}{\smaller[1]{Ref}} &
        \rcell[width=\linewidth]{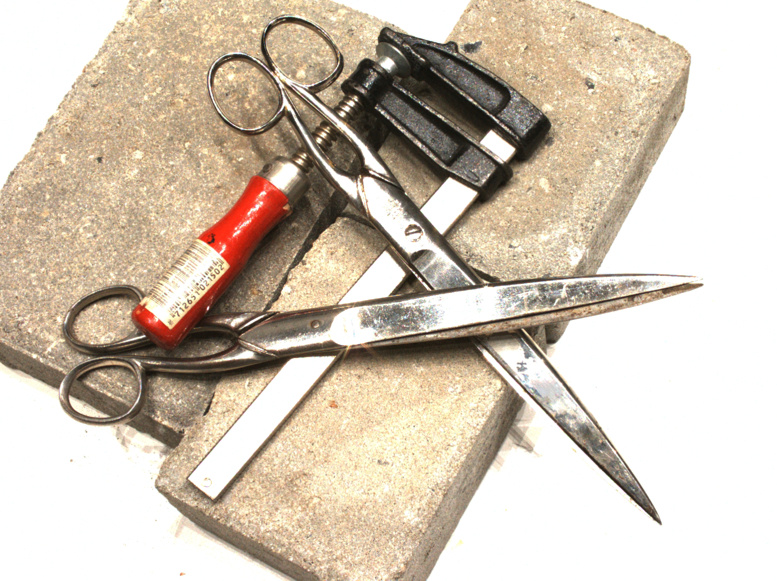} \\
        \noalign{\vskip 2\tabcolsep}
        \rotatebox{90}{\makecell{\smaller[1]{\ourshort{}}}} &
        \pcell[width=\linewidth]{11.3}{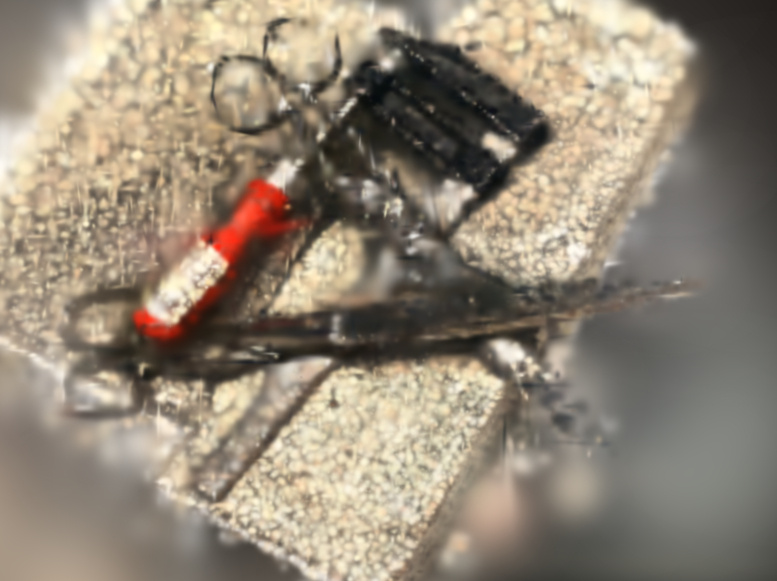} &
        \pcell[width=\linewidth]{18.1}{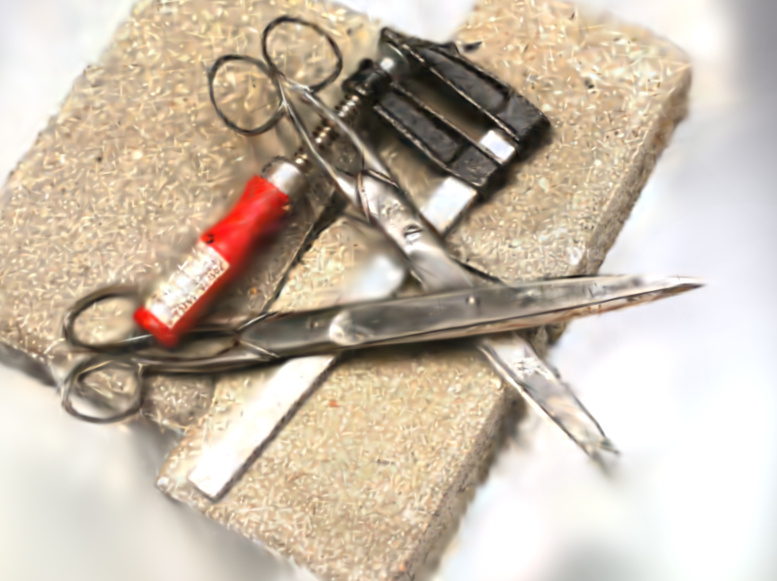} &
        \pcell[width=\linewidth]{21.8}{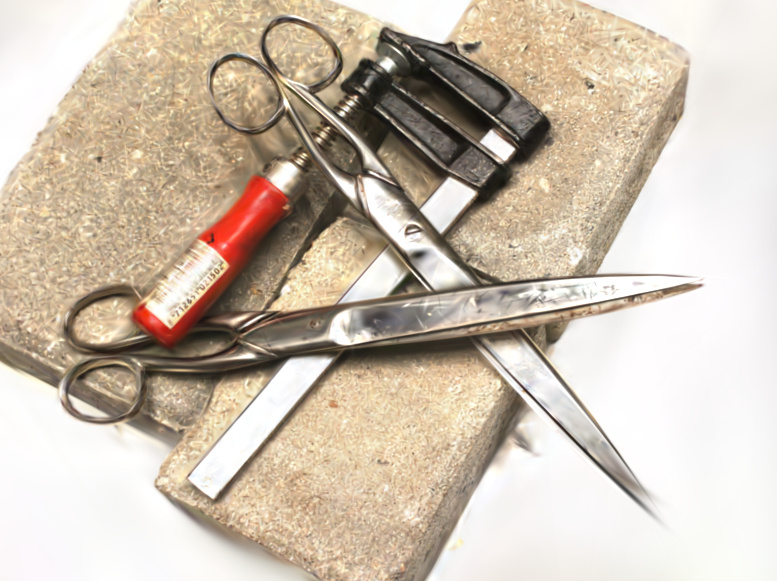} &
        \pcell[width=\linewidth]{24.0}{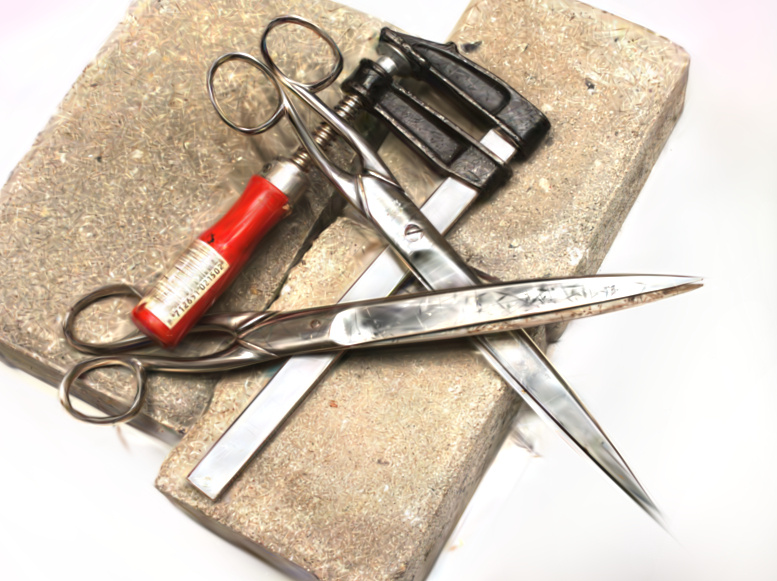} &
        \pcell[width=\linewidth]{24.3}{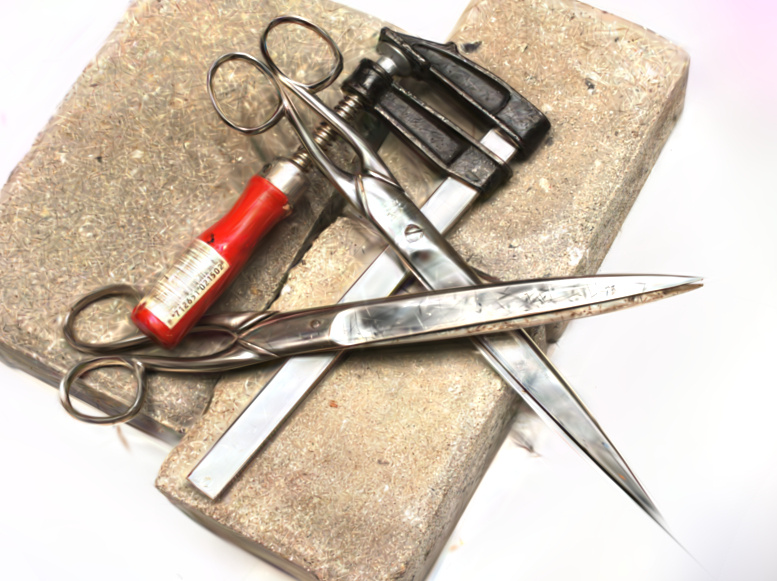} &
         &
         \\
        \noalign{\vskip 2\tabcolsep}
        \rotatebox{90}{\makecell{\smaller[1]{\adam{} error}}} &
        \rcell[width=\linewidth]{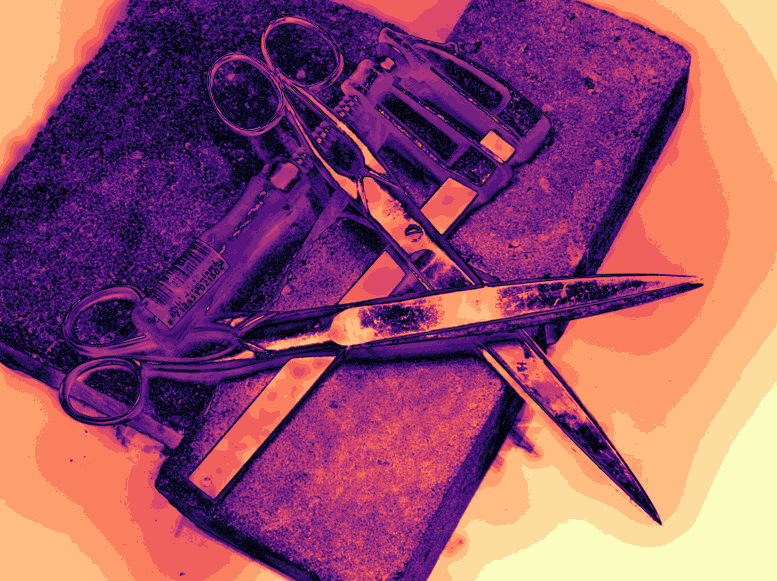} &
        \rcell[width=\linewidth]{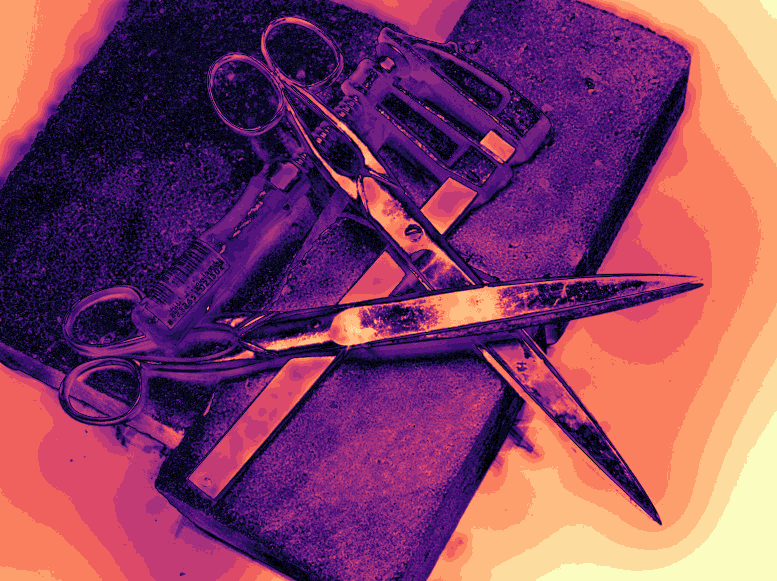} &
        \rcell[width=\linewidth]{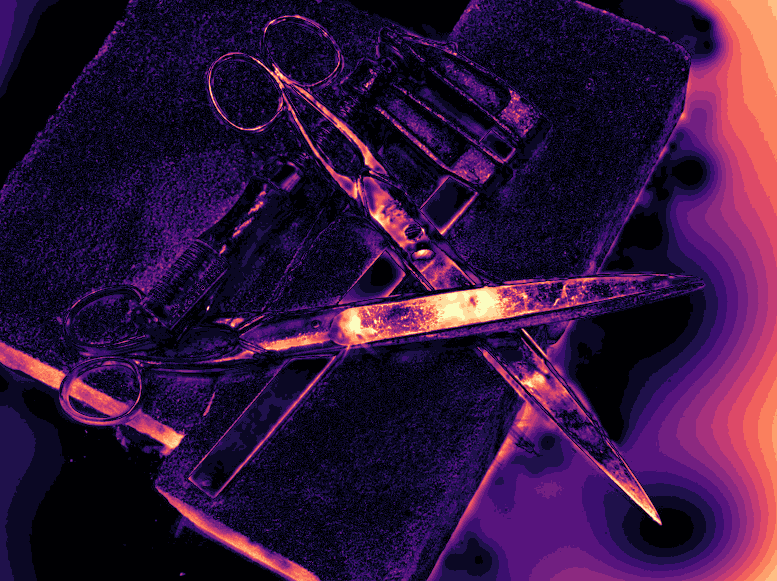} &
        \rcell[width=\linewidth]{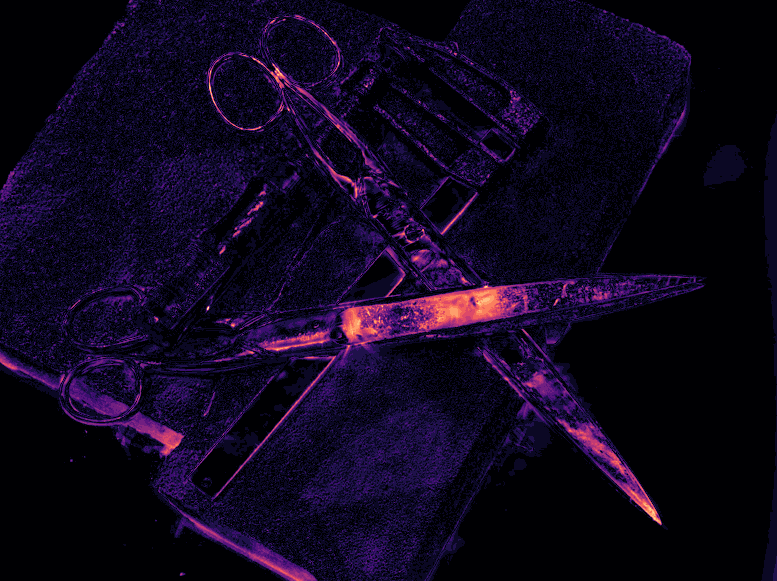} &
        \rcell[width=\linewidth]{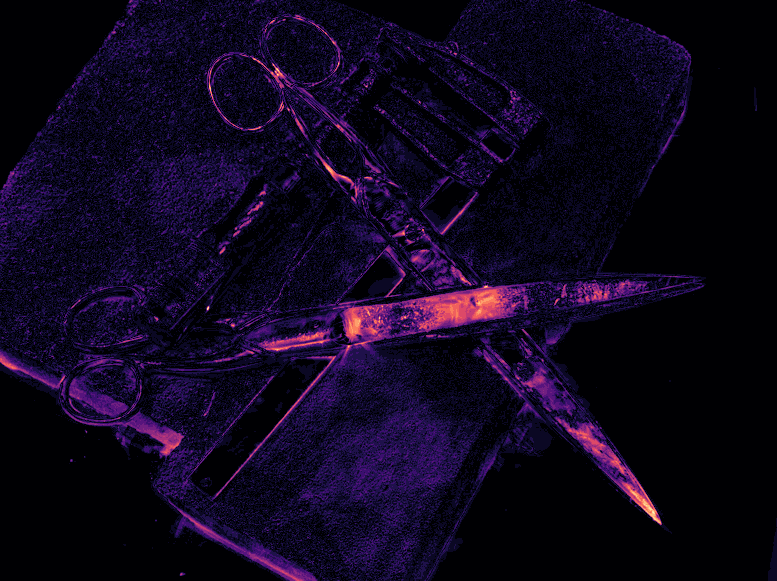} &
        \rotatebox{90}{\smaller[1]{Init error}} &
        \rcell[width=\linewidth]{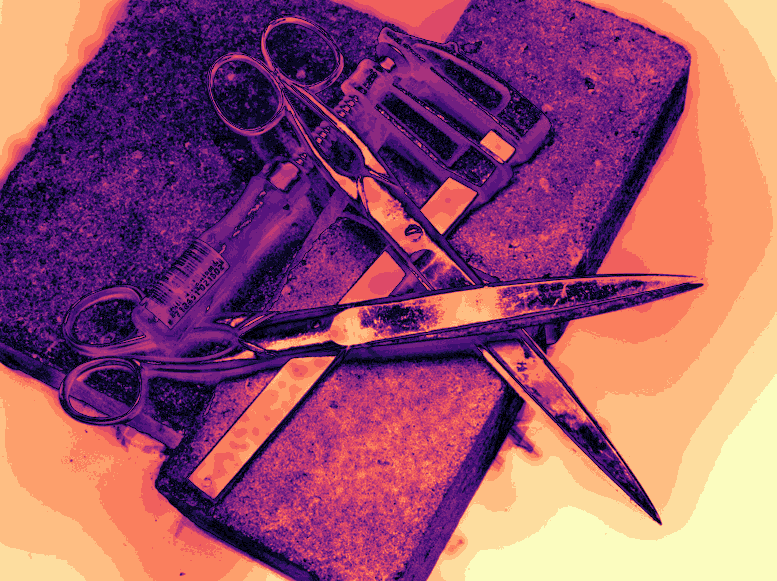} \\
        \noalign{\vskip 2\tabcolsep}
        \rotatebox{90}{\makecell{\smaller[1]{\ourshort{} error}}} &
        \rcell[width=\linewidth]{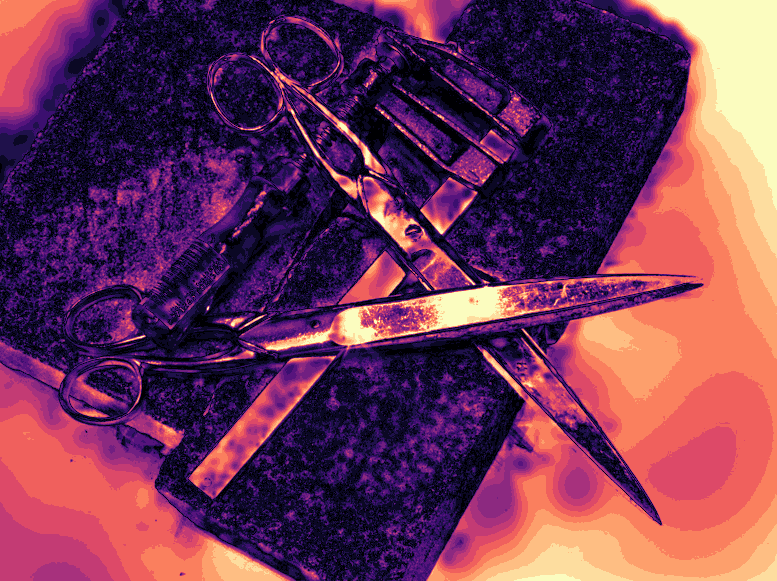} &
        \rcell[width=\linewidth]{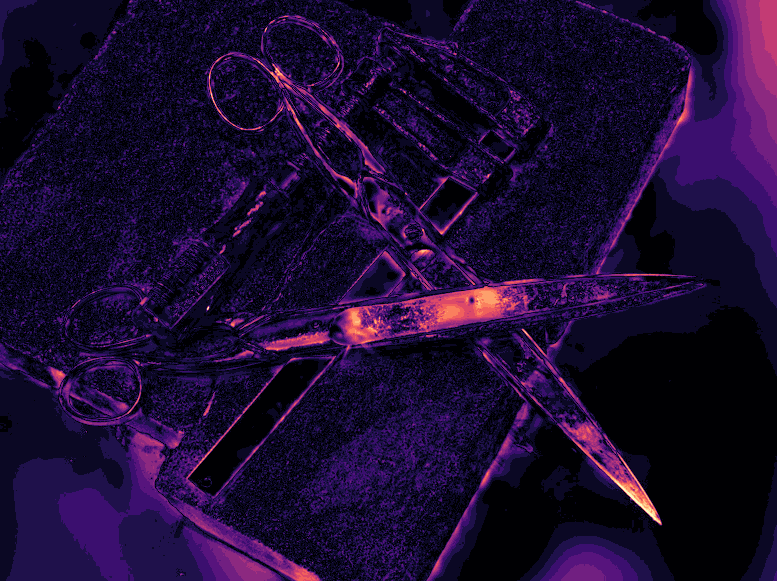} &
        \rcell[width=\linewidth]{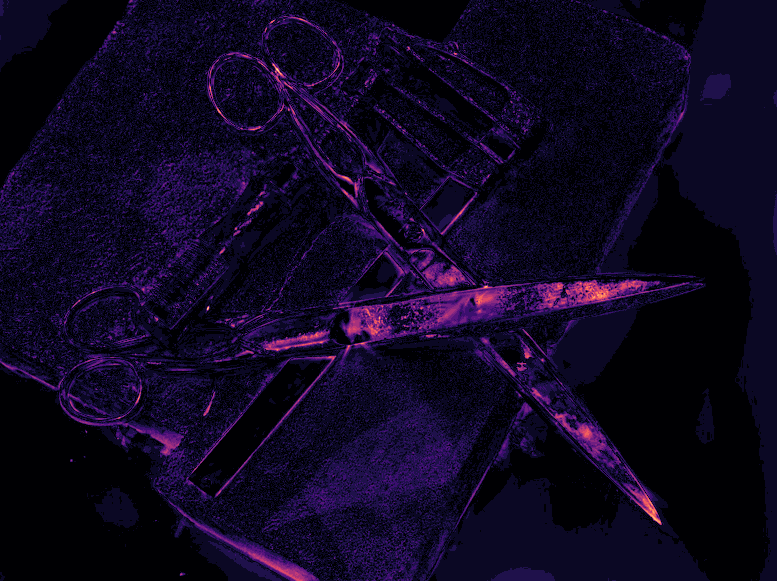} &
        \rcell[width=\linewidth]{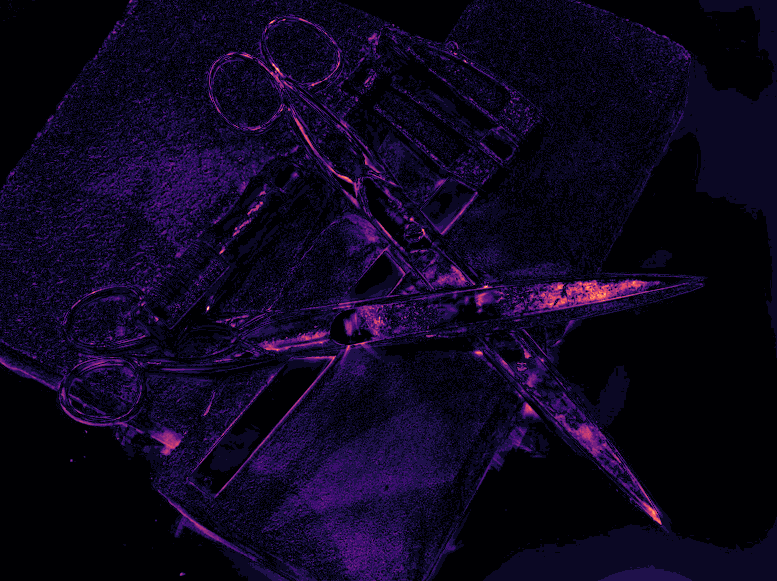} &
        \rcell[width=\linewidth]{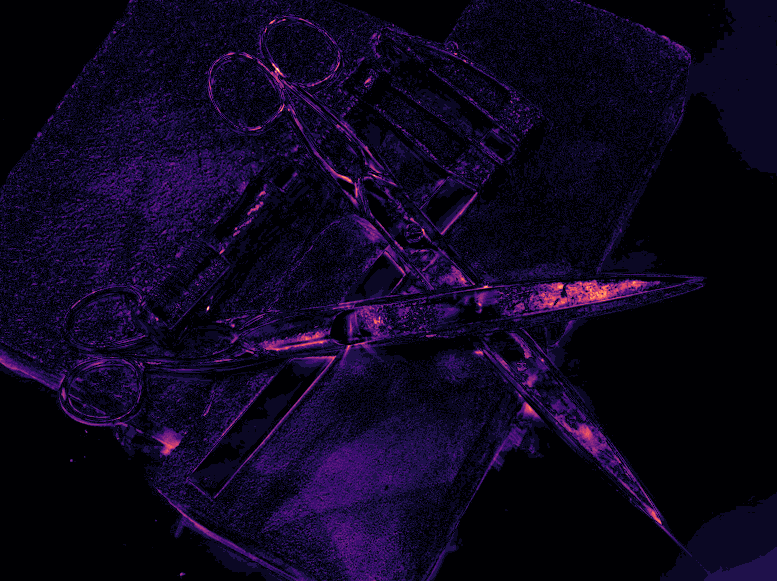} &
         &
         \\
    \end{tabular}
}
    \caption{
        \textbf{Zero-shot Generalization to Mip-NeRF360 and DTU}. Scene reconstructions from Mip-NeRF360~\cite{barron2022mipnerf360} (top, $840{\times}1297$) and DTU~\cite{Aanes2016IJCV} (bottom, $581{\times}777$), both in the dense-view setting: SfM initialization, every $8$th view held out as a target, and $8$ of the remaining context views sampled per iteration.
        \ourshort{} resolves the scene within the first ten iterations, a state \adam{} reaches an order of magnitude later.
    }
    \label{fig:mipnerf-dtu-comp}
\end{figure*}

\clearpage
}
\afterpage{
    \clearpage

\begin{figure*}[p]
    \centering

    \resizebox{1.0\linewidth}{!}{%
    \setlength{\tabcolsep}{0.002\linewidth} %
    \renewcommand{\arraystretch}{0} %

    \begin{tabular}{@{}%
        >{\centering\arraybackslash}m{0.03\linewidth}
        >{\centering\arraybackslash}m{0.24\linewidth}
        >{\centering\arraybackslash}m{0.24\linewidth}
        >{\centering\arraybackslash}m{0.24\linewidth}
        >{\centering\arraybackslash}m{0.24\linewidth}
        >{\centering\arraybackslash}m{0.24\linewidth}
        >{\centering\arraybackslash}m{0.02\linewidth}
        >{\centering\arraybackslash}m{0.24\linewidth}
    @{}}
         & $t = 2$ & $t = 10$ & $t = 100$ & $t = 1000$ & $t = 2000$ &  &  \\[3pt]

        \rotatebox{90}{\makecell{\smaller[1]{\lob{}}}} &
        \pcell[width=\linewidth]{8.1}{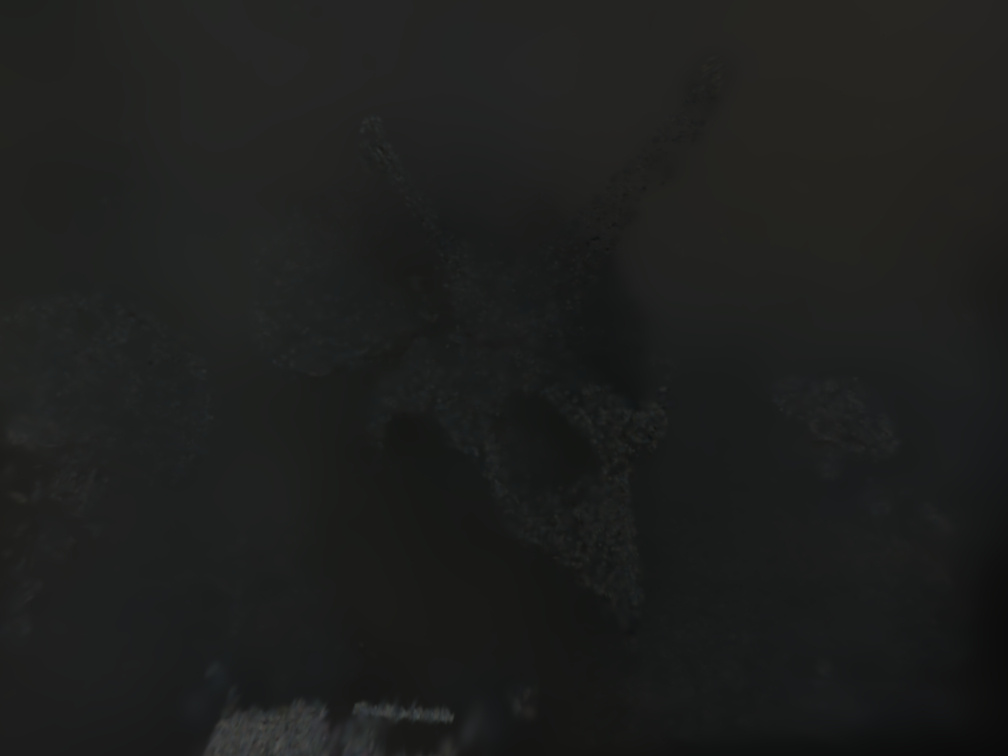} &
        \pcell[width=\linewidth]{18.4}{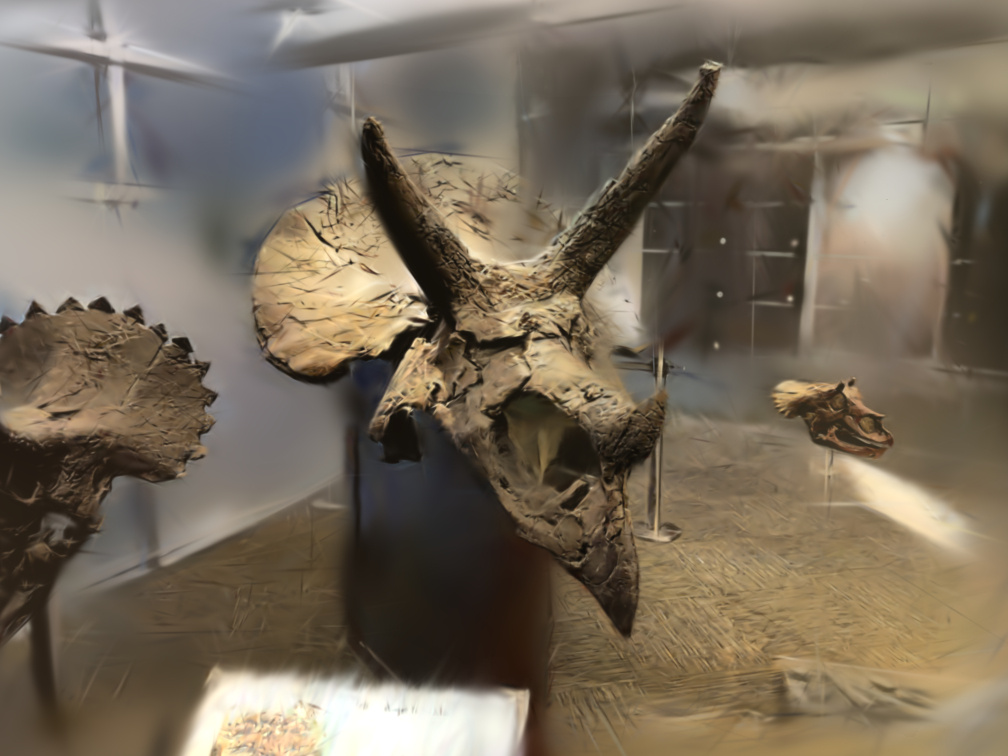} &
        \pcell[width=\linewidth]{23.7}{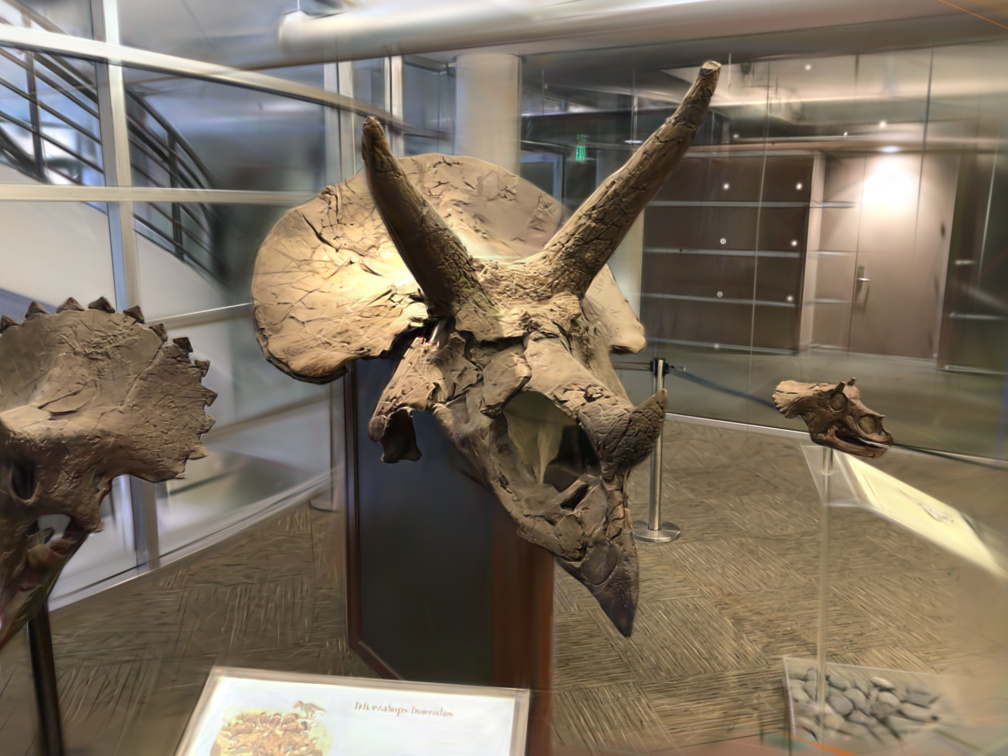} &
        \pcell[width=\linewidth]{24.3}{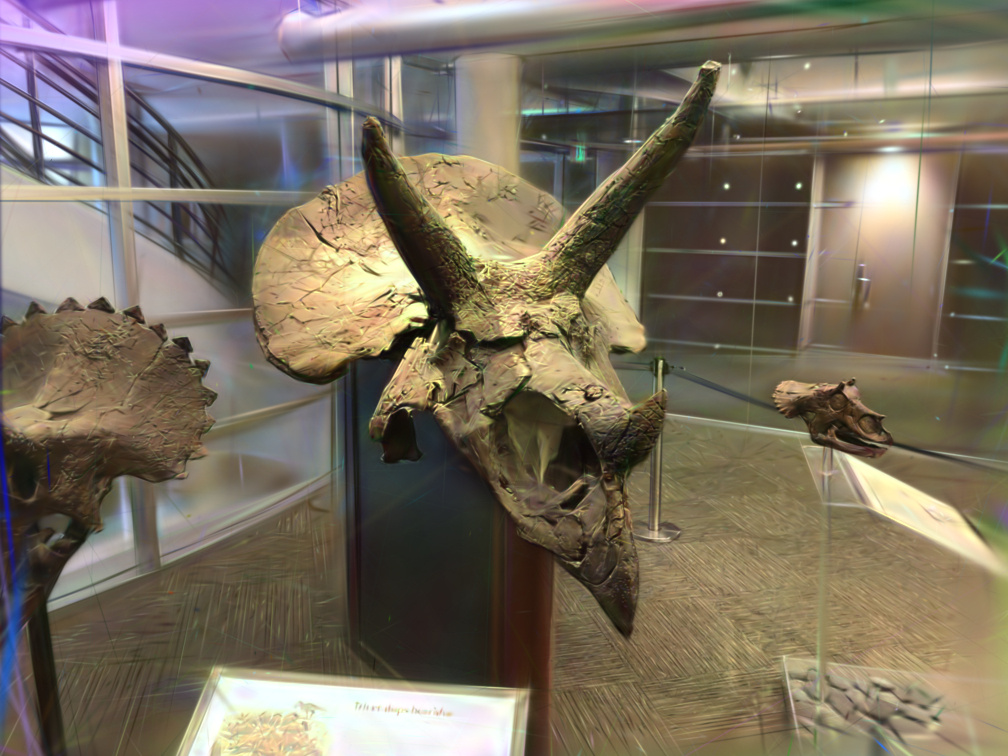} &
        \pcell[width=\linewidth]{24.0}{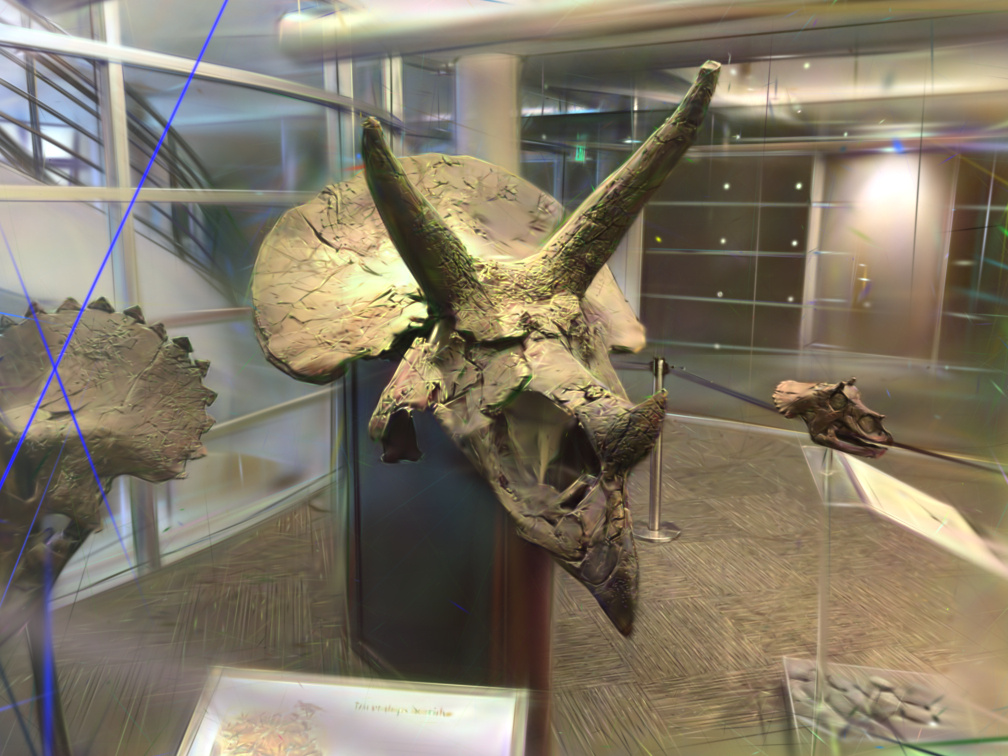} &
        \rotatebox{90}{\smaller[1]{Init}} &
        \pcell[width=\linewidth]{6.0}{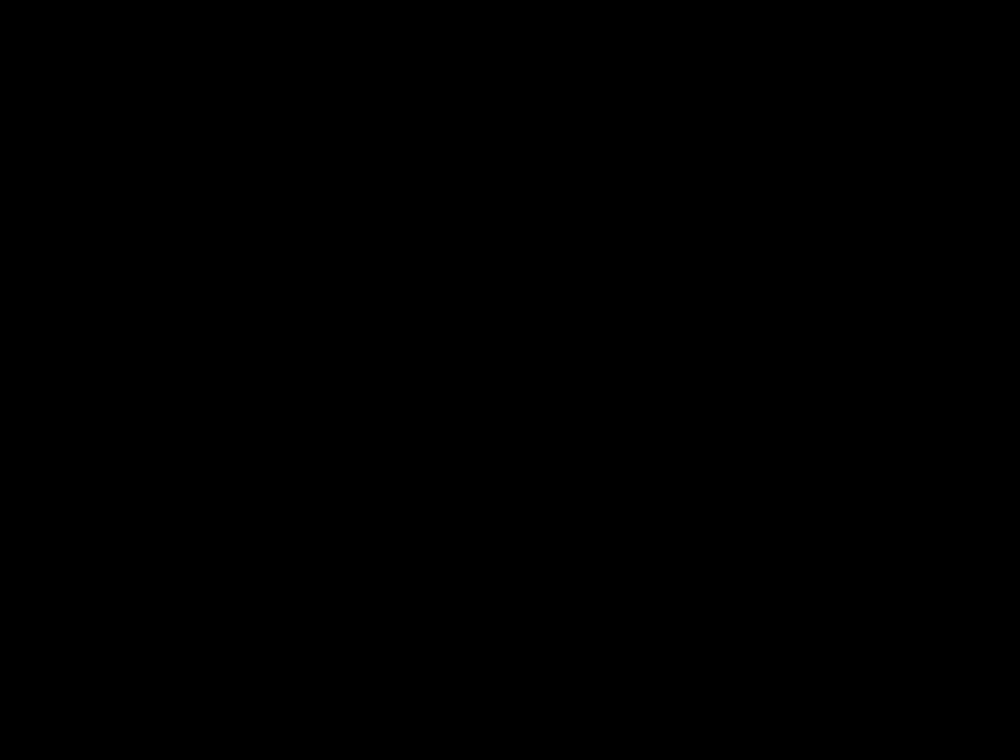} \\
        \noalign{\vskip 2\tabcolsep}
        \rotatebox{90}{\makecell{\smaller[1]{\adam{}}}} &
        \pcell[width=\linewidth]{6.0}{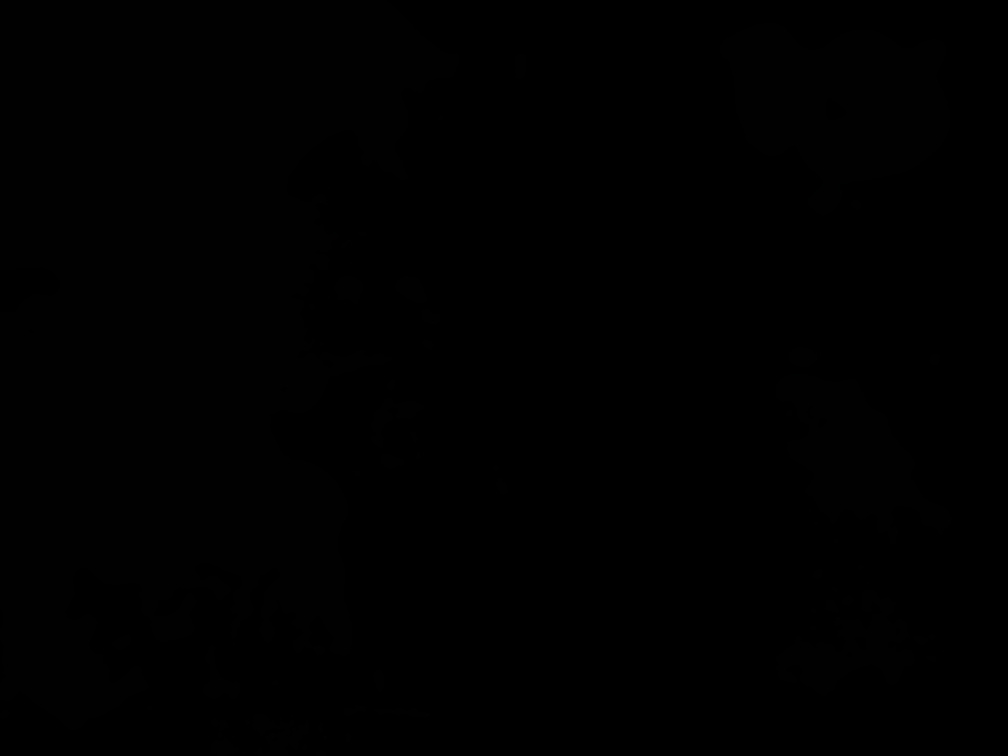} &
        \pcell[width=\linewidth]{6.1}{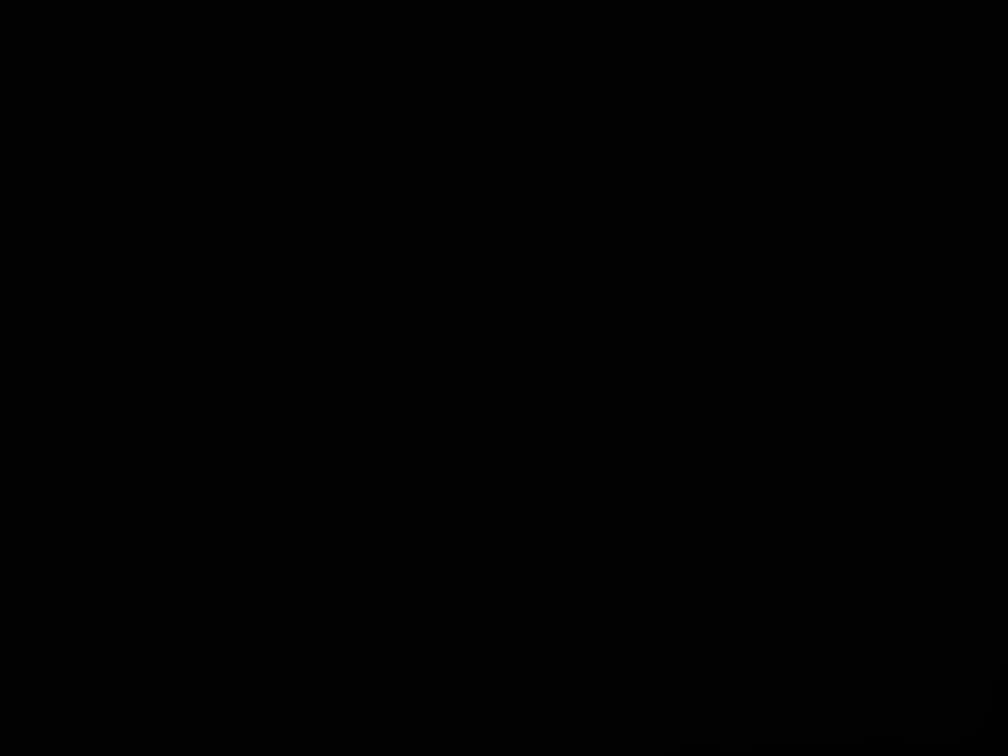} &
        \pcell[width=\linewidth]{8.3}{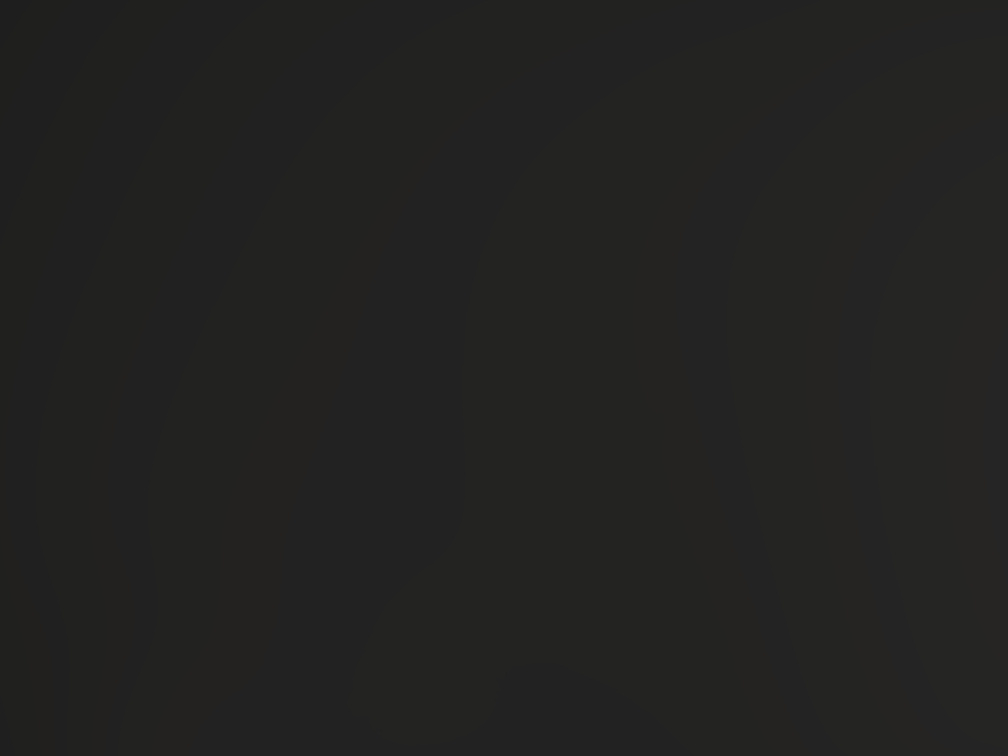} &
        \pcell[width=\linewidth]{16.7}{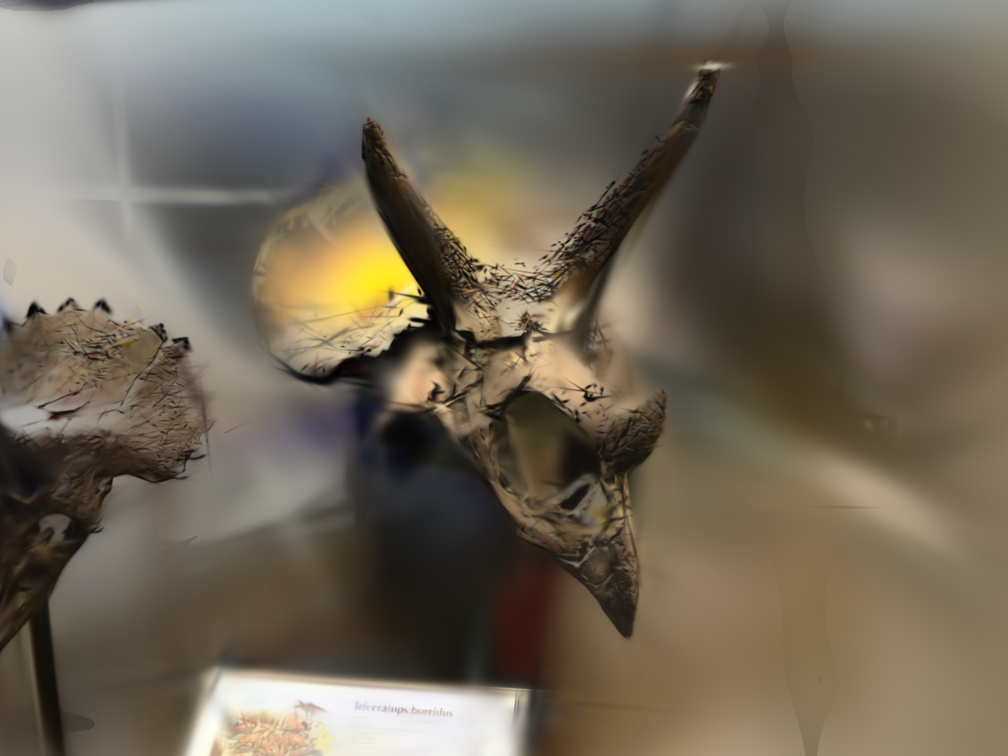} &
        \pcell[width=\linewidth]{21.8}{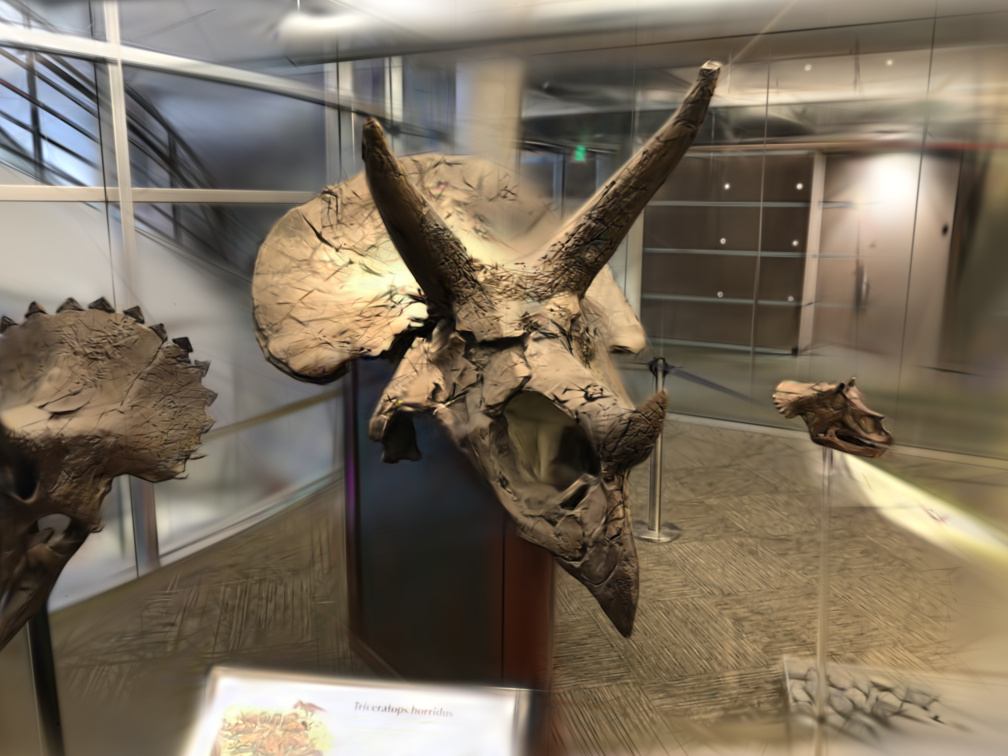} &
        \rotatebox{90}{\smaller[1]{Ref}} &
        \rcell[width=\linewidth]{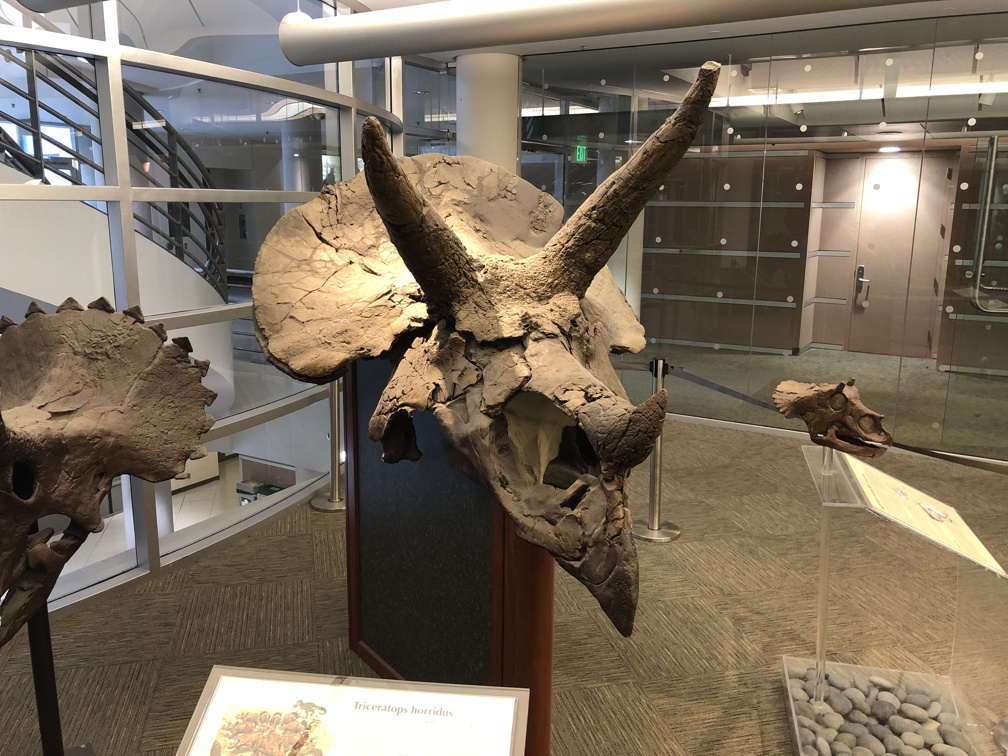} \\
        \noalign{\vskip 2\tabcolsep}
        \rotatebox{90}{\makecell{\smaller[1]{\ourshort{}}}} &
        \pcell[width=\linewidth]{7.4}{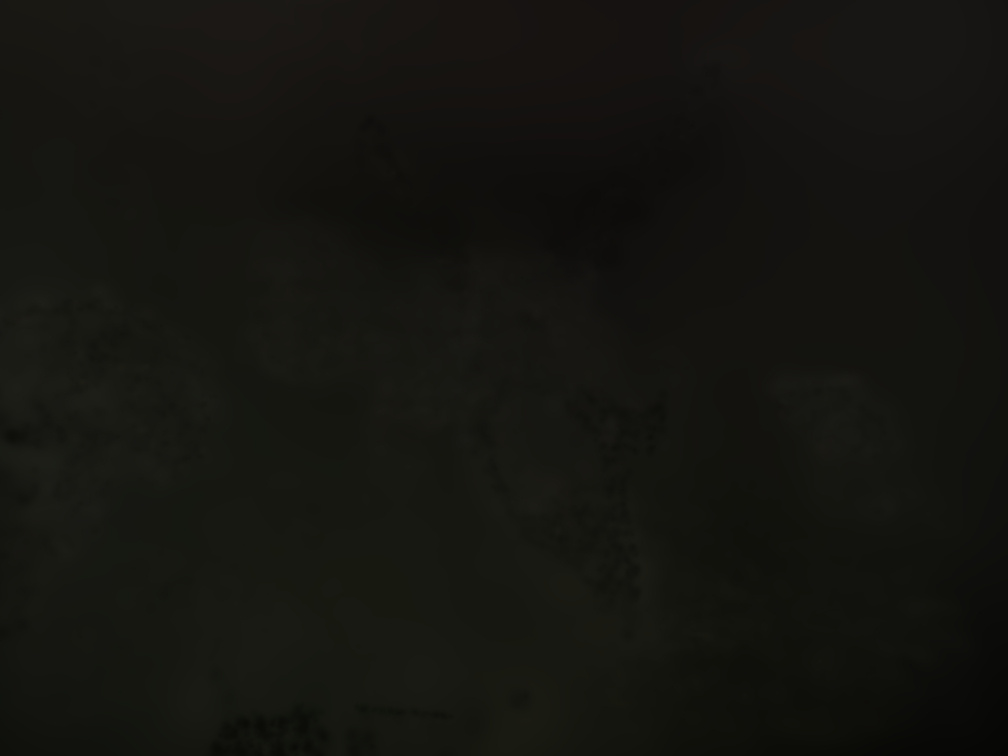} &
        \pcell[width=\linewidth]{16.6}{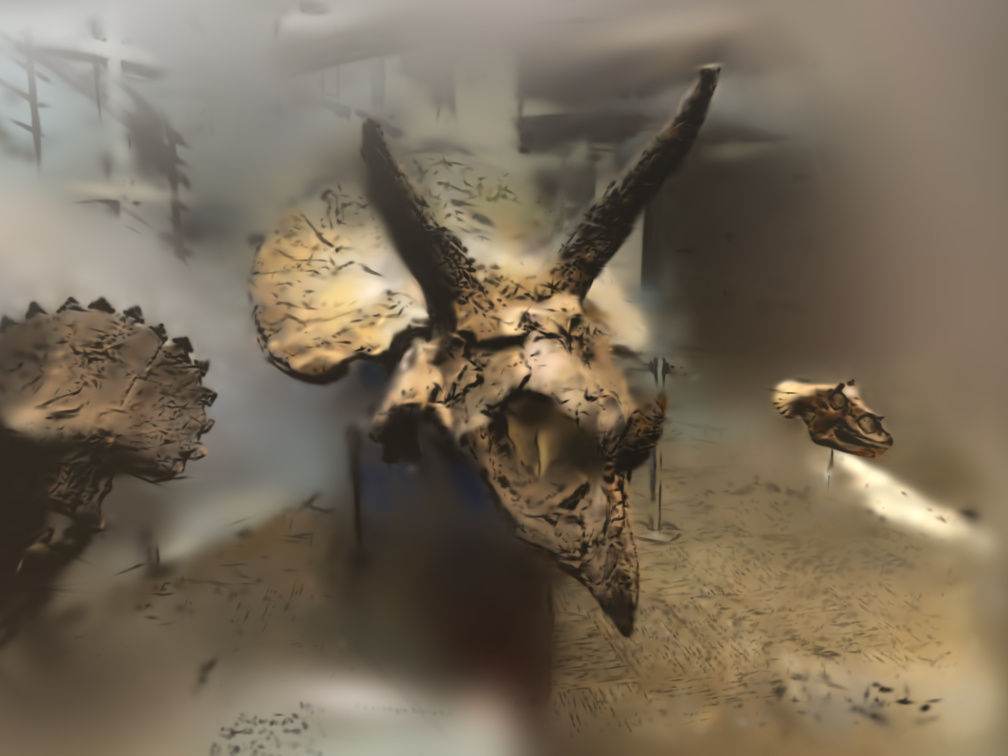} &
        \pcell[width=\linewidth]{23.5}{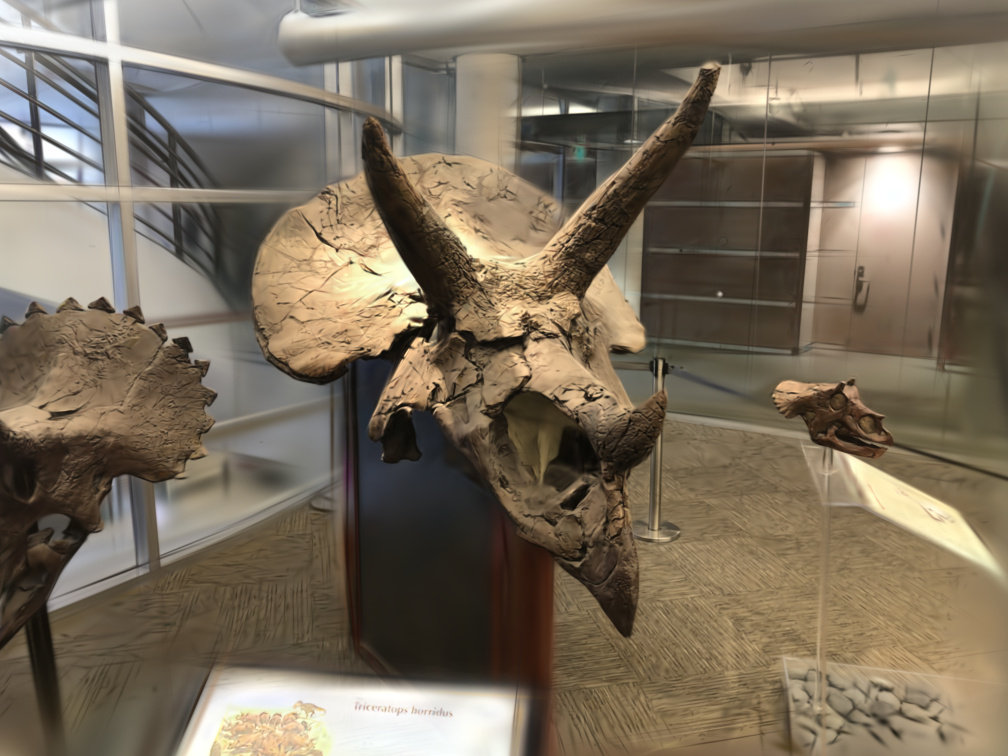} &
        \pcell[width=\linewidth]{25.7}{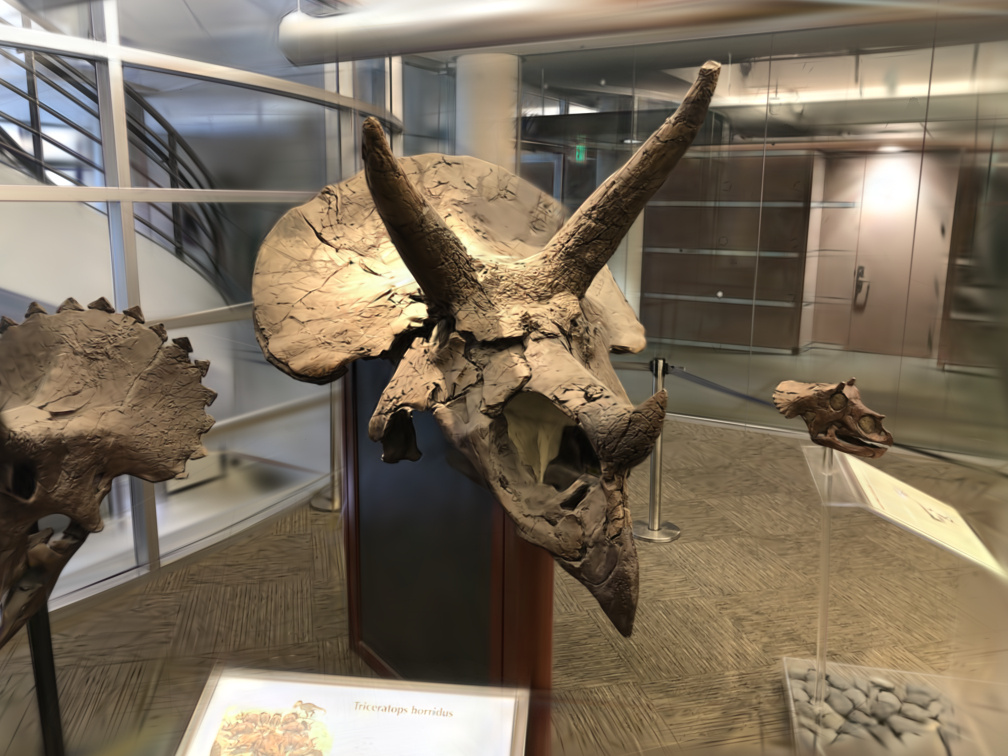} &
        \pcell[width=\linewidth]{26.1}{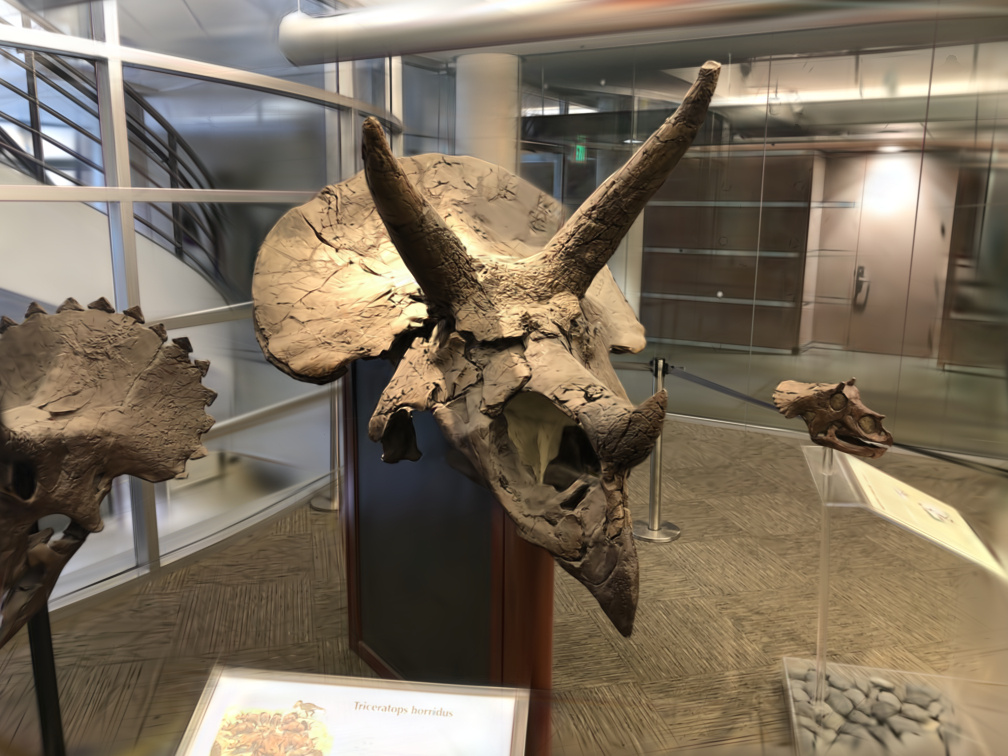} &
         &
         \\
        \noalign{\vskip 2\tabcolsep}
        \rotatebox{90}{\makecell{\smaller[1]{\adam{} error}}} &
        \rcell[width=\linewidth]{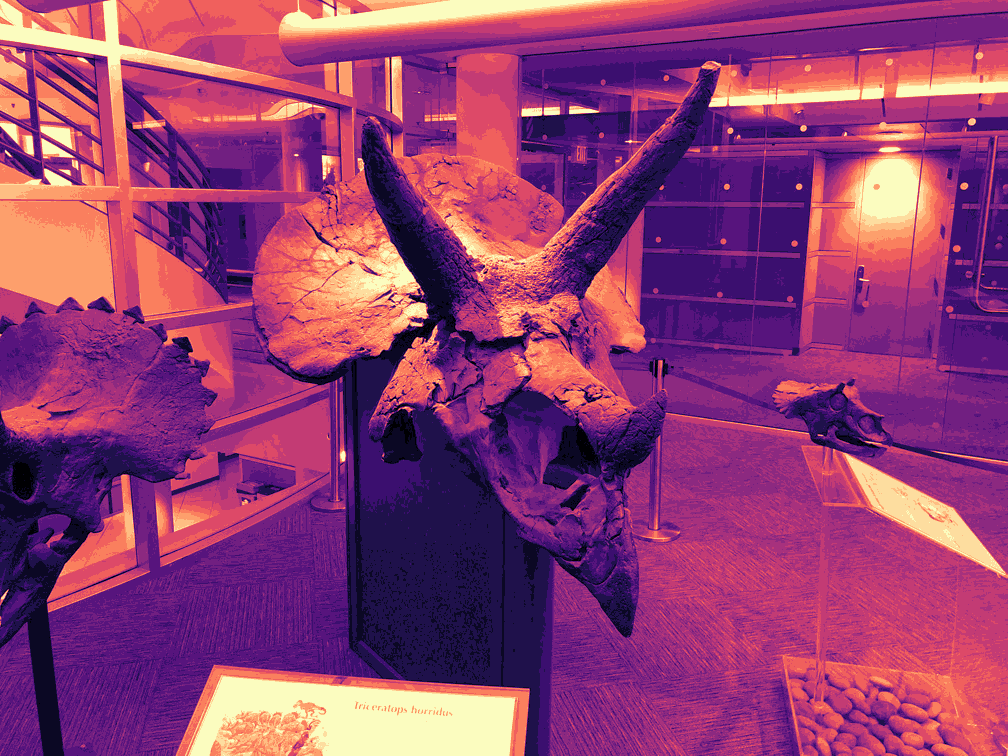} &
        \rcell[width=\linewidth]{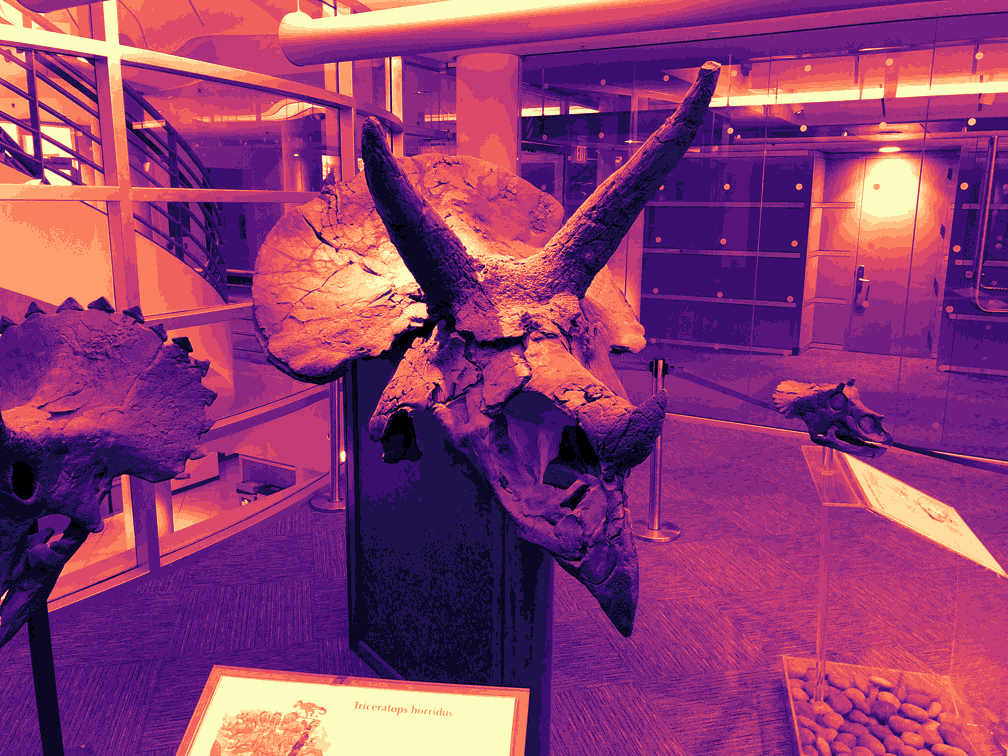} &
        \rcell[width=\linewidth]{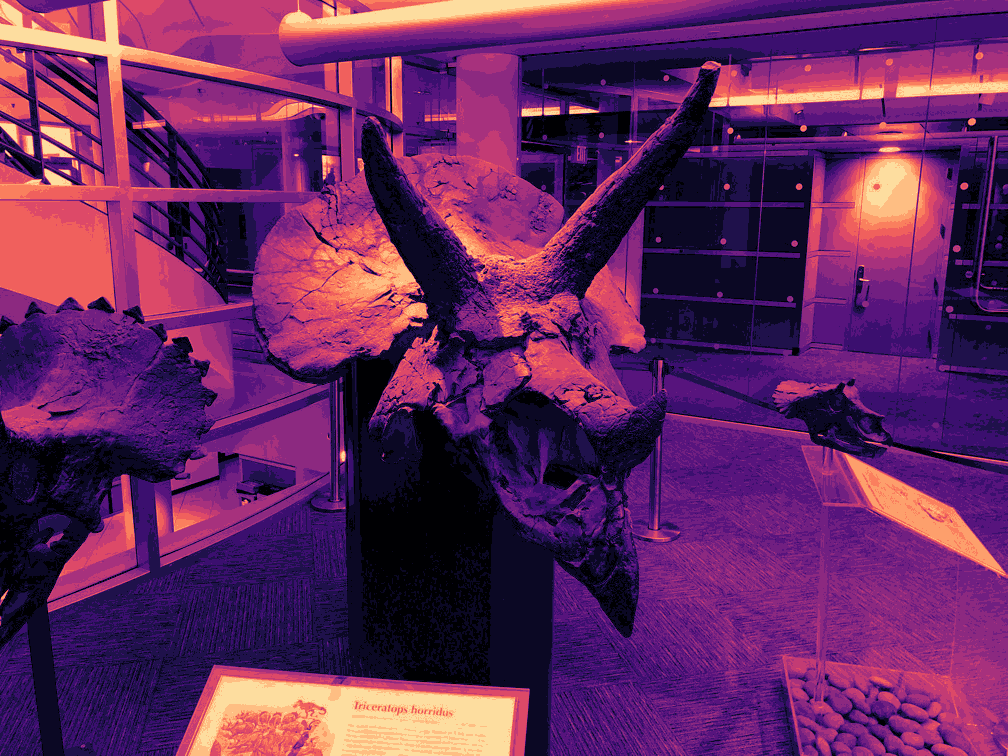} &
        \rcell[width=\linewidth]{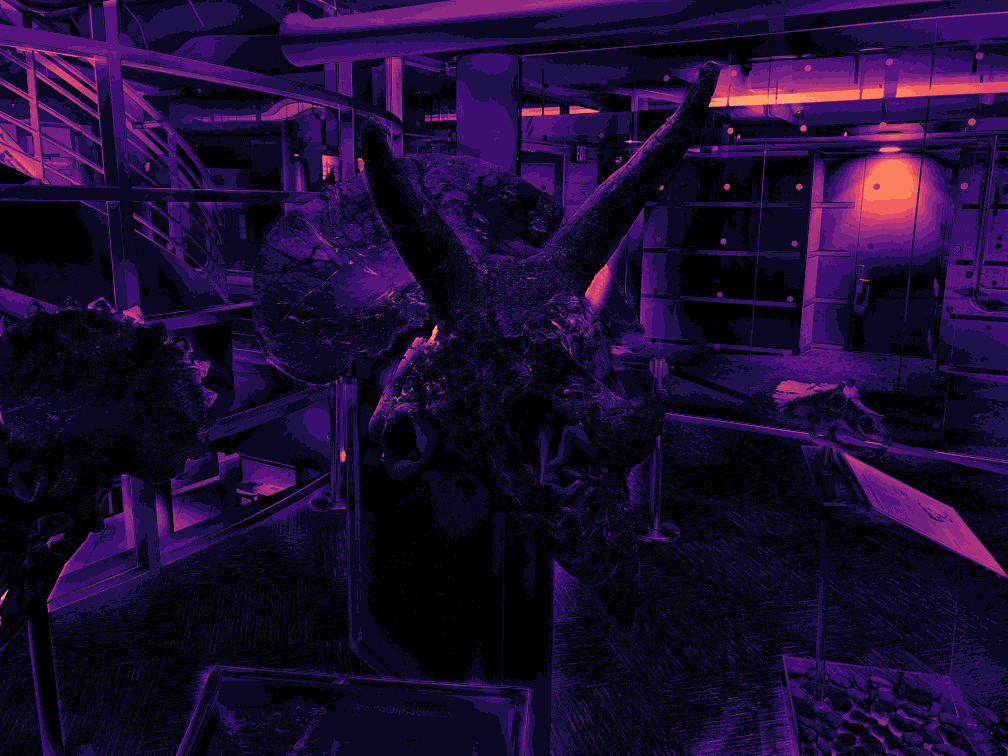} &
        \rcell[width=\linewidth]{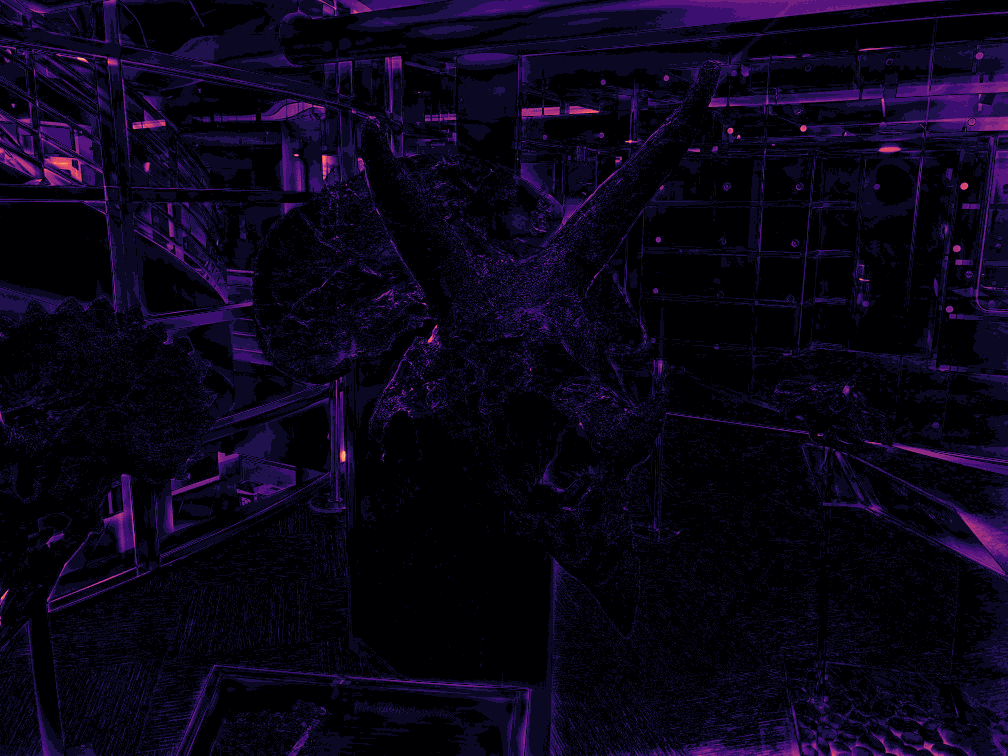} &
        \rotatebox{90}{\smaller[1]{Init error}} &
        \rcell[width=\linewidth]{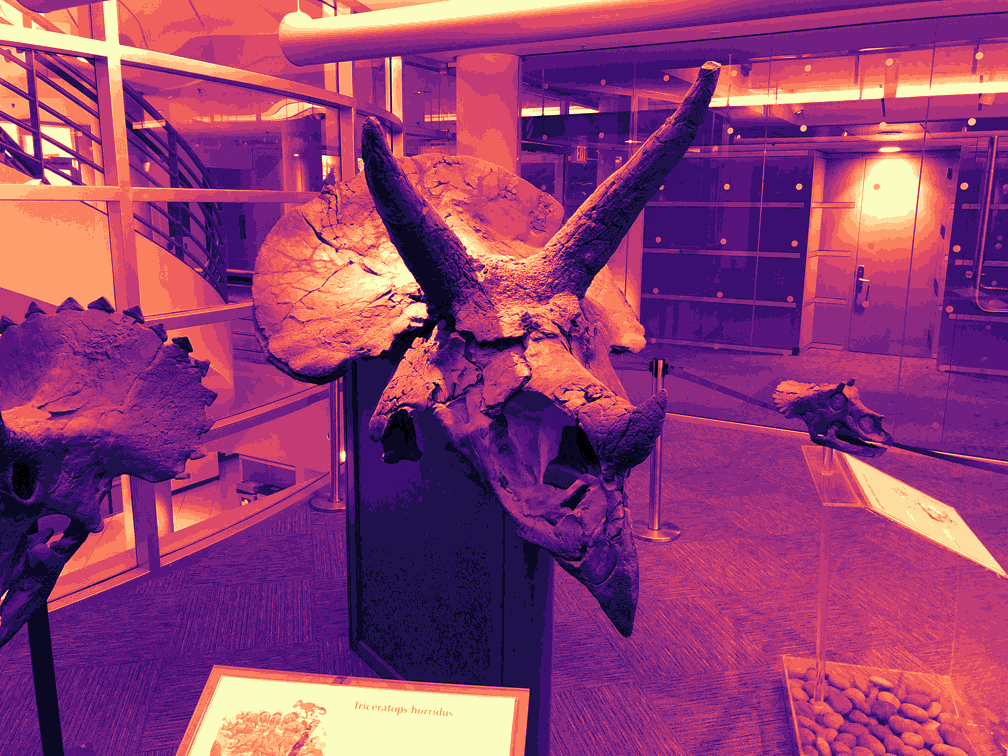} \\
        \noalign{\vskip 2\tabcolsep}
        \rotatebox{90}{\makecell{\smaller[1]{\ourshort{} error}}} &
        \rcell[width=\linewidth]{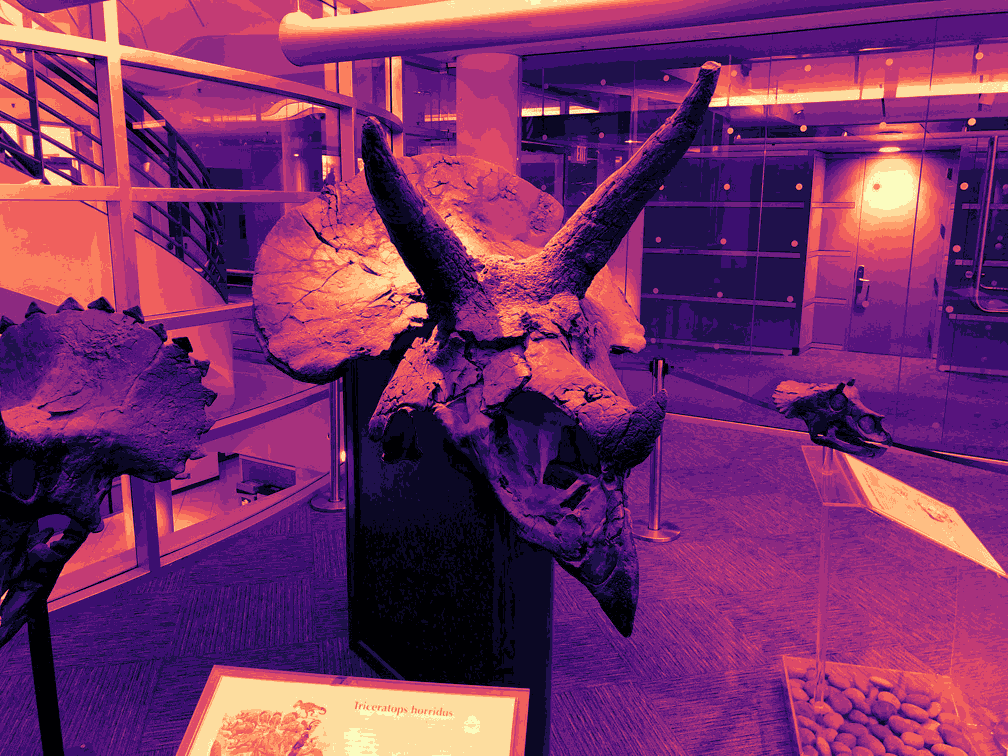} &
        \rcell[width=\linewidth]{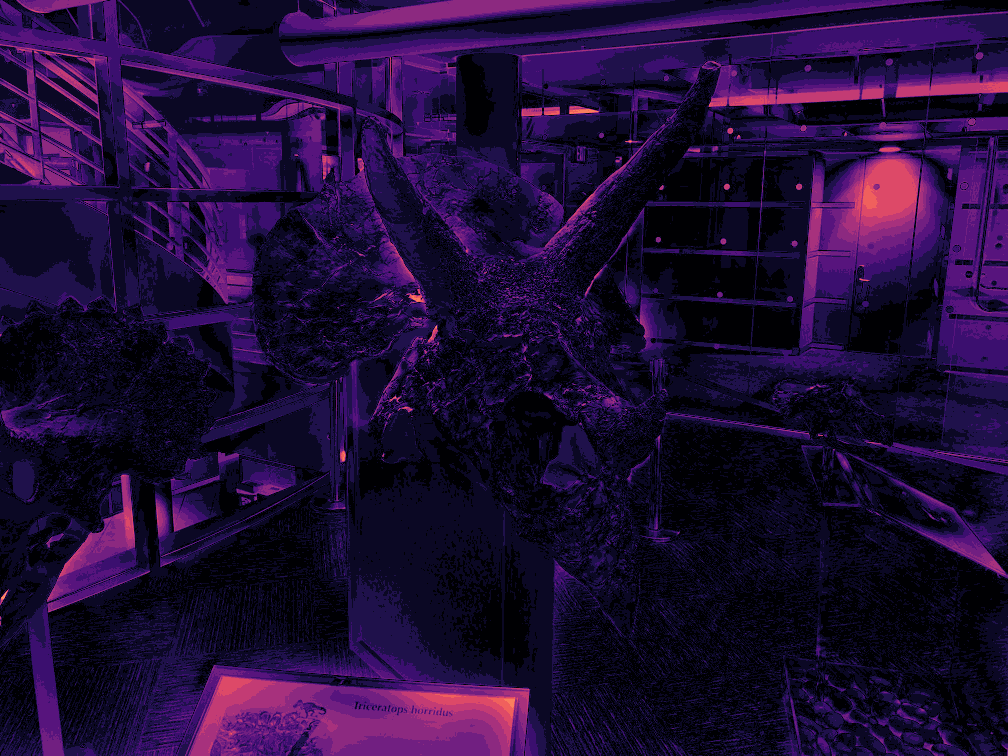} &
        \rcell[width=\linewidth]{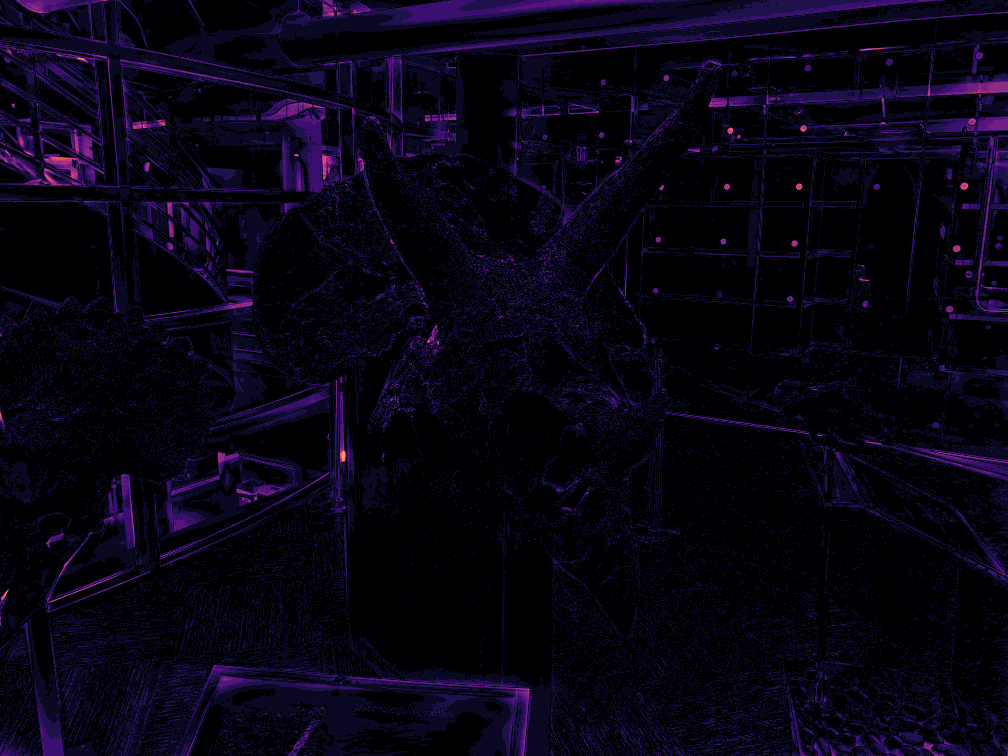} &
        \rcell[width=\linewidth]{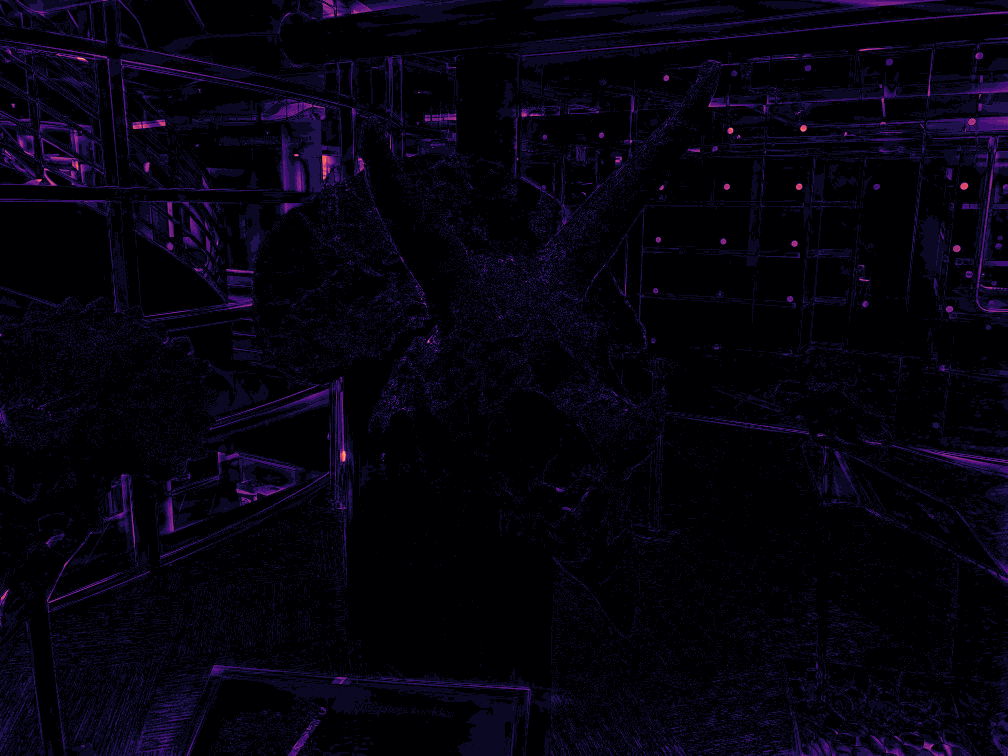} &
        \rcell[width=\linewidth]{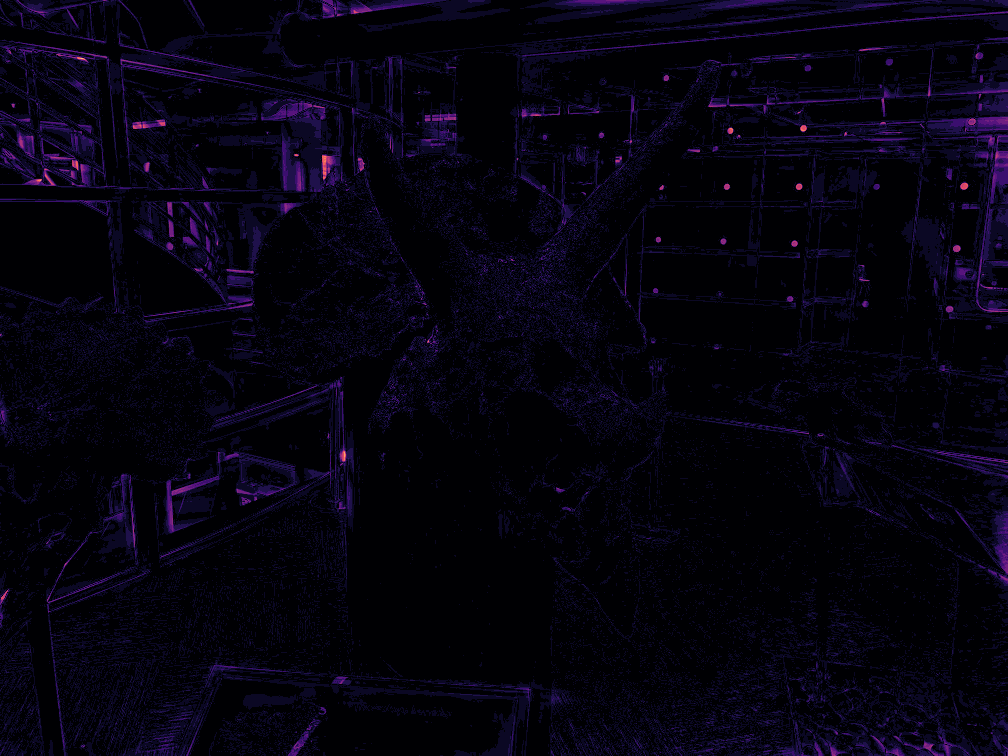} &
         &
         \\
        \noalign{\vskip 2\tabcolsep}
        \rotatebox{90}{\makecell{\smaller[1]{\lob{}}}} &
        \pcell[width=\linewidth]{18.8}{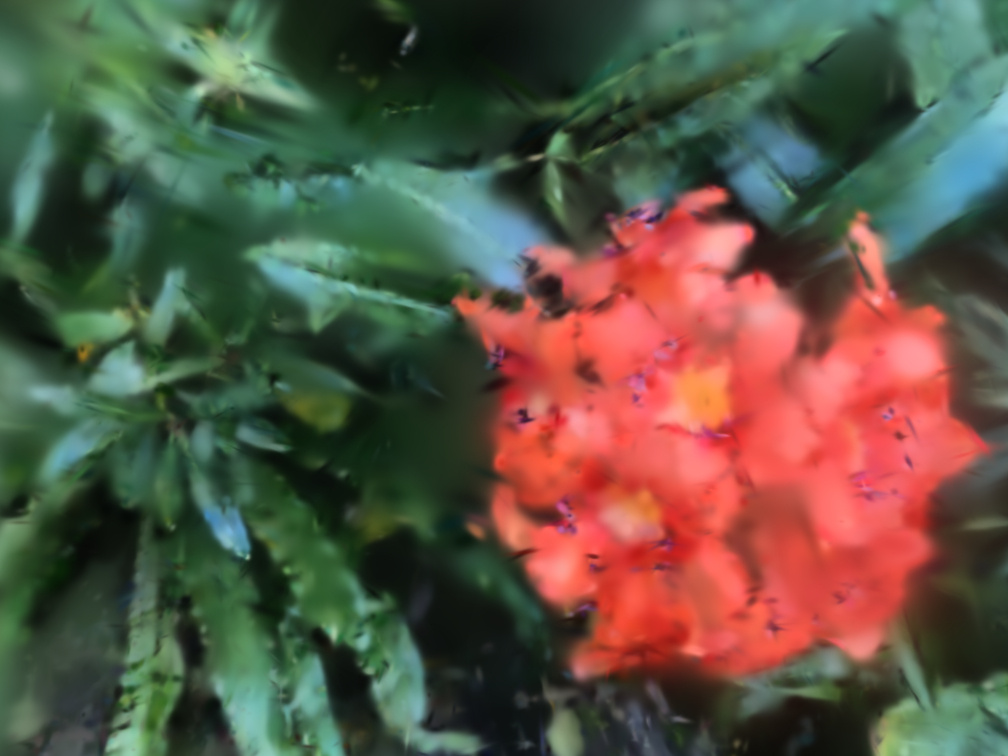} &
        \pcell[width=\linewidth]{22.6}{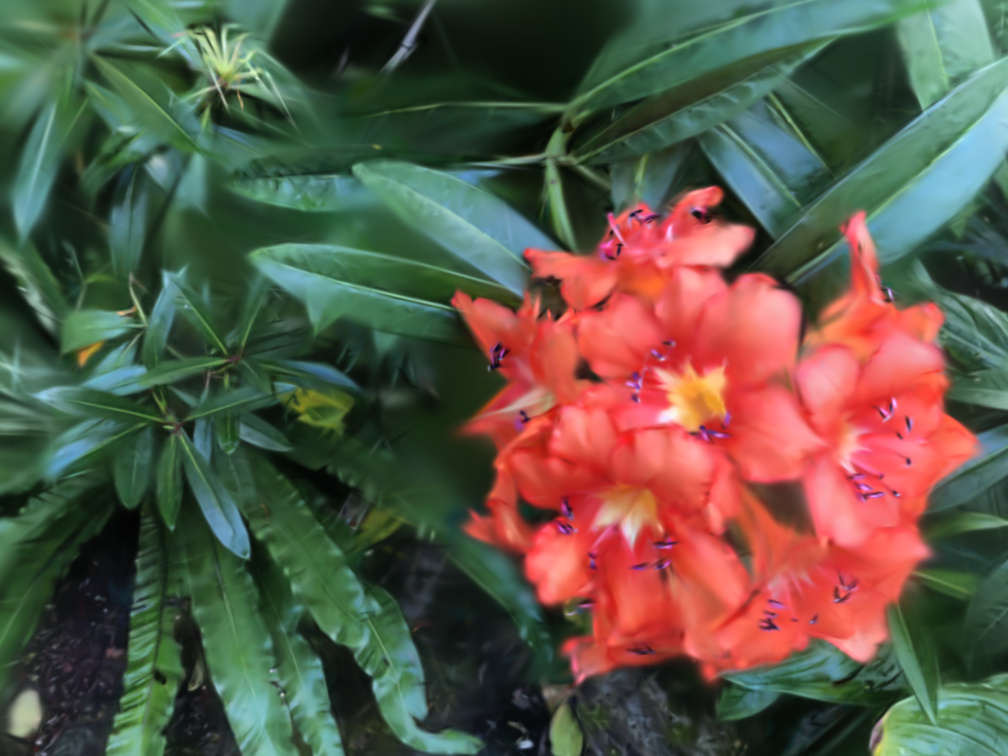} &
        \pcell[width=\linewidth]{24.3}{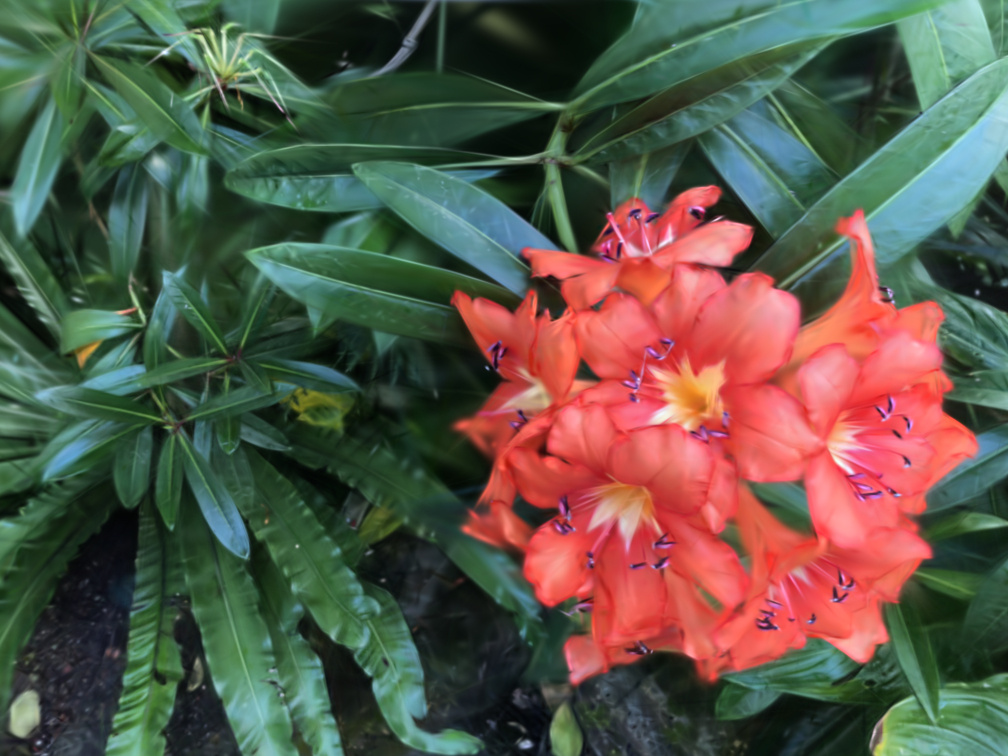} &
        \pcell[width=\linewidth]{23.7}{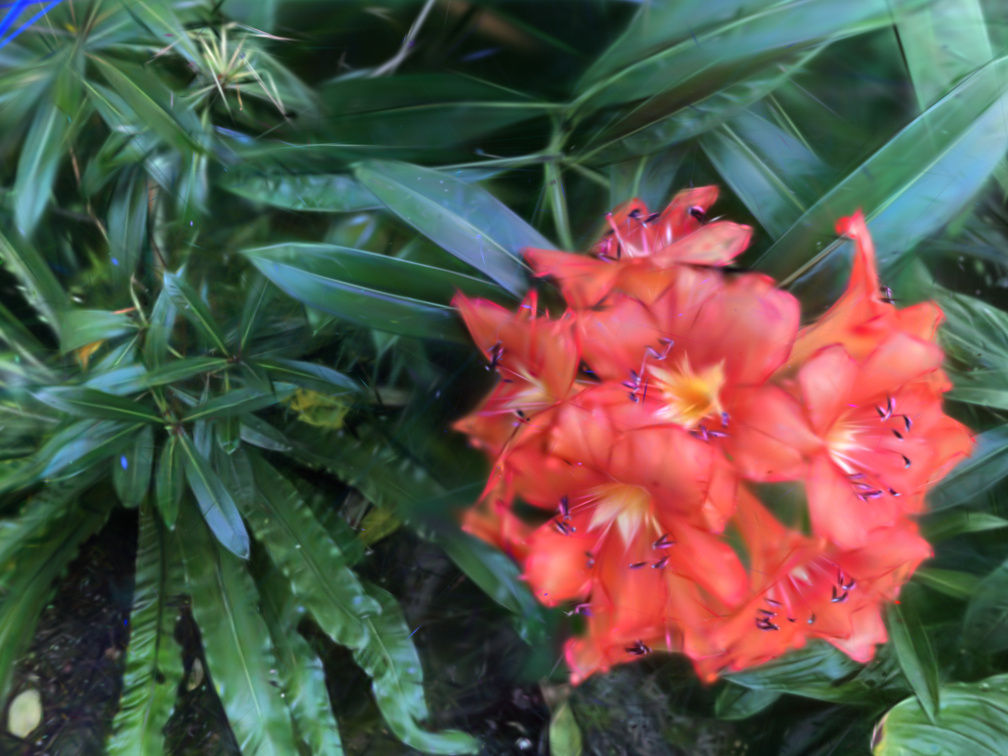} &
        \pcell[width=\linewidth]{23.7}{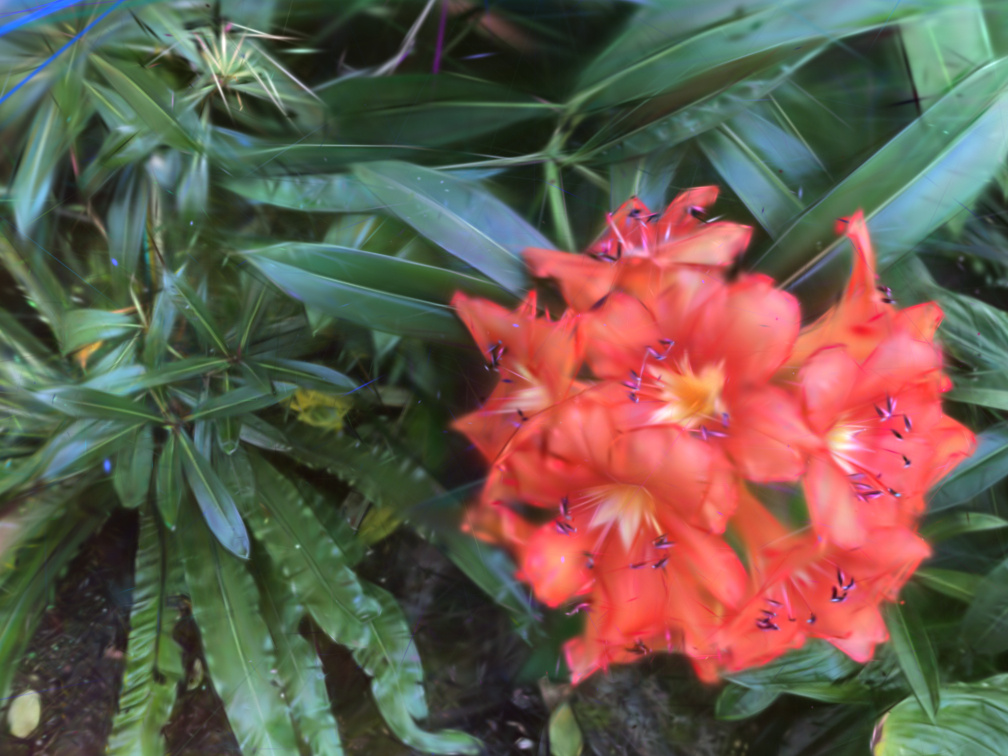} &
        \rotatebox{90}{\smaller[1]{Init}} &
        \pcell[width=\linewidth]{14.4}{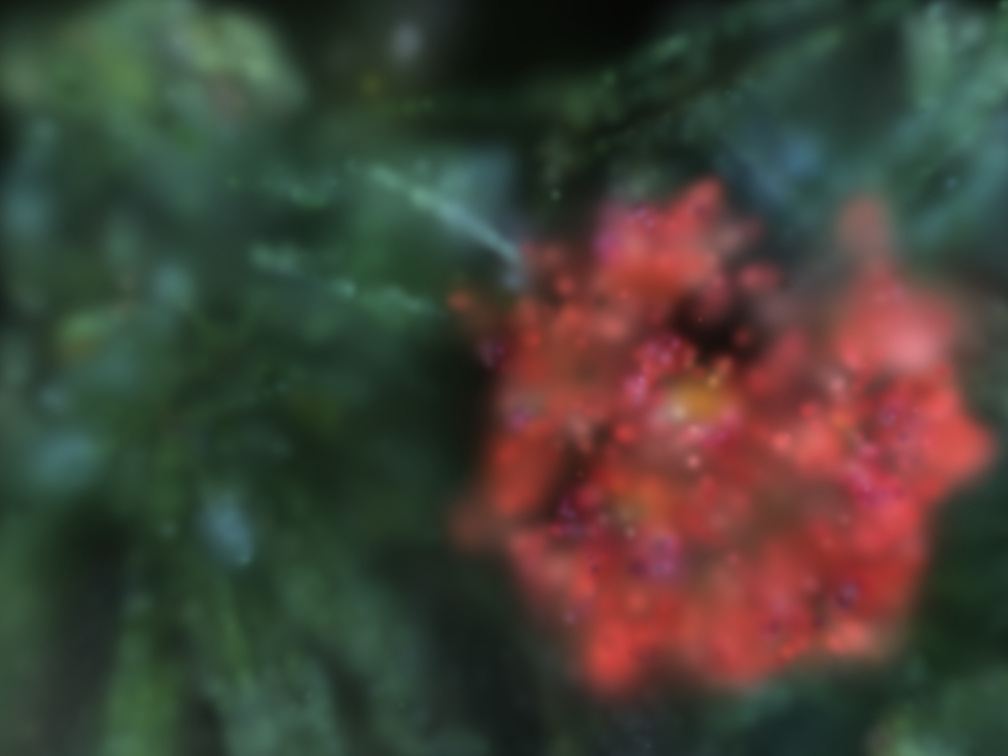} \\
        \noalign{\vskip 2\tabcolsep}
        \rotatebox{90}{\makecell{\smaller[1]{\adam{}}}} &
        \pcell[width=\linewidth]{15.0}{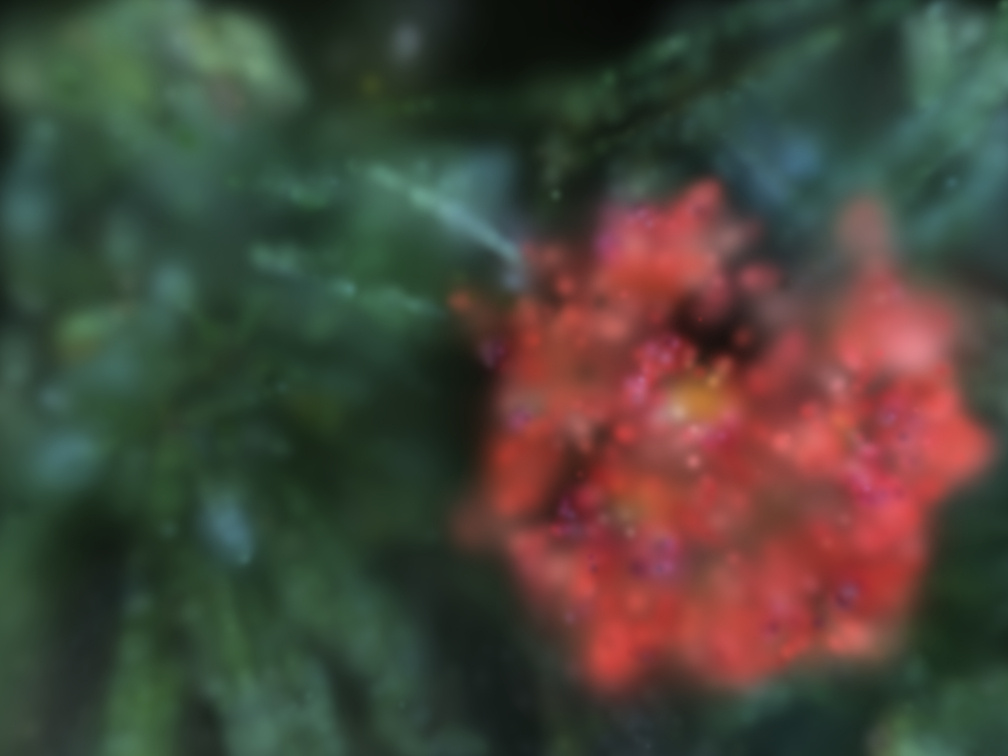} &
        \pcell[width=\linewidth]{16.6}{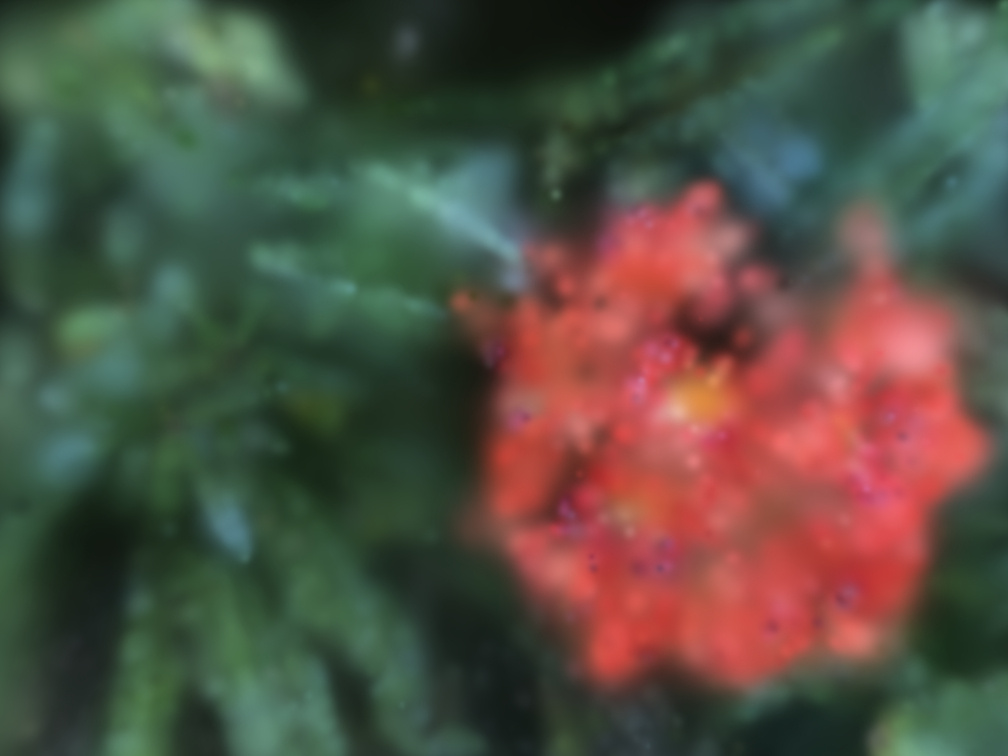} &
        \pcell[width=\linewidth]{20.6}{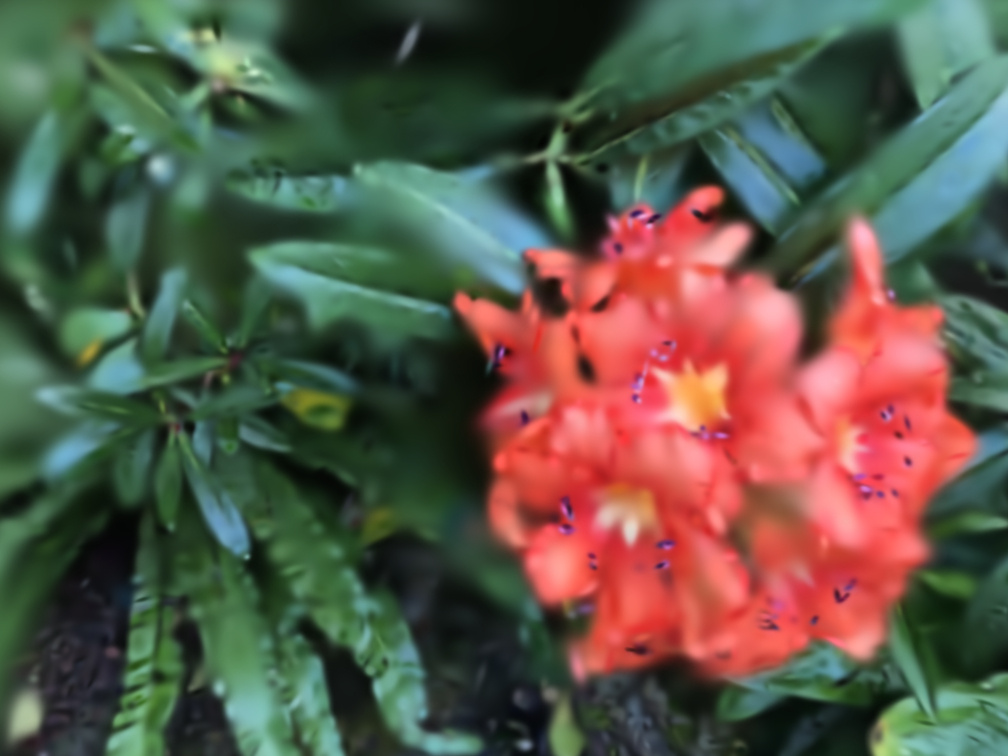} &
        \pcell[width=\linewidth]{24.4}{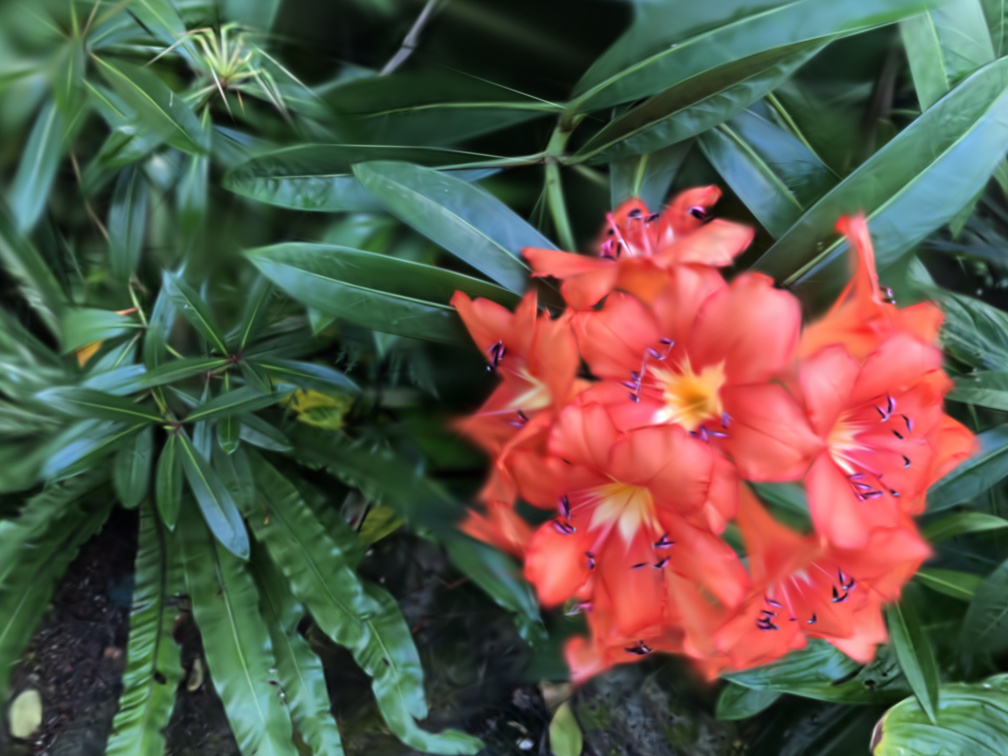} &
        \pcell[width=\linewidth]{24.9}{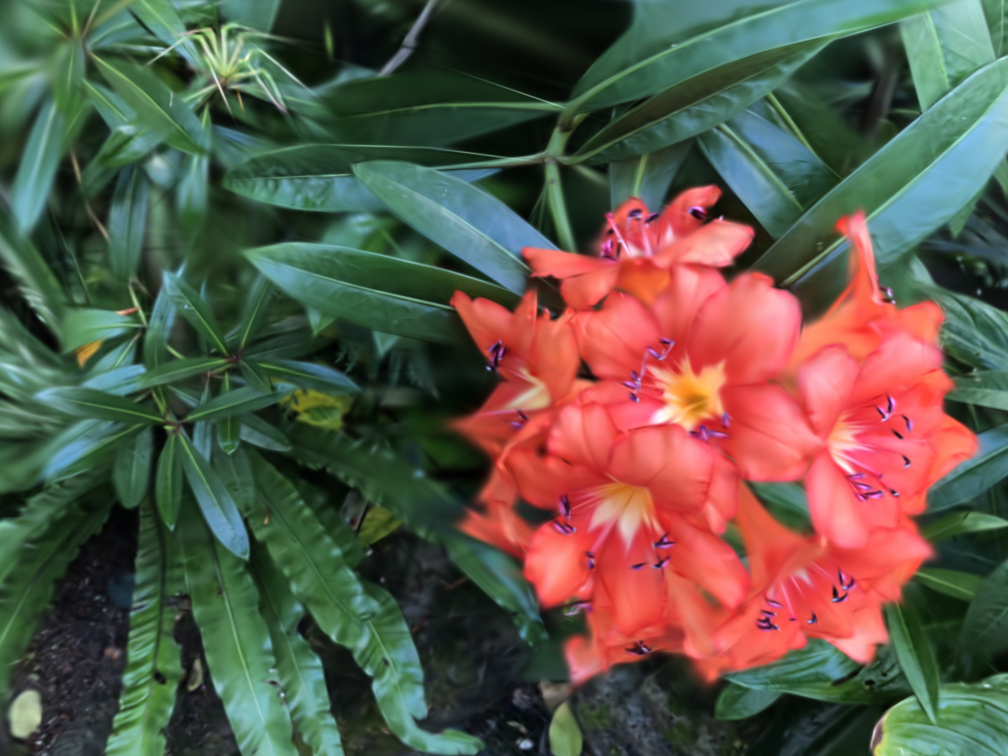} &
        \rotatebox{90}{\smaller[1]{Ref}} &
        \rcell[width=\linewidth]{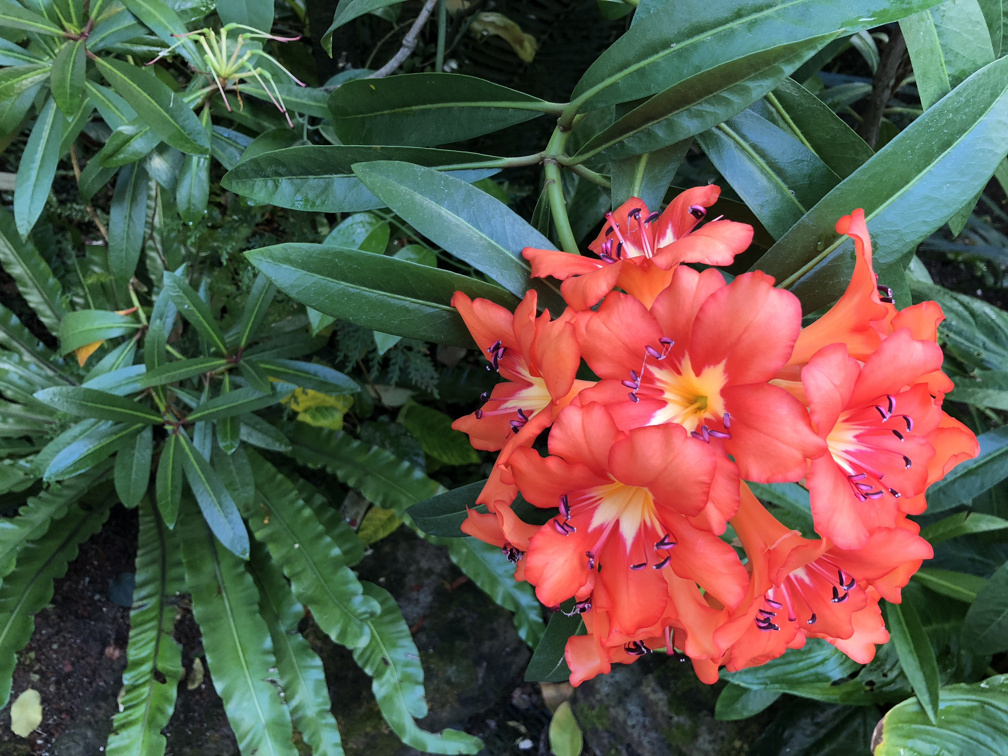} \\
        \noalign{\vskip 2\tabcolsep}
        \rotatebox{90}{\makecell{\smaller[1]{\ourshort{}}}} &
        \pcell[width=\linewidth]{18.4}{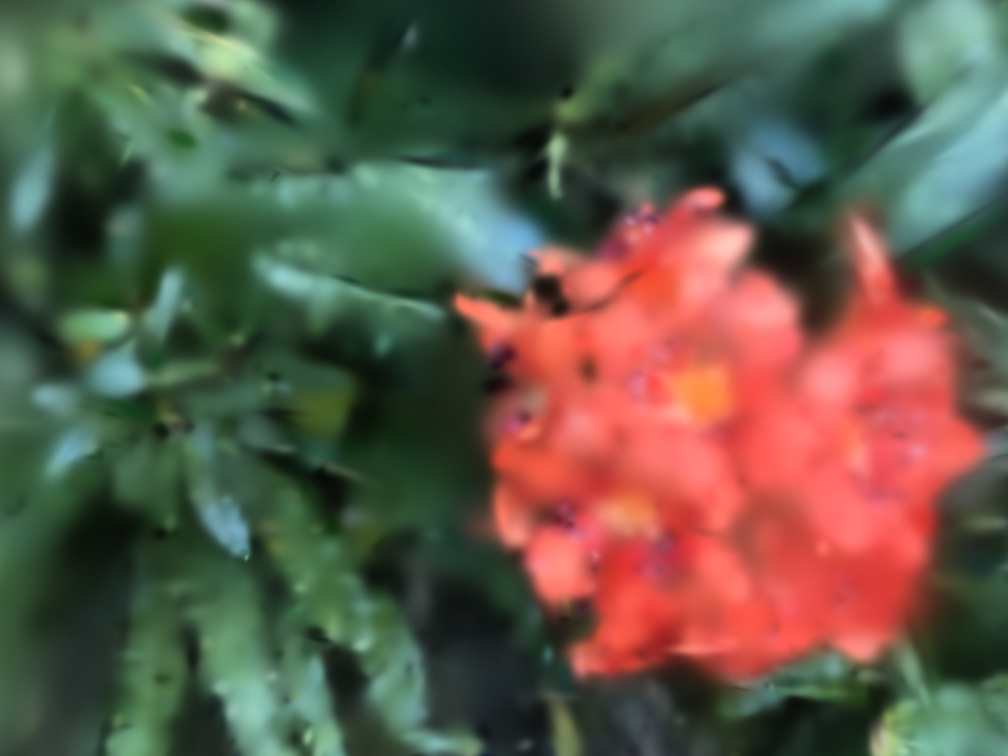} &
        \pcell[width=\linewidth]{22.0}{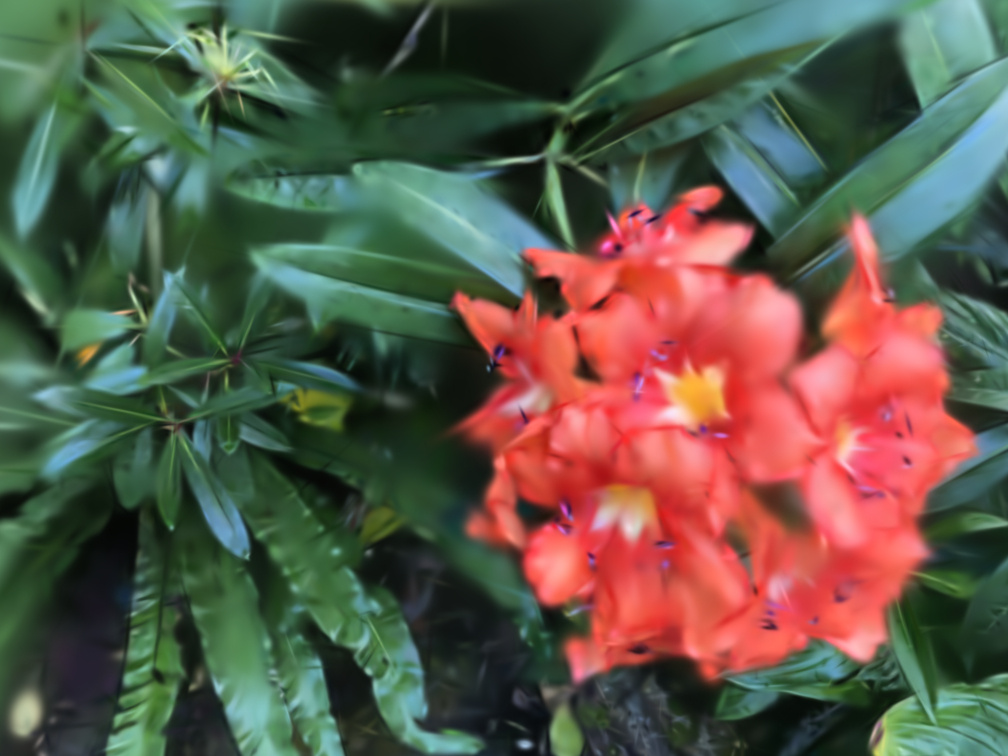} &
        \pcell[width=\linewidth]{25.1}{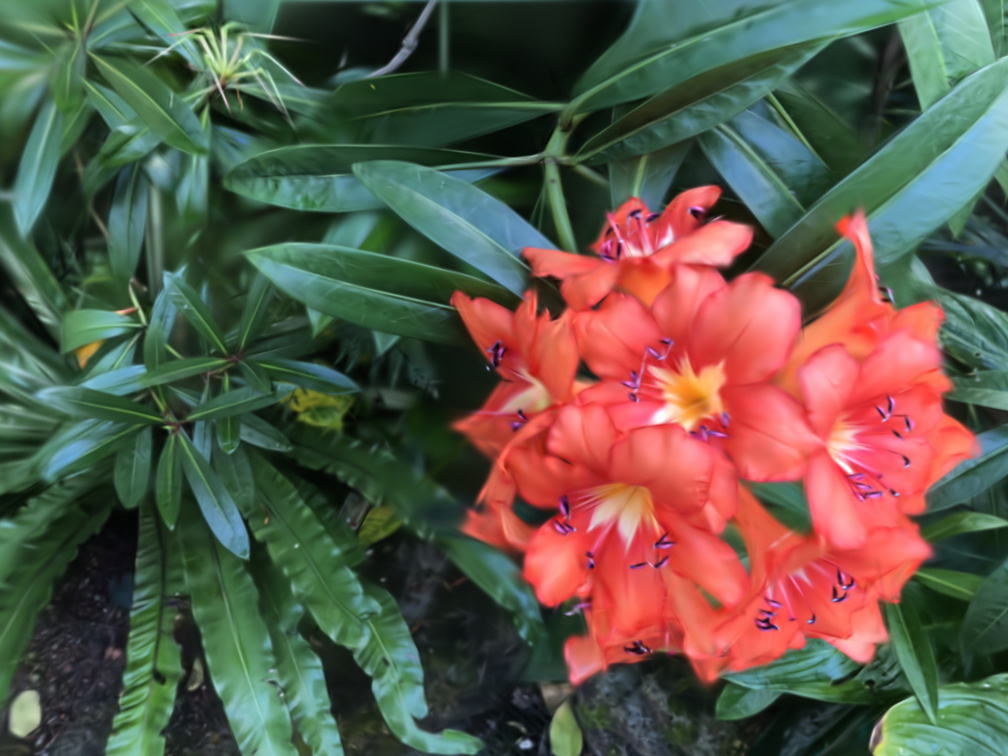} &
        \pcell[width=\linewidth]{26.3}{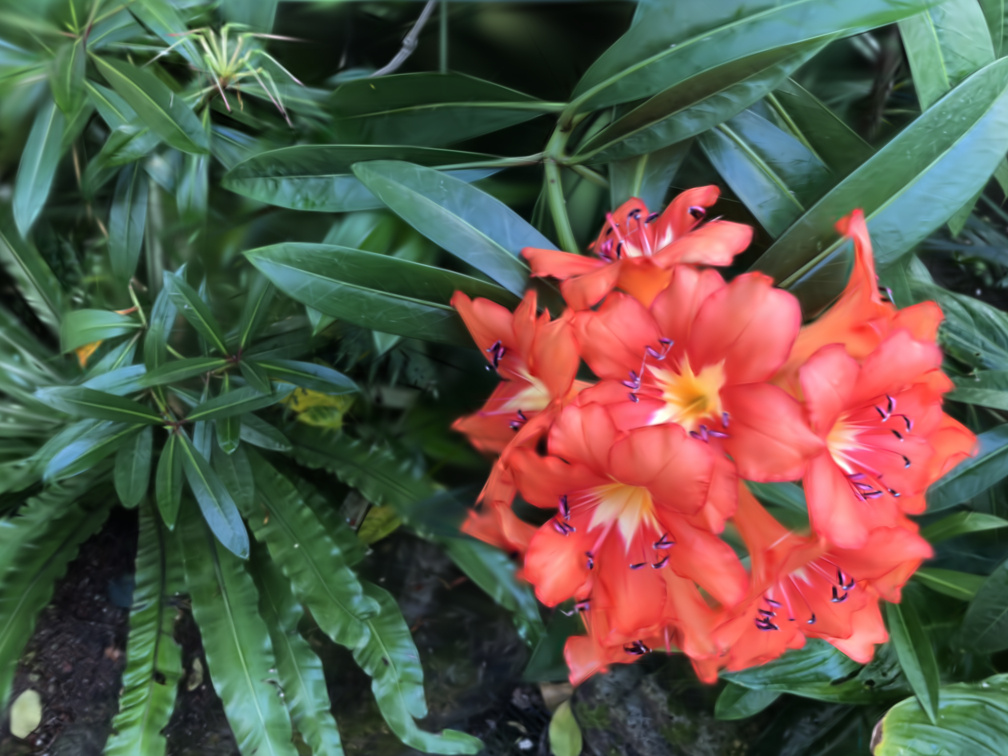} &
        \pcell[width=\linewidth]{26.2}{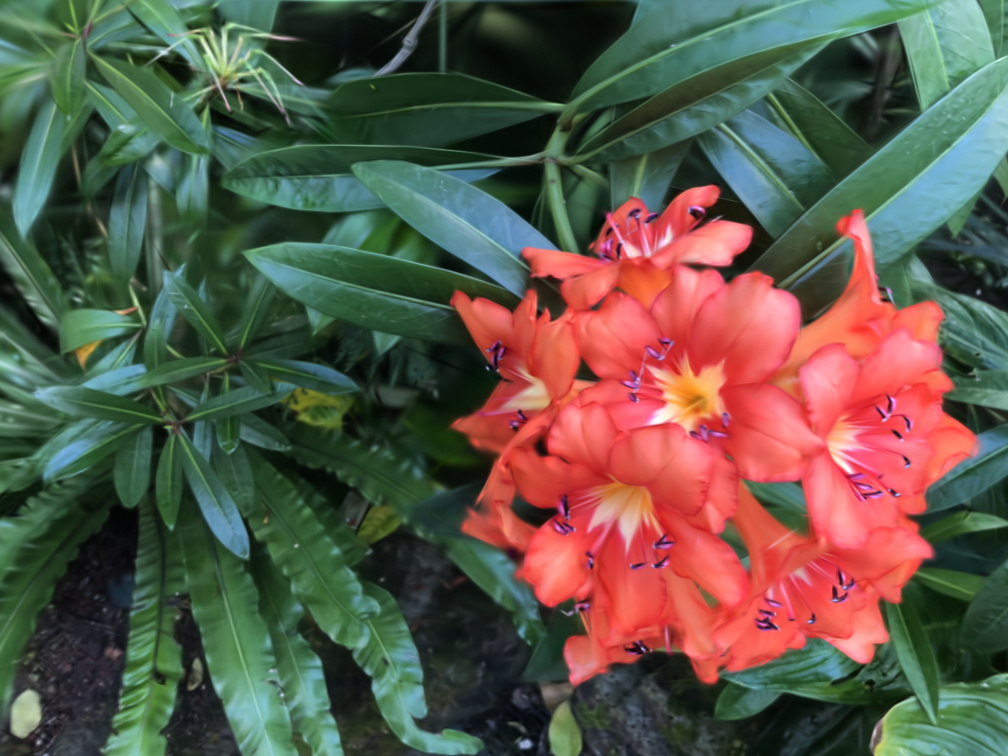} &
         &
         \\
        \noalign{\vskip 2\tabcolsep}
        \rotatebox{90}{\makecell{\smaller[1]{\adam{} error}}} &
        \rcell[width=\linewidth]{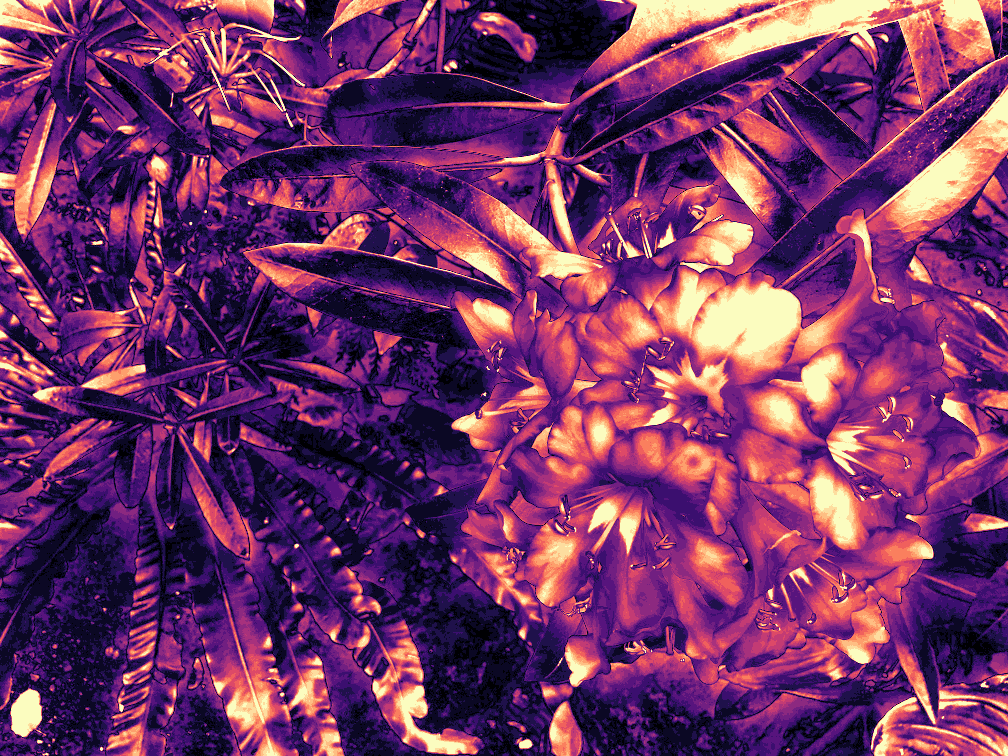} &
        \rcell[width=\linewidth]{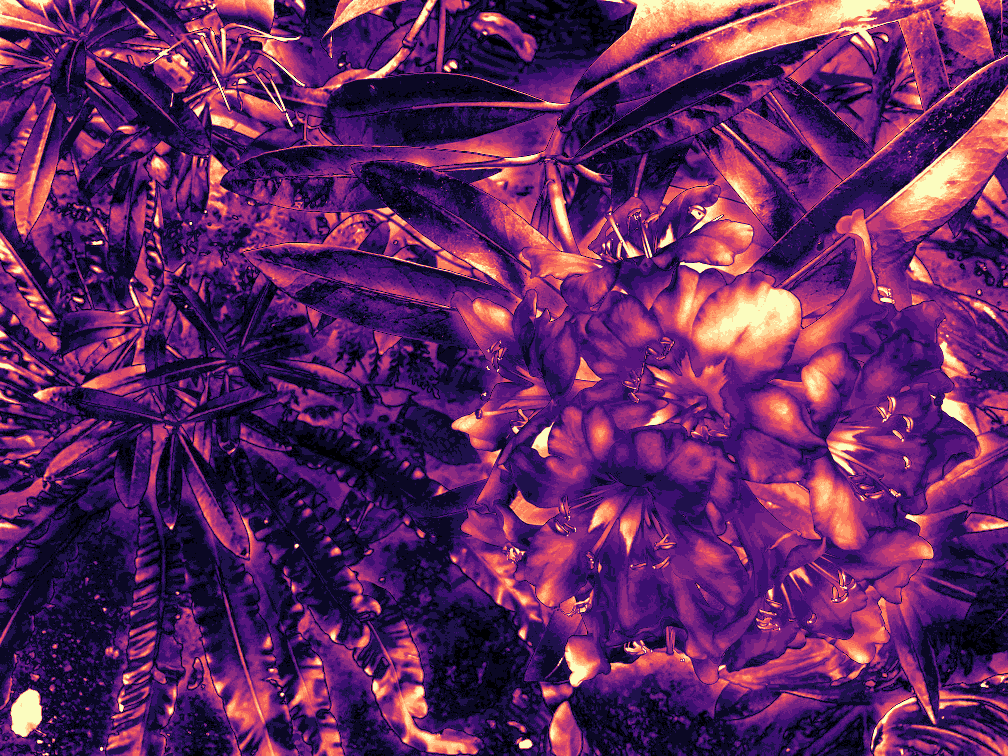} &
        \rcell[width=\linewidth]{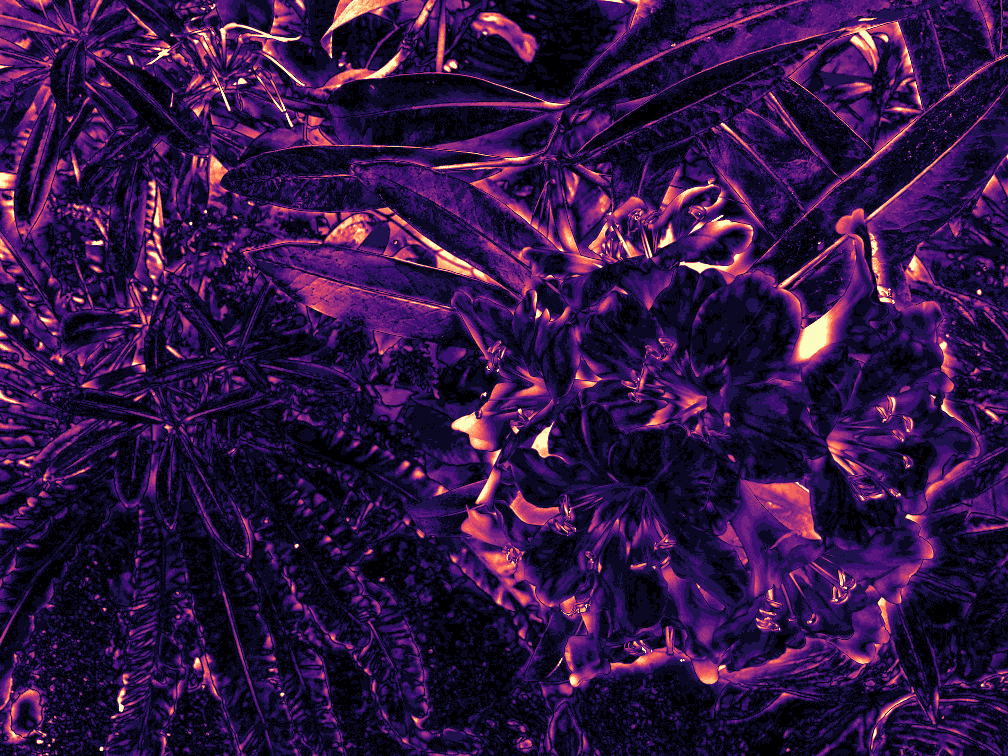} &
        \rcell[width=\linewidth]{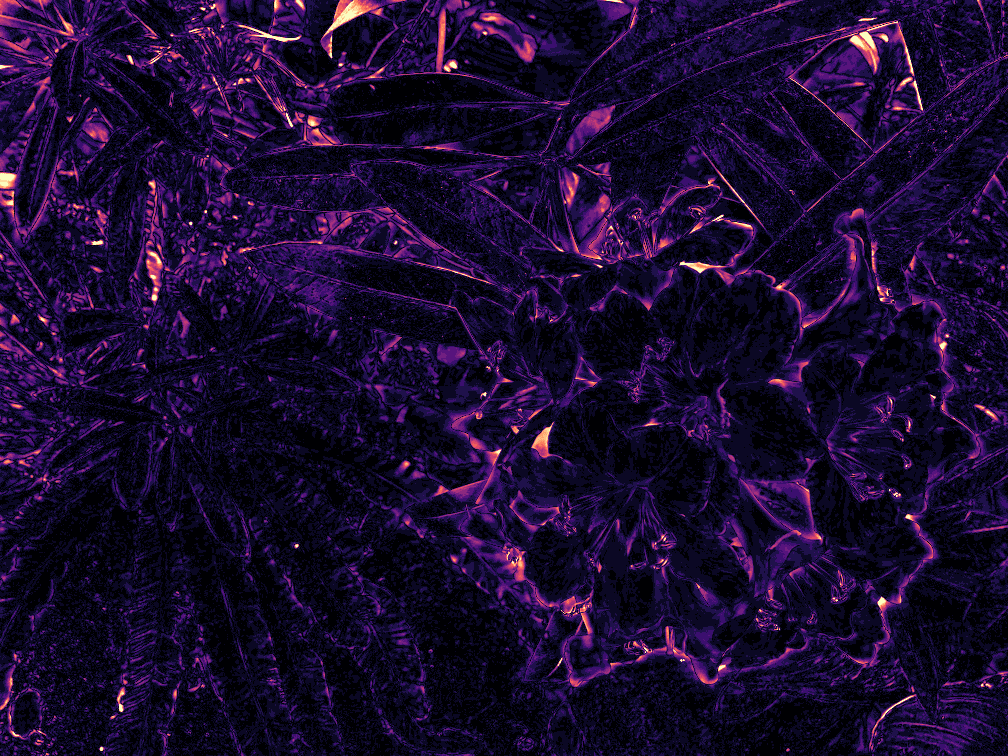} &
        \rcell[width=\linewidth]{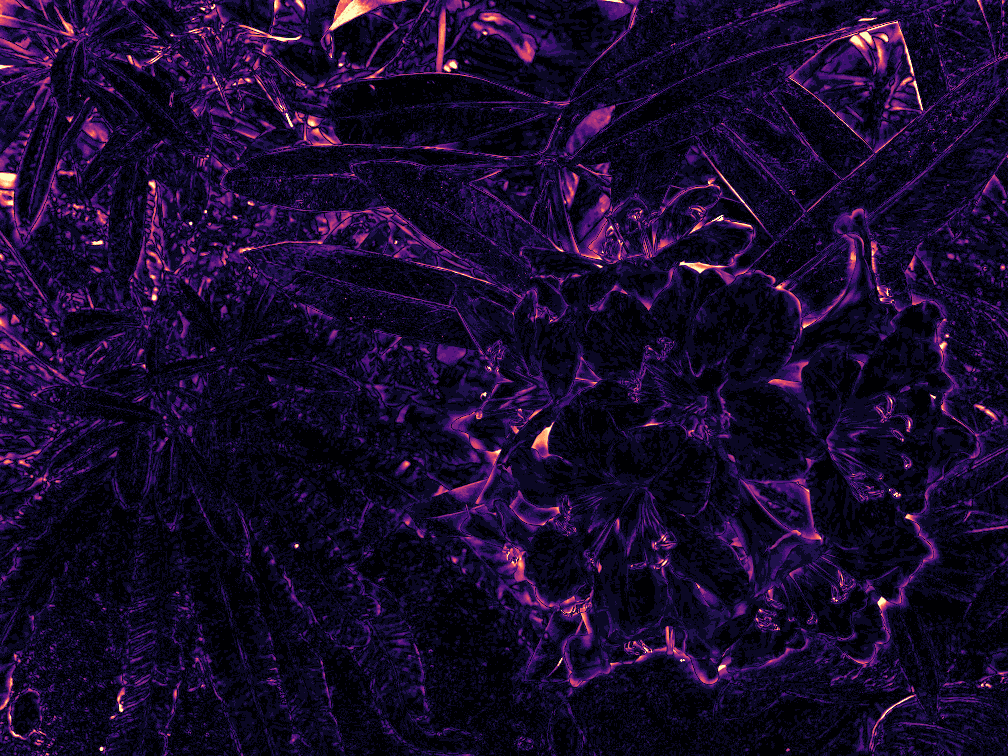} &
        \rotatebox{90}{\smaller[1]{Init error}} &
        \rcell[width=\linewidth]{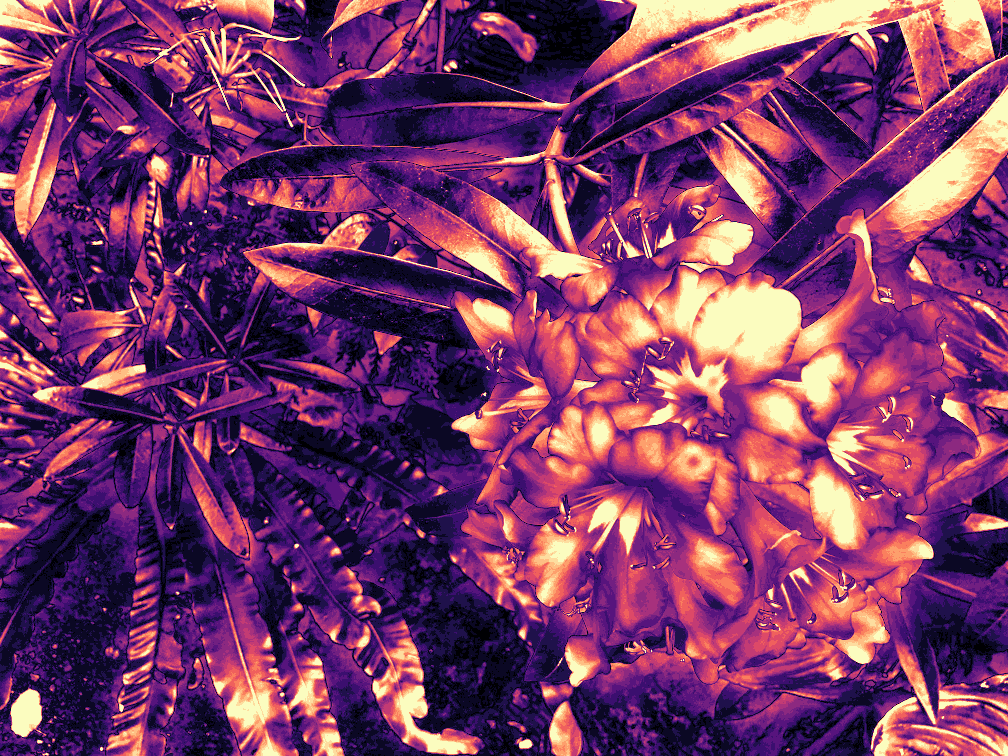} \\
        \noalign{\vskip 2\tabcolsep}
        \rotatebox{90}{\makecell{\smaller[1]{\ourshort{} error}}} &
        \rcell[width=\linewidth]{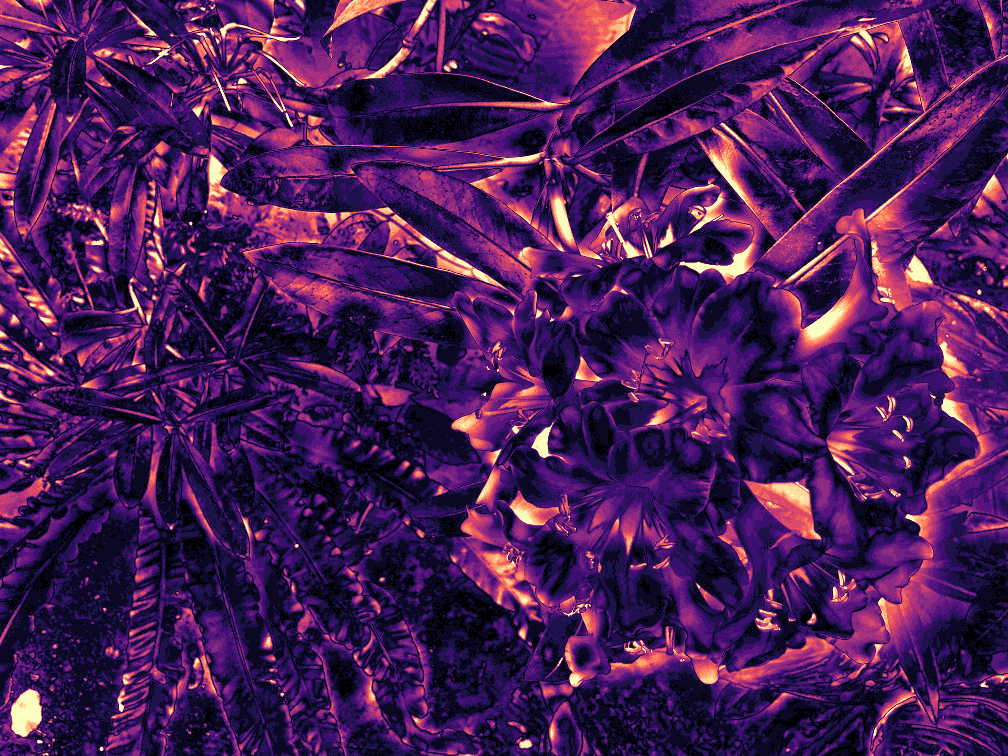} &
        \rcell[width=\linewidth]{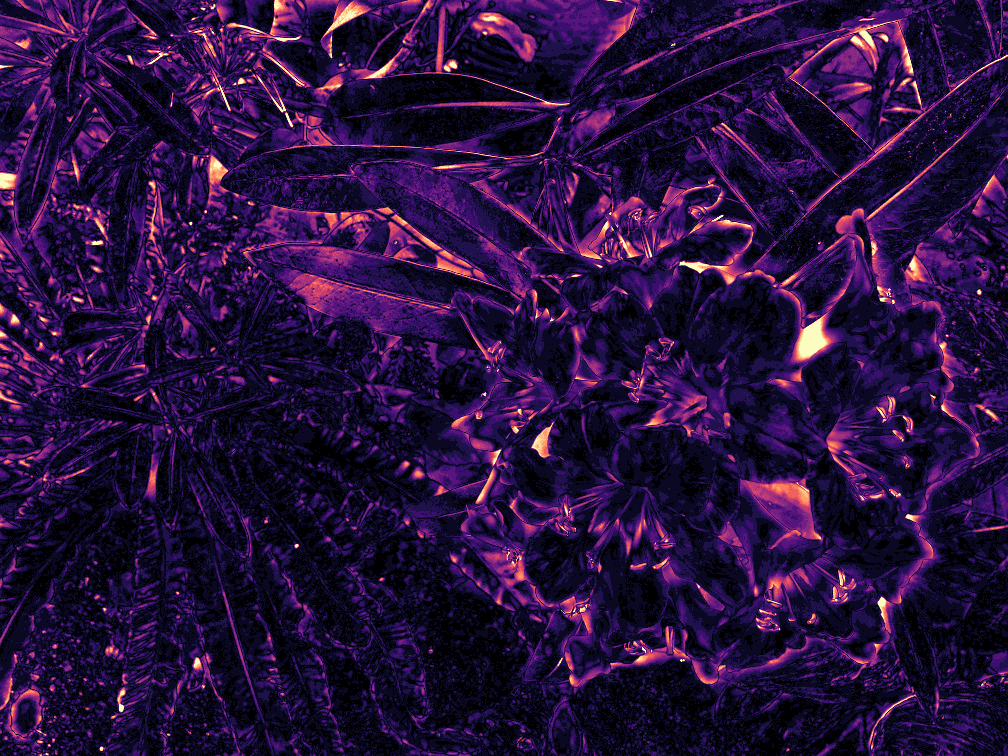} &
        \rcell[width=\linewidth]{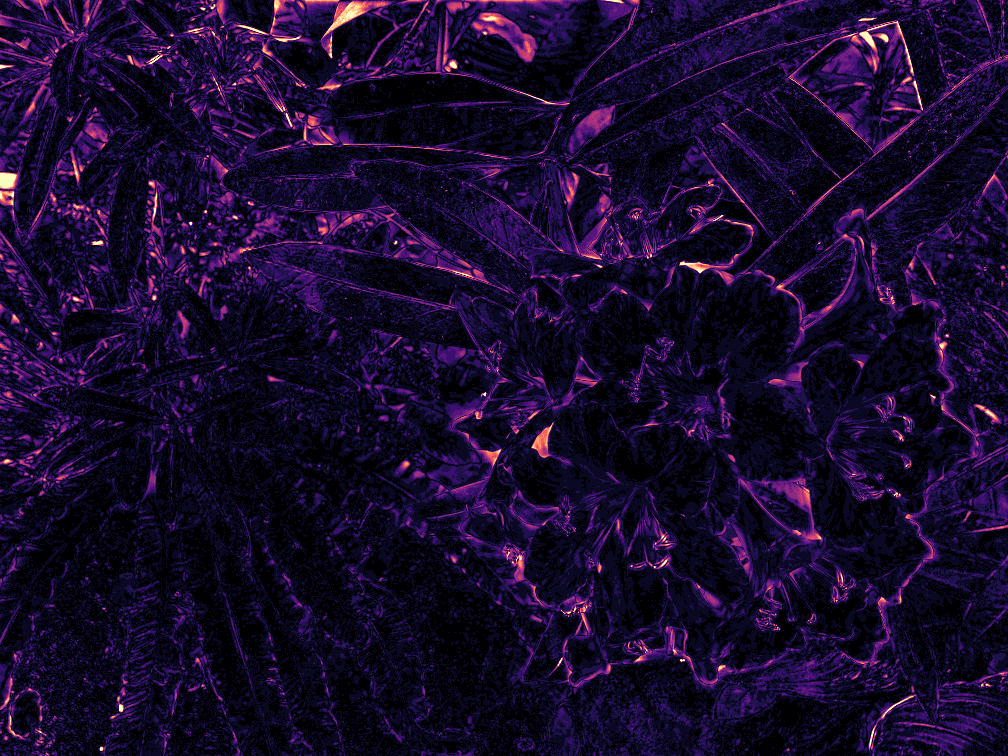} &
        \rcell[width=\linewidth]{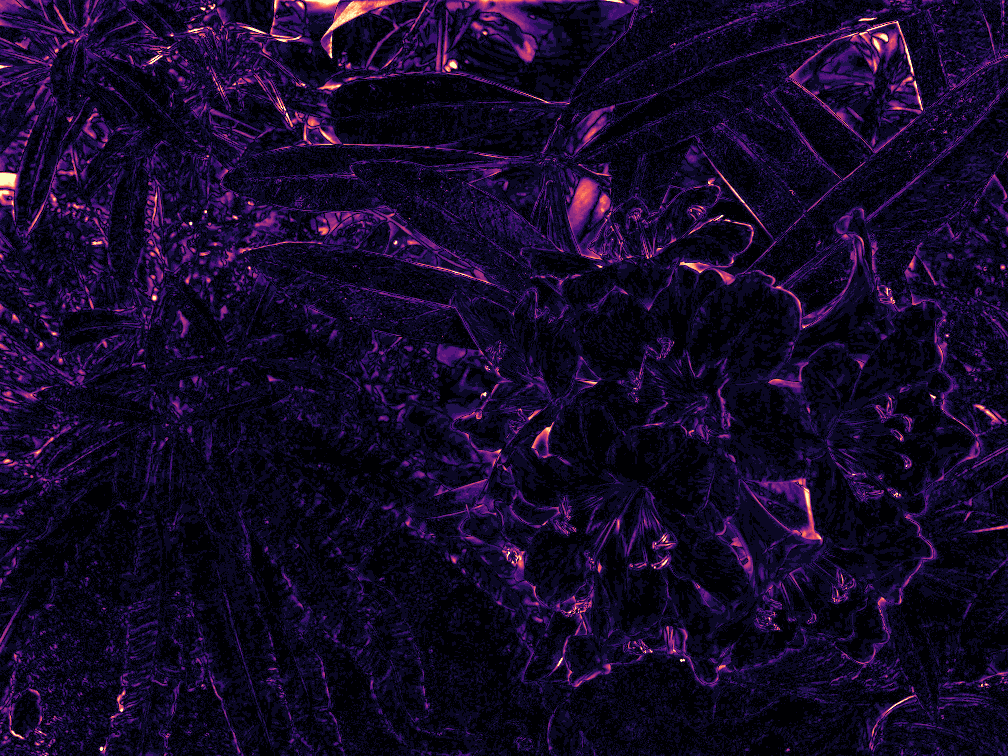} &
        \rcell[width=\linewidth]{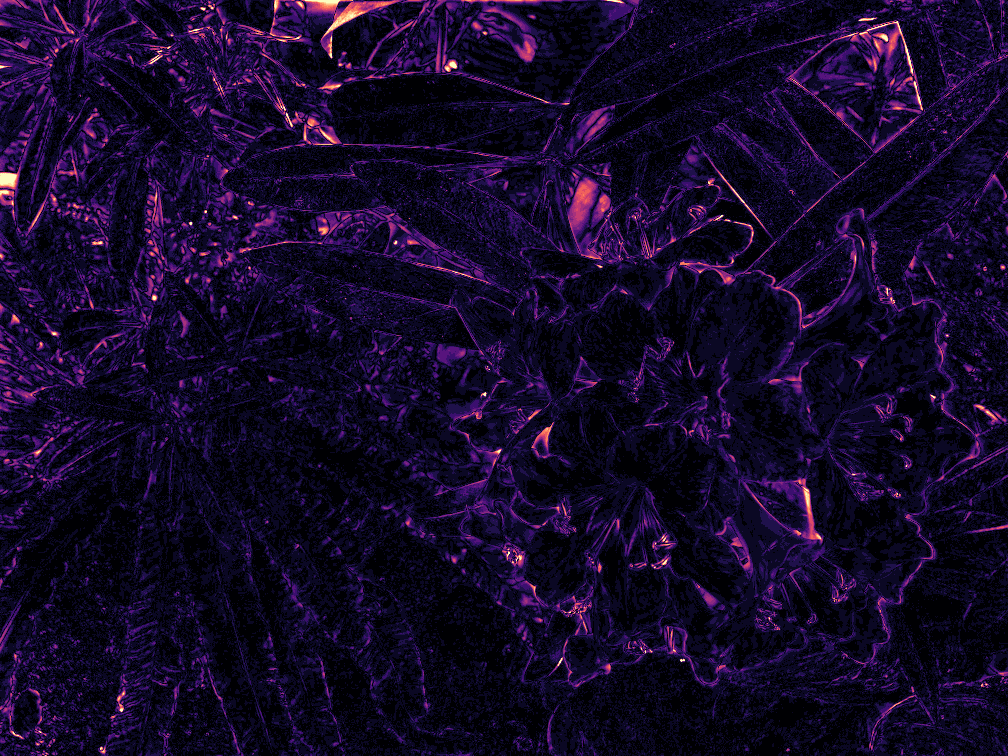} &
         &
         \\
    \end{tabular}
}
    \caption{
        \textbf{Zero-shot Generalization to LLFF}. Scene reconstructions from LLFF~\cite{mildenhall2019llff} in the $\sim$ 20 to 60 views, zero-shot high-resolution setting ($756{\times}1008$).
        We discuss the black SfM initialization, resulting from the original COLMAP reconstruction provided with the dataset, in \cref{subsec:supp-results-dense}.
    }
    \label{fig:llff-comp}
\end{figure*}

\clearpage
}
\afterpage{
    \clearpage

\begin{figure*}[p]
    \centering

    \resizebox{1.0\linewidth}{!}{%
    \setlength{\tabcolsep}{0.002\linewidth} %
    \renewcommand{\arraystretch}{0} %

    \begin{tabular}{@{}%
        >{\centering\arraybackslash}m{0.03\linewidth}
        >{\centering\arraybackslash}m{0.24\linewidth}
        >{\centering\arraybackslash}m{0.24\linewidth}
        >{\centering\arraybackslash}m{0.24\linewidth}
        >{\centering\arraybackslash}m{0.24\linewidth}
        >{\centering\arraybackslash}m{0.24\linewidth}
        >{\centering\arraybackslash}m{0.02\linewidth}
        >{\centering\arraybackslash}m{0.24\linewidth}
    @{}}
         & $t = 2$ & $t = 10$ & $t = 100$ & $t = 1000$ & $t = 2000$ &  &  \\[3pt]

        \rotatebox{90}{\makecell{\smaller[1]{ReSplat~\cite{xu2025resplat}}}} &
        \pcell[width=\linewidth]{25.8}{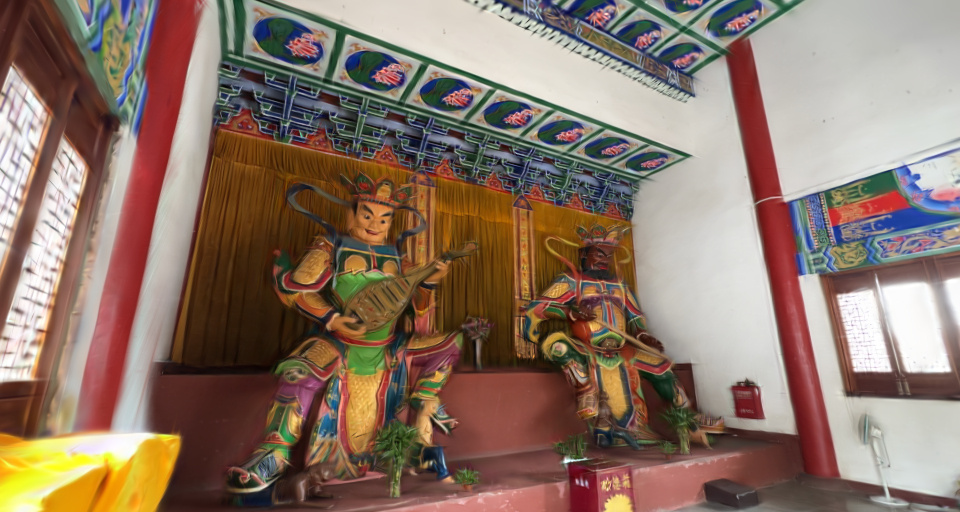} &
        \pcell[width=\linewidth]{26.9}{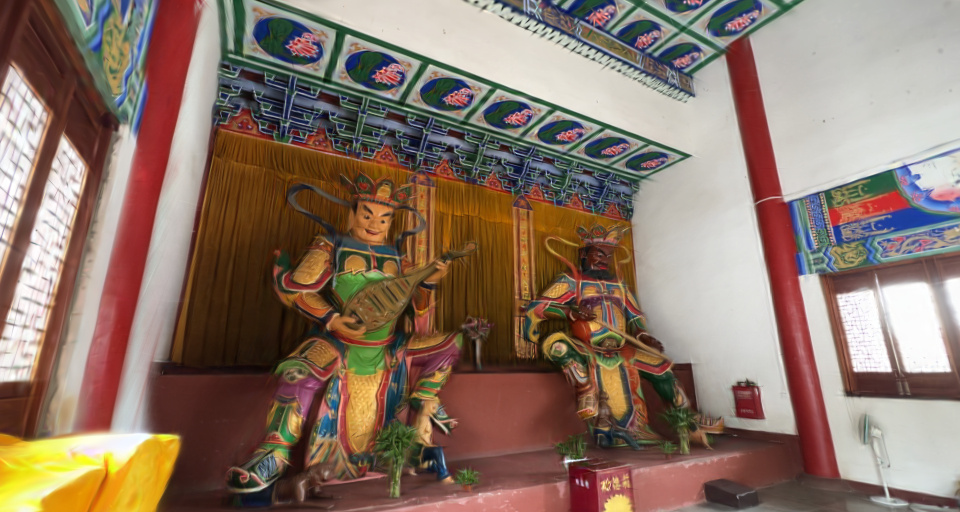} &
        \pcell[width=\linewidth]{19.6}{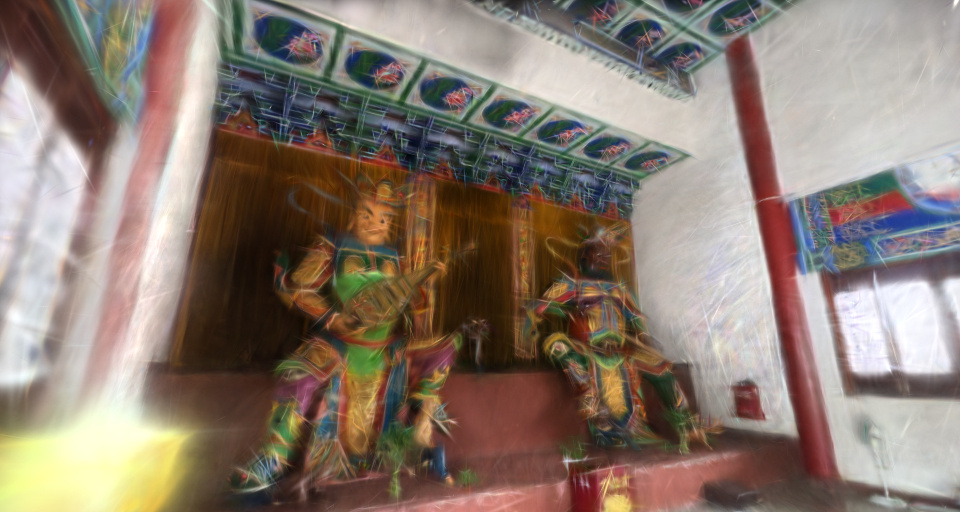} &
        \pcell[width=\linewidth]{5.3}{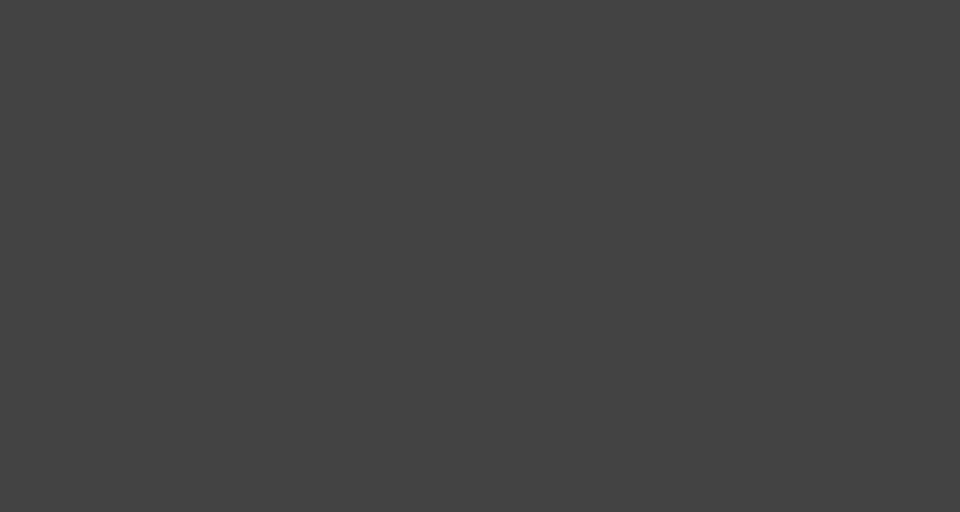} &
        \pcell[width=\linewidth]{5.2}{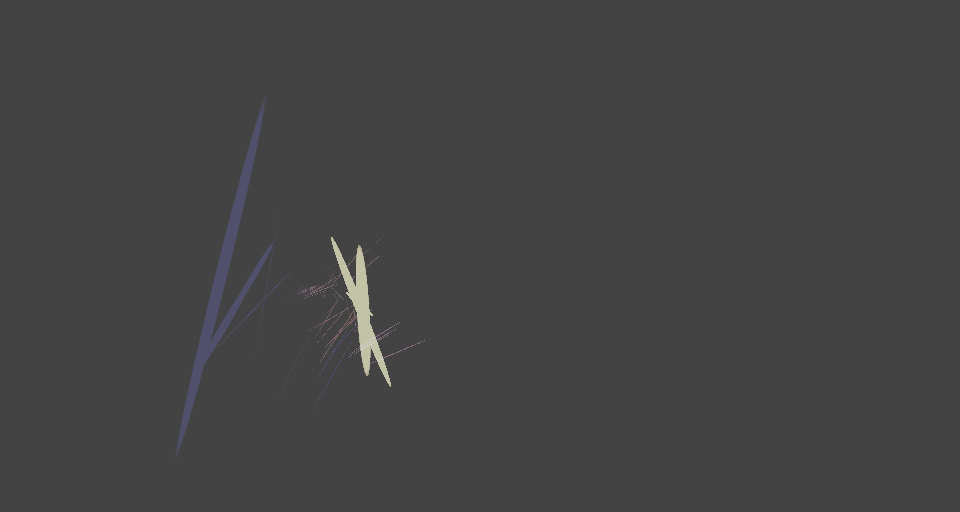} &
        \rotatebox{90}{\smaller[1]{Init}} &
        \pcell[width=\linewidth]{21.1}{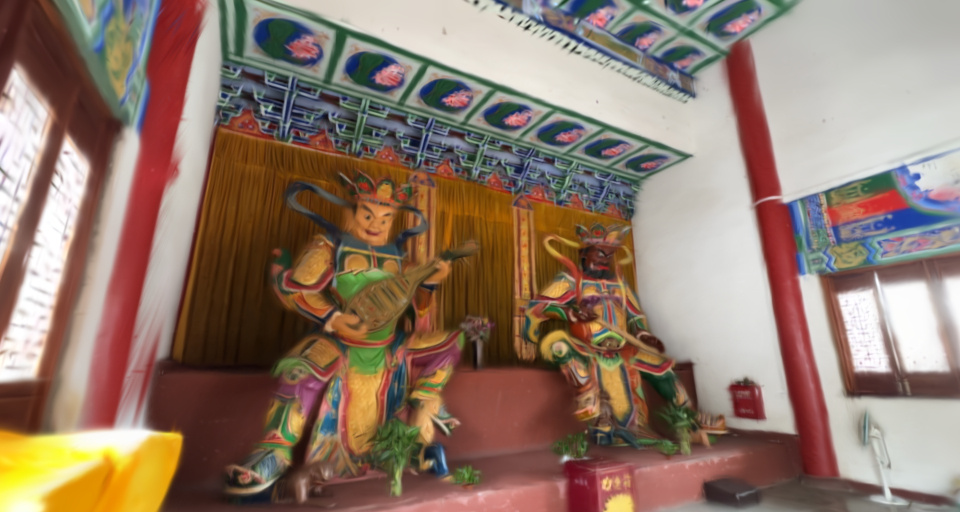} \\
        \noalign{\vskip 2\tabcolsep}
        \rotatebox{90}{\makecell{\smaller[1]{\lob{}}}} &
        \pcell[width=\linewidth]{26.7}{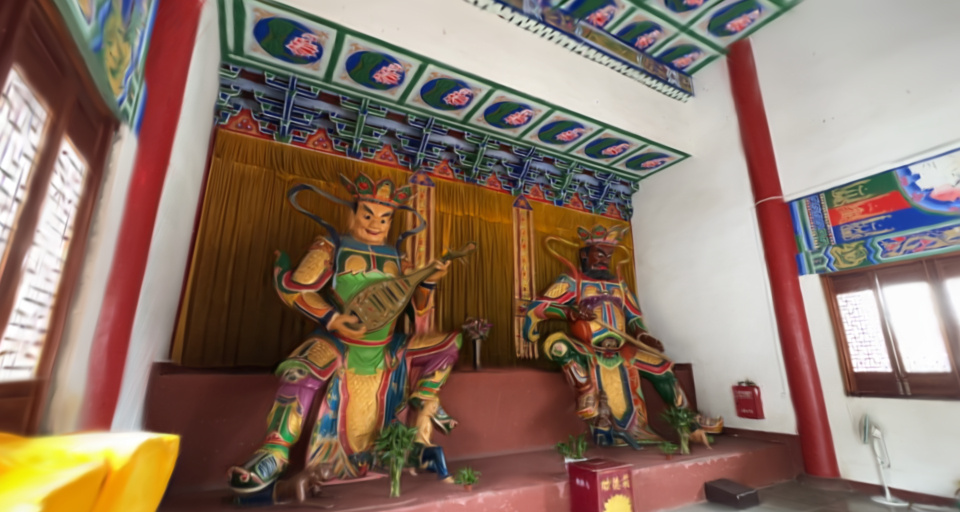} &
        \pcell[width=\linewidth]{32.9}{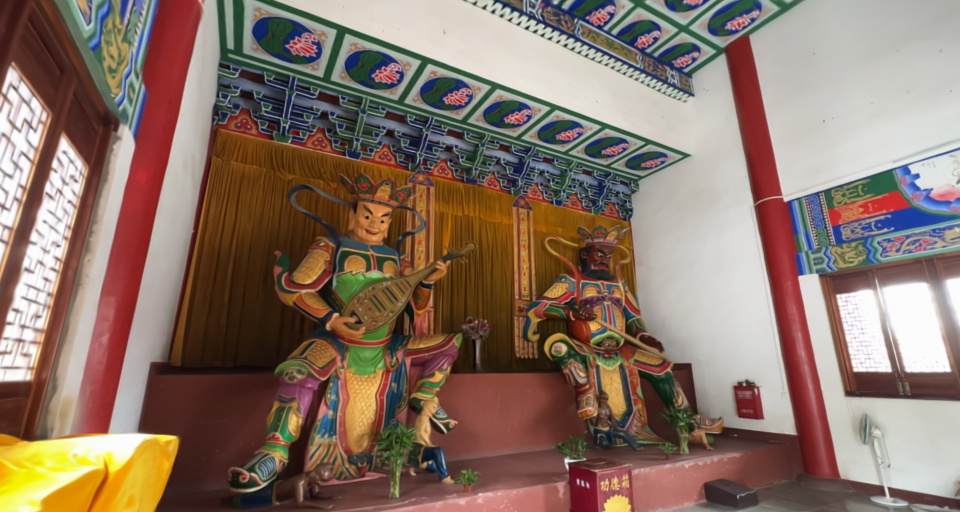} &
        \pcell[width=\linewidth]{34.8}{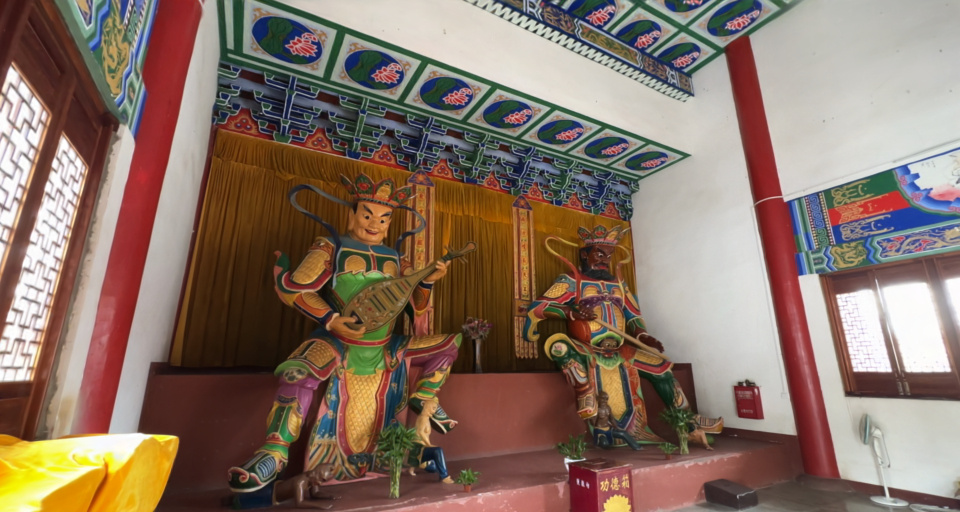} &
        \pcell[width=\linewidth]{31.1}{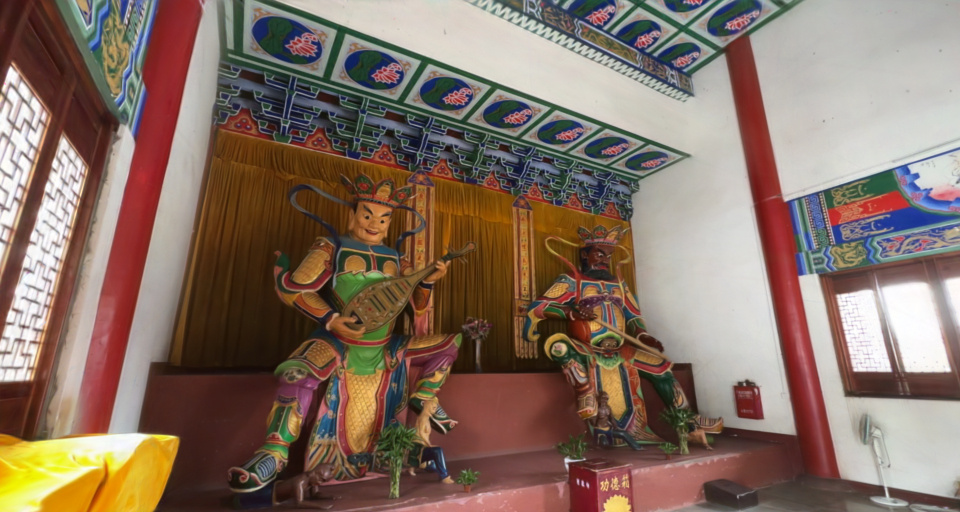} &
        \pcell[width=\linewidth]{33.0}{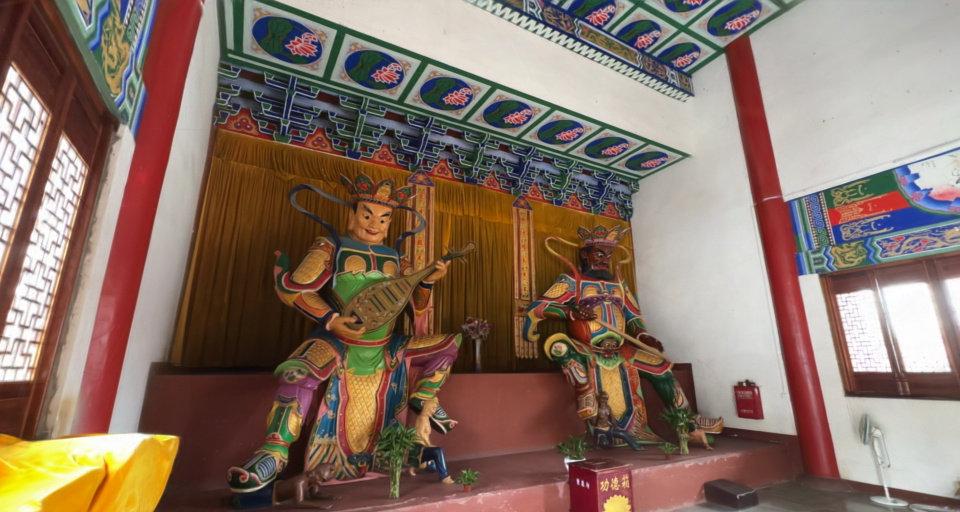} &
        \rotatebox{90}{\smaller[1]{Ref}} &
        \rcell[width=\linewidth]{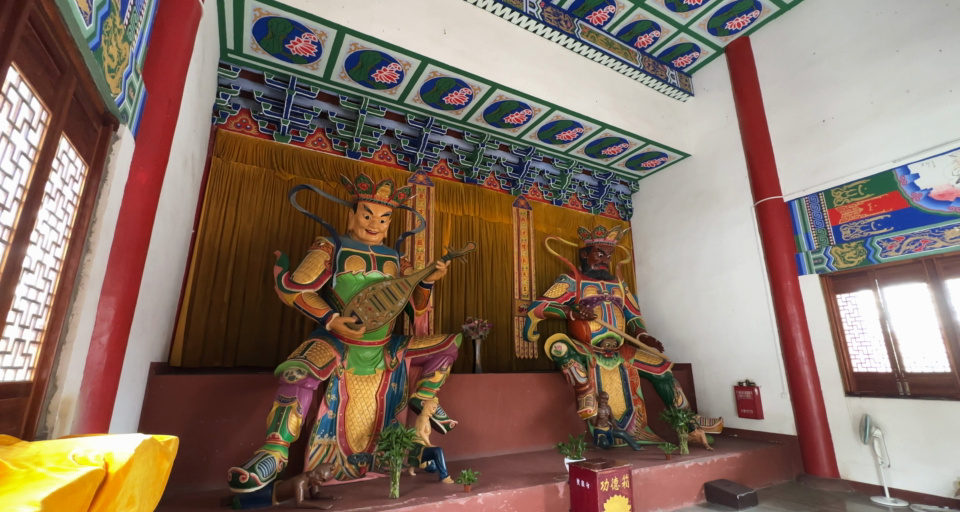} \\
        \noalign{\vskip 2\tabcolsep}
        \rotatebox{90}{\makecell{\smaller[1]{\adam{}}}} &
        \pcell[width=\linewidth]{21.8}{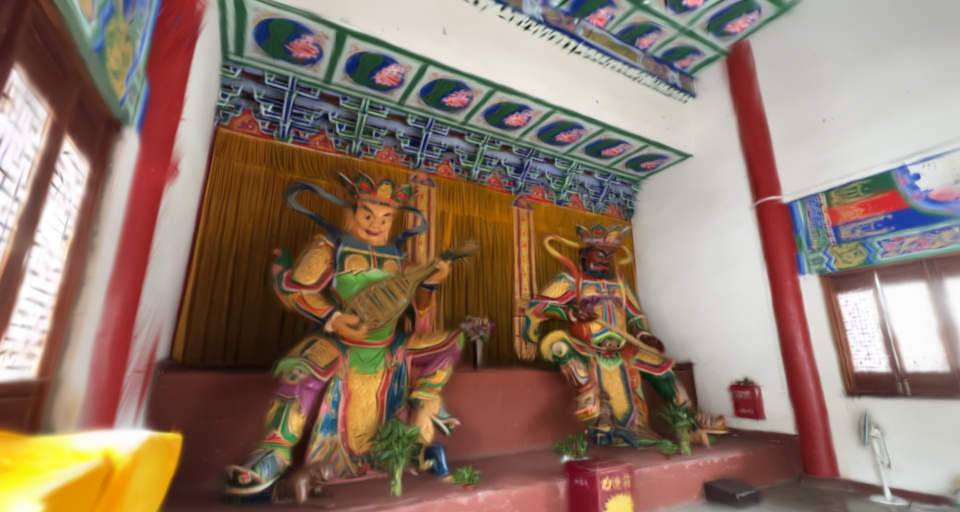} &
        \pcell[width=\linewidth]{24.3}{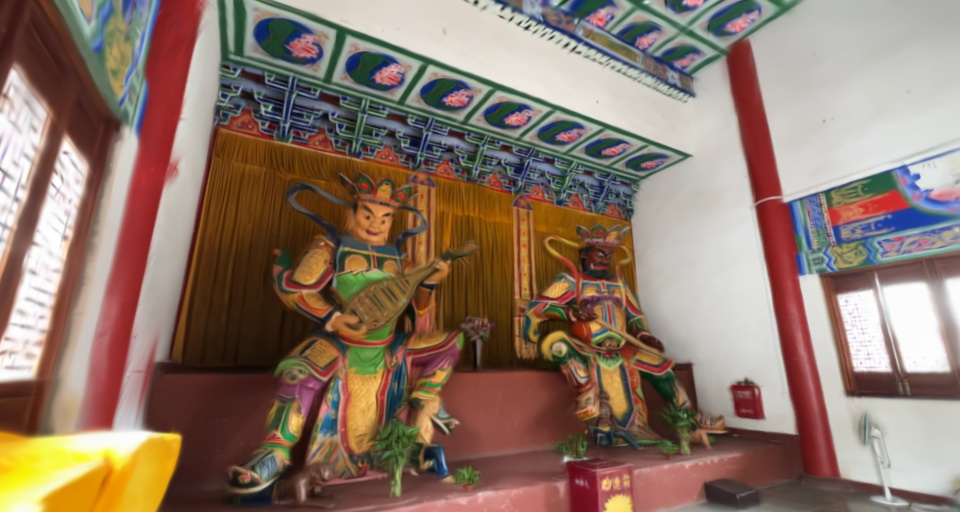} &
        \pcell[width=\linewidth]{33.0}{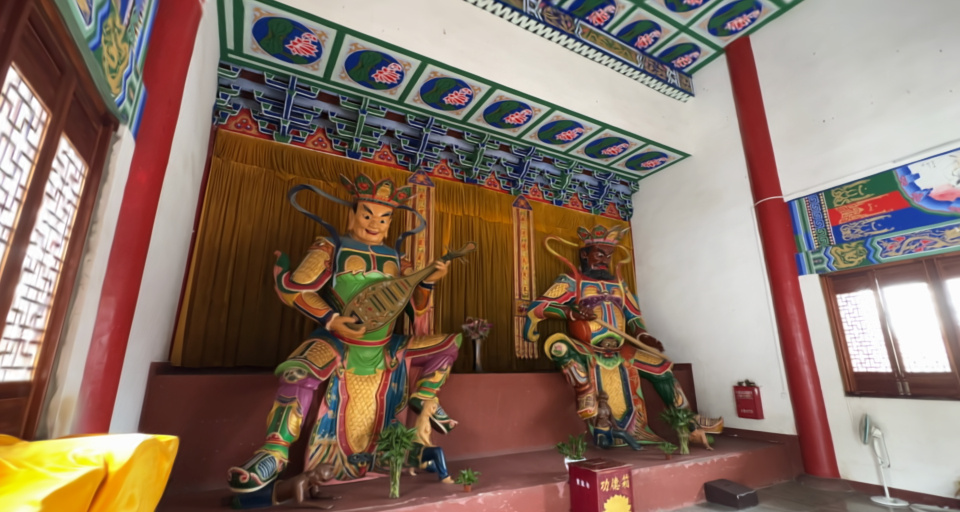} &
        \pcell[width=\linewidth]{36.0}{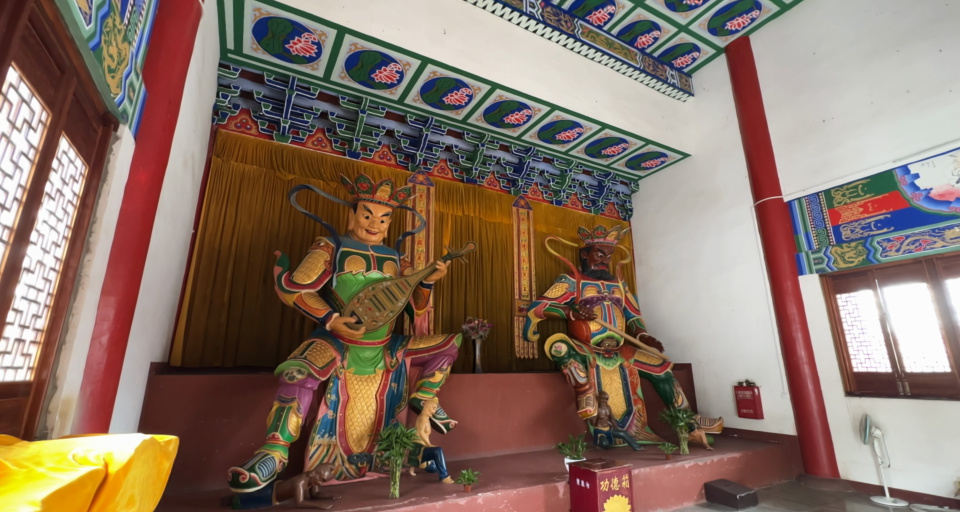} &
        \pcell[width=\linewidth]{36.1}{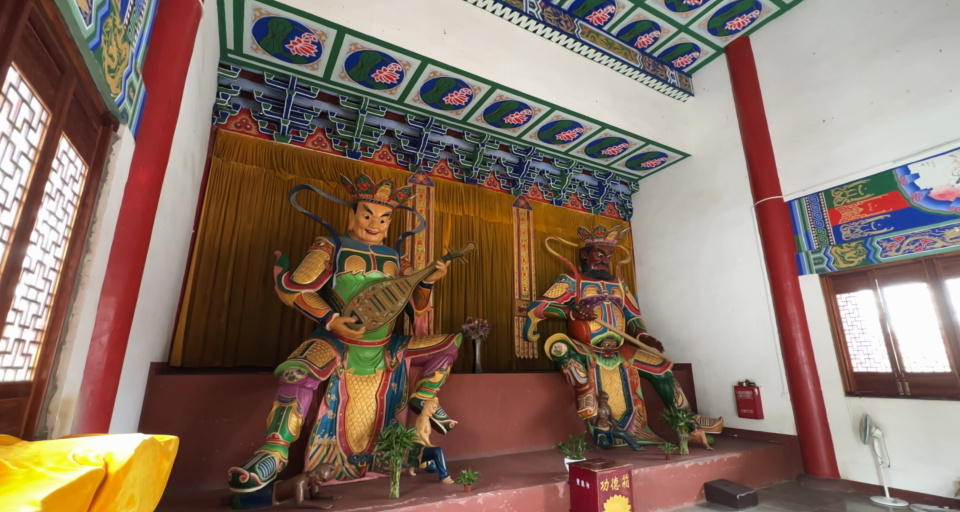} &
         &
         \\
        \noalign{\vskip 2\tabcolsep}
        \rotatebox{90}{\makecell{\smaller[1]{\ourshort{}}}} &
        \pcell[width=\linewidth]{26.9}{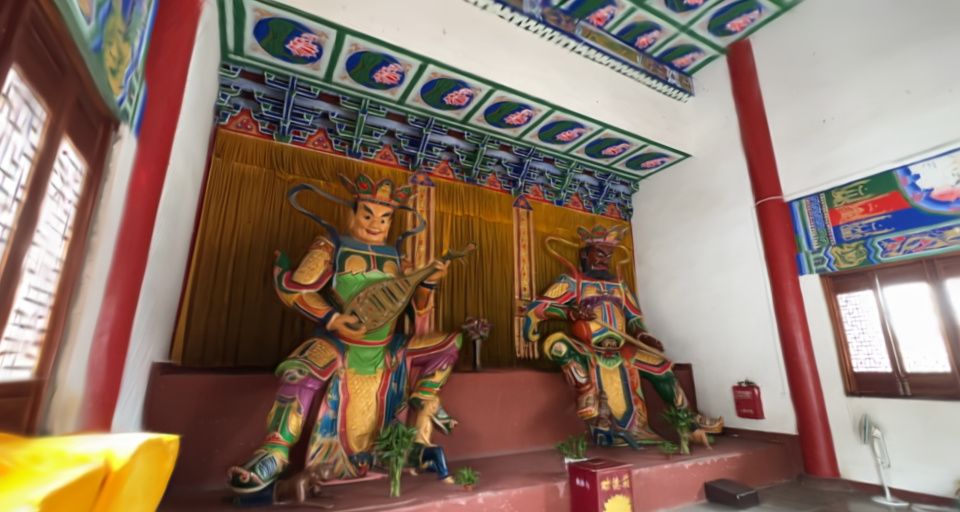} &
        \pcell[width=\linewidth]{32.7}{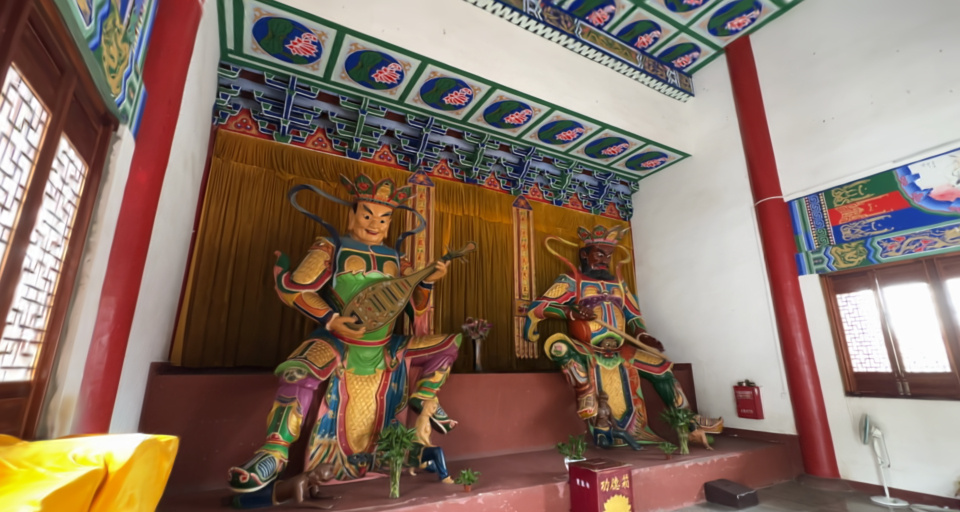} &
        \pcell[width=\linewidth]{35.5}{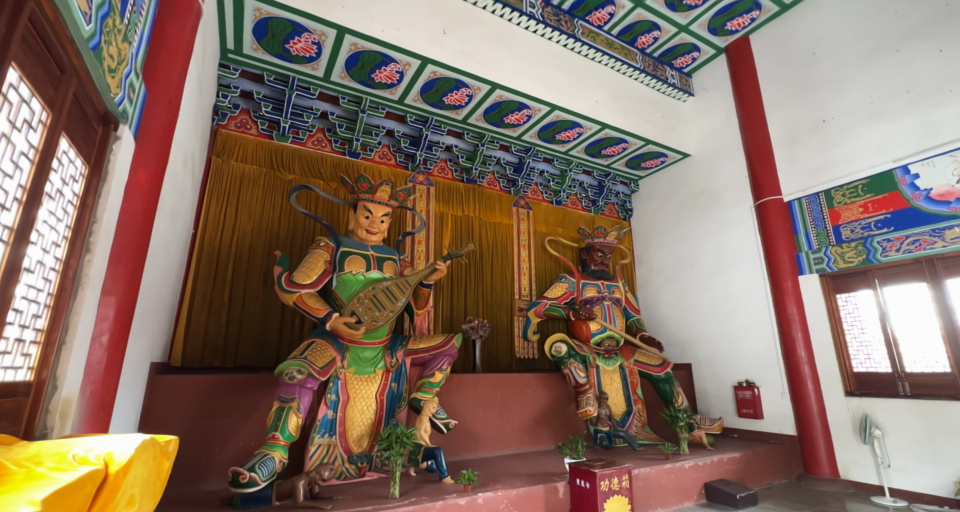} &
        \pcell[width=\linewidth]{36.3}{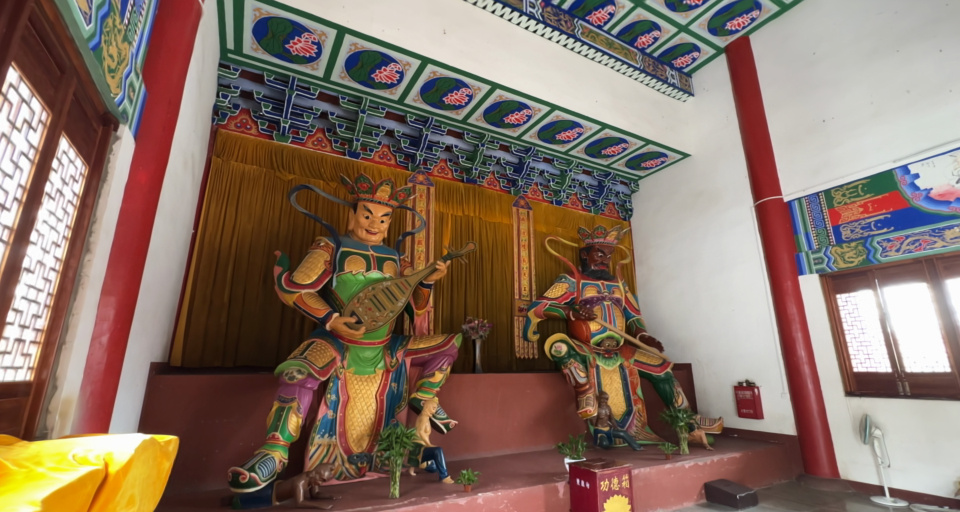} &
        \pcell[width=\linewidth]{36.4}{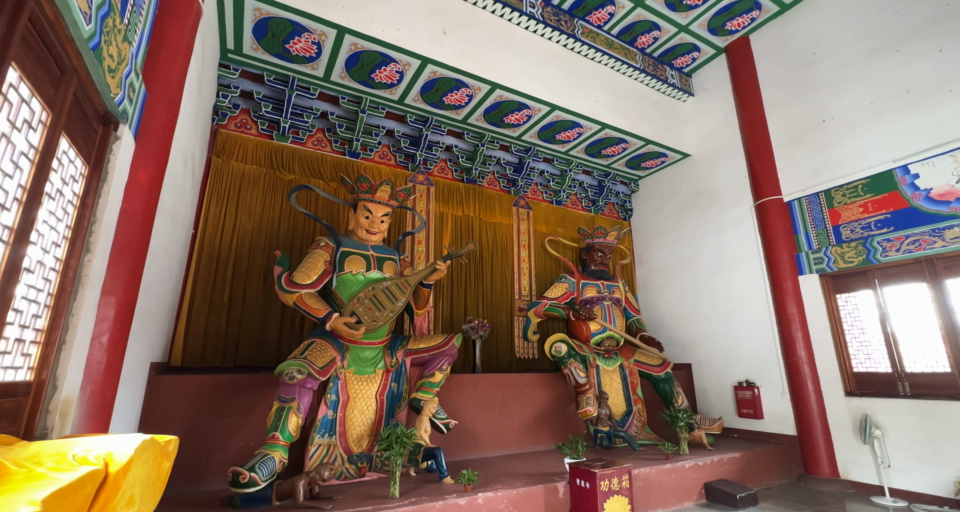} &
         &
         \\
        \noalign{\vskip 2\tabcolsep}
        \rotatebox{90}{\makecell{\smaller[1]{\adam{} error}}} &
        \rcell[width=\linewidth]{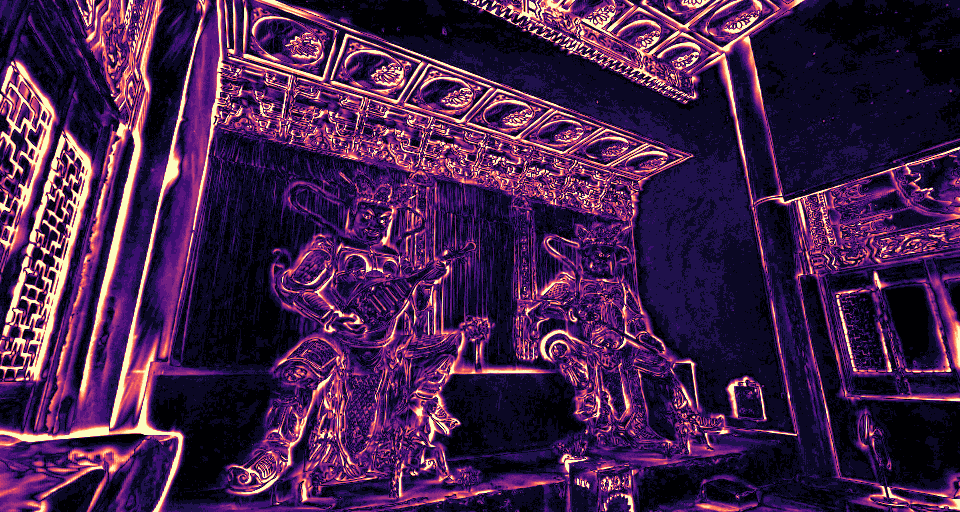} &
        \rcell[width=\linewidth]{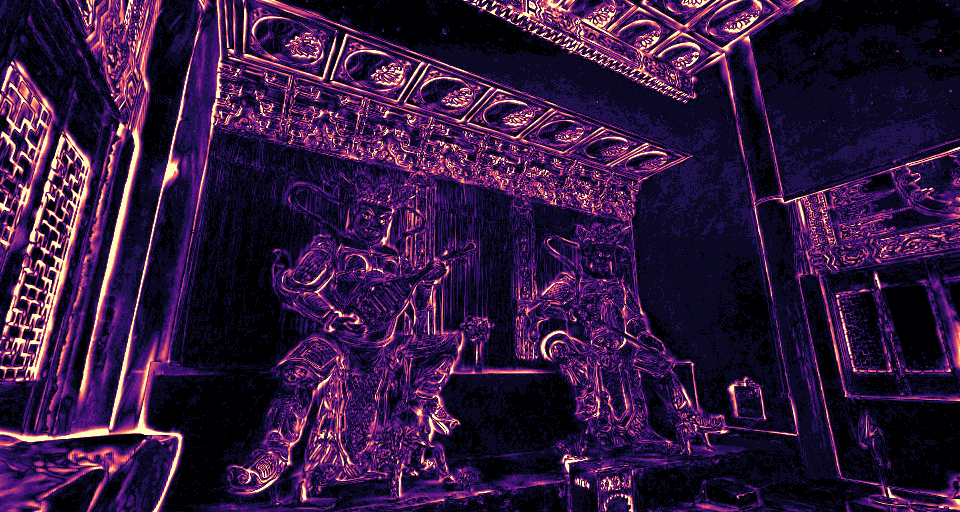} &
        \rcell[width=\linewidth]{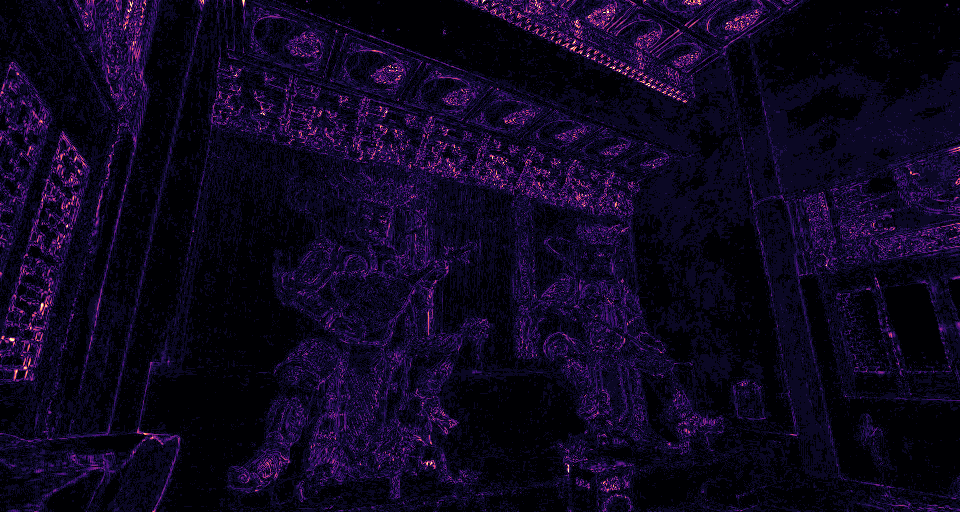} &
        \rcell[width=\linewidth]{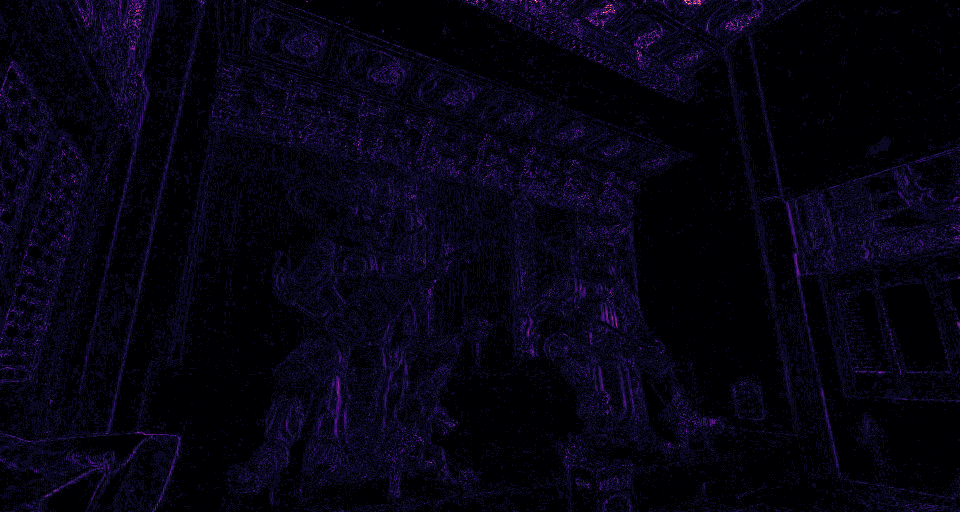} &
        \rcell[width=\linewidth]{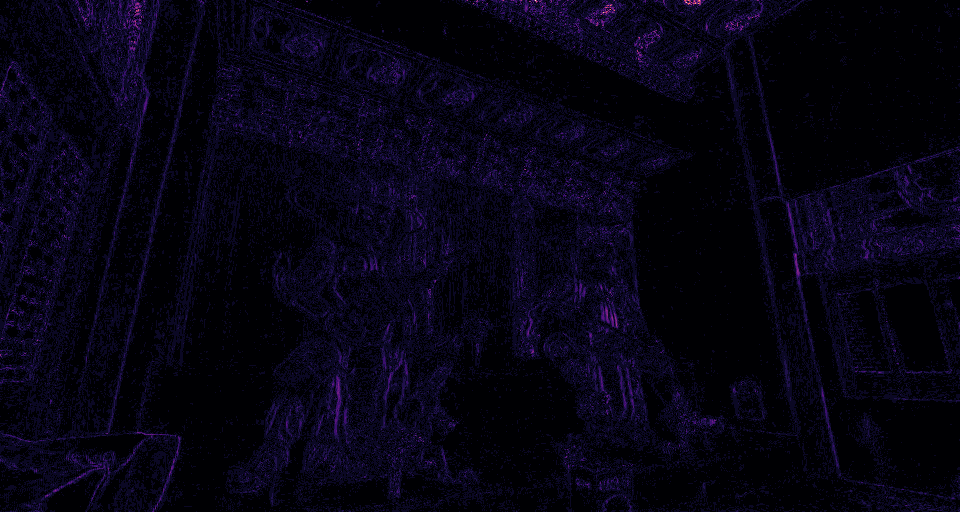} &
        \rotatebox{90}{\smaller[1]{Init error}} &
        \rcell[width=\linewidth]{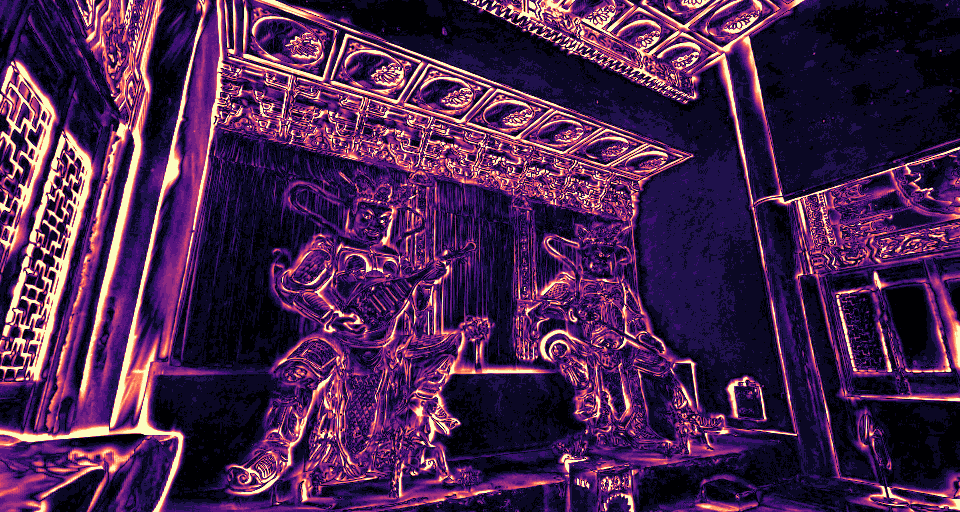} \\
        \noalign{\vskip 2\tabcolsep}
        \rotatebox{90}{\makecell{\smaller[1]{\ourshort{} error}}} &
        \rcell[width=\linewidth]{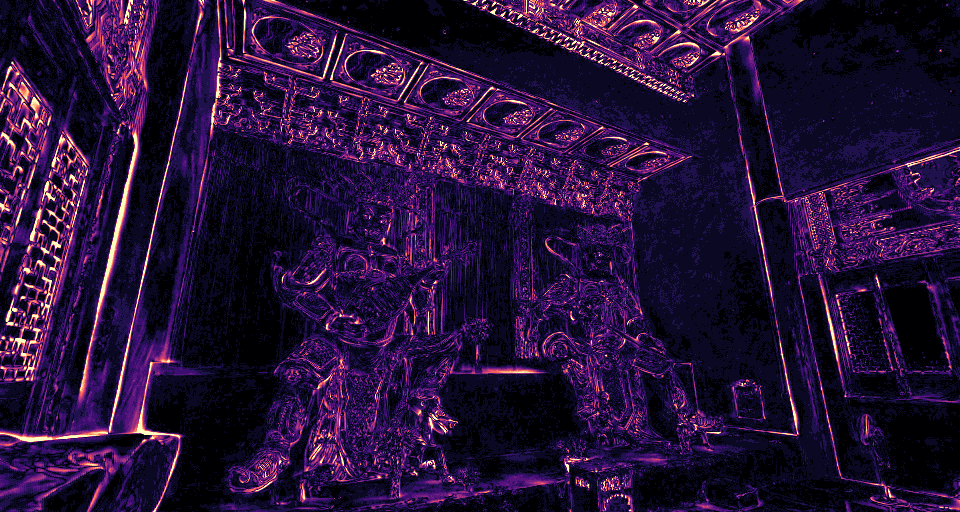} &
        \rcell[width=\linewidth]{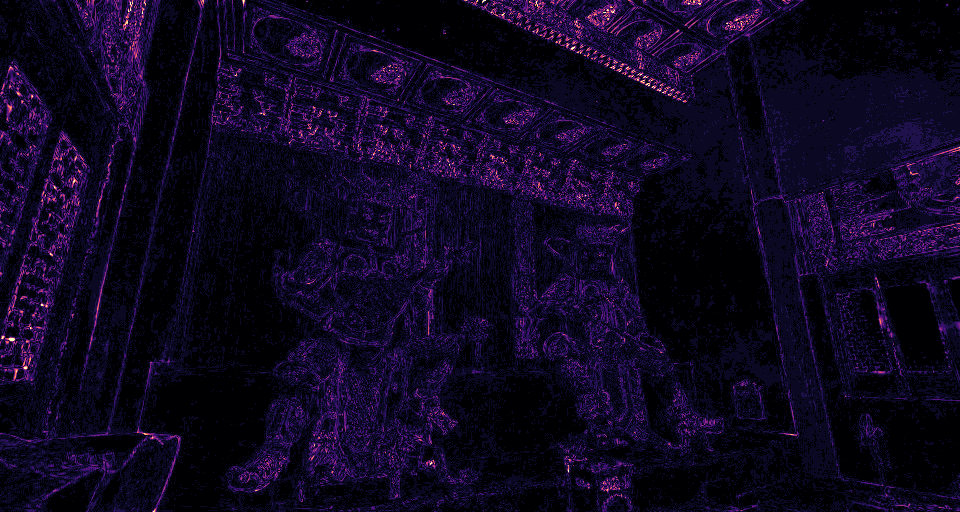} &
        \rcell[width=\linewidth]{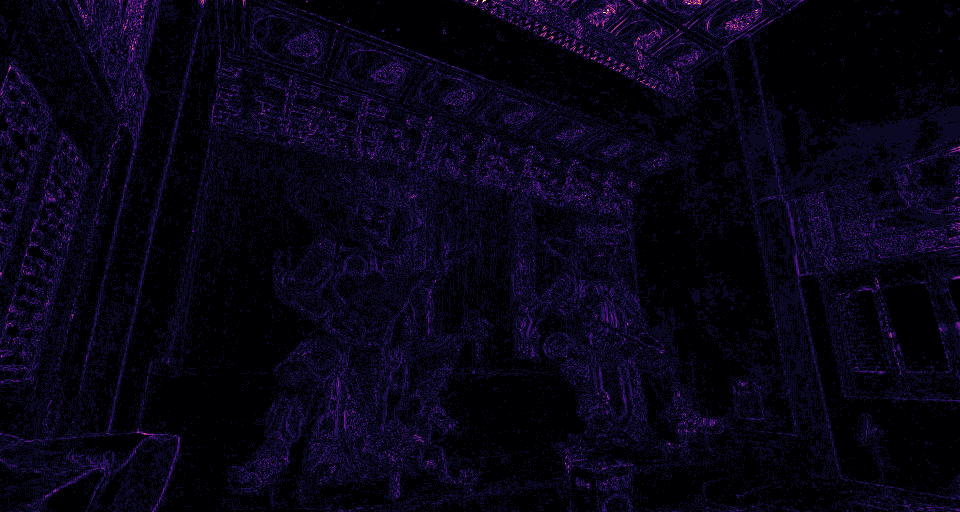} &
        \rcell[width=\linewidth]{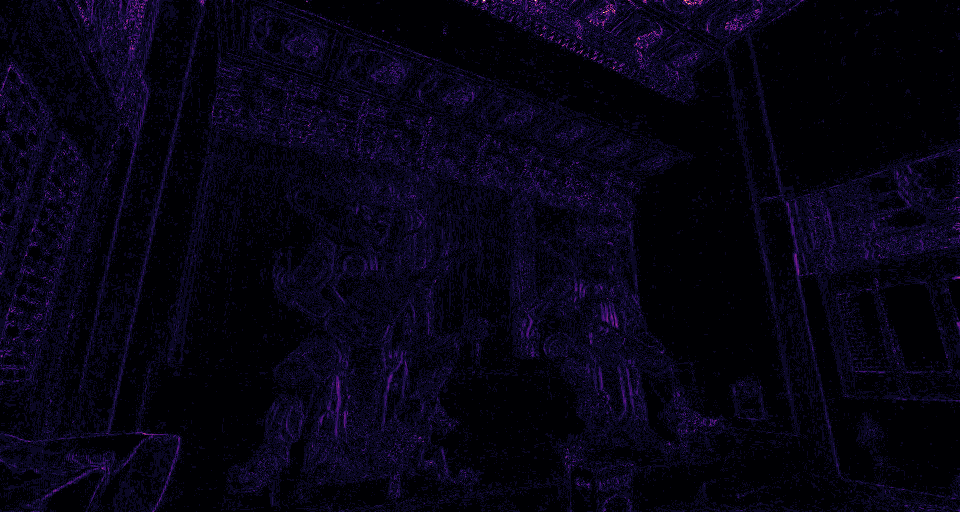} &
        \rcell[width=\linewidth]{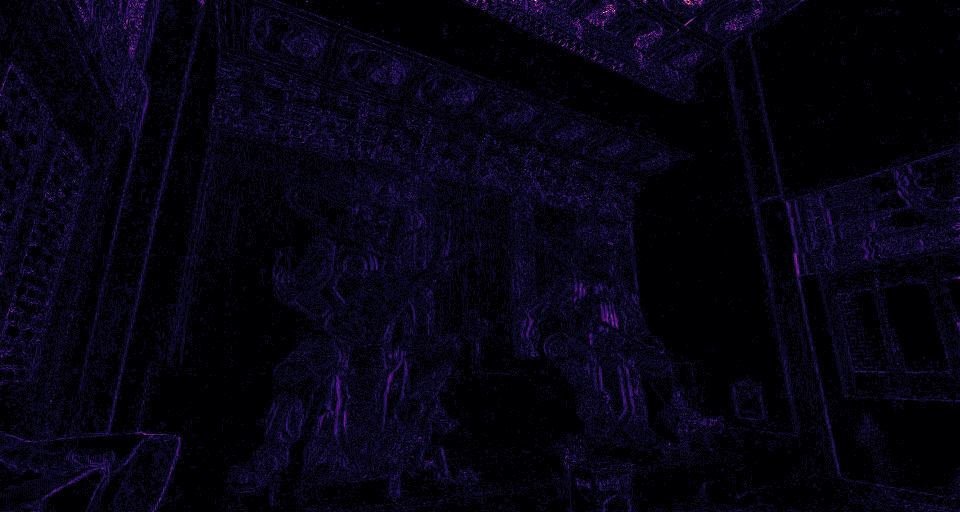} &
         &
         \\
        \noalign{\vskip 2\tabcolsep}
        \rotatebox{90}{\makecell{\smaller[1]{ReSplat~\cite{xu2025resplat}}}} &
        \pcell[width=\linewidth]{23.7}{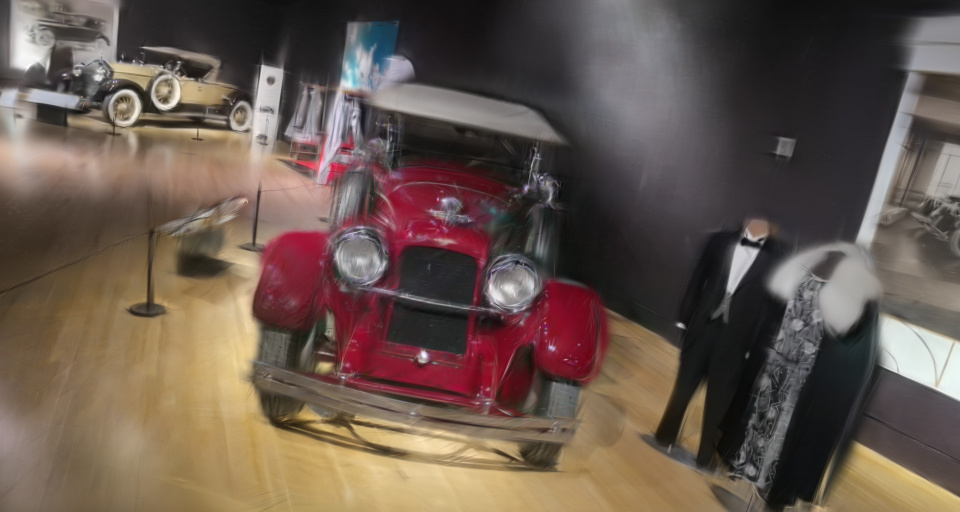} &
        \pcell[width=\linewidth]{23.8}{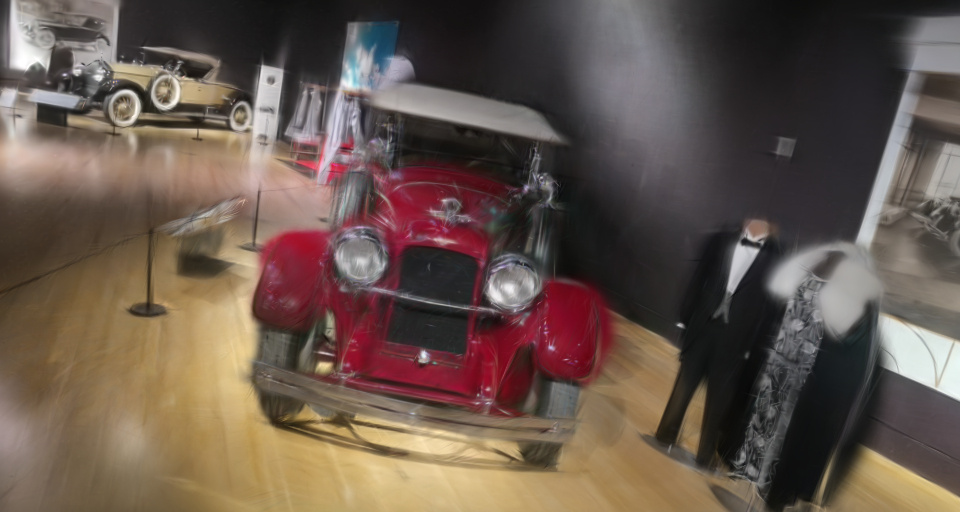} &
        \pcell[width=\linewidth]{15.5}{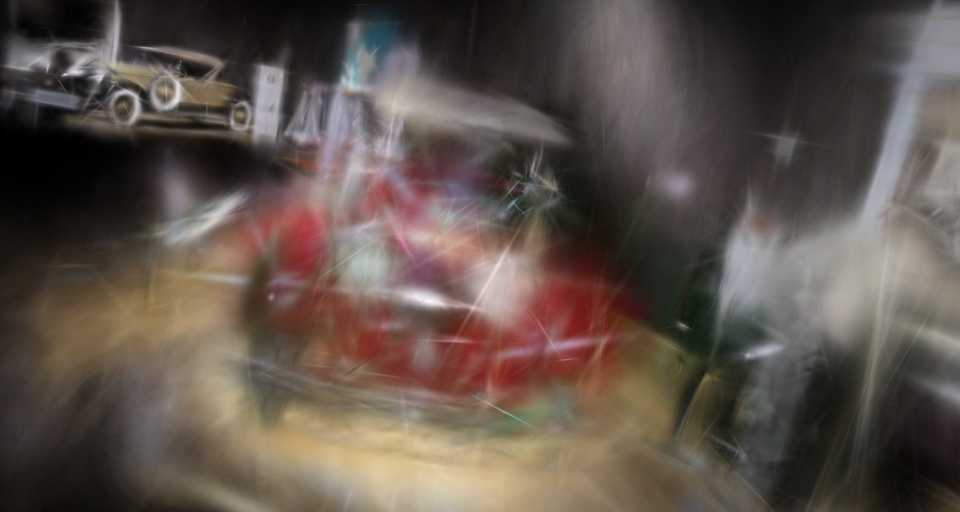} &
        \rcell[width=\linewidth]{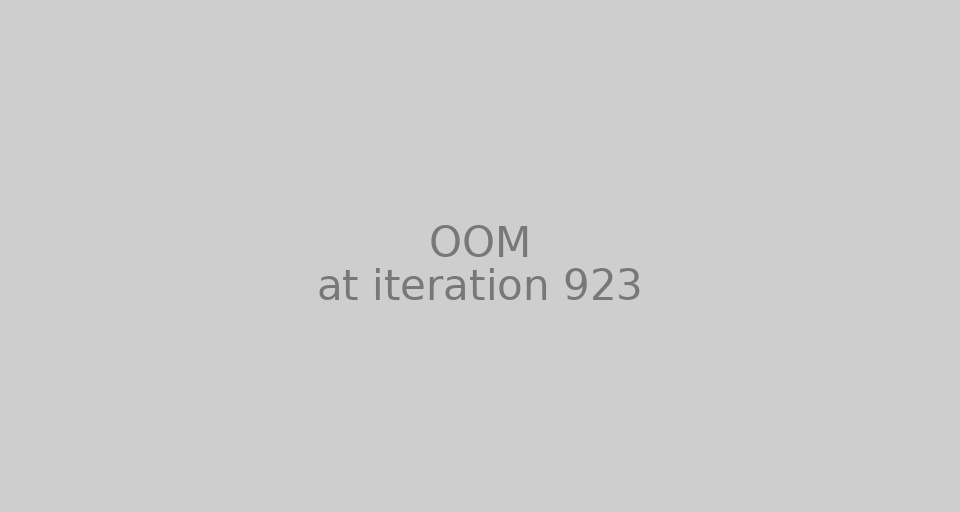} &
        \rcell[width=\linewidth]{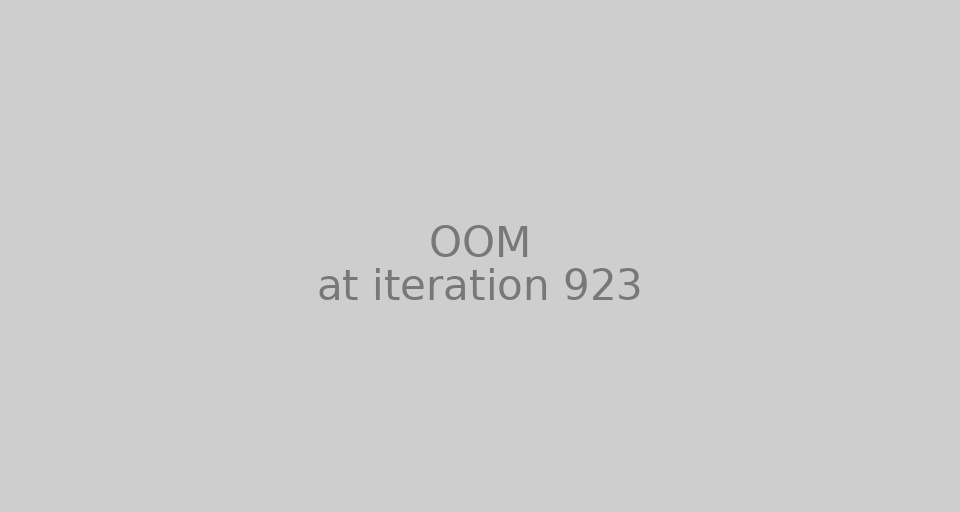} &
        \rotatebox{90}{\smaller[1]{Init}} &
        \pcell[width=\linewidth]{21.3}{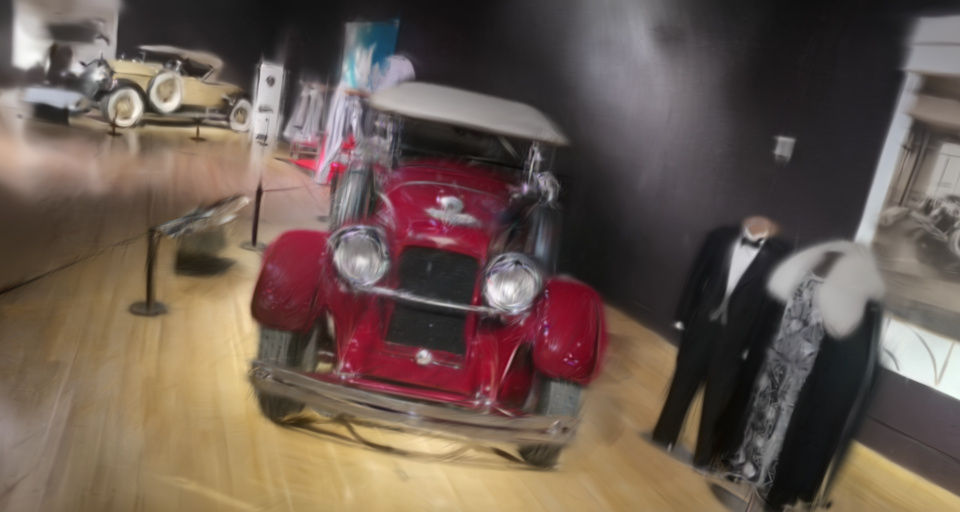} \\
        \noalign{\vskip 2\tabcolsep}
        \rotatebox{90}{\makecell{\smaller[1]{\lob{}}}} &
        \pcell[width=\linewidth]{26.3}{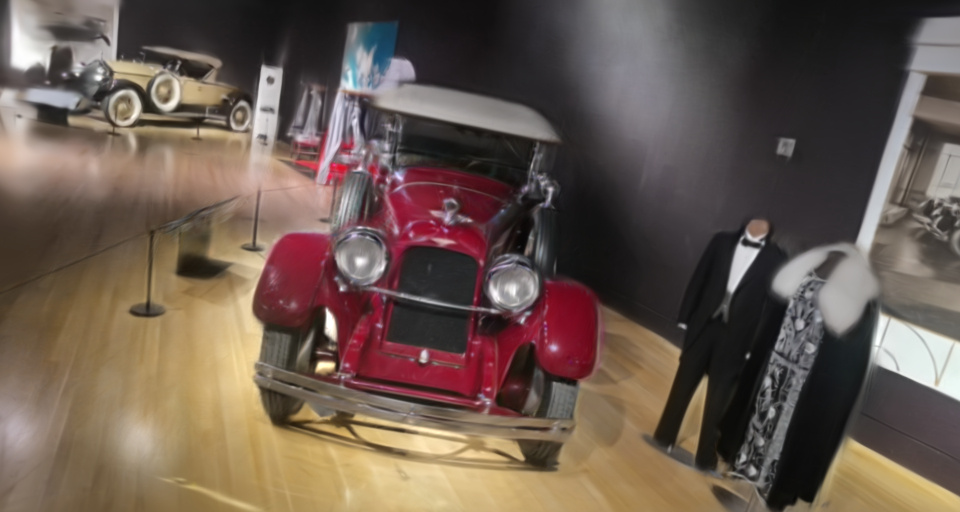} &
        \pcell[width=\linewidth]{30.3}{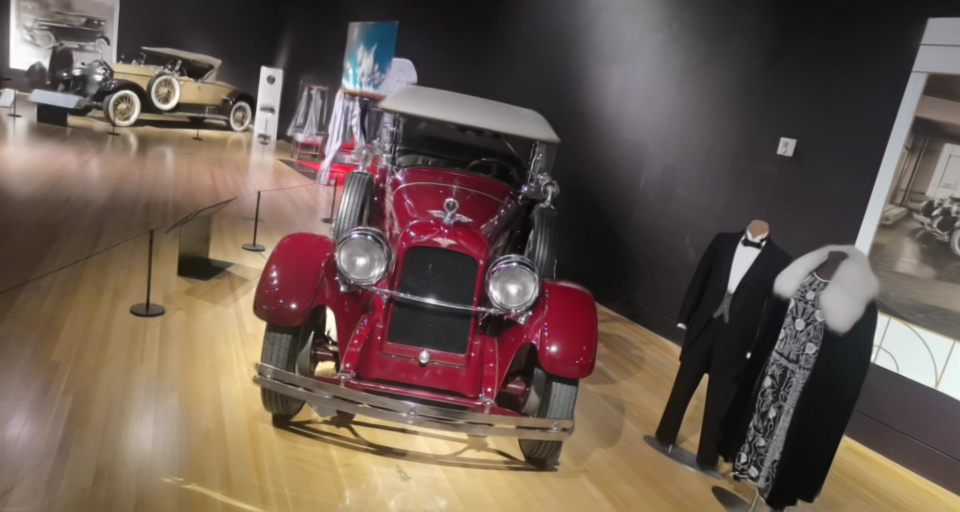} &
        \pcell[width=\linewidth]{31.4}{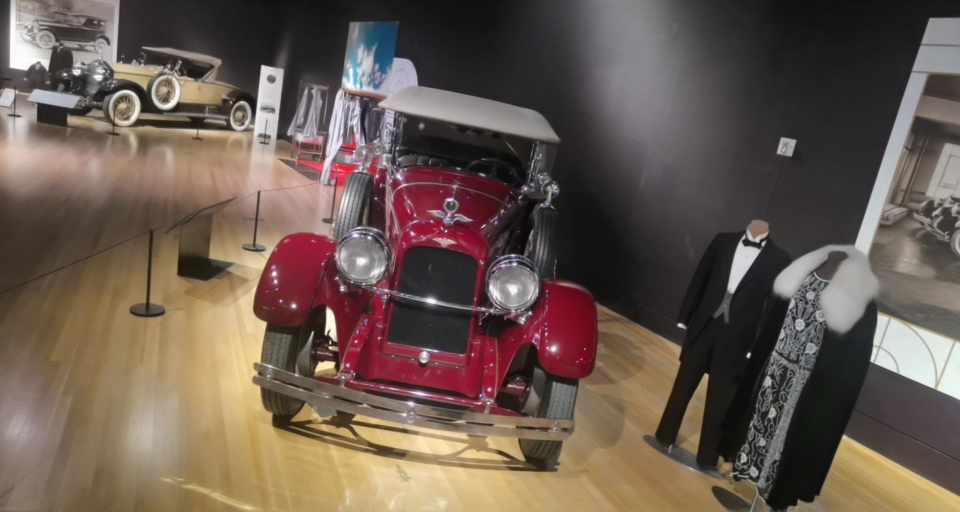} &
        \pcell[width=\linewidth]{29.1}{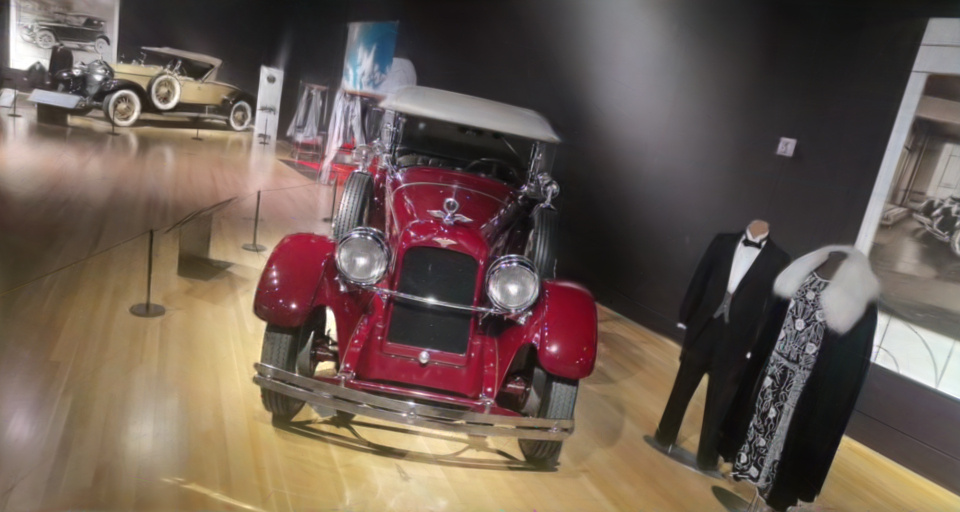} &
        \pcell[width=\linewidth]{28.2}{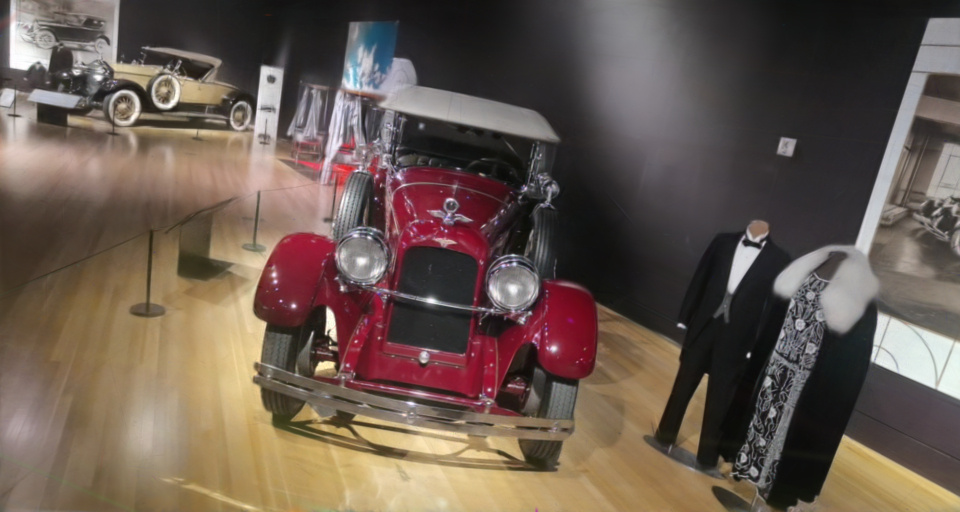} &
        \rotatebox{90}{\smaller[1]{Ref}} &
        \rcell[width=\linewidth]{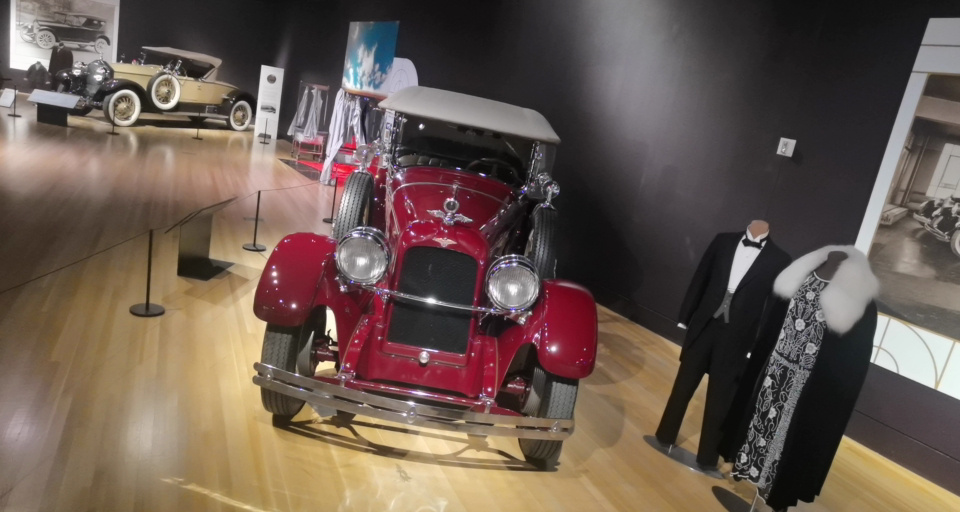} \\
        \noalign{\vskip 2\tabcolsep}
        \rotatebox{90}{\makecell{\smaller[1]{\adam{}}}} &
        \pcell[width=\linewidth]{22.0}{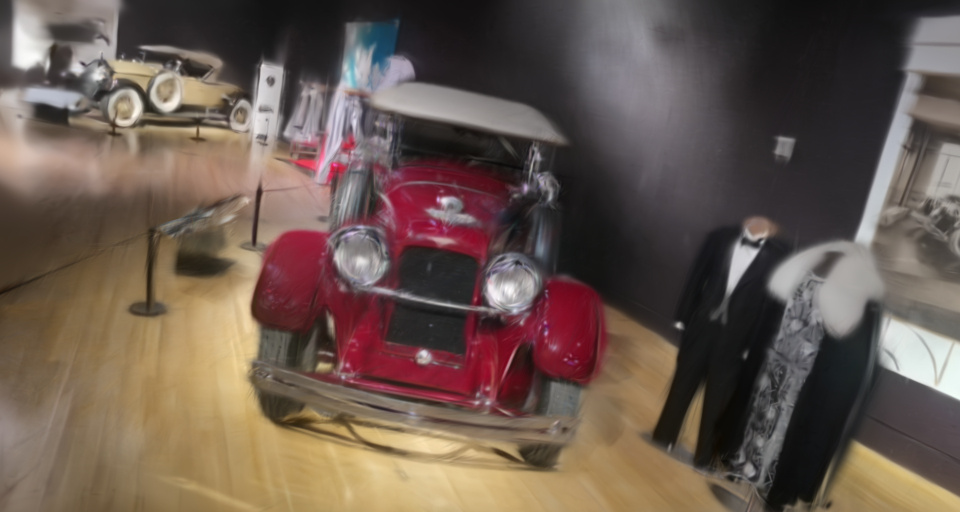} &
        \pcell[width=\linewidth]{24.3}{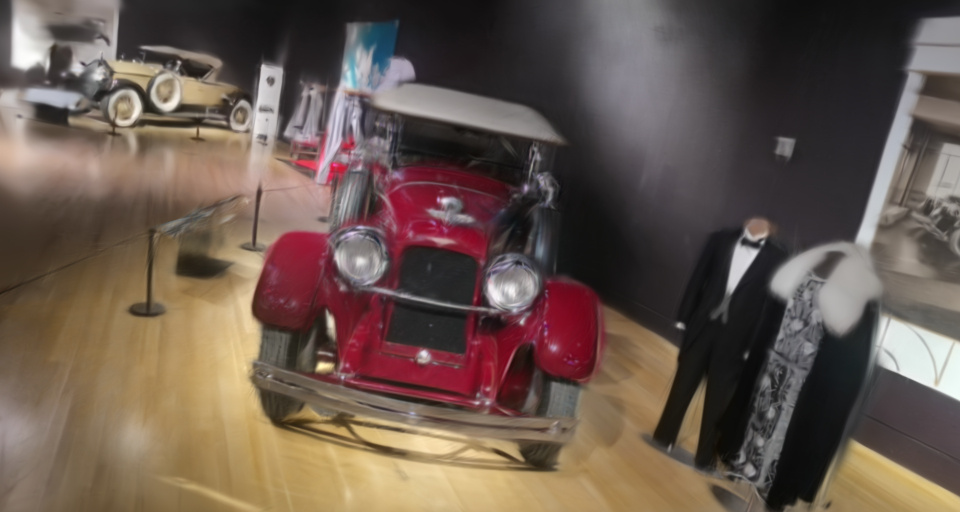} &
        \pcell[width=\linewidth]{30.0}{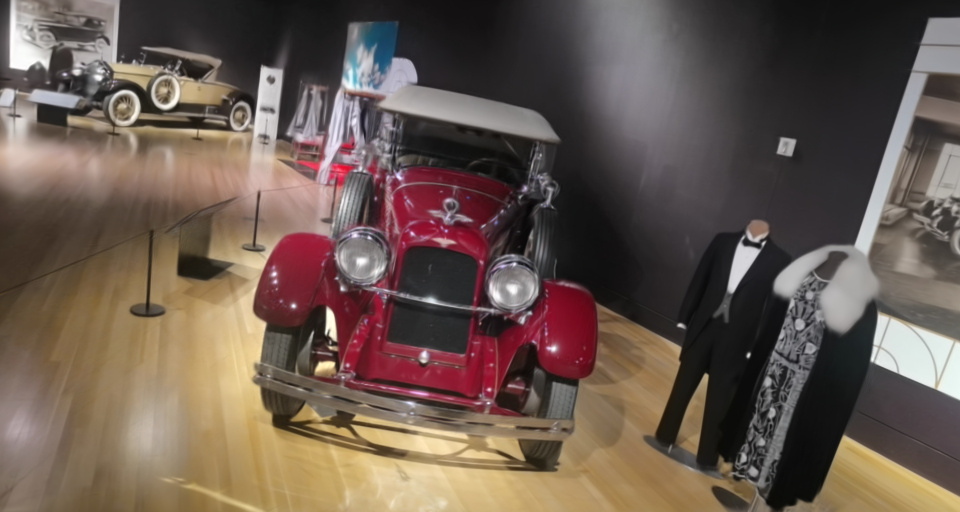} &
        \pcell[width=\linewidth]{31.5}{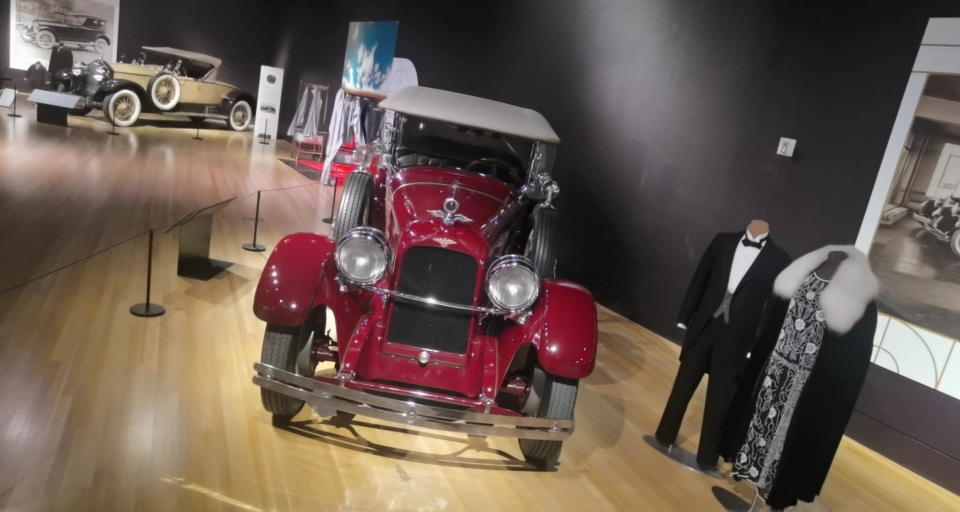} &
        \pcell[width=\linewidth]{31.5}{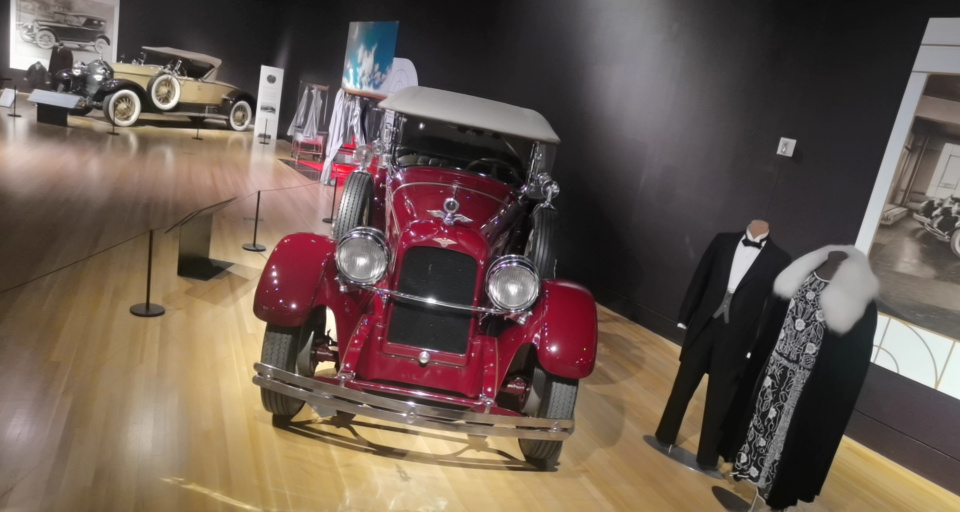} &
         &
         \\
        \noalign{\vskip 2\tabcolsep}
        \rotatebox{90}{\makecell{\smaller[1]{\ourshort{}}}} &
        \pcell[width=\linewidth]{26.3}{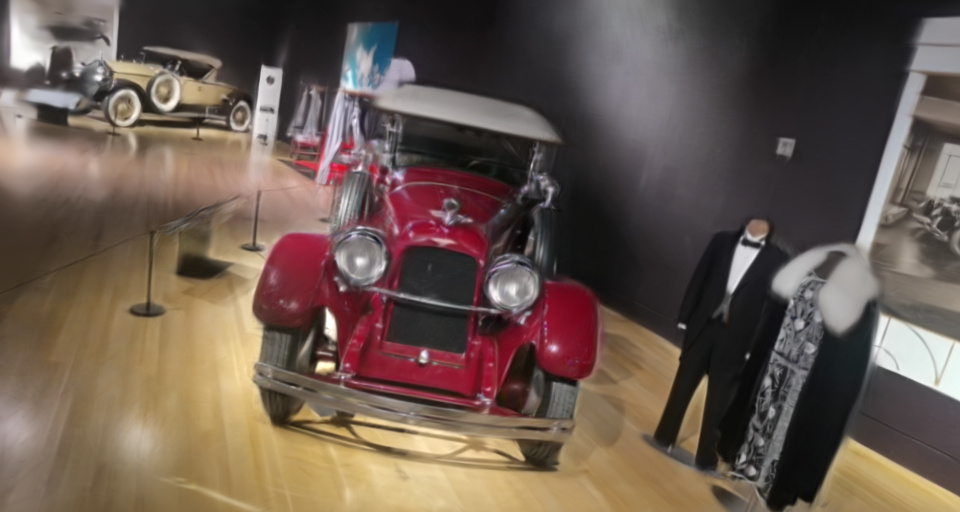} &
        \pcell[width=\linewidth]{30.1}{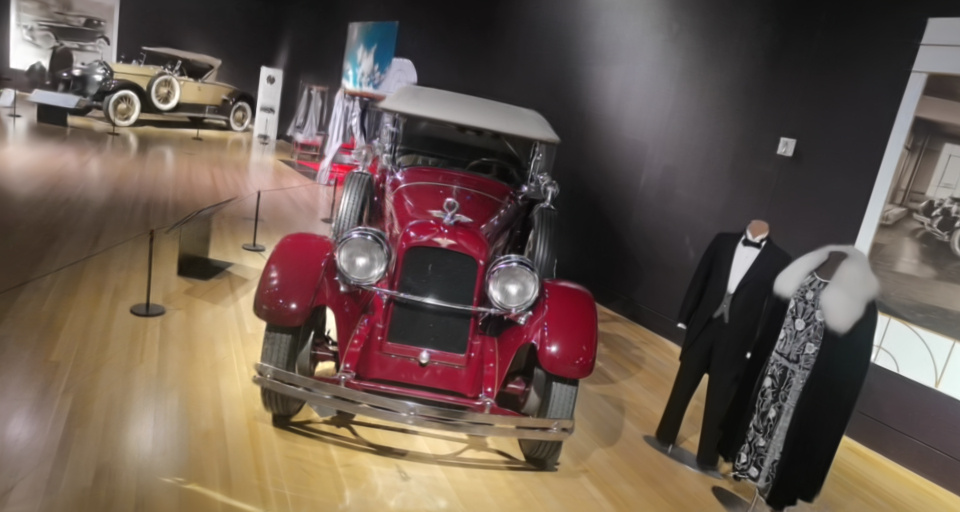} &
        \pcell[width=\linewidth]{31.5}{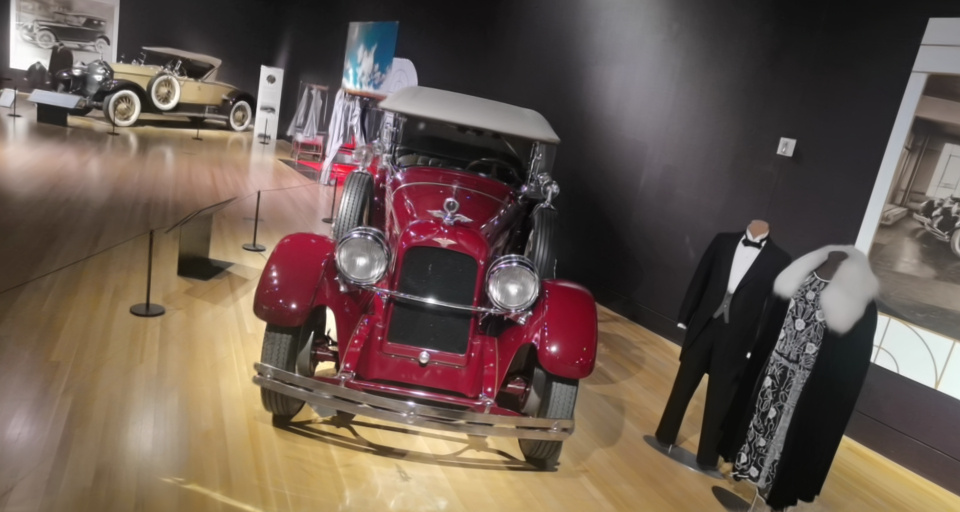} &
        \pcell[width=\linewidth]{31.8}{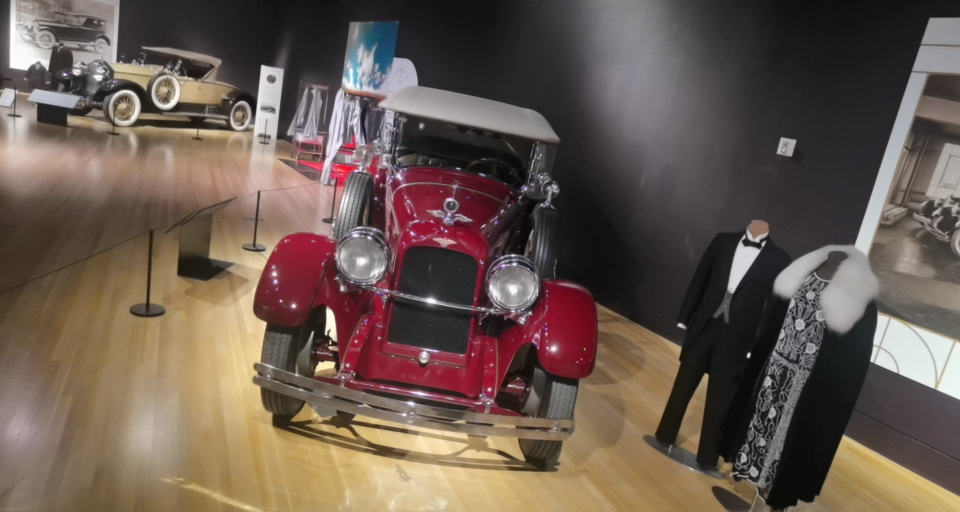} &
        \pcell[width=\linewidth]{31.8}{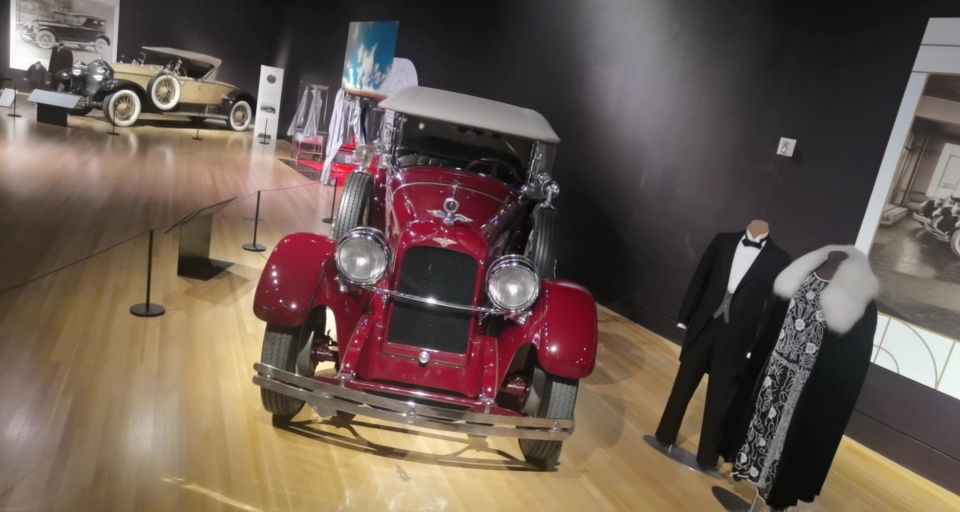} &
         &
         \\
        \noalign{\vskip 2\tabcolsep}
        \rotatebox{90}{\makecell{\smaller[1]{\adam{} error}}} &
        \rcell[width=\linewidth]{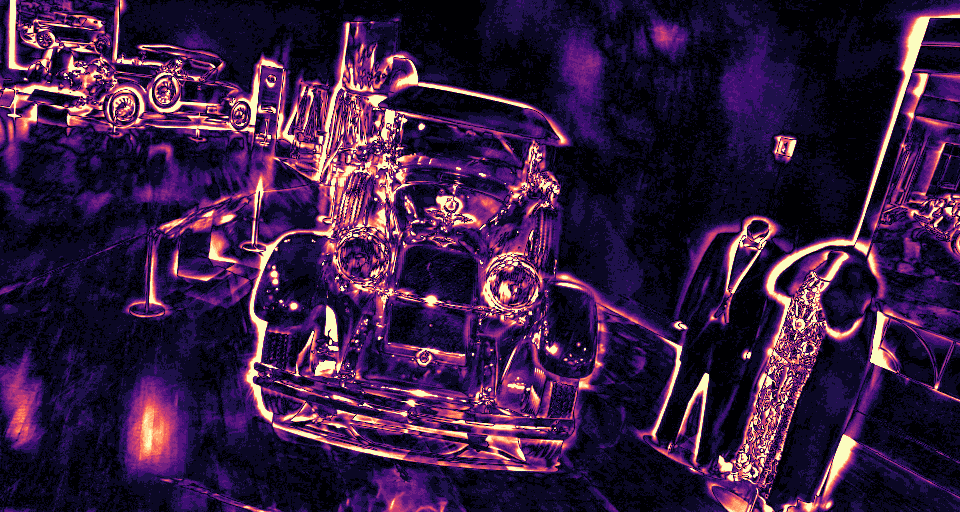} &
        \rcell[width=\linewidth]{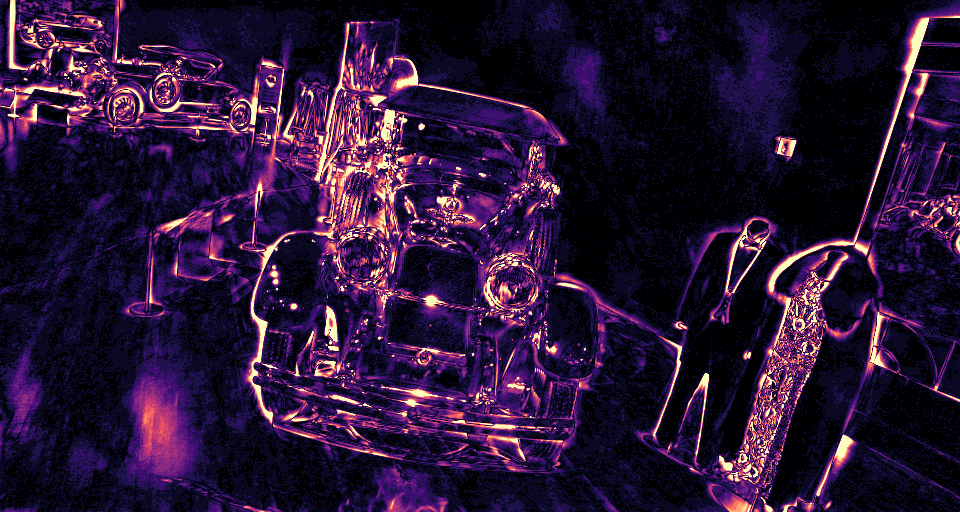} &
        \rcell[width=\linewidth]{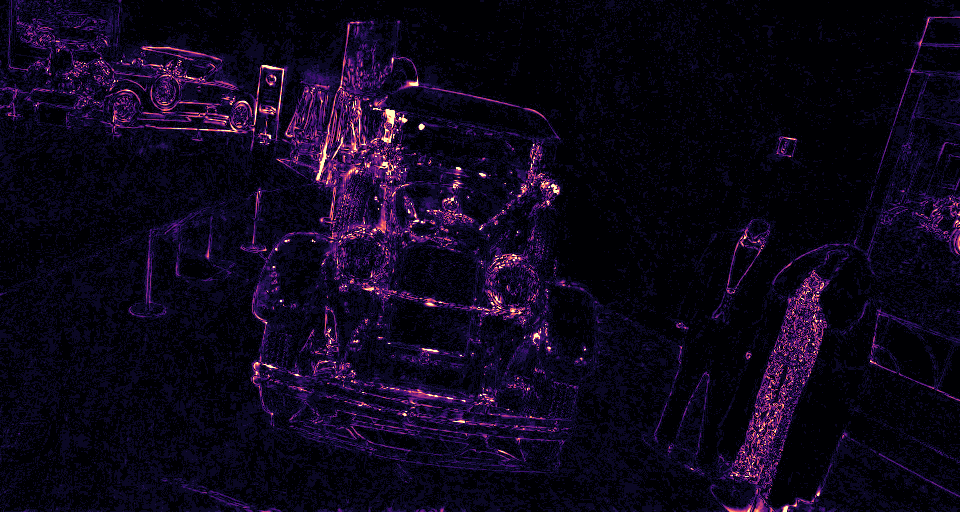} &
        \rcell[width=\linewidth]{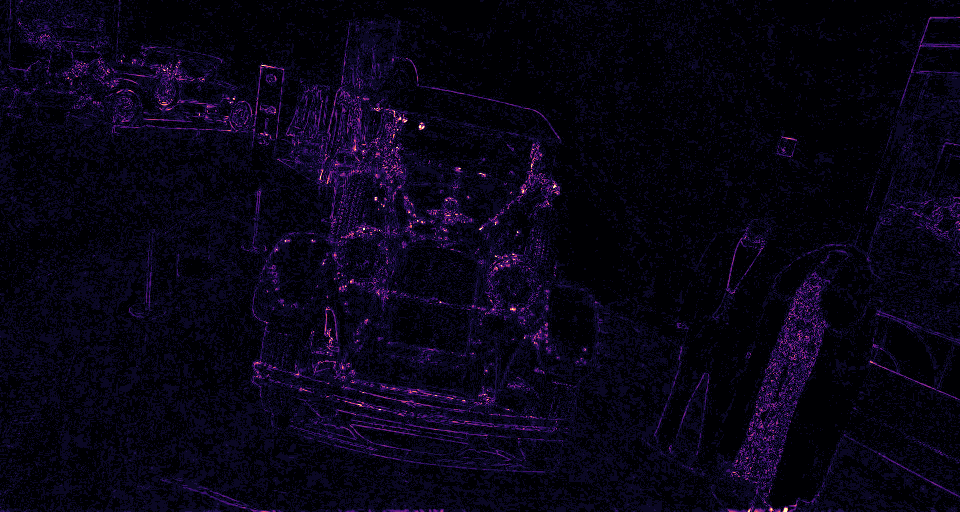} &
        \rcell[width=\linewidth]{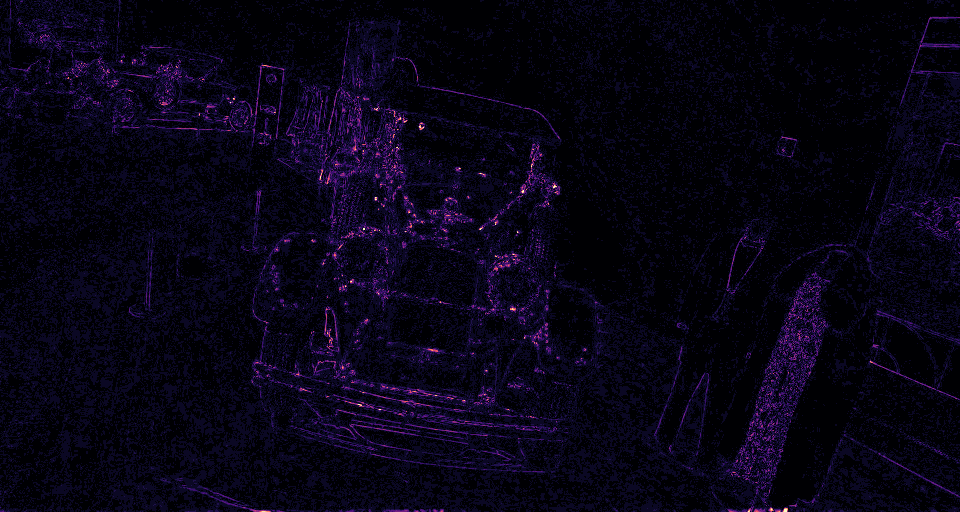} &
        \rotatebox{90}{\smaller[1]{Init error}} &
        \rcell[width=\linewidth]{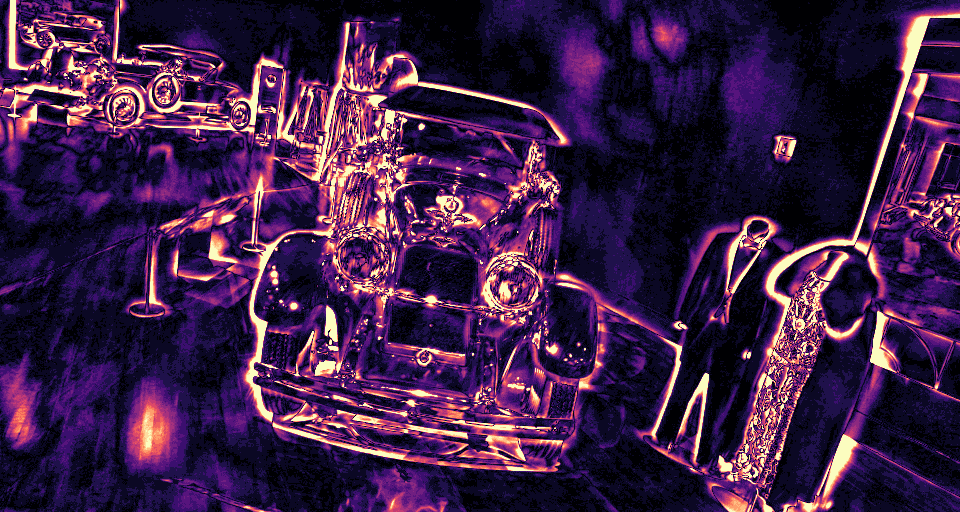} \\
        \noalign{\vskip 2\tabcolsep}
        \rotatebox{90}{\makecell{\smaller[1]{\ourshort{} error}}} &
        \rcell[width=\linewidth]{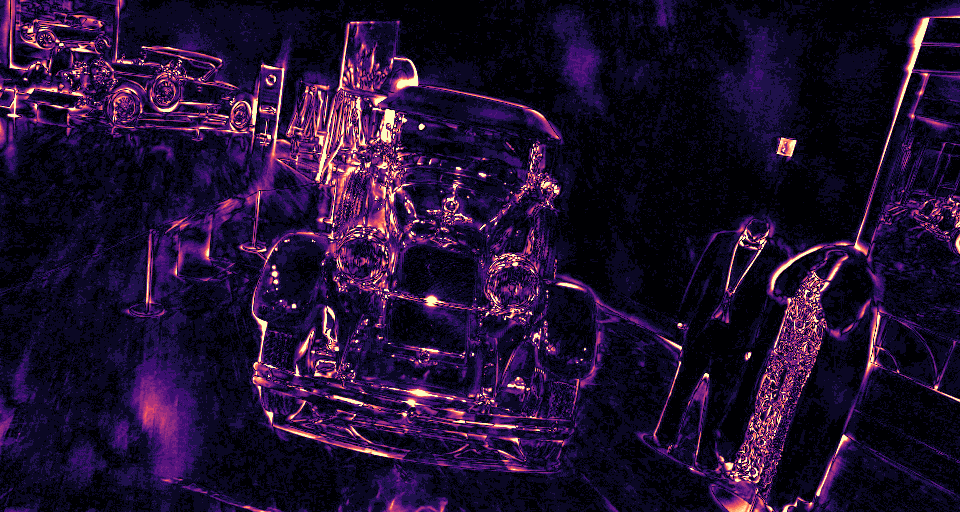} &
        \rcell[width=\linewidth]{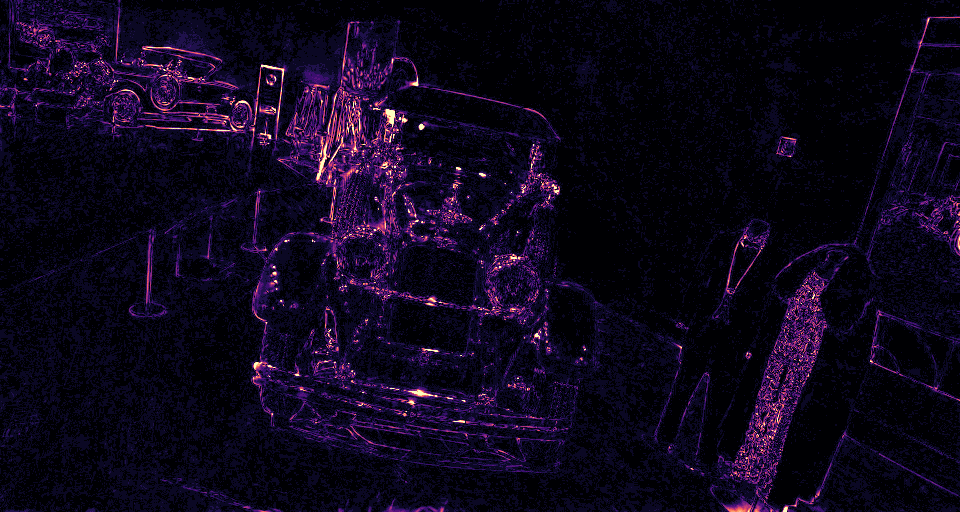} &
        \rcell[width=\linewidth]{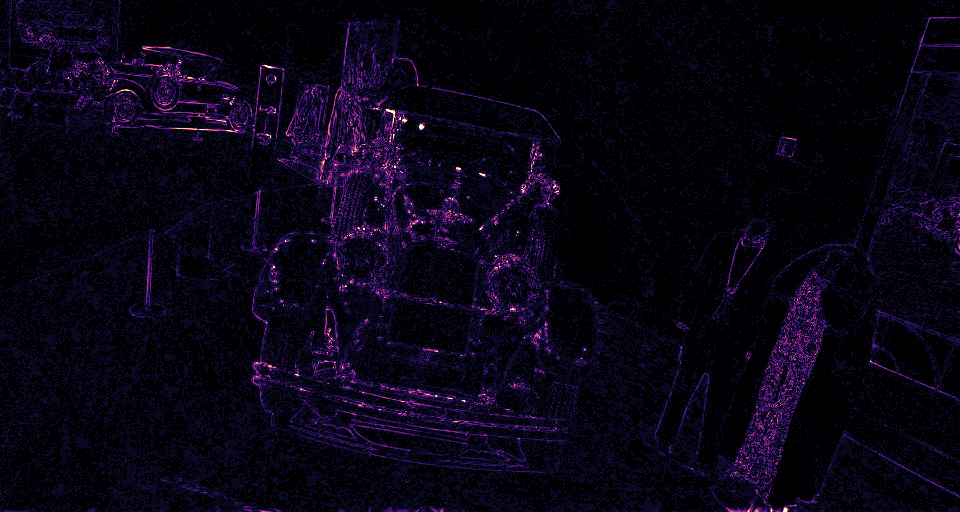} &
        \rcell[width=\linewidth]{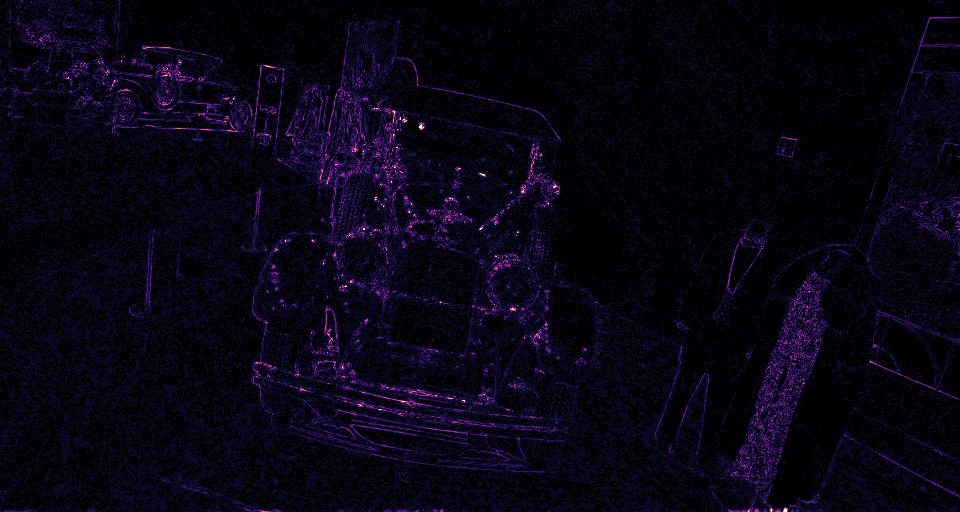} &
        \rcell[width=\linewidth]{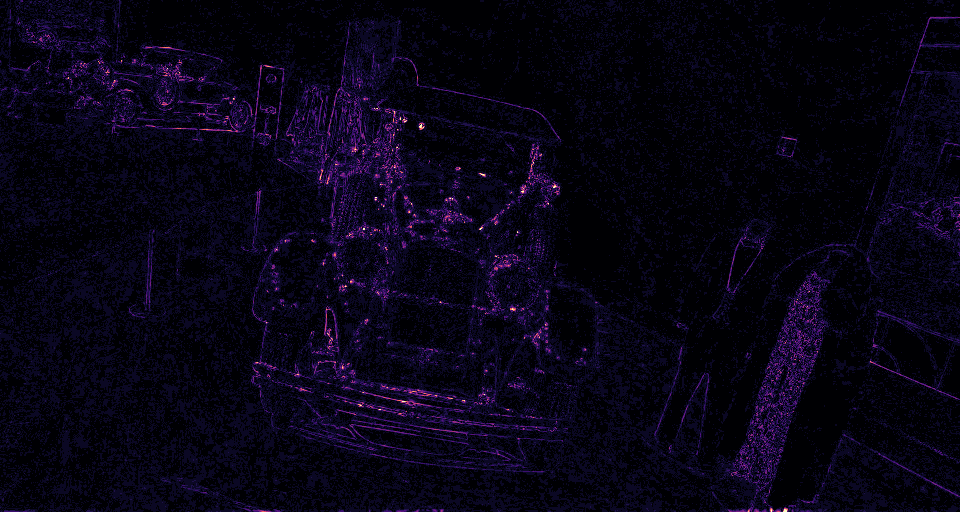} &
         &
         \\
    \end{tabular}
}
    \caption{
        \textbf{Additional Sparse-View Results on DL3DV~\cite{ling2024dl3dv}}. Scene reconstructions in the $8$ views, zero-shot high-resolution setting ($512 \times 960$), across varied scene types. \ourshort{} recovers sharp detail in far fewer iterations than \adam{} and the sparse baselines.
        On the second scene, ReSplat diverges and its expanding Gaussians exceed the available GPU memory at iteration $923$, so no reconstruction exists for the last two columns.
    }
    \label{fig:dl3dv-comp}
\end{figure*}

\clearpage
}
\afterpage{
    \clearpage

\begin{figure*}[p]
    \centering

    \resizebox{1.0\linewidth}{!}{%
    \setlength{\tabcolsep}{0.002\linewidth} %
    \renewcommand{\arraystretch}{0} %

    \begin{tabular}{@{}%
        >{\centering\arraybackslash}m{0.03\linewidth}
        >{\centering\arraybackslash}m{0.24\linewidth}
        >{\centering\arraybackslash}m{0.24\linewidth}
        >{\centering\arraybackslash}m{0.24\linewidth}
        >{\centering\arraybackslash}m{0.24\linewidth}
        >{\centering\arraybackslash}m{0.24\linewidth}
        >{\centering\arraybackslash}m{0.02\linewidth}
        >{\centering\arraybackslash}m{0.24\linewidth}
    @{}}
         & $t = 2$ & $t = 10$ & $t = 100$ & $t = 1000$ & $t = 2000$ &  &  \\[3pt]

        \rotatebox{90}{\makecell{\smaller[1]{ReSplat~\cite{xu2025resplat}}}} &
        \pcell[width=\linewidth]{19.5}{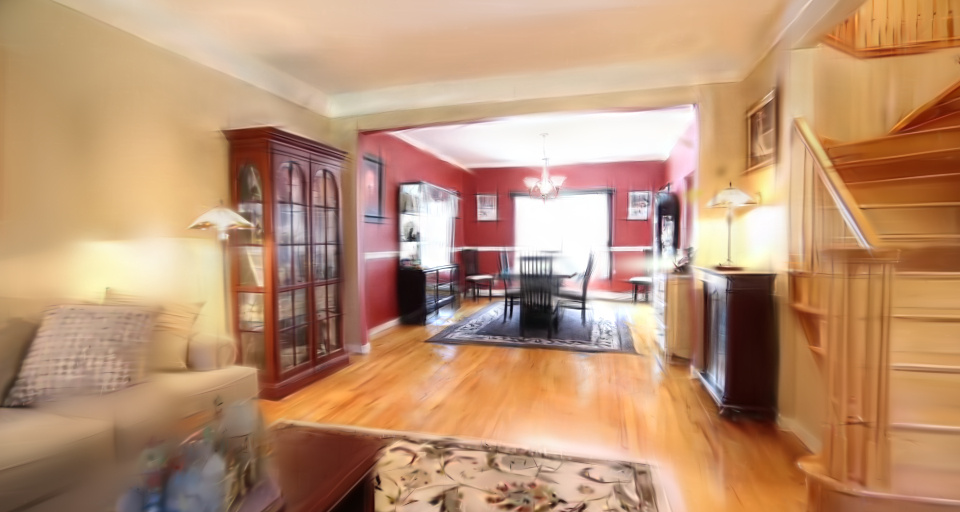} &
        \pcell[width=\linewidth]{19.7}{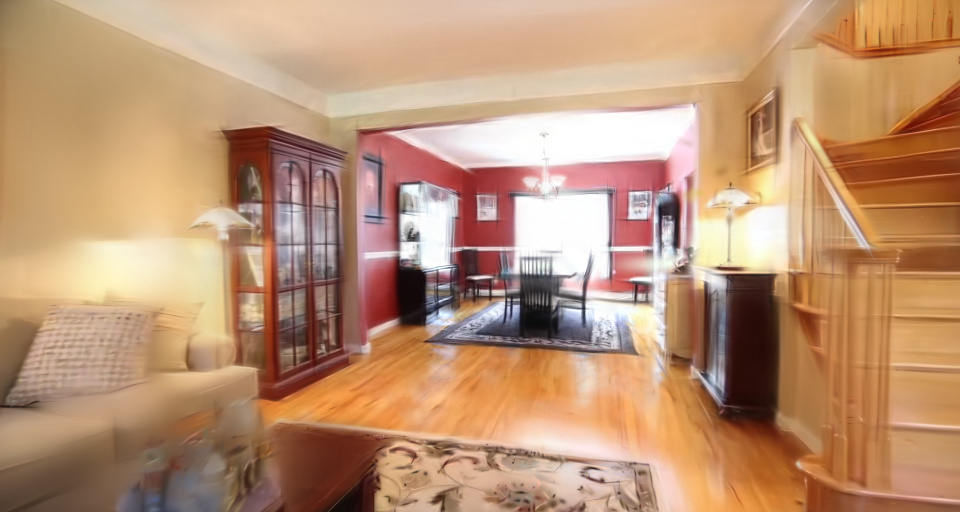} &
        \pcell[width=\linewidth]{11.8}{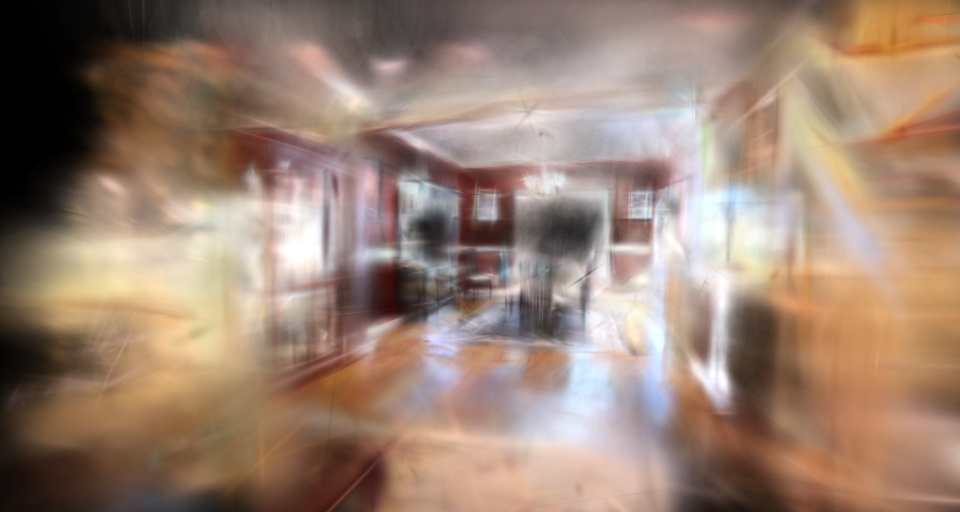} &
        \pcell[width=\linewidth]{4.7}{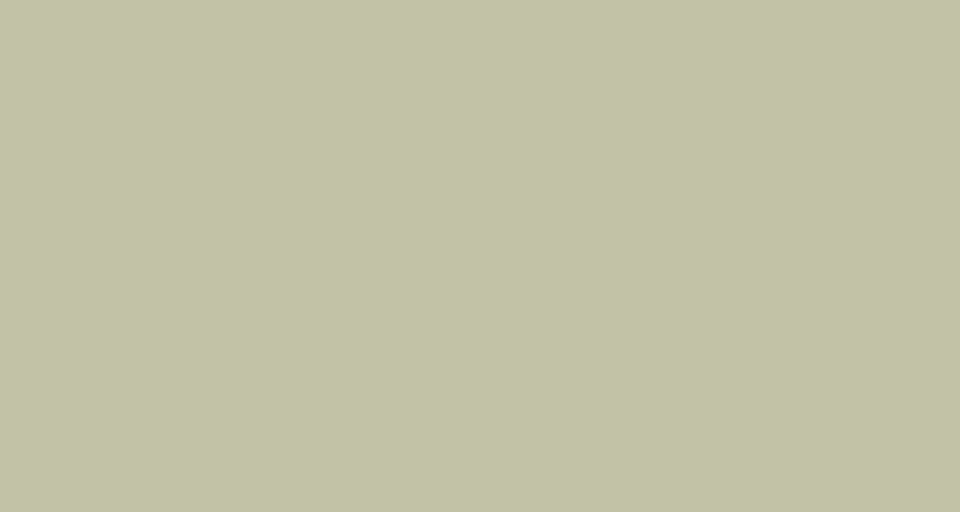} &
        \pcell[width=\linewidth]{4.0}{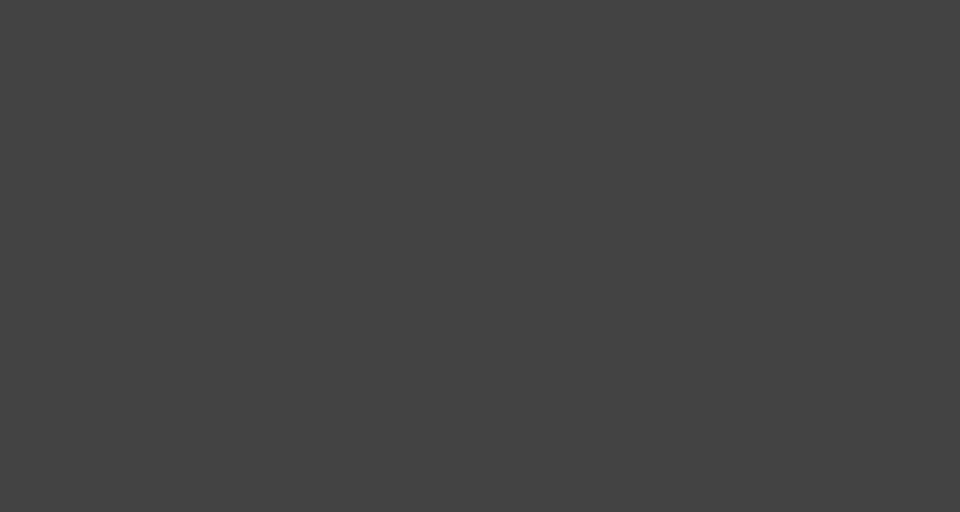} &
        \rotatebox{90}{\smaller[1]{Init}} &
        \pcell[width=\linewidth]{17.7}{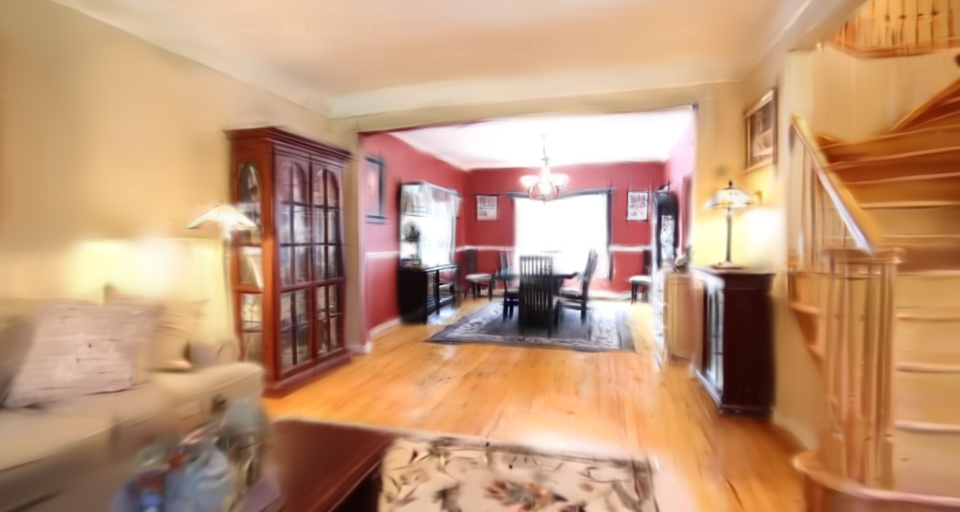} \\
        \noalign{\vskip 2\tabcolsep}
        \rotatebox{90}{\makecell{\smaller[1]{\lob{}}}} &
        \pcell[width=\linewidth]{21.8}{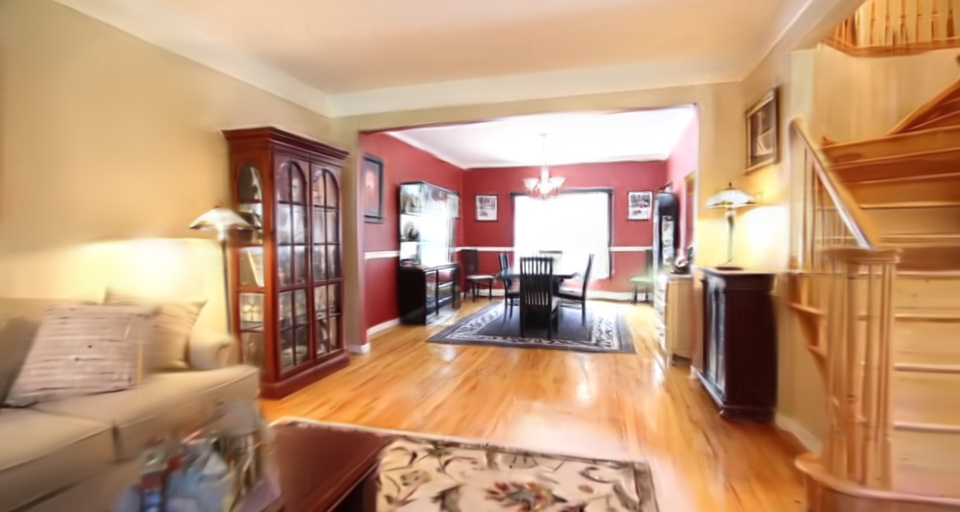} &
        \pcell[width=\linewidth]{25.9}{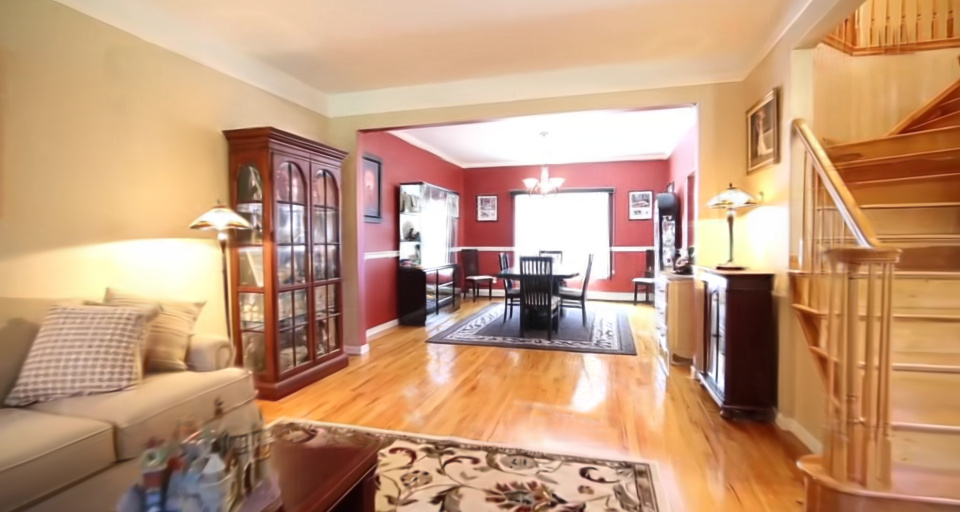} &
        \pcell[width=\linewidth]{26.8}{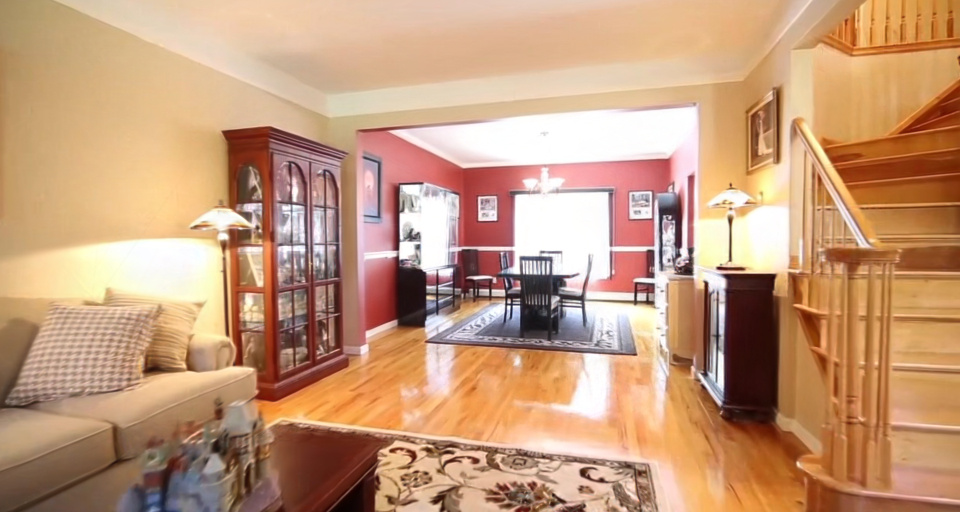} &
        \rcell[width=\linewidth]{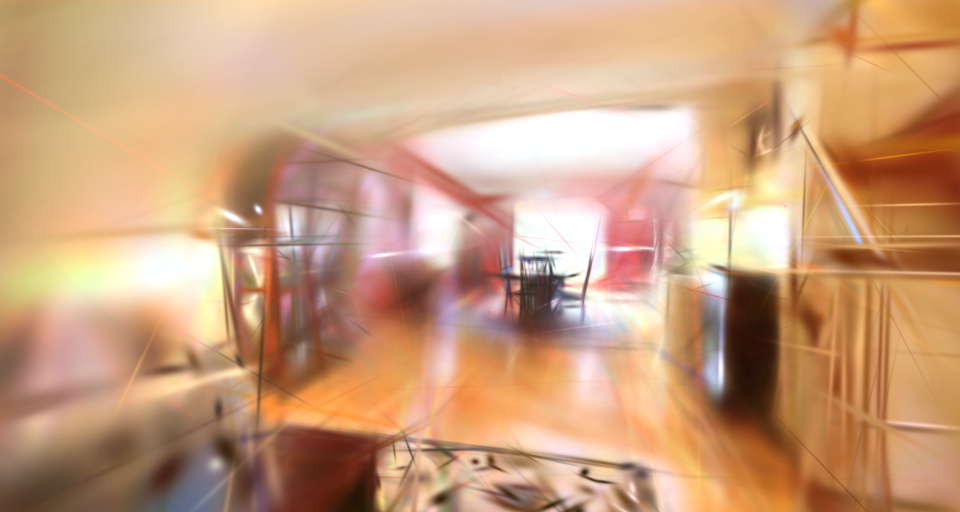} &
        \rcell[width=\linewidth]{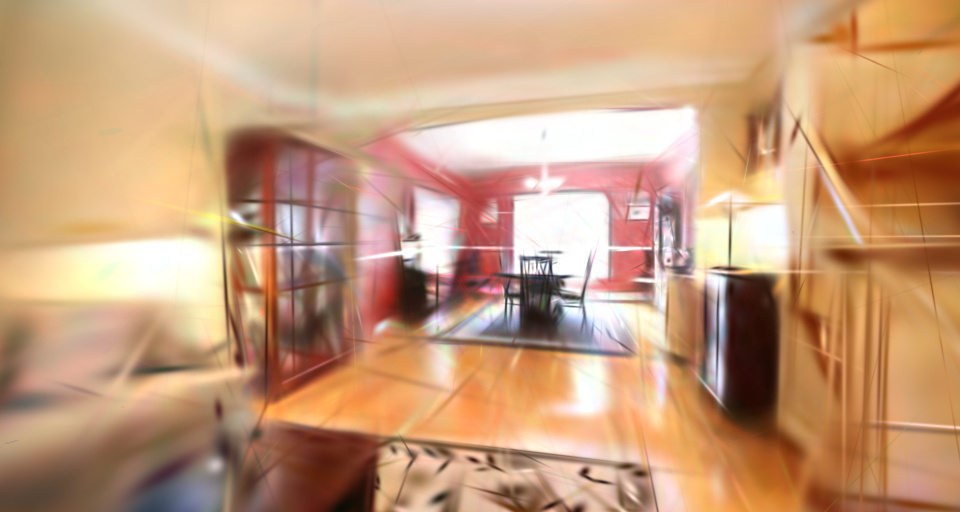} &
        \rotatebox{90}{\smaller[1]{Ref}} &
        \rcell[width=\linewidth]{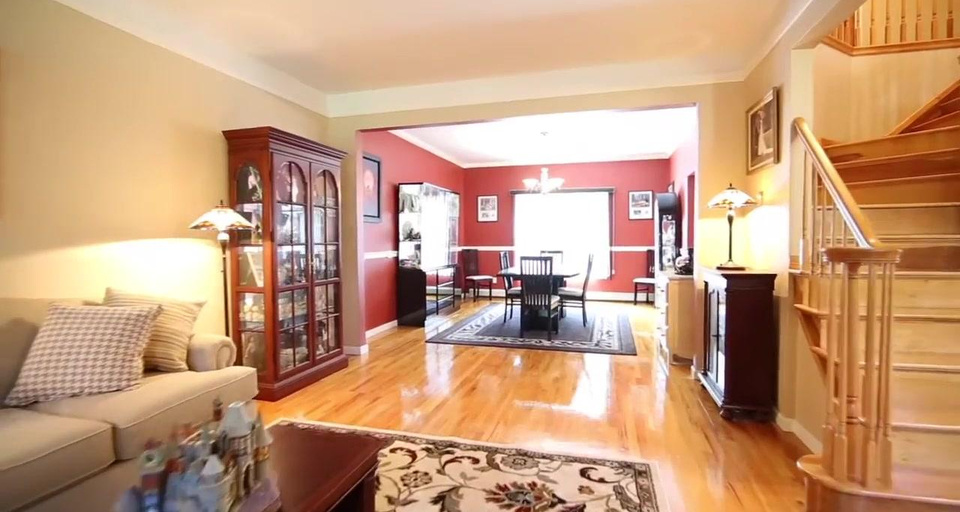} \\
        \noalign{\vskip 2\tabcolsep}
        \rotatebox{90}{\makecell{\smaller[1]{\adam{}}}} &
        \pcell[width=\linewidth]{18.3}{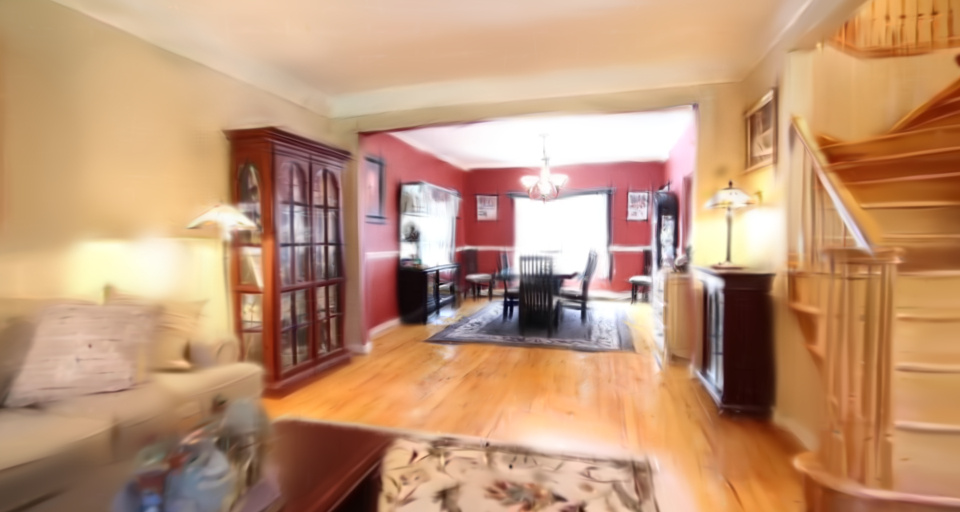} &
        \pcell[width=\linewidth]{20.2}{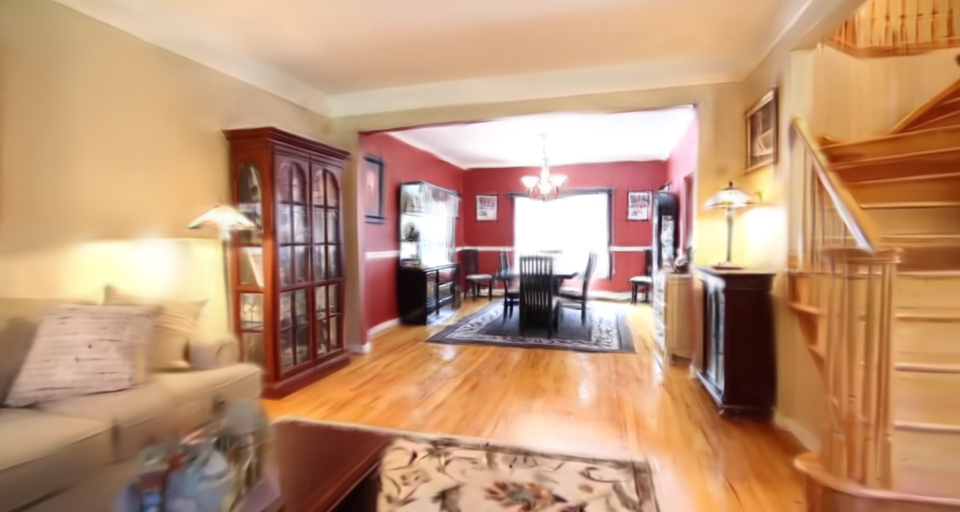} &
        \pcell[width=\linewidth]{25.8}{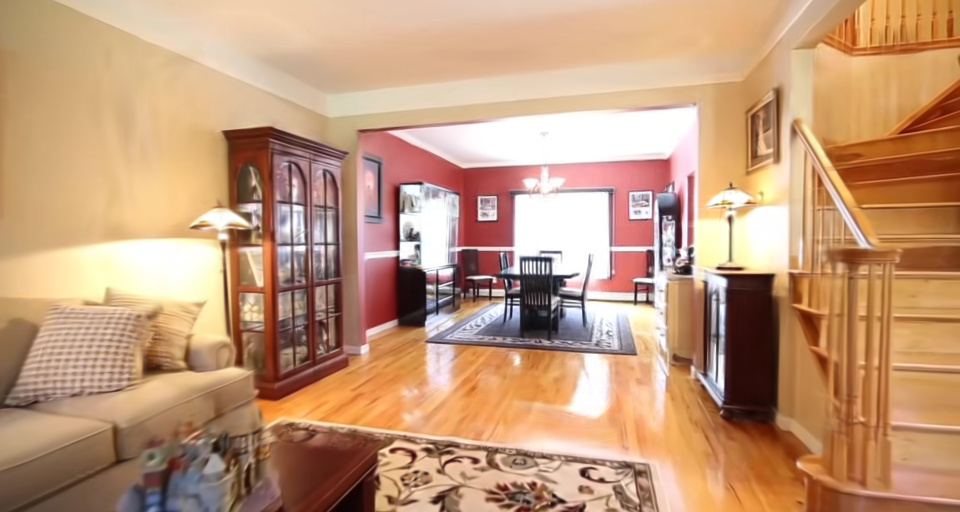} &
        \pcell[width=\linewidth]{26.5}{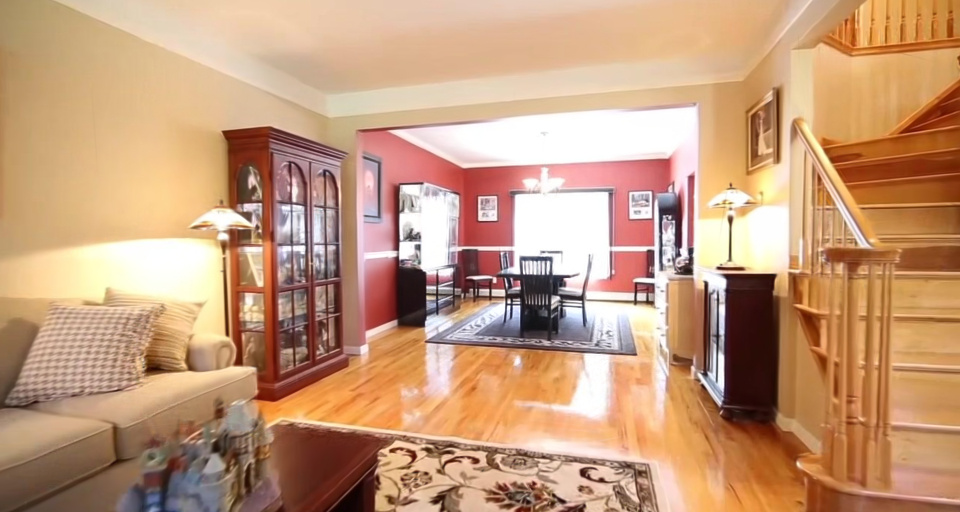} &
        \pcell[width=\linewidth]{26.4}{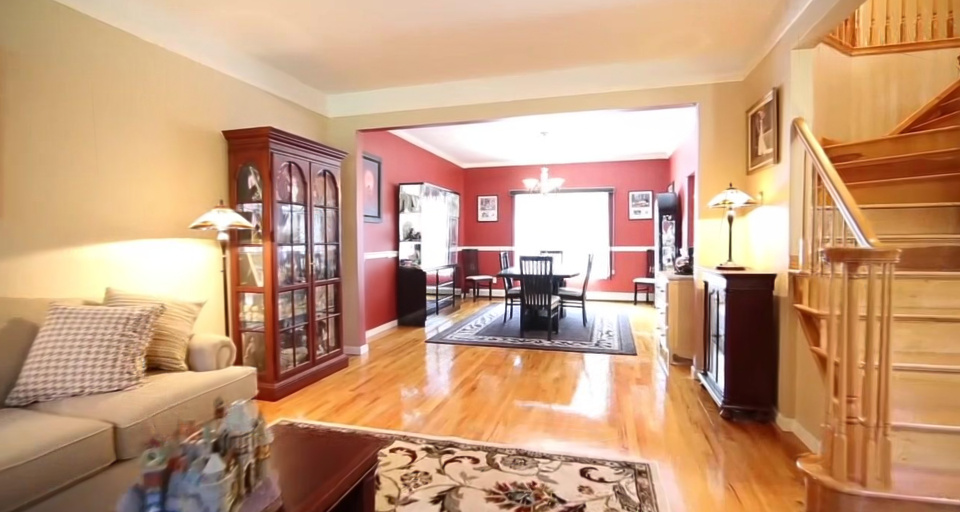} &
         &
         \\
        \noalign{\vskip 2\tabcolsep}
        \rotatebox{90}{\makecell{\smaller[1]{\ourshort{}}}} &
        \pcell[width=\linewidth]{18.8}{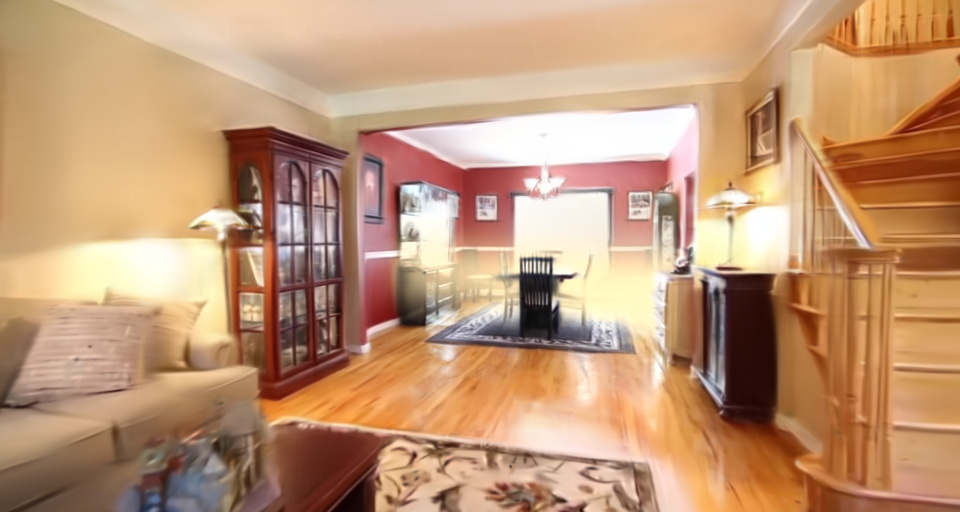} &
        \pcell[width=\linewidth]{24.3}{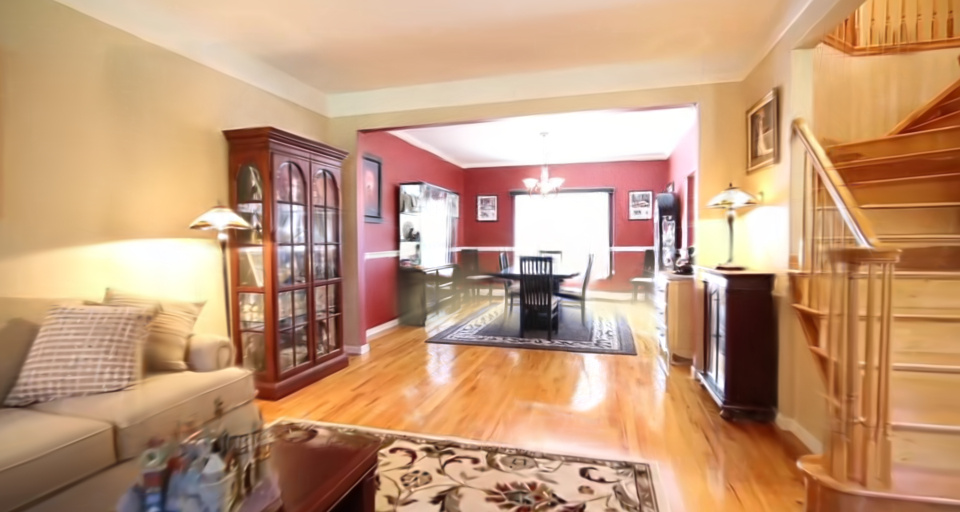} &
        \pcell[width=\linewidth]{27.0}{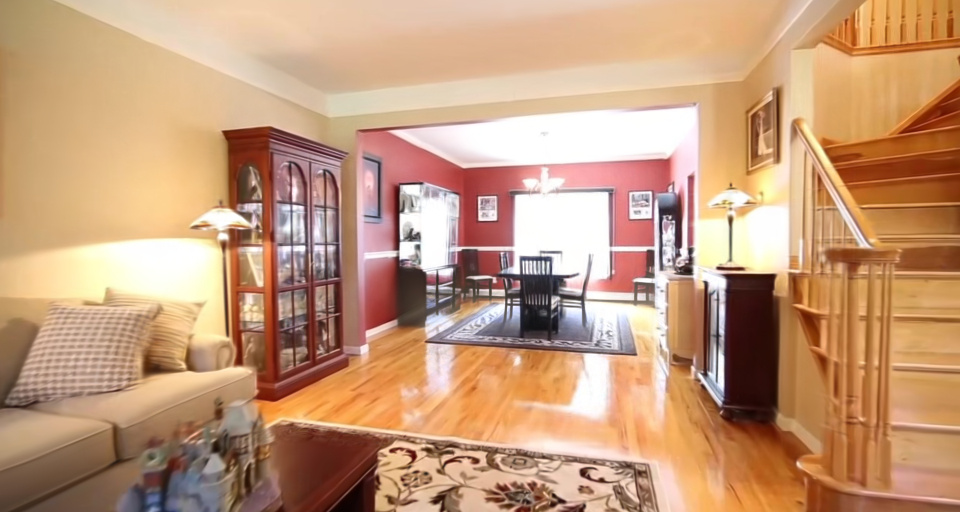} &
        \pcell[width=\linewidth]{27.4}{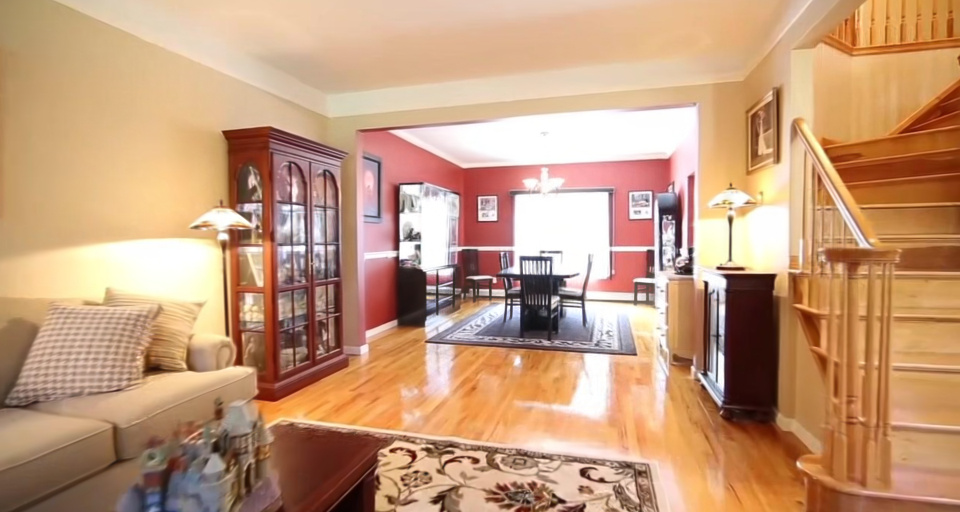} &
        \pcell[width=\linewidth]{27.4}{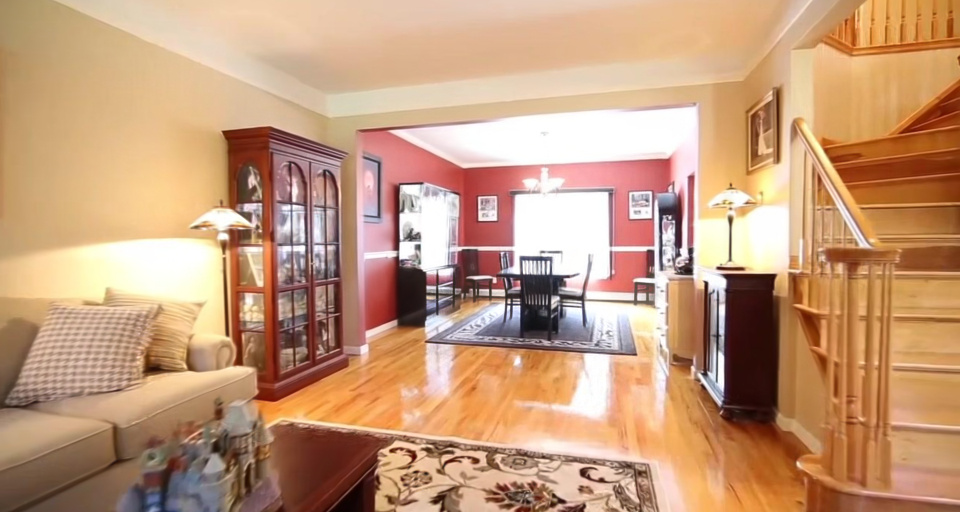} &
         &
         \\
        \noalign{\vskip 2\tabcolsep}
        \rotatebox{90}{\makecell{\smaller[1]{\adam{} error}}} &
        \rcell[width=\linewidth]{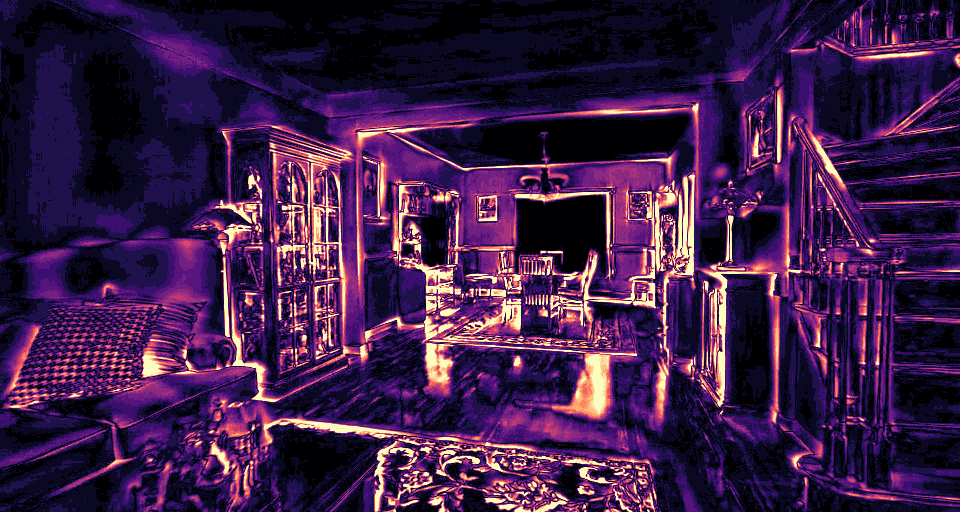} &
        \rcell[width=\linewidth]{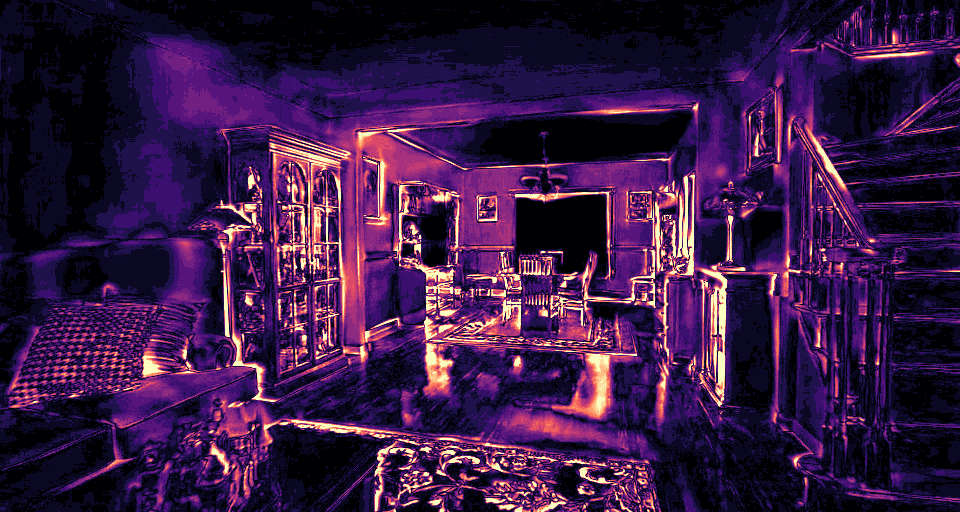} &
        \rcell[width=\linewidth]{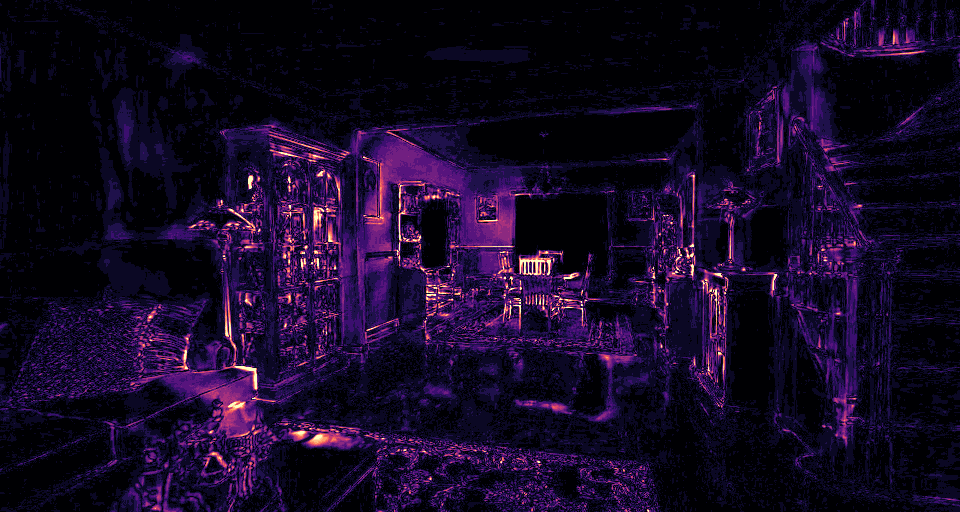} &
        \rcell[width=\linewidth]{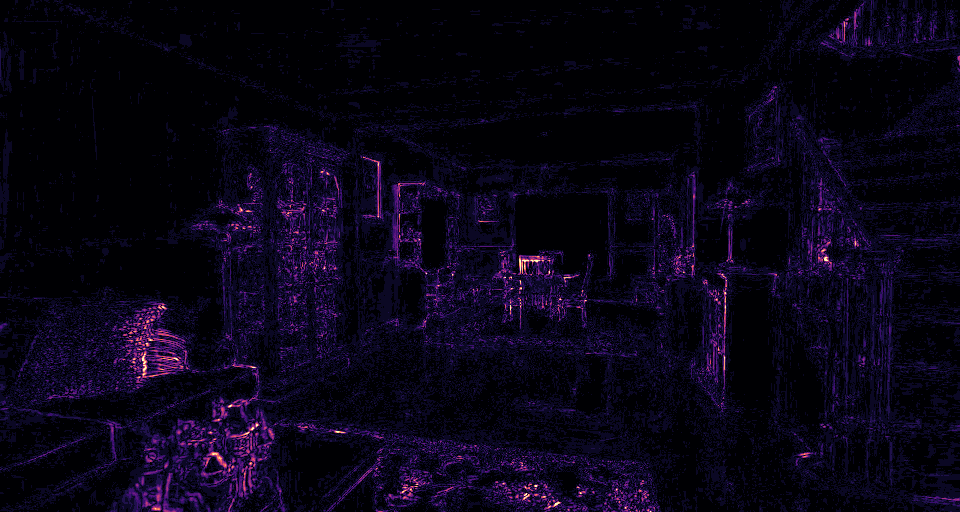} &
        \rcell[width=\linewidth]{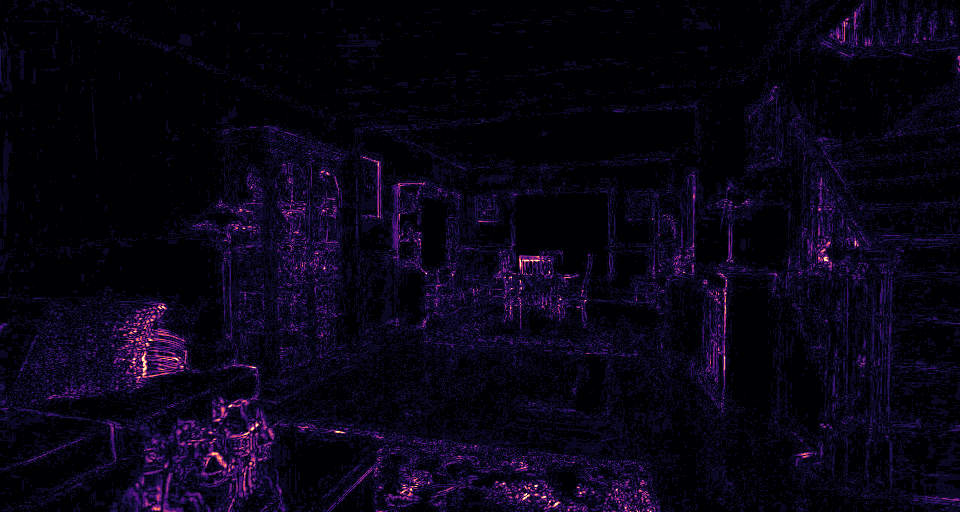} &
        \rotatebox{90}{\smaller[1]{Init error}} &
        \rcell[width=\linewidth]{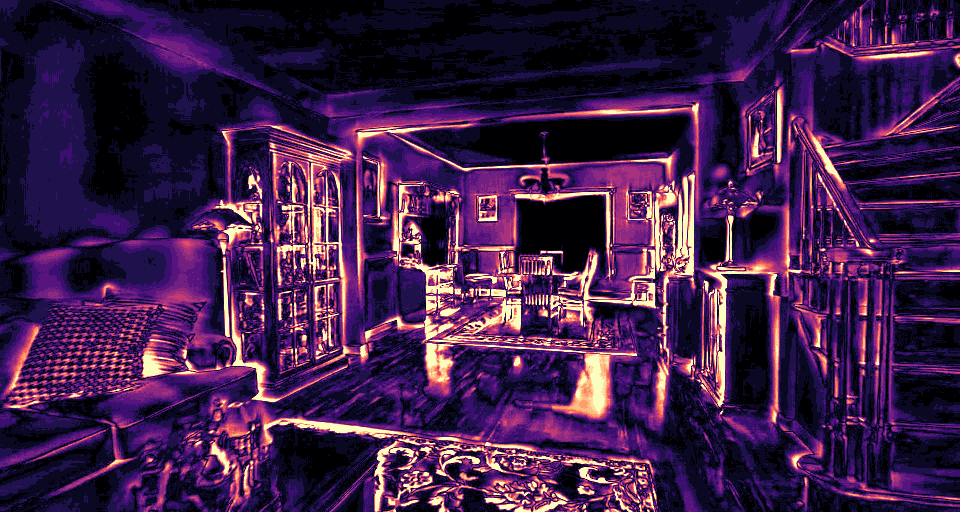} \\
        \noalign{\vskip 2\tabcolsep}
        \rotatebox{90}{\makecell{\smaller[1]{\ourshort{} error}}} &
        \rcell[width=\linewidth]{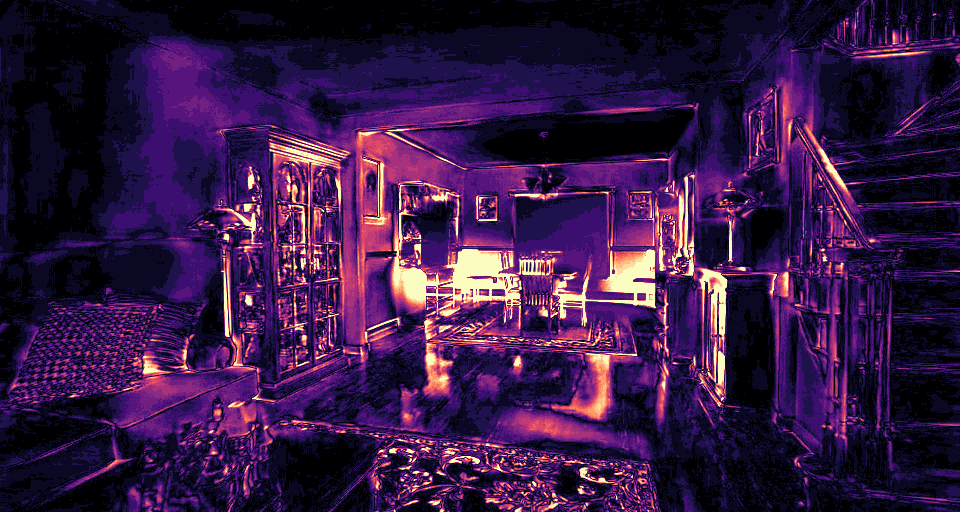} &
        \rcell[width=\linewidth]{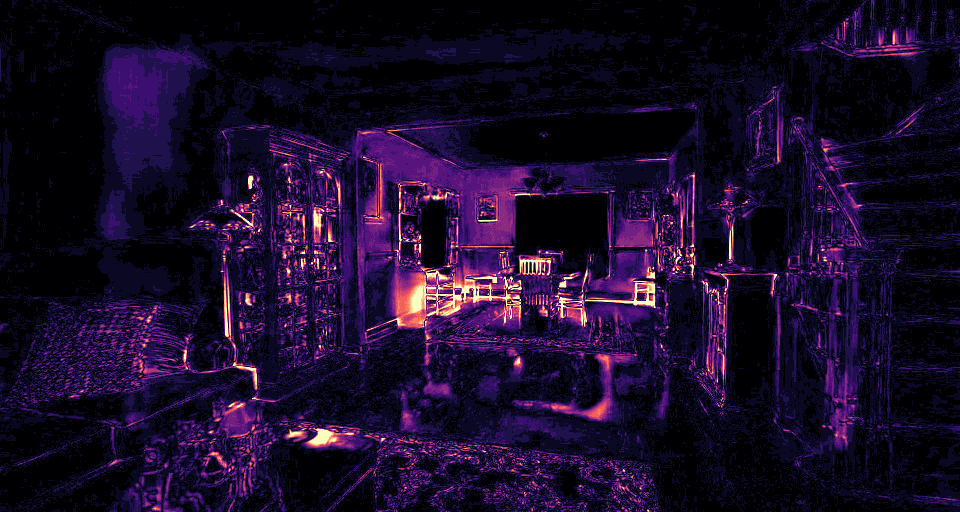} &
        \rcell[width=\linewidth]{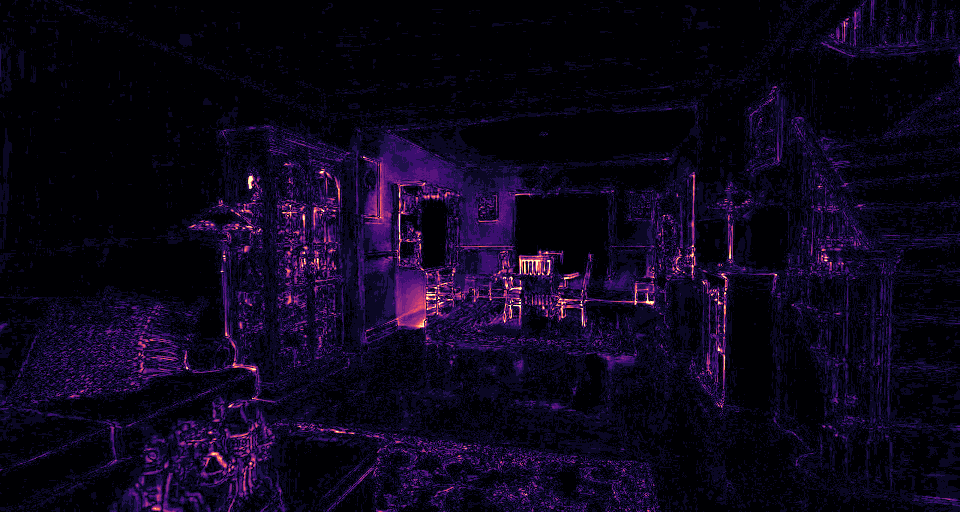} &
        \rcell[width=\linewidth]{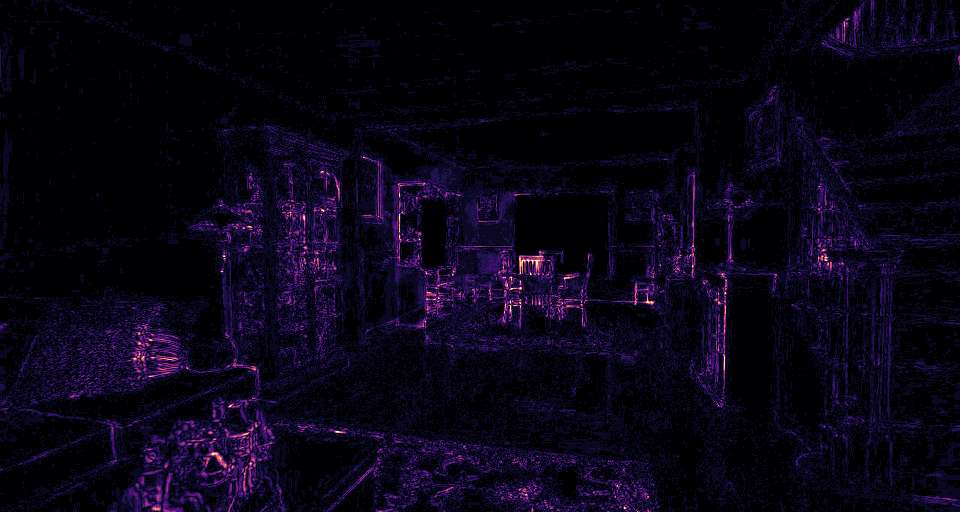} &
        \rcell[width=\linewidth]{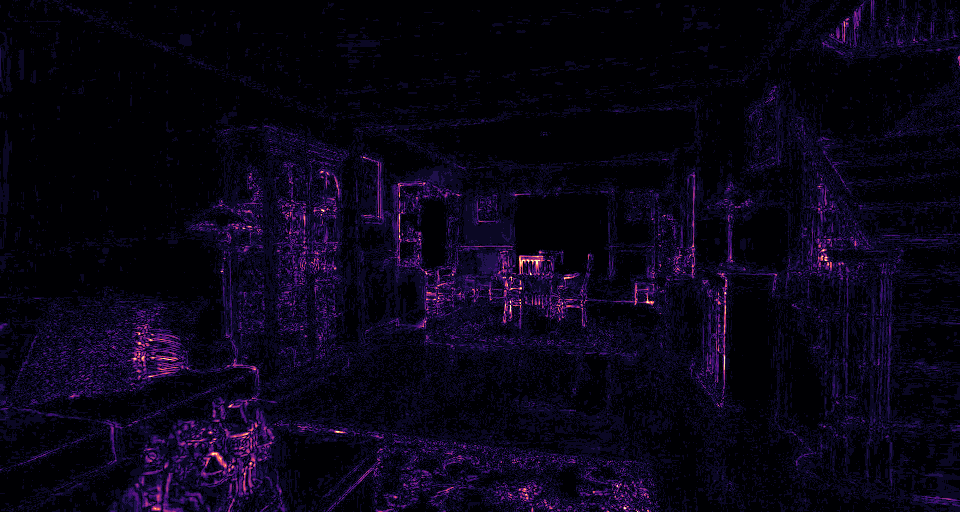} &
         &
         \\
        \noalign{\vskip 2\tabcolsep}
        \rotatebox{90}{\makecell{\smaller[1]{ReSplat~\cite{xu2025resplat}}}} &
        \pcell[width=\linewidth]{18.5}{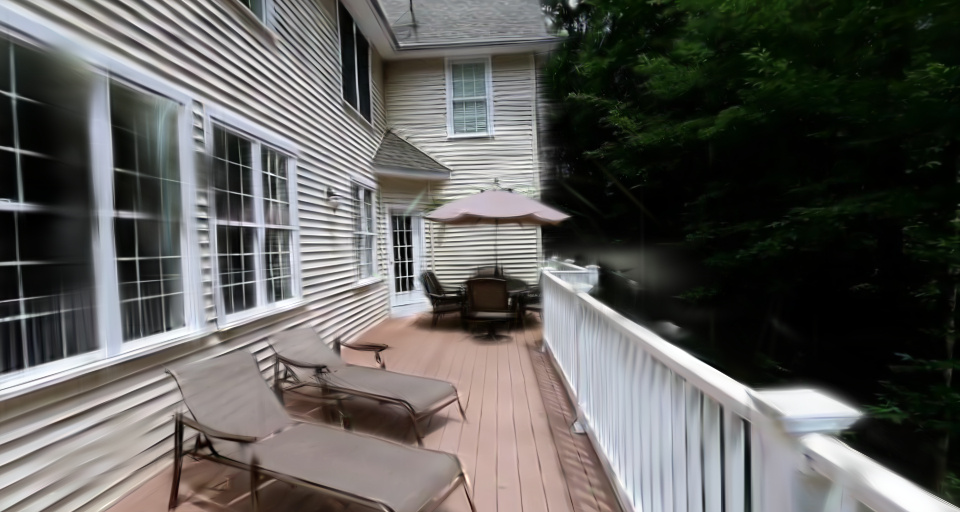} &
        \pcell[width=\linewidth]{18.8}{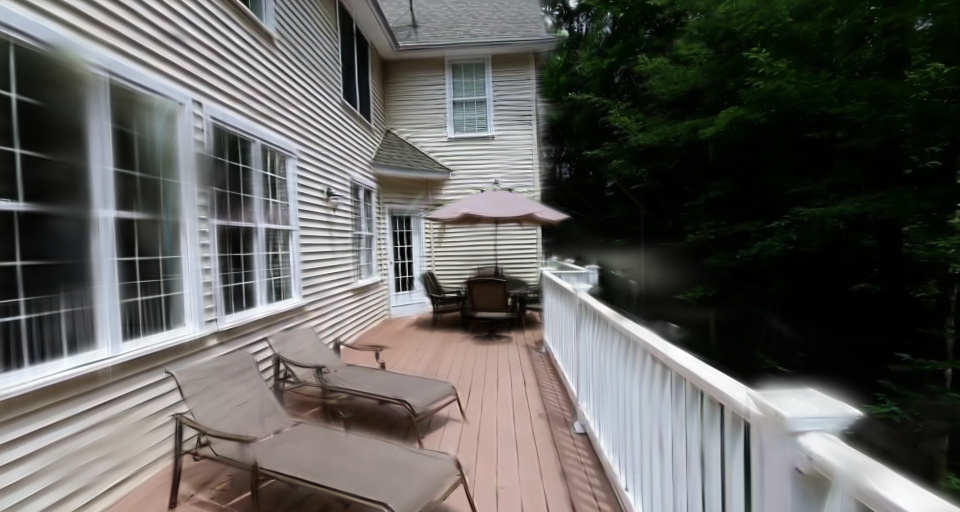} &
        \pcell[width=\linewidth]{10.0}{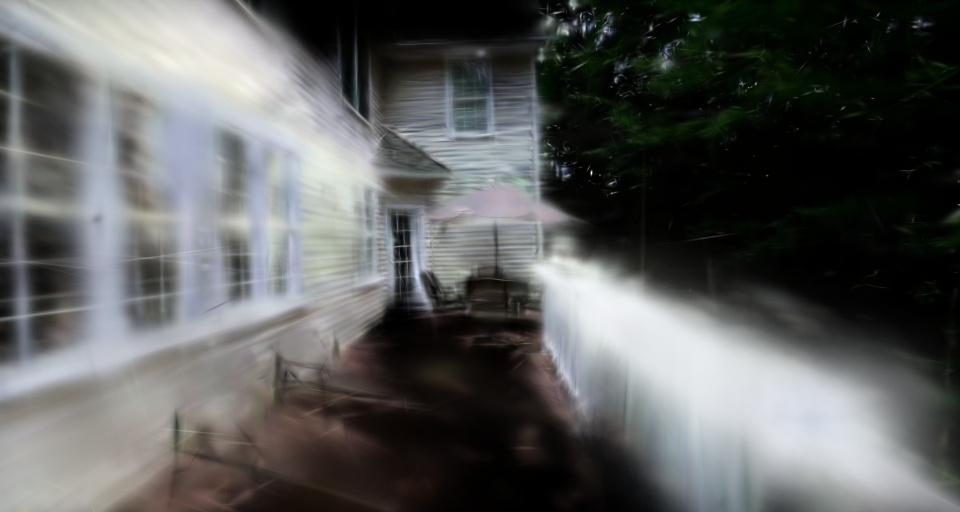} &
        \pcell[width=\linewidth]{5.7}{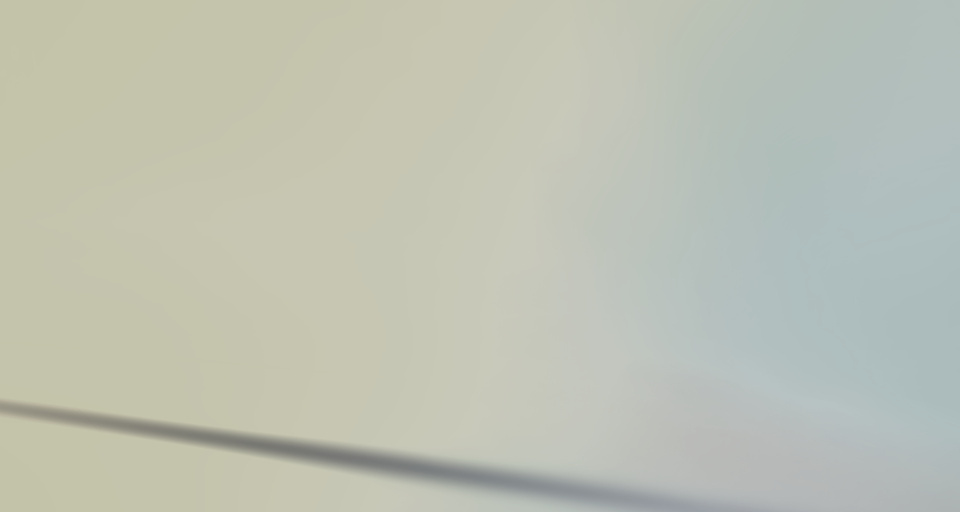} &
        \pcell[width=\linewidth]{5.4}{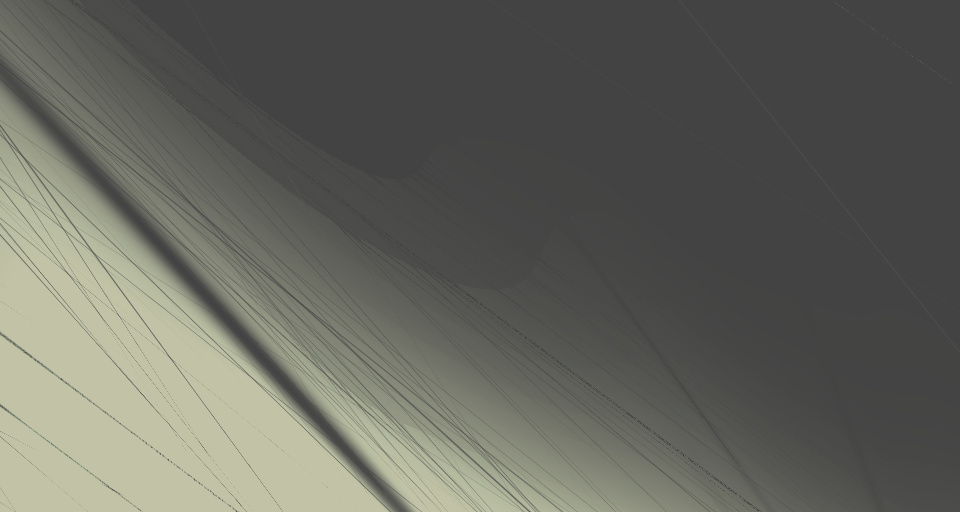} &
        \rotatebox{90}{\smaller[1]{Init}} &
        \pcell[width=\linewidth]{15.9}{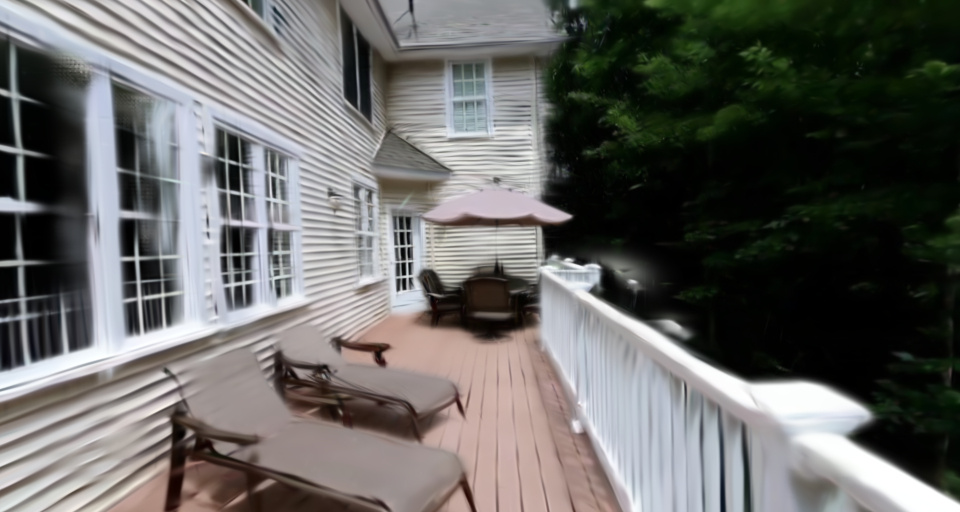} \\
        \noalign{\vskip 2\tabcolsep}
        \rotatebox{90}{\makecell{\smaller[1]{\lob{}}}} &
        \pcell[width=\linewidth]{20.5}{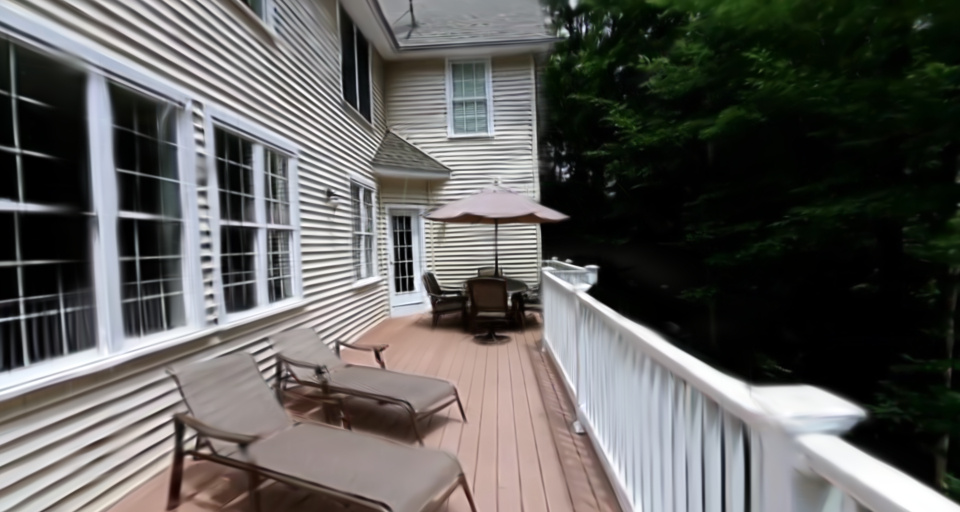} &
        \pcell[width=\linewidth]{23.6}{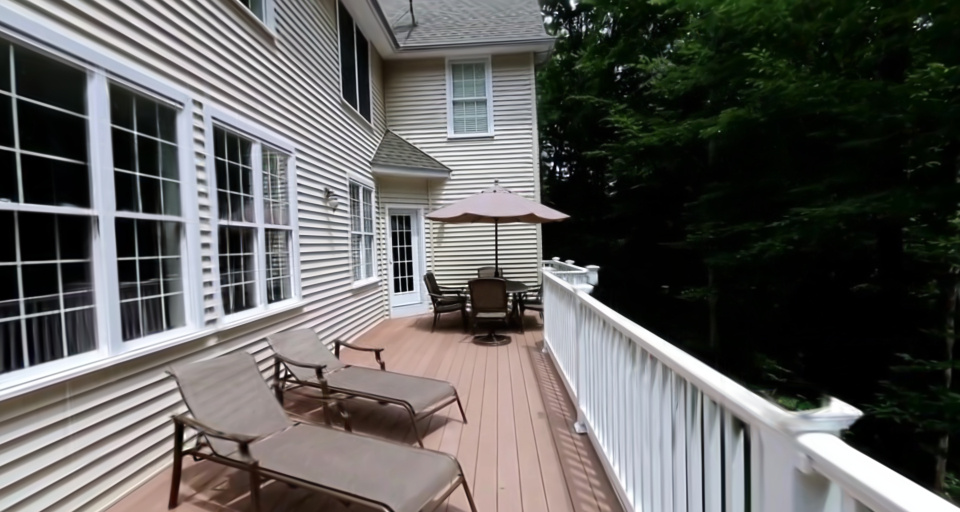} &
        \pcell[width=\linewidth]{24.3}{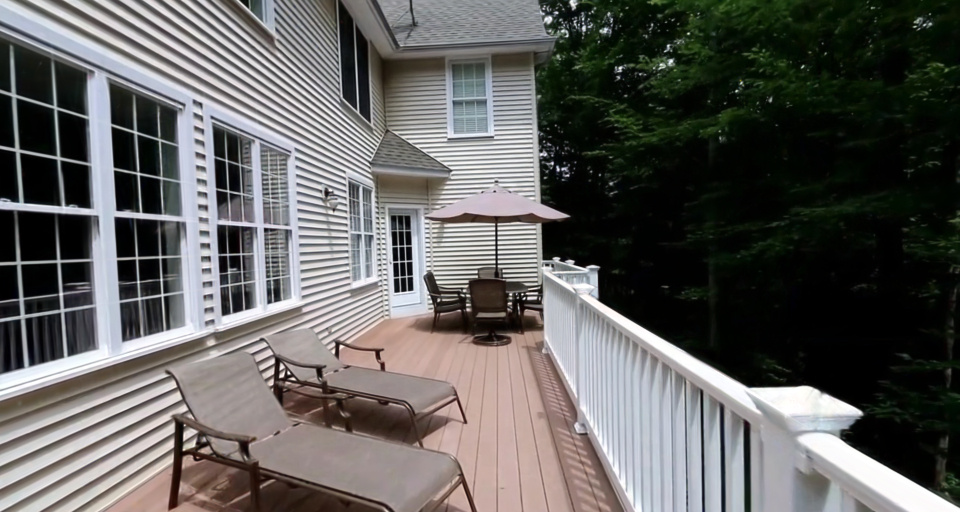} &
        \rcell[width=\linewidth]{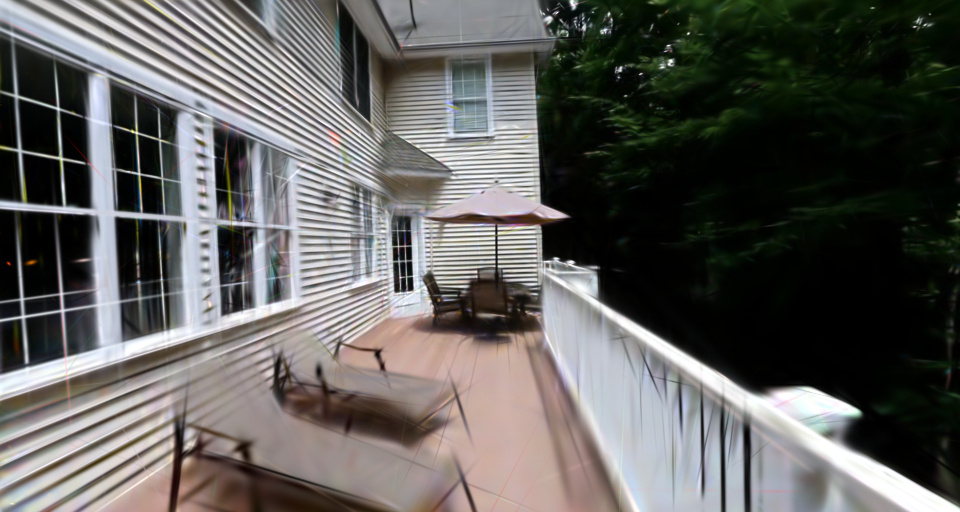} &
        \rcell[width=\linewidth]{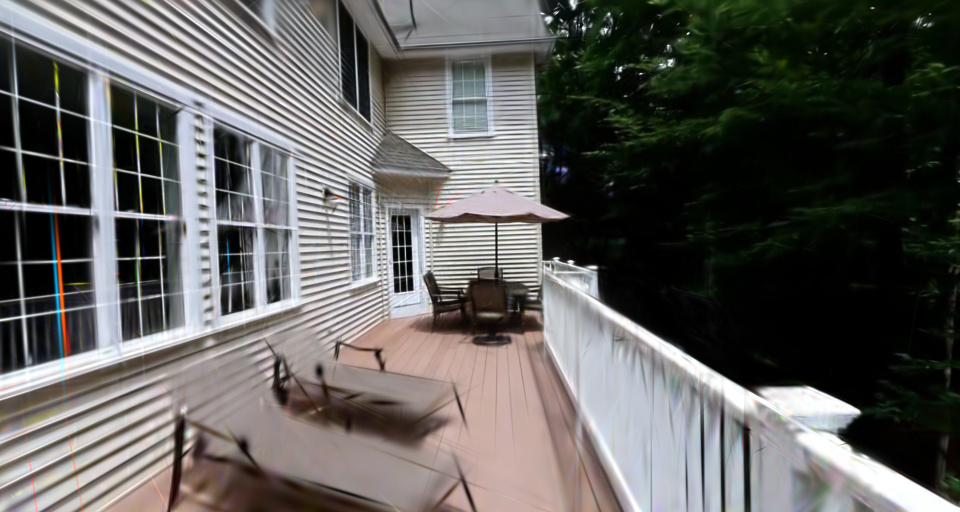} &
        \rotatebox{90}{\smaller[1]{Ref}} &
        \rcell[width=\linewidth]{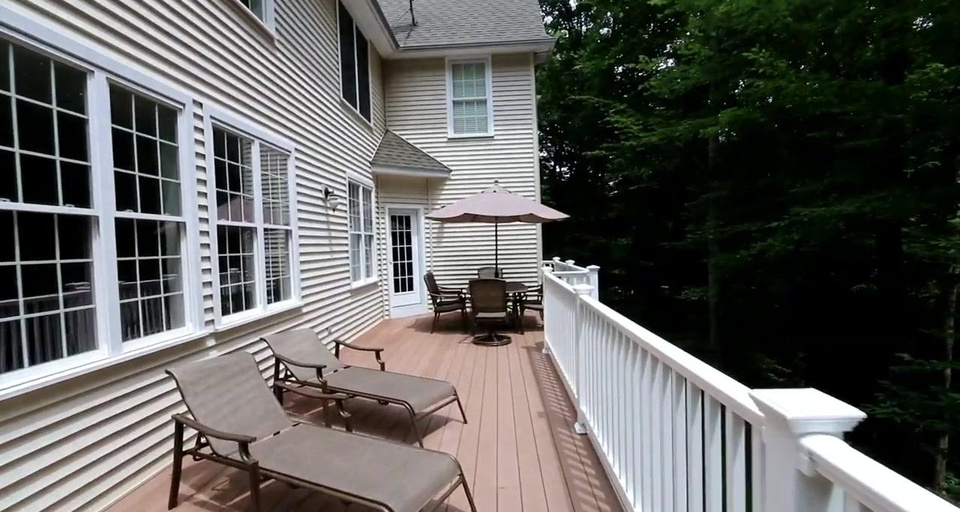} \\
        \noalign{\vskip 2\tabcolsep}
        \rotatebox{90}{\makecell{\smaller[1]{\adam{}}}} &
        \pcell[width=\linewidth]{16.5}{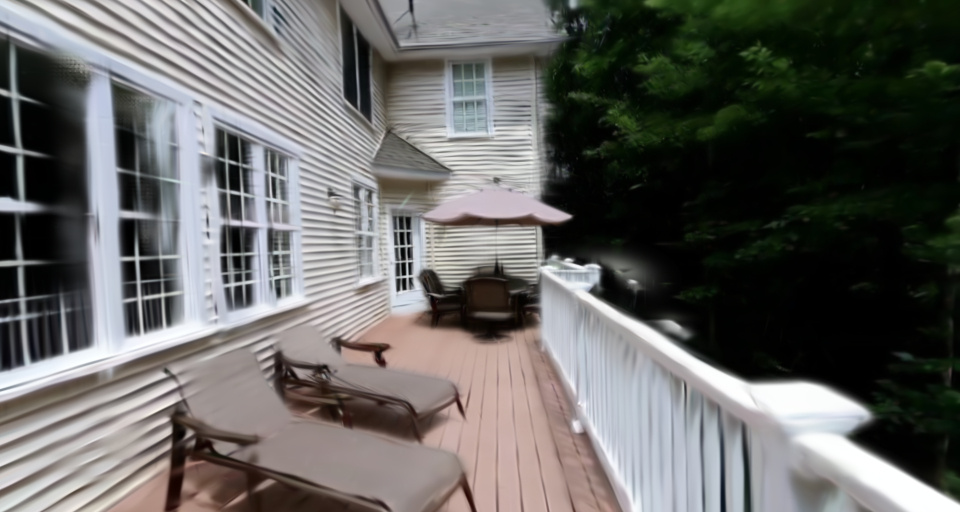} &
        \pcell[width=\linewidth]{18.6}{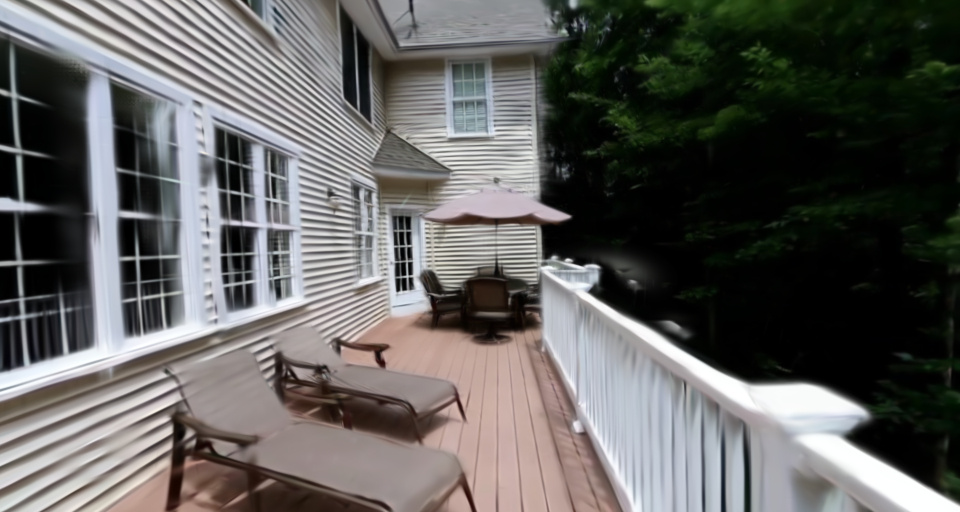} &
        \pcell[width=\linewidth]{23.6}{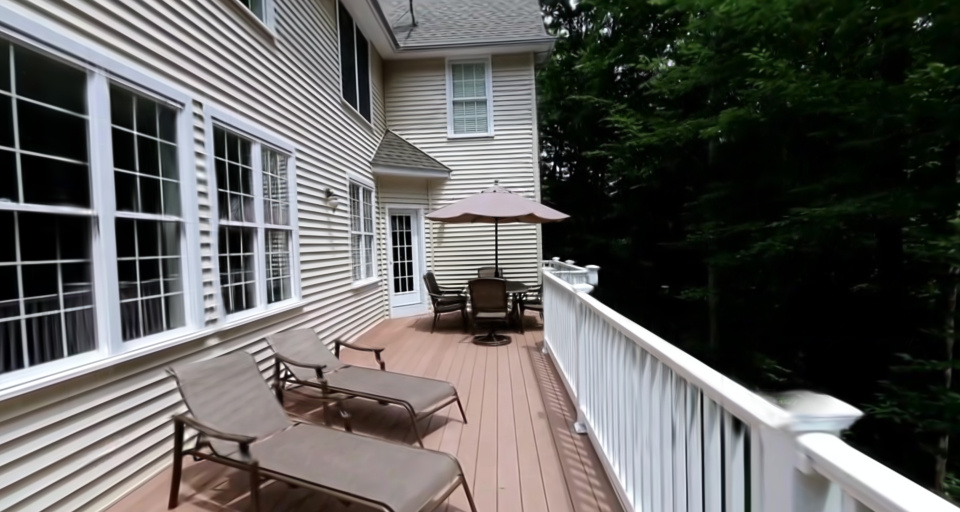} &
        \pcell[width=\linewidth]{24.2}{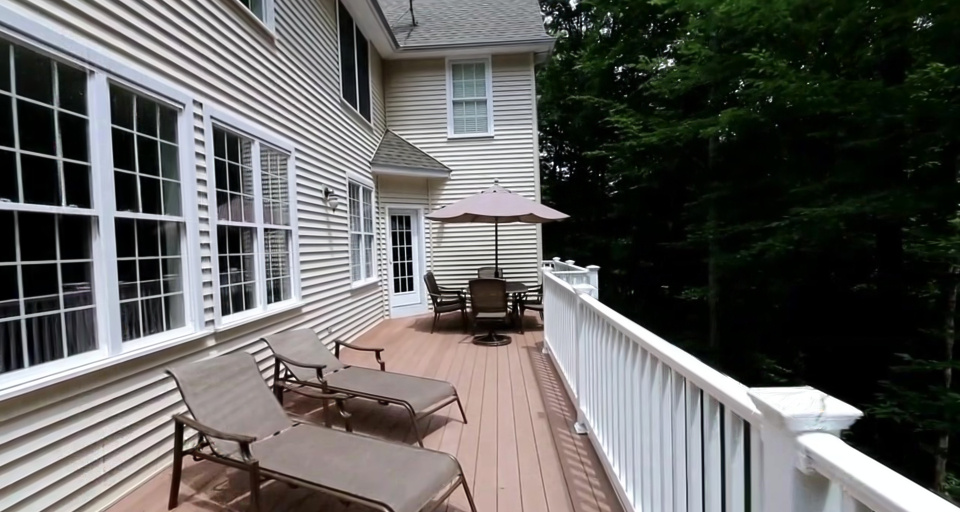} &
        \pcell[width=\linewidth]{24.1}{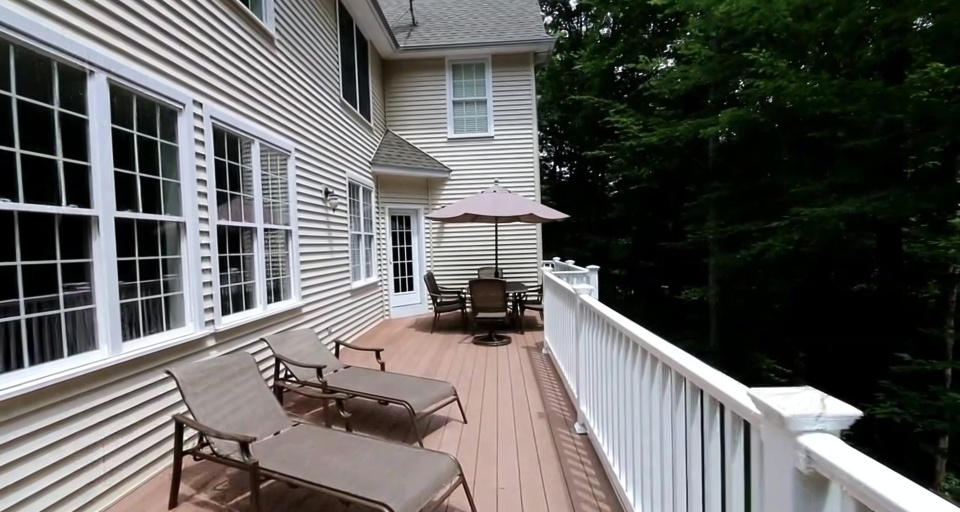} &
         &
         \\
        \noalign{\vskip 2\tabcolsep}
        \rotatebox{90}{\makecell{\smaller[1]{\ourshort{}}}} &
        \pcell[width=\linewidth]{20.5}{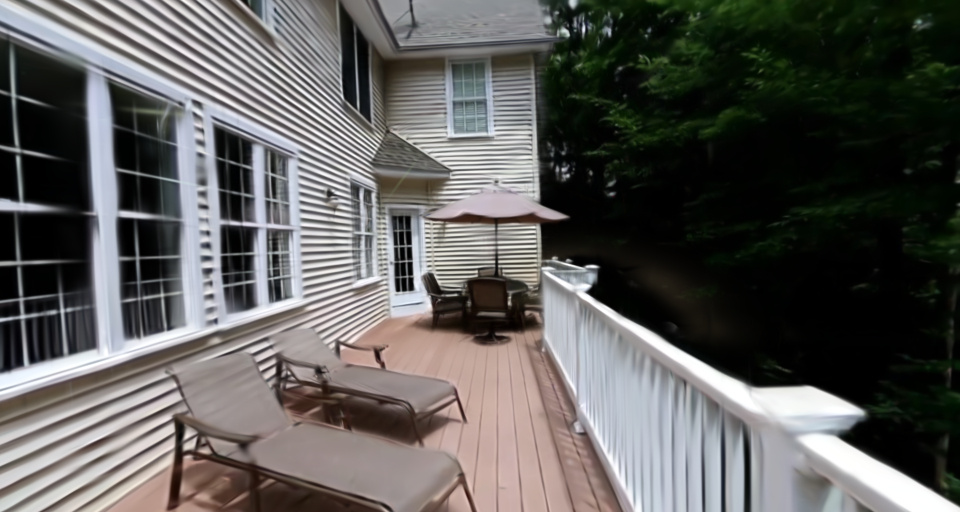} &
        \pcell[width=\linewidth]{23.4}{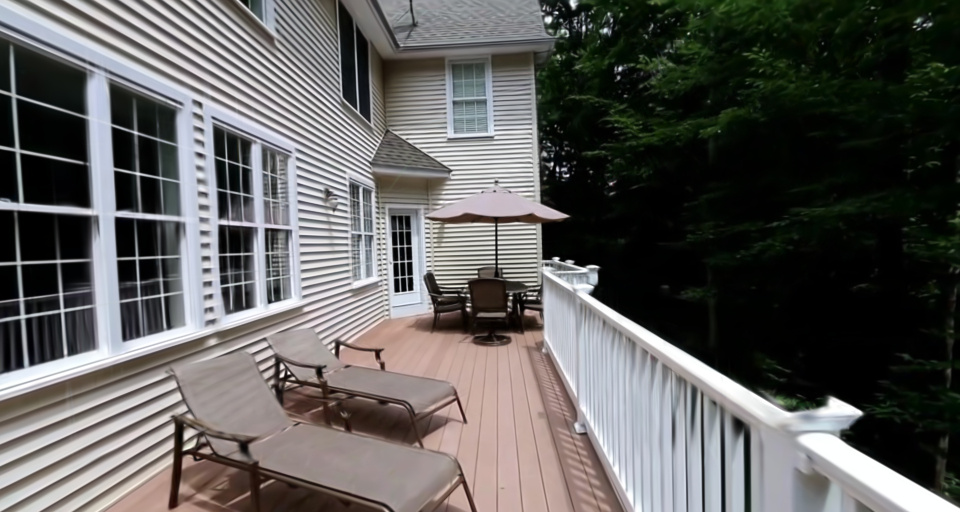} &
        \pcell[width=\linewidth]{24.4}{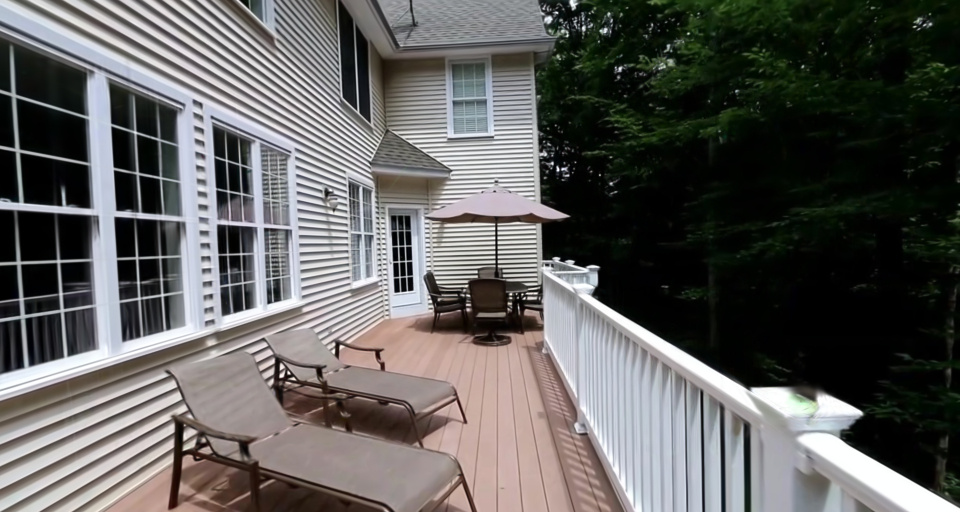} &
        \pcell[width=\linewidth]{24.7}{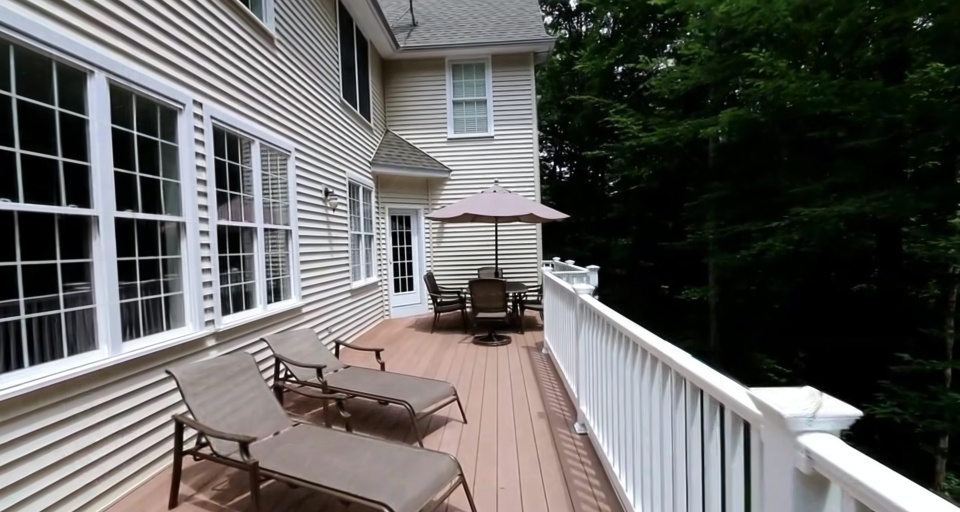} &
        \pcell[width=\linewidth]{24.7}{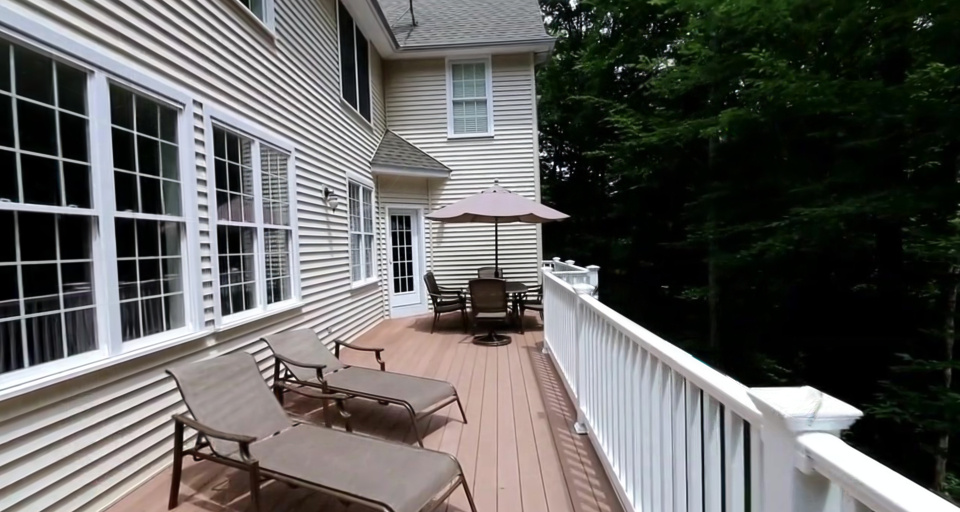} &
         &
         \\
        \noalign{\vskip 2\tabcolsep}
        \rotatebox{90}{\makecell{\smaller[1]{\adam{} error}}} &
        \rcell[width=\linewidth]{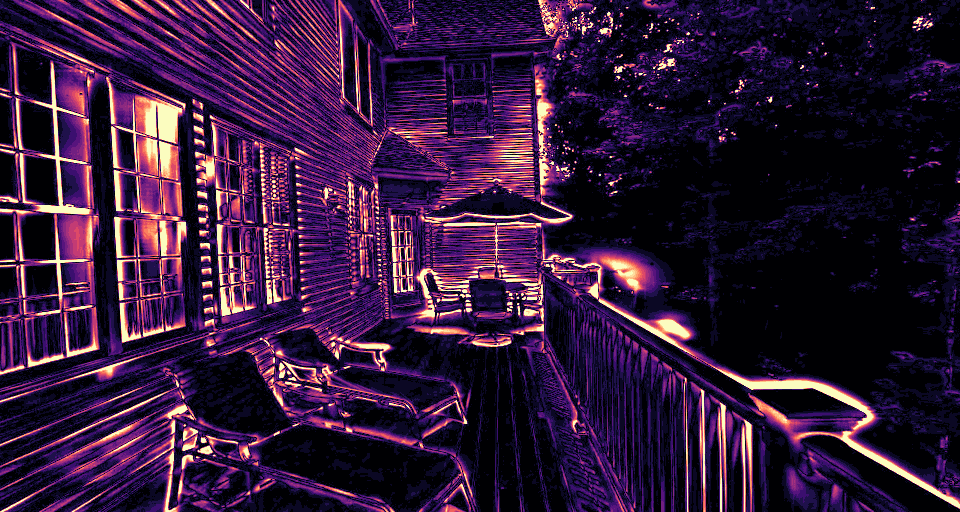} &
        \rcell[width=\linewidth]{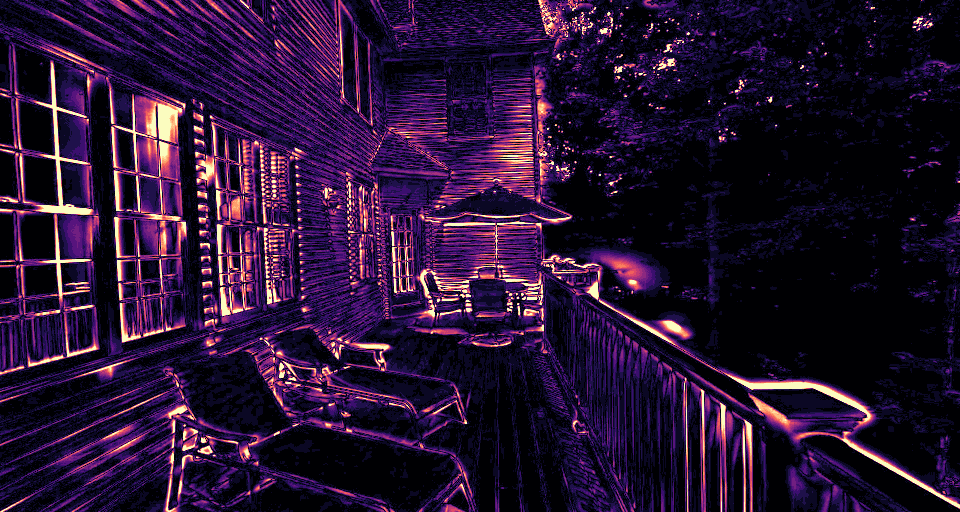} &
        \rcell[width=\linewidth]{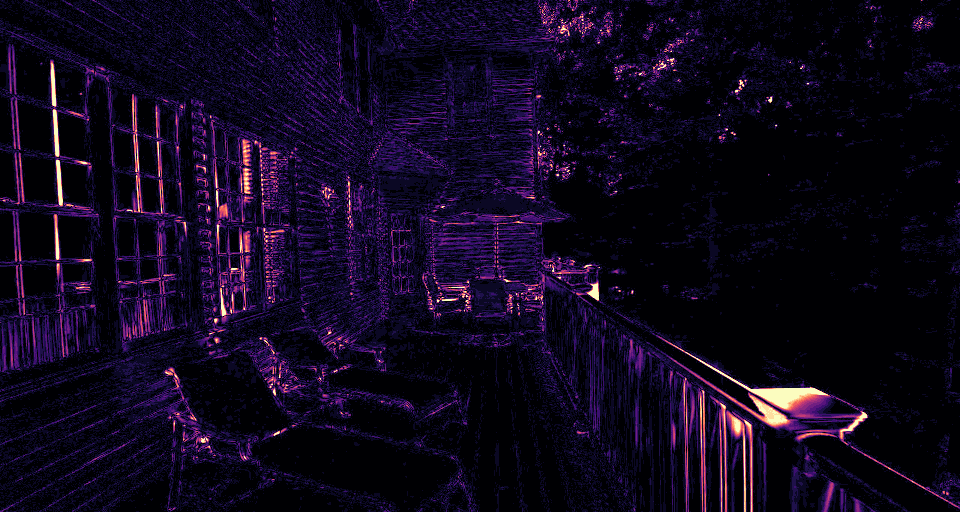} &
        \rcell[width=\linewidth]{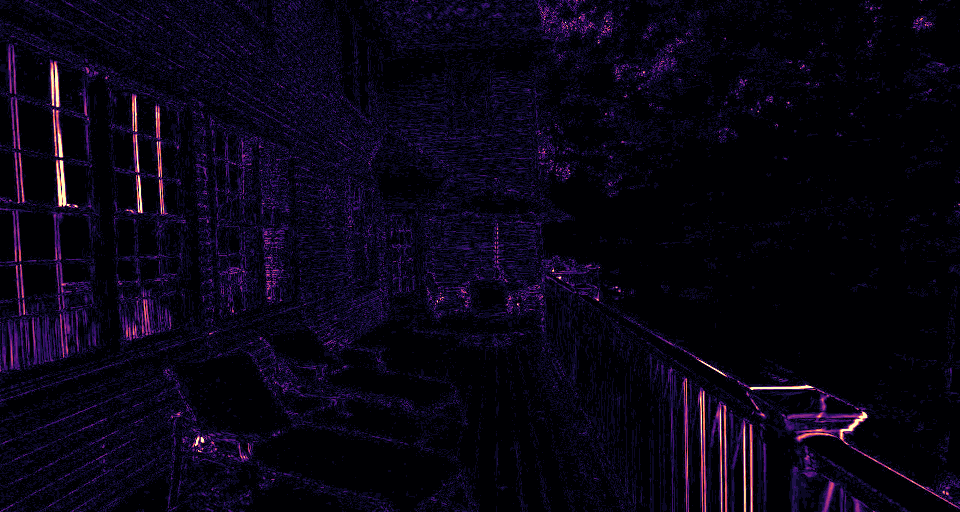} &
        \rcell[width=\linewidth]{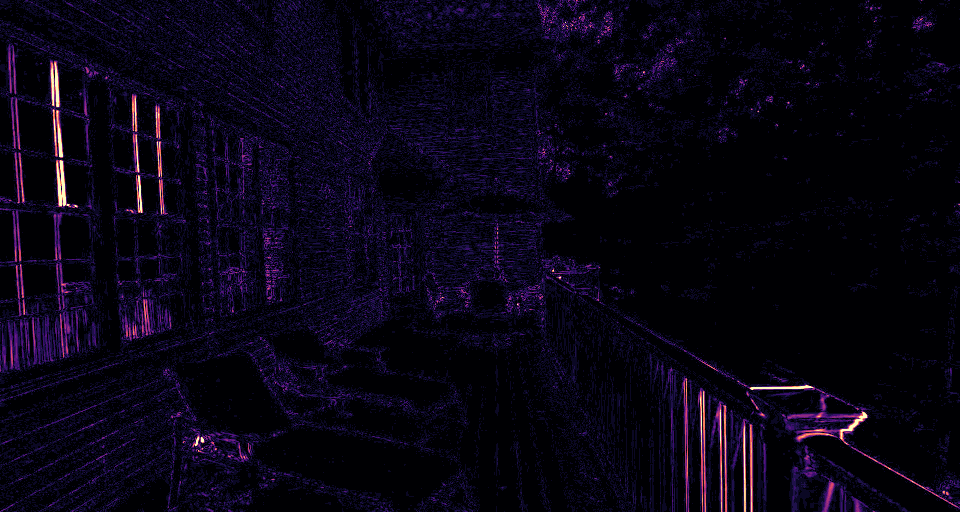} &
        \rotatebox{90}{\smaller[1]{Init error}} &
        \rcell[width=\linewidth]{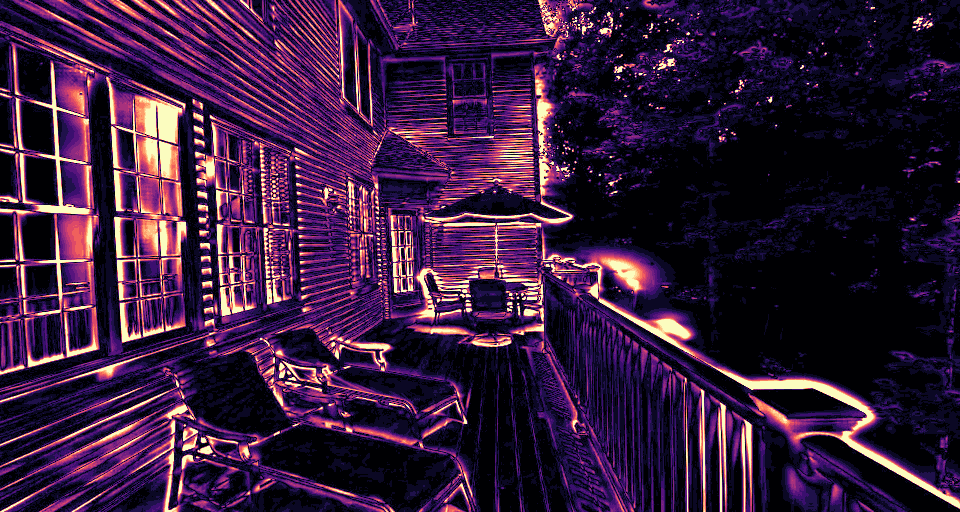} \\
        \noalign{\vskip 2\tabcolsep}
        \rotatebox{90}{\makecell{\smaller[1]{\ourshort{} error}}} &
        \rcell[width=\linewidth]{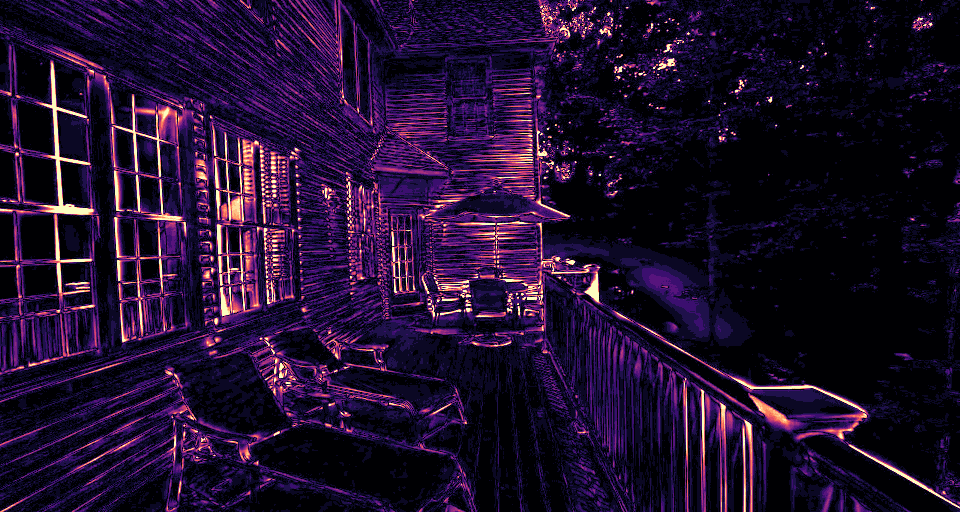} &
        \rcell[width=\linewidth]{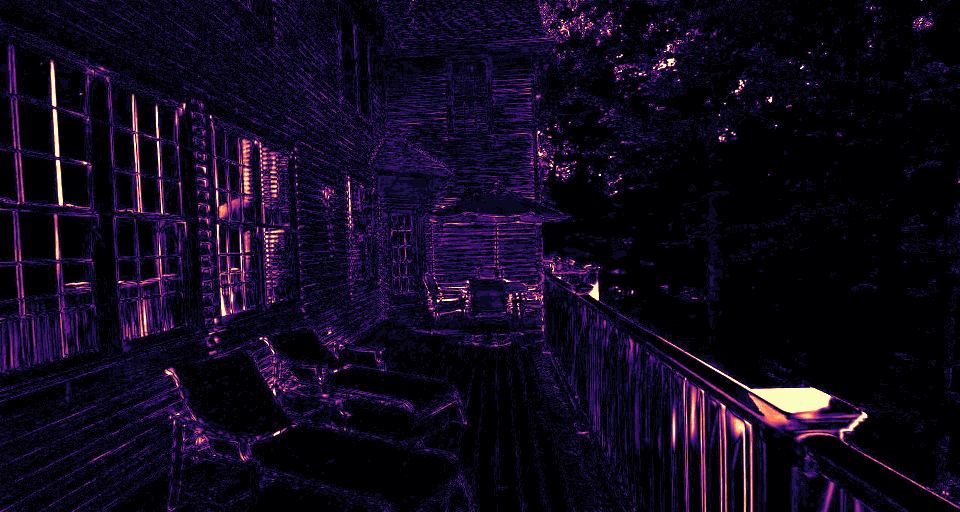} &
        \rcell[width=\linewidth]{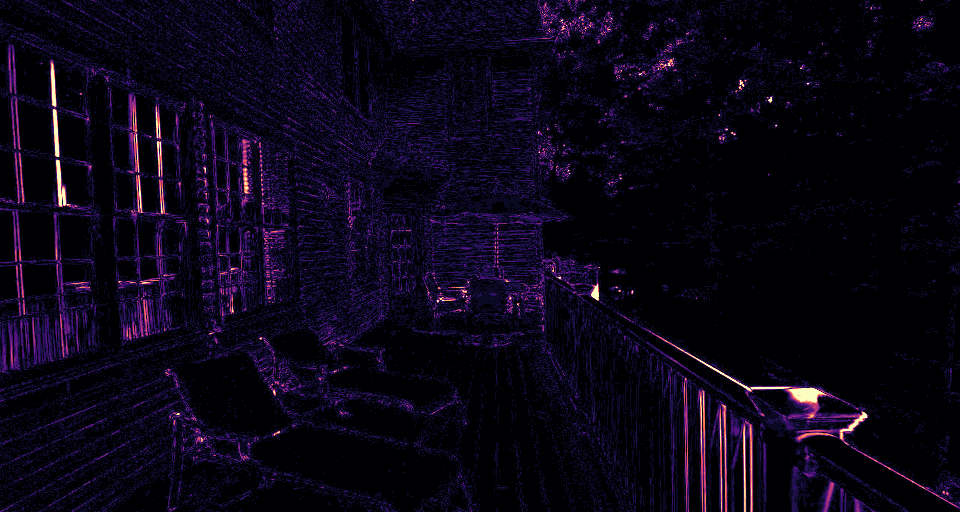} &
        \rcell[width=\linewidth]{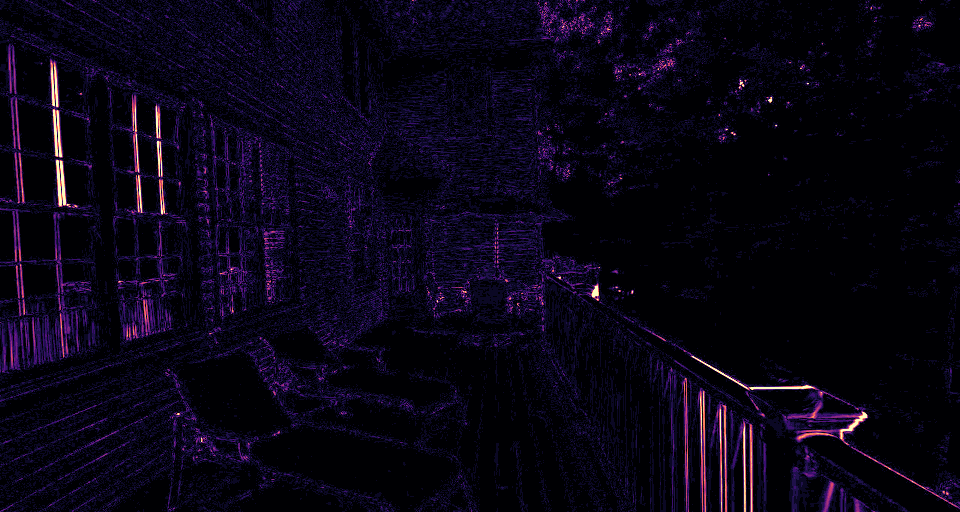} &
        \rcell[width=\linewidth]{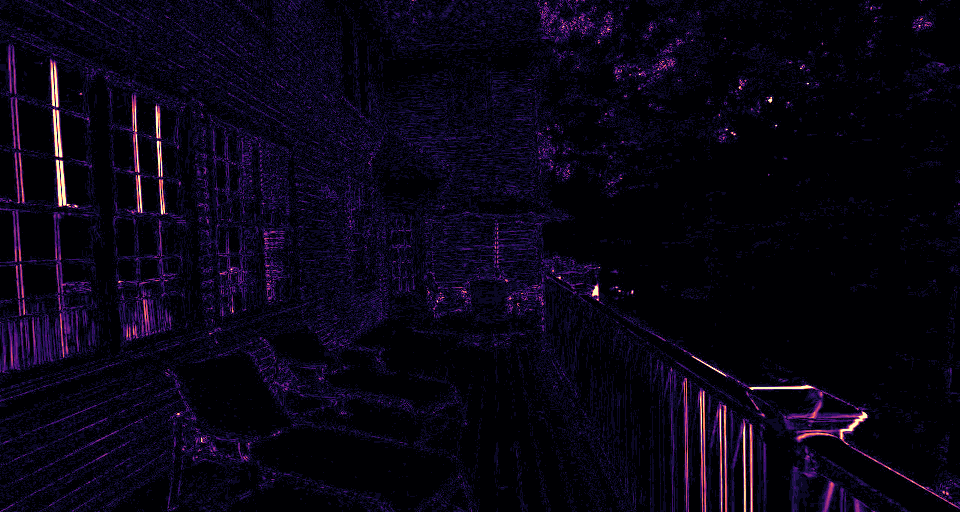} &
         &
         \\
    \end{tabular}
}
    \caption{
        \textbf{Zero-shot Generalization to RealEstate10k}. Scene reconstructions from RealEstate10K~\cite{Zhou2018SIGGRAPH} in the $8$ views, zero-shot high resolution setting ($512 \times 960$).
    }
    \label{fig:re10k-comp}
\end{figure*}

\clearpage
}

\section{Analysis of the learned optimizer}
\label{sec:app-analysis}

All analyses in this section were produced with the dedicated sparse checkpoint \ourssparse{} (\cref{subsec:supp-dedicated-vs-unified}).

\subsection{Statistical analysis}
\label{subsec:supp-statistical}

\begin{figure}[t]
    \centering
    \begin{subfigure}{0.80\linewidth}
        \centering
        \includegraphics[width=\linewidth]{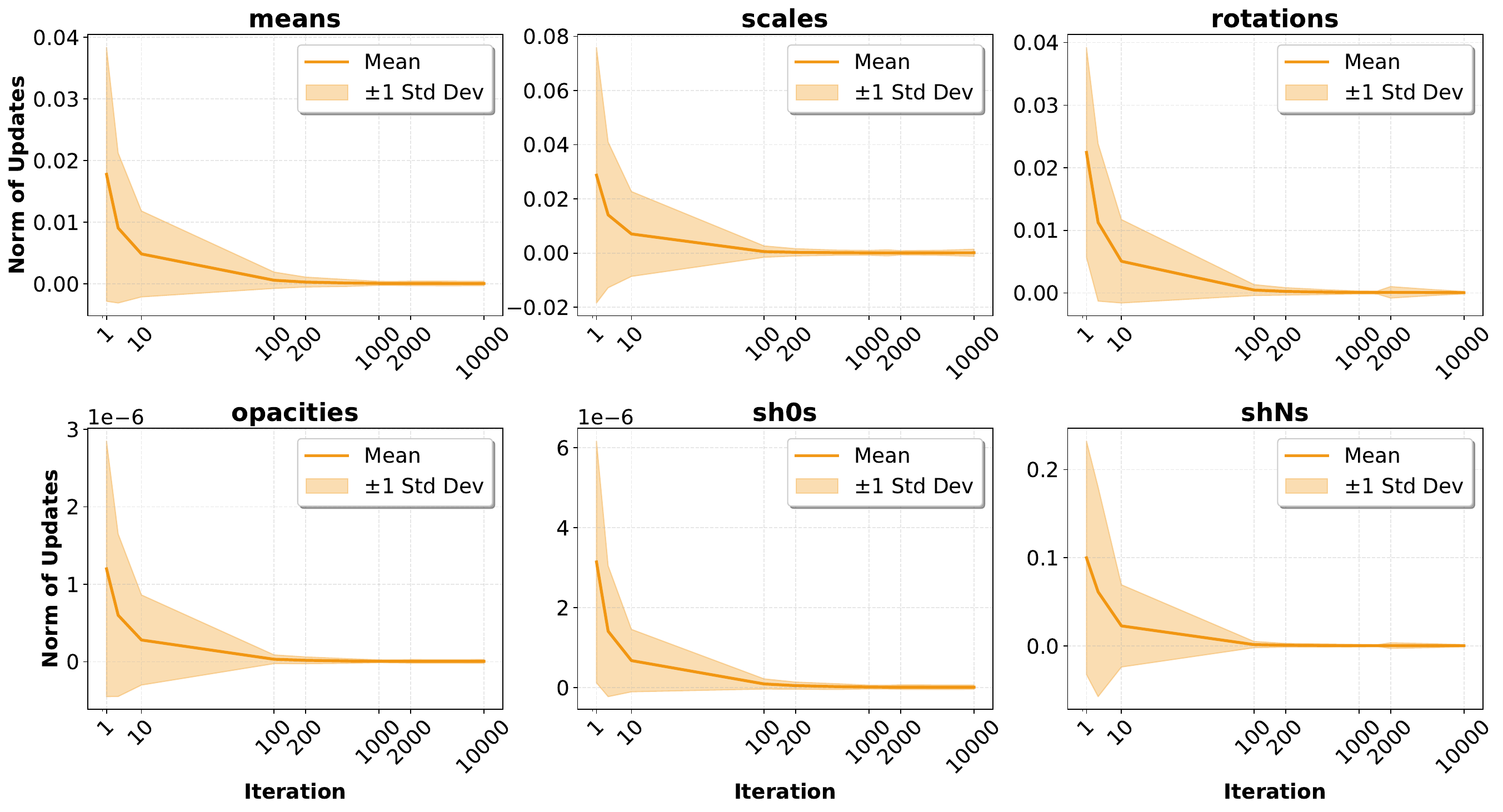}
        \caption{\ourssparse{}: Update norms}
        \label{fig:ours_updates_norm}
    \end{subfigure}

    \vspace{1em}

    \begin{subfigure}{0.80\linewidth}
        \centering
        \includegraphics[width=\linewidth]{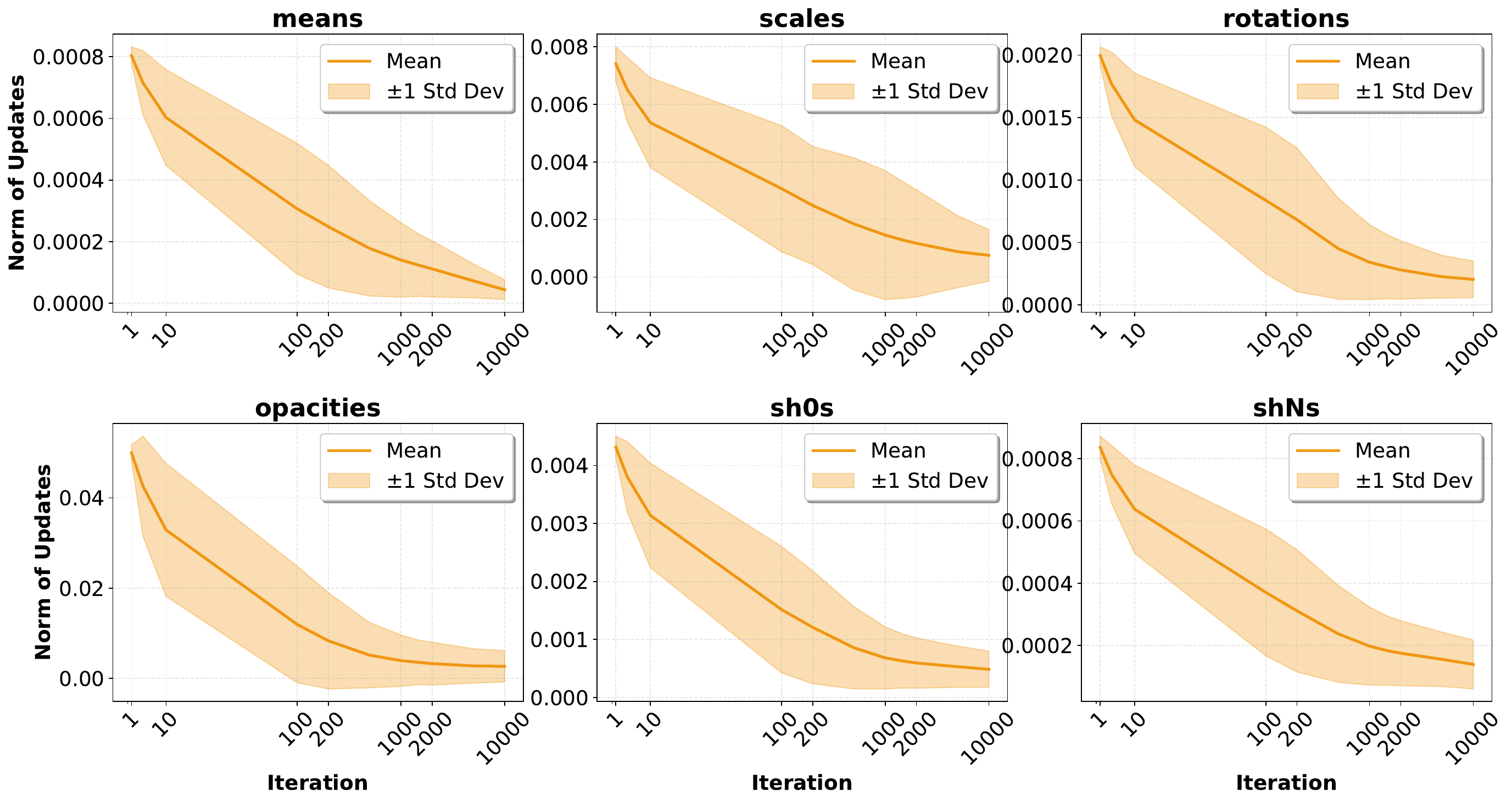}
        \caption{\adam{}: Update norms}
        \label{fig:3dgs_updates_norm}
    \end{subfigure}

    \vspace{1em}

    \begin{subfigure}{0.80\linewidth}
        \centering
        \includegraphics[width=0.65\linewidth]{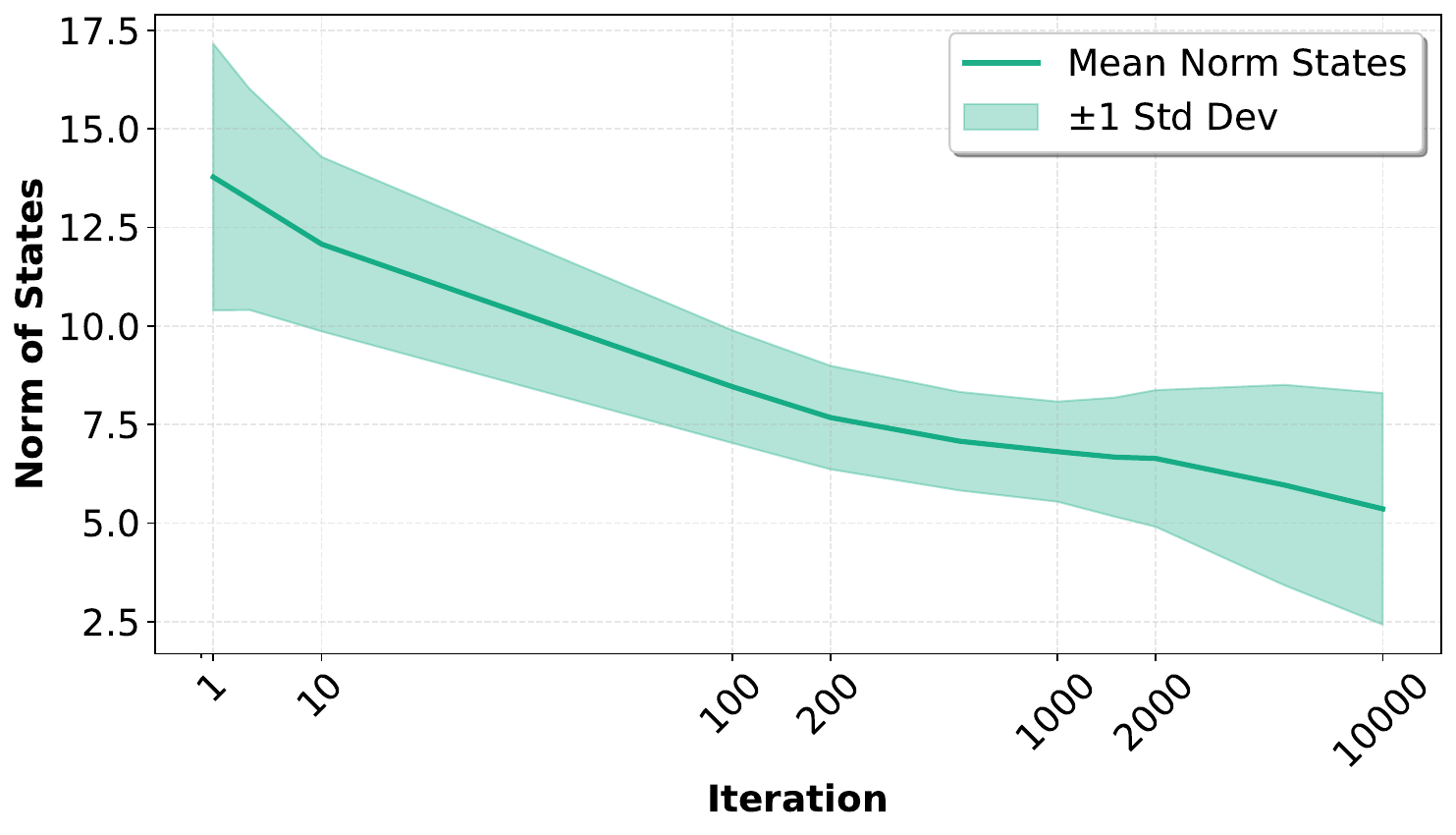}
        \caption{\ourssparse{}: Latent state norms}
        \label{fig:state-norm}
    \end{subfigure}

    \caption{\textbf{Optimization dynamics and state norms.} 
    \textbf{(a, b)} Gaussian update norms for \ourssparse{} and \adam{} over $10{,}000$ optimization steps. At each iteration, we report the update norm for each parameter group, averaged across 10 scenes from the DL3DV test set in the sparse low-resolution setting (8 views, $256 \times 448$).
    \textbf{(c)} Corresponding latent state norms for \ourssparse{} averaged across all Gaussians and scenes.
    See \cref{subsec:supp-statistical} for discussion.}
    \label{fig:combined_norms_analysis}
\end{figure}

\begin{figure}[h]
    \centering

    \begin{subfigure}{\linewidth}
        \centering
        \includegraphics[width=\linewidth]{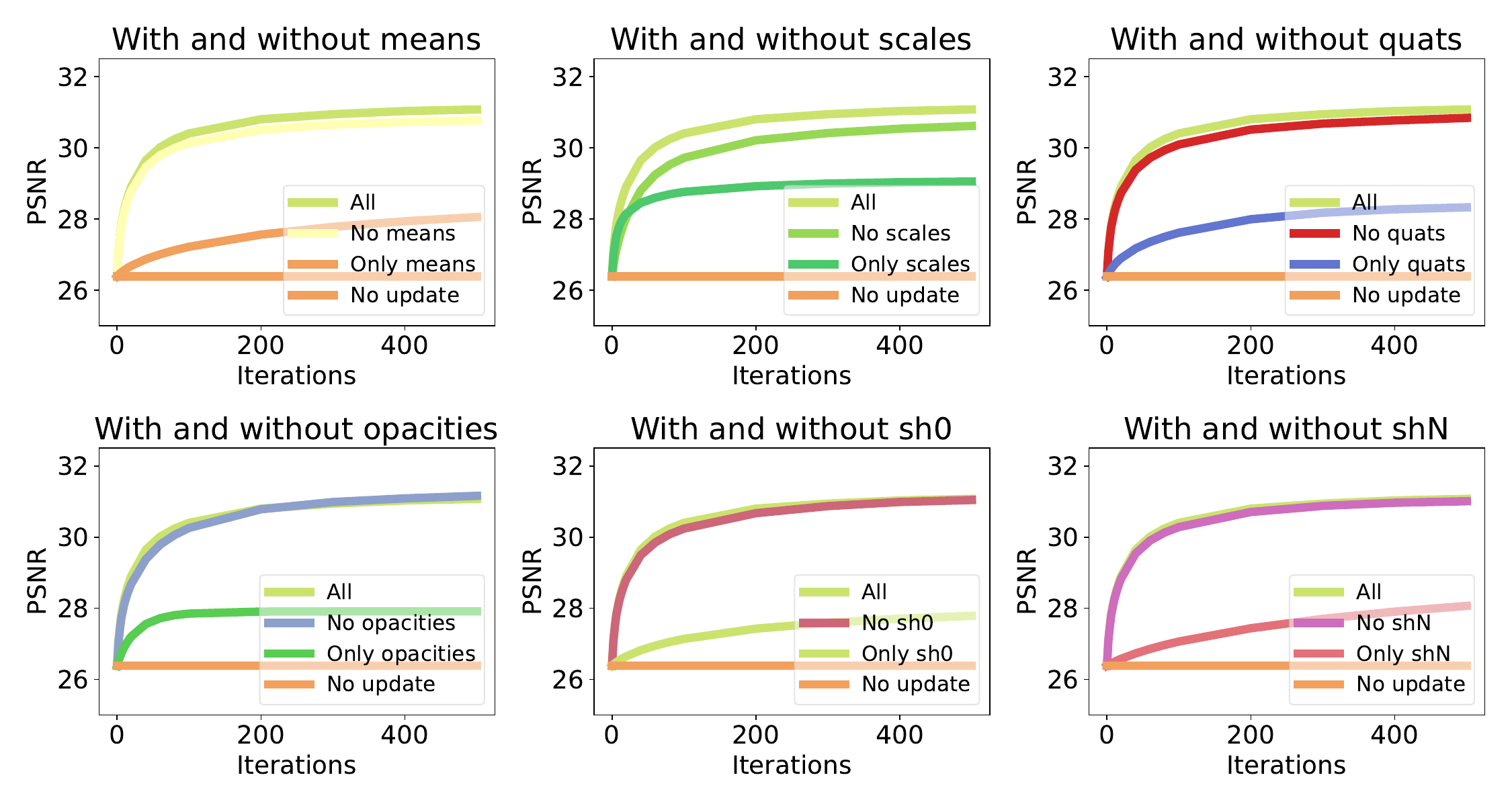}
        \caption{Adam: per-parameter update analysis.}
        \label{fig:adam-separate}
    \end{subfigure}

    \vspace{1em}

    \begin{subfigure}{\linewidth}
        \centering
        \includegraphics[width=\linewidth]{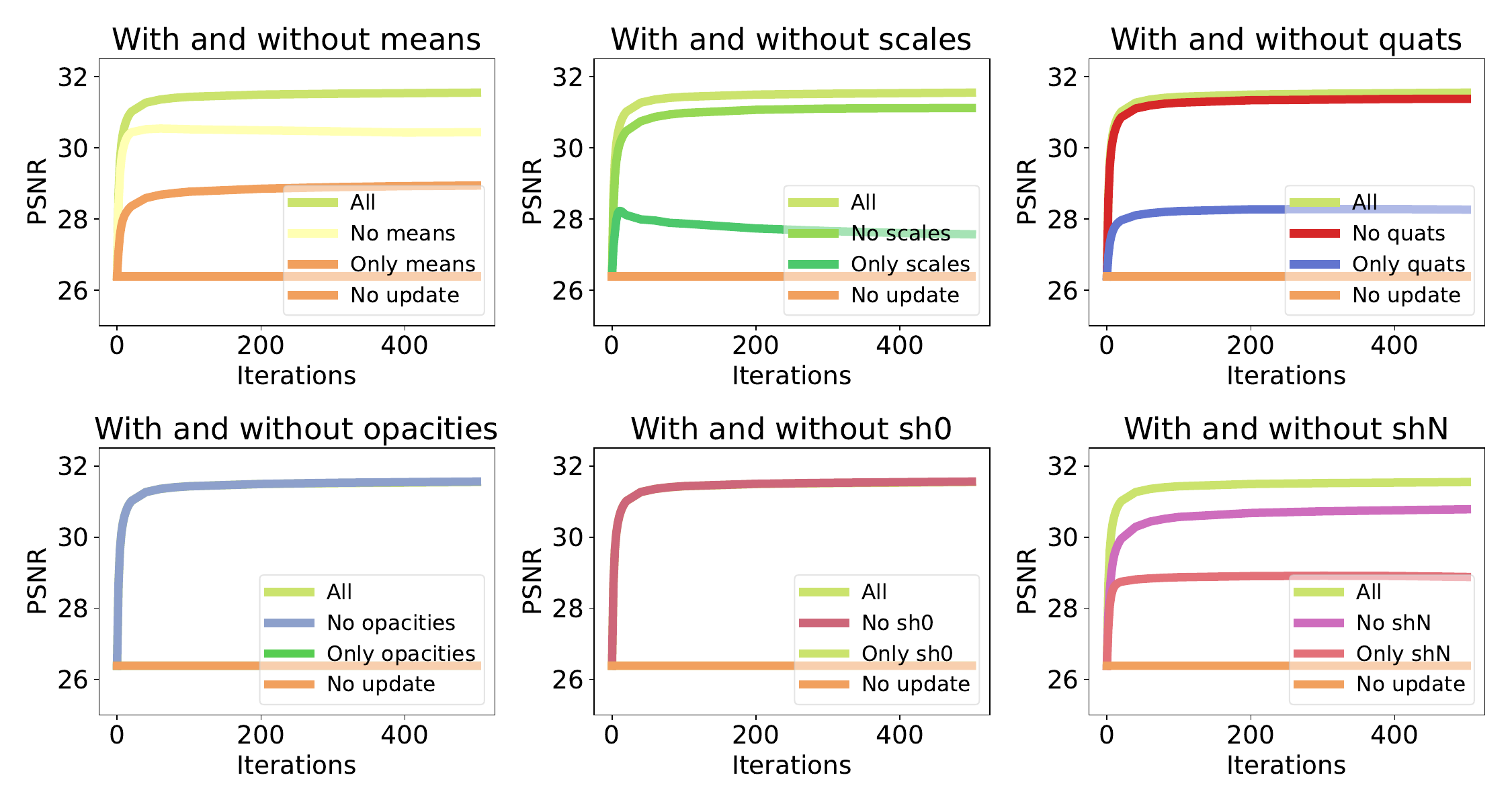}
        \caption{\ourssparse{}: per-parameter update analysis.}
        \label{fig:clogs-separate}
    \end{subfigure}

    \caption{\textbf{Per-parameter updates contributions.} Analysis of the contribution of each parameter in Adam optimization (top) and \ourssparse{} optimization (bottom). Each plot shows four configurations: (1) full optimization (all parameters updated), (2) updates applied to all parameters except the selected one, (3) updates applied only to the selected parameter, and (4) frozen parameters (no updates).
    Results are computed over 10 scenes from the DL3DV test set in the sparse low-resolution setting (8 views, $256 \times 448$ resolution).
    See \cref{subsec:supp-per-param} for discussion.}
    \label{fig:seperate-updates}
\end{figure}

In \cref{fig:combined_norms_analysis} we show the average norm of the updates for each group of parameters over 10,000 optimization steps for 10 scenes from the DL3DV test set.
We report the update norms for both \adam{} and our learned optimizer in \cref{fig:ours_updates_norm,fig:3dgs_updates_norm}.
As can be seen, the magnitudes of the updates are much larger during the early iterations, when our learned optimizer achieves its performance advantage over Adam, and gradually decay towards zero as optimization progresses.
\cref{fig:state-norm} further illustrates the decay of the internal state of the learned optimizer.
Results are computed over 10 scenes from the DL3DV test set in the sparse low-resolution setting (8 views, $256 \times 448$ resolution).

\subsection{Per-parameter contribution}
\label{subsec:supp-per-param}
In this section, we present an analysis of how each parameter group contributes to the behavior of \adam{} and \ourssparse{} during optimization.
\cref{fig:seperate-updates} compares how optimization progresses when individual parameter groups are selectively included or excluded during training. 
For both Adam (\cref{fig:adam-separate}) and \ourssparse{} (\cref{fig:clogs-separate}), each subplot isolates a single parameter: means, scales, quaternions, opacities, the first SH channel (sh0), and the remaining SH channels (shN), and evaluates four conditions: updating all parameters, freezing only that parameter group, updating only that parameter group, and performing no updates. 
Note that although \ourssparse{} now updates each parameter independently (or excludes different parameters), it was trained to \emph{jointly} update all parameters.
All of the experiments were conducted on 10 scenes of DL3DV test set in the sparse low resolution.

Across all parameters and across both optimizers, the ``All'' configuration reliably yields the fastest PSNR improvement, confirming that none of the parameter updates introduce negative effects.
For Adam, scales emerge as the most influential parameter: removing them noticeably slows convergence, while optimizing only the scales brings significant improvements. 
Opacities, sh0, and shN have little impact when removed, though optimizing only these groups still yields slight improvements over the no-update baseline. 
For \ourssparse{}, opacities and sh0 have little influence, whereas the means and shN contribute substantially. Removing the latter severely degrades convergence, and optimizing only these parameters yields clear gains. 
Since excluding opacities and sh0 does not reduce the maximum PSNR in these scenes, the meta-learner appears to have inferred that these parameters are good enough at initialization and should not be optimized.

\begin{figure}[h]
    \centering

    \begin{subfigure}{0.7\linewidth}
        \centering
        \includegraphics[width=\linewidth]{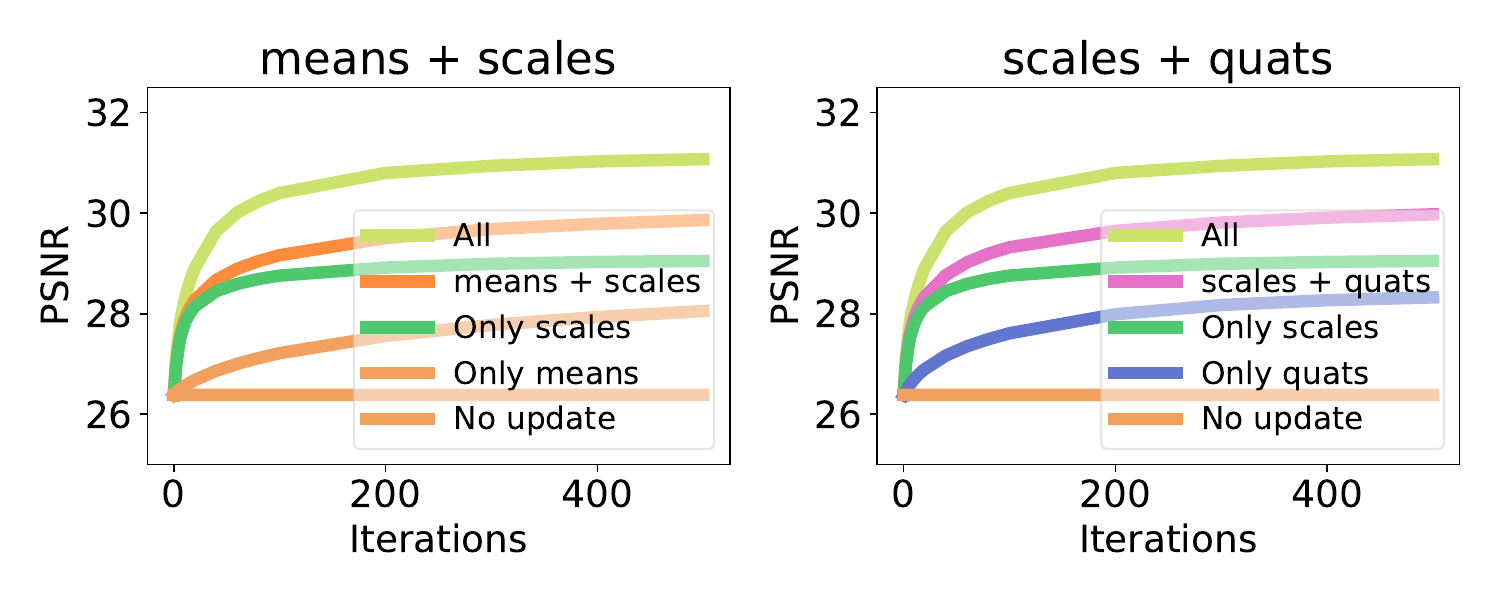}
        \caption{Adam: contribution of the combination of parameters.}
        \label{fig:adam-two-updates}
    \end{subfigure}

    \vspace{1em}

    \begin{subfigure}{0.7\linewidth}
        \centering
        \includegraphics[width=\linewidth]{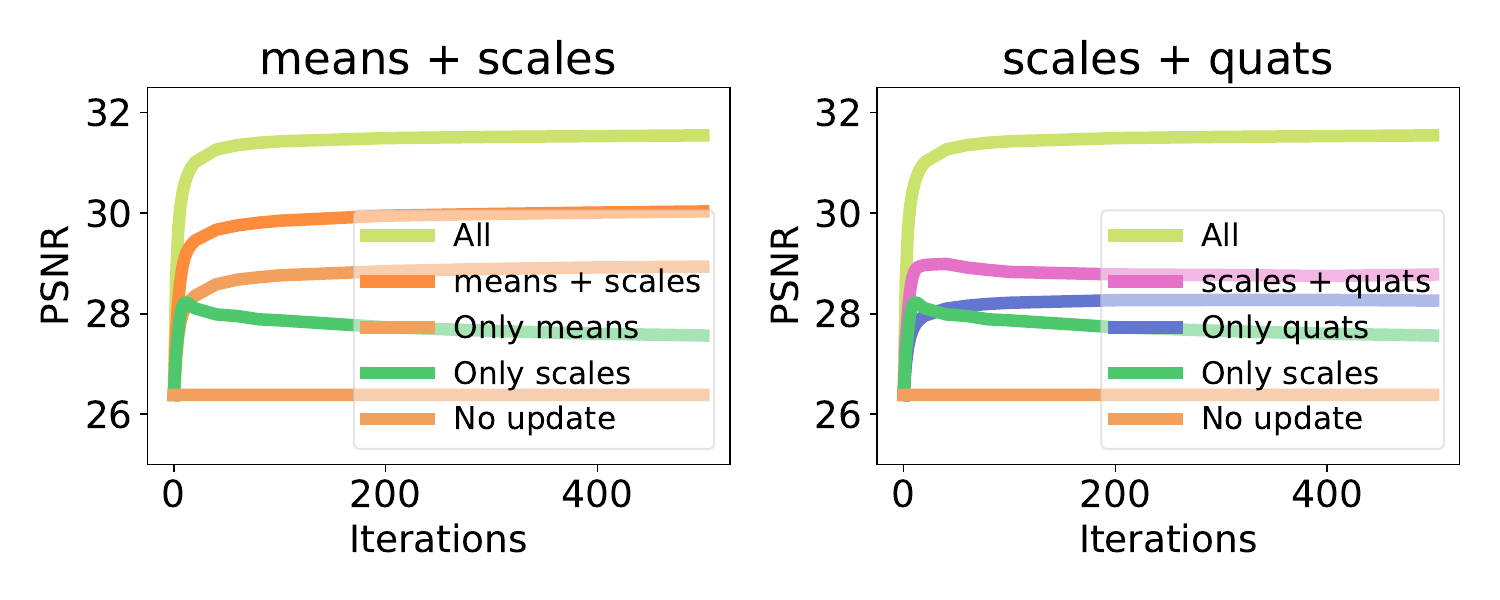}
        \caption{\ourssparse{}: contribution of the combination of parameters.}
        \label{fig:clogs-two-updates}
    \end{subfigure}

    \caption{
    \textbf{Joint parameters updates contributions.}
    Analysis of the joint contribution of two parameters in Adam (top) and \ourssparse{} (bottom). Each plot includes five configurations: (1) full optimization with all parameters updated, (2) updates applied only to the selected parameter pair, (3--4) updates applied to just one of the two parameters, and (5) all parameters frozen.
    For Adam, the results suggest (as expected) that parameters do not influence each other.
    In contrast, for \ourssparse{}, the combination of means and scales shows a positive interaction between the two parameters.
    Results are computed over 10 scenes from the DL3DV test set in the sparse low-resolution setting (8 views, $256 \times 448$ resolution).
    See \cref{subsec:supp-per-param} for discussion.}
    \label{fig:two-updates}
\end{figure}

In \cref{fig:two-updates}, we further analyze the interactions between parameter updates. For the combination of scales and rotations (quats), both Adam and \ourssparse{} behave additively, as the combined trend roughly mirrors the sum of their individual contributions.
However, for means and scales in \ourssparse{}, although updating only scales degrades performance from an early iteration, jointly updating means and scales continues to improve results throughout training. This suggests a dependency between these updates in \ourssparse{} that is not present in standard optimizers.

\begin{figure}[h]
    \centering
    \includegraphics[width=\linewidth]{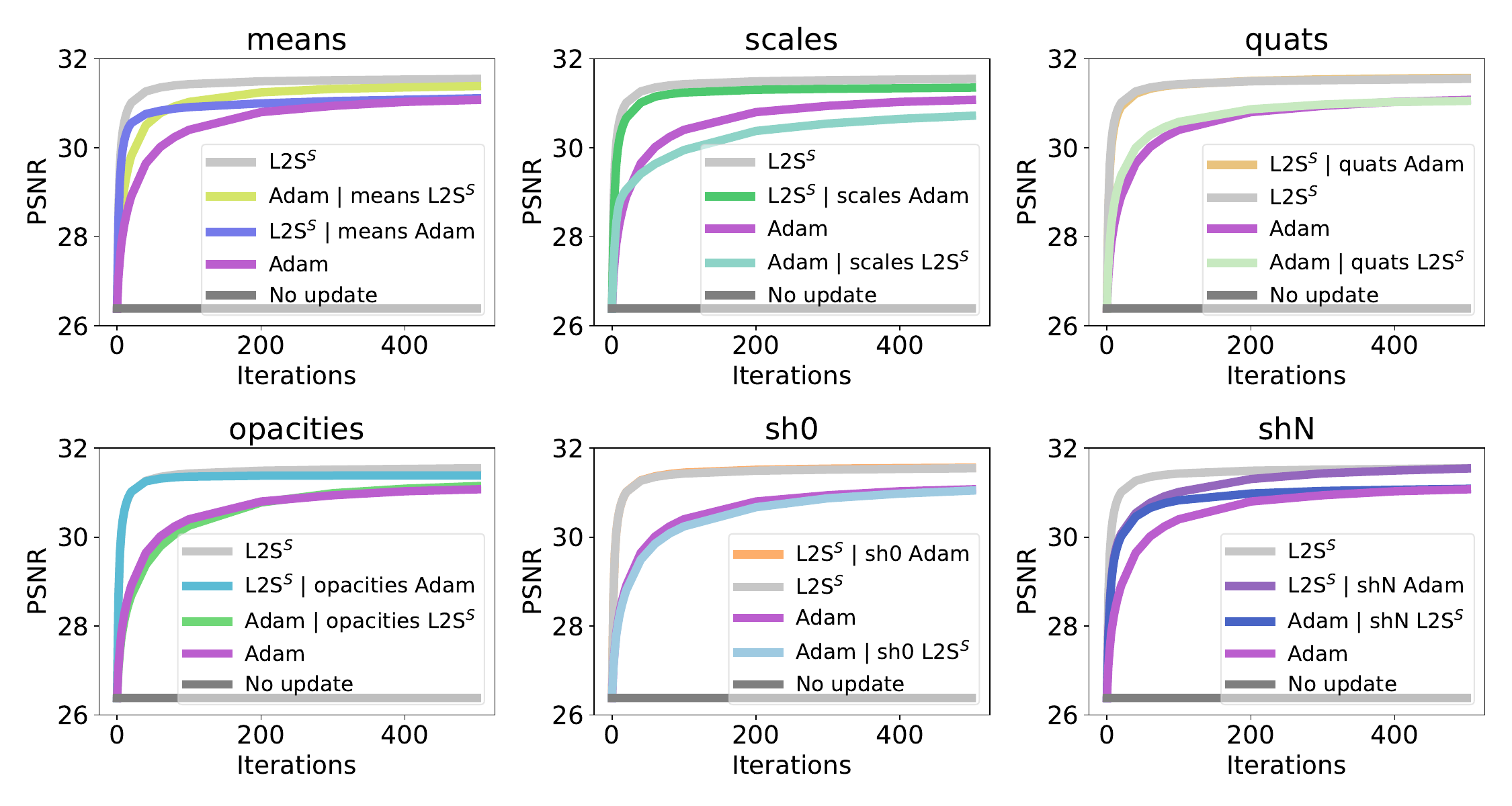}
    \caption{
    \textbf{Per-parameter updates swap.}
    Analysis of swapping parameter updates between Adam and \ourssparse{}. Each plot shows five configurations: (1) full optimization with \ourssparse{}, (2) \ourssparse{} with one parameter updated using Adam, (3) Adam with one parameter updated using \ourssparse{}, (4) full optimization with Adam, and (5) all parameters frozen. 
    The results indicate that the updates for the means and shN are better handled by \ourssparse{}: incorporating \ourssparse{} updates into Adam improves Adam's performance, whereas replacing \ourssparse{} updates with Adam's versions degrades performance. 
    For the scales, the evidence is mixed, as transferring this update in either direction worsens results. 
    For the remaining parameters, swapping updates appears to make little difference.
    Results are computed over 10 scenes from the DL3DV test set in the sparse low-resolution setting (8 views, $256 \times 448$ resolution).
    See \cref{subsec:supp-per-param} for discussion.}
    \label{fig:adam-clogs-switch-updates}
\end{figure}

In \cref{fig:adam-clogs-switch-updates}, we illustrate the effect of swapping update steps between Adam and \ourssparse{}. The results show that \ourssparse{} provides more effective updates for the means and shN parameters. Replacing Adam's updates with \ourssparse{}-style updates for these parameters consistently improves Adam's performance, whereas applying Adam's updates within \ourssparse{} leads to degradation. In contrast, the behavior of the scale parameters is less clear: transferring their updates between the two algorithms deteriorates performance in both directions, suggesting that neither method has a clear advantage for this parameter type. For the remaining parameters, swapping updates produces no change, indicating that Adam and \ourssparse{} behave similarly for those components.

\clearpage

\begin{table*}[!t]
\fontsize{9pt}{10pt}\selectfont   %
\adjustbox{max width=\textwidth,max totalheight=\textheight}{%
\begin{tabular}{ll}
\toprule
\textbf{Symbol} & \textbf{Description} \\
\midrule
\multicolumn{2}{l}{\textit{3D Gaussian representation}} \\
$\gaus = \{\gaus_m\}_{m=1}^G$ & Set of $G$ 3D Gaussians, also referred to as a matrix $\gaus \in  \mathbb{R}^{G \times p}$ \\
$\gaus_m = \{\bp_m, \bq_m, \bs_m, \alpha_m, \sh_m\}$ & Parameters of Gaussian $m$, also referred to as a vector $\gaus_m \in  \mathbb{R}^{p}$ \\
$\bp_m \in \mathbb{R}^3$ & 3D mean (center) of Gaussian $m$ \\
$\bq_m \in \mathbb{R}^4$ & Rotation (quaternion) of Gaussian $m$ \\
$\bs_m \in \mathbb{R}^3$ & Scale vector of Gaussian $m$ \\
$\alpha_m \in [0,1]$ & Opacity of Gaussian $m$ \\
$\sh_m \in \mathbb{R}^{d \times 3}$ & Spherical harmonics coefficients of Gaussian $m$ \\
$p$ & Number of parameters per Gaussian. In our case, $p=59$ \\
$G$ & Total number of Gaussians \\

\midrule
\multicolumn{2}{l}{\textit{Scene and rendering}} \\
$\cV$ & Set of $N$ views $\cV_i$ \\
$\cV_i$ & View $i$, including intrinsics, rotation, and translation $(\bK_i, \bR_i, \bt_i)$ \\
$\tilde{\bI}_i(\gaus)$ & Rendered image given Gaussians $\gaus$ and viewpoint $\cV_i$ \\
$\cL(\bI_i, \tilde{\bI}_i(\gaus))$ & Reconstruction loss for view $\cV_i$ given the Gaussians $\gaus$ \\

\midrule
\multicolumn{2}{l}{\textit{Optimization}} \\
$\tinner$ & Inner optimization step \\
$\gaus_{\tinner}$ & Gaussian set at optimization step $\tinner$ \\

$\ggaus = \nabla_{\gaus_{\tinner}} \cL_{\inner}(\bI_i, \tilde{\bI}_i(\gaus_{\tinner}))$ & Gradient of the inner loss w.r.t. the Gaussian parameters at $\tinner$\\
$f(\cdot)$ & Standard optimizer (e.g., SGD, Adam) \\
$f_{\btheta}(\cdot)$ & Learned optimizer parameterized by weights $\btheta$ \\
$\btheta$ & Parameters of the learned optimizer network \\
$\eta_{\tinner}$ & Learning rate or step size at inner step $\tinner$ \\
$\mathbf{m}_t, \mathbf{v}_t$ & First and second moment estimates in Adam \\
$\beta_1, \beta_2$ & Exponential decay rates for moment estimation \\
$\epsilon$ & Numerical stability constant in Adam \\
$\varepsilon$ & Vanishing-gradient threshold in the low-visibility loss. In our case, $\varepsilon = 10^{-8}$ \\

\midrule
\multicolumn{2}{l}{\textit{Meta-learning training}} \\
$\tmeta$ & Meta optimization step \\
$\cV^j$ & Scene $j$ sampled from the training dataset \\
$\gaus_{\tinner}^j(\btheta_{\tmeta})$ & State of Gaussians for scene $j$ after $\tinner$  updates using the learned optimizer $\btheta_{\tmeta}$ at $\tmeta$ \\
$\bI^j$ & Ground-truth RGB images of a set of views from scene $j$ \\
$\tilde{\bI}^j(\gaus^j_{\tinner}(\btheta_{\tmeta}))$ & Rendering of a set of images from scene $j$ given Gaussians $\gaus^j_{\tinner}$ \\
$\cL_{\meta}(\bI^j, \tilde{\bI}^j(\gaus^j_{\tinner+\tau}(\btheta_{\tmeta})))$ & Meta loss after $\tau$ unrolled steps \\
$\gtheta = \nabla_{\btheta_{\tmeta}} \cL_{\meta}$ & Gradient of the meta-loss w.r.t. the learned optimizer weights \\
$\tau$ & Number of unrolled inner steps per meta-iteration \\
$\lambda_{\text{render}}, \lambda_{\text{lvs}}, \lambda_{\text{stab}}$ & Weights of the three meta-loss terms, set to $1$, $2$ and $1$ \\
$V$ & Total number of training scenes \\

\midrule
\multicolumn{2}{l}{\textit{Model architecture}}\\
$\ggaust{t} \in \nR^{G \times p}$ & Per-Gaussian gradients at iteration $t$, the input to the network \\
$\aggaust{t} \in \nR^{G \times p}$ & Adam-normalized gradients, obtained from $\ggaust{t}$ inside the network (\cref{eq:adam-normalization}) \\
$\gaust \in \nR^{G \times p}$ & Gaussian parameters at iteration $t$ \\
$\bs_t \in \nR^{G \times 256}$ & Latent states of Gaussians at iteration $t$ \\
$\bs_{t+1}$ & Updated latent states before scaling \\
$\brho_\bs \in \nR^{G}$ & Predicted per-Gaussian scaling coefficients for latent states \\
$\tilde{\bs}_{t+1}$ & Scaled latent states after applying $\brho_\bs$ \\
$\bO_{\gaust} \in \nR^{G \times (p+1)}$ & Raw parameter updates predicted by the network \\
$\tilde{\Delta}_{\gaust} \in \nR^{G \times p}$ & Direction of parameter updates (unit-length) \\
$\brho_{\Delta_t} \in \nR^{G}$ & Learned per-Gaussian update magnitude \\
$\Delta_{\gaust}$ & Final Gaussian parameter updates \\

\midrule
\multicolumn{2}{l}{\textit{Checkpoint buffer and training dynamics}} \\
$p_{\text{buffer}}$ & Probability of sampling a scene state from the checkpoint buffer \\
$p_{\text{push}}$ & Probability of storing a newly optimized scene into the buffer \\
$p_{\text{push-back}}$ & Probability of re-storing a previously sampled scene \\
$\cB$ & Checkpoint buffer containing intermediate Gaussian states \\
$\tau_a$ & Number of optimizer rollout inner steps before storing to checkpoint buffer \\
\bottomrule
\end{tabular}
}
\vspace{10pt}
\caption{Summary of notations used in this paper.}
\label{tab:notation}
\end{table*}

\end{document}